\documentclass{article}

\usepackage{microtype}
\usepackage{graphicx}
\usepackage{subcaption}
\usepackage{booktabs} %

\usepackage{hyperref}

\usepackage[accepted]{icml2026}

\usepackage{amsmath}
\usepackage{amssymb}
\usepackage{mathtools}
\usepackage{amsthm}
\usepackage{bbm}

\usepackage[nameinlink,capitalize,noabbrev]{cleveref}
\crefformat{equation}{{\color{mydarkblue}(#2\textcolor{mydarkblue}{#1}#3)}} %
\usepackage{thmtools}

\usepackage{multirow}

\usepackage{enumitem}

\usepackage{soul}

\usepackage{xspace}
\newcommand{\PINNfluence}{\textsc{PINN\-fluence}\xspace}
\newcommand{\Inf}{\operatorname{Inf}}
\newcommand{\bx}{x}
\newcommand{\bz}{z}

\newcommand{\norm}[1]{|\! | #1 |\! |}

\usepackage{stmaryrd}
\usepackage{trimclip}

\makeatletter
\DeclareRobustCommand{\shortto}{%
  \mathrel{\mathpalette\short@to\relax}%
}

\newcommand{\short@to}[2]{%
  \mkern2mu
  \clipbox{{.5\width} 0 0 0}{$\m@th#1\vphantom{+}{\shortrightarrow}$}%
  }
\makeatother

\theoremstyle{plain}
\newtheorem{theorem}{Theorem}[section]

\newtheorem{lemma}[theorem]{Lemma}
\newtheorem{corollary}[theorem]{Corollary}
\theoremstyle{definition}
\newtheorem{definition}[theorem]{Definition}

\theoremstyle{remark}

\usepackage{threeparttable}

\usepackage[textsize=tiny]{todonotes}

\icmltitlerunning{PINNfluence: Interpreting PINNs through Influence Functions}

\begin{document}

\twocolumn[
  \icmltitle{PINNfluence: Interpreting PINNs through Influence Functions}

  \icmlsetsymbol{equal}{*}

  \begin{icmlauthorlist}
    \icmlauthor{Aleksander Krasowski}{equal,hhi}
    \icmlauthor{Jonas Naujoks}{equal,hhi}
    \icmlauthor{Moritz Weckbecker}{hhi}
    \icmlauthor{Galip Ü. Yolcu}{hhi}
    \icmlauthor{Thomas Wiegand}{hhi,tub,bifold}
    \icmlauthor{Sebastian Lapuschkin}{hhi,xaidublin}
    \icmlauthor{Wojciech Samek}{hhi,tub,bifold}
    \icmlauthor{René P. Klausen}{hhi}
  \end{icmlauthorlist}

    \icmlaffiliation{hhi}{Department of Artificial Intelligence, Fraunhofer HHI, Berlin, Germany}
    \icmlaffiliation{tub}{Department of Electrical Engineering, Technische Universität Berlin, Berlin, Germany}
    \icmlaffiliation{bifold}{BIFOLD -- Berlin Institute for the Foundations of Learning and Data, Berlin, Germany}
    \icmlaffiliation{xaidublin}{Centre of eXplainable Artificial Intelligence, Technological University Dublin, Dublin, Ireland}

  \icmlcorrespondingauthor{Aleksander Krasowski}{aleksander.krasowski@hhi.fraunhofer.de}
  \icmlcorrespondingauthor{Jonas Naujoks}{jonas.naujoks@hhi.fraunhofer.de}
  \icmlcorrespondingauthor{René P. Klausen}{rene.pascal.klausen@hhi.fraunhofer.de}

  \icmlkeywords{Physics-Informed Machine Learning, Physics-Informed Neural Networks, PINNs, Explainable AI, XAI, Interpretability, ICML}

  \vskip 0.3in
]

\printAffiliationsAndNotice{\icmlEqualContribution}

\begin{abstract}

Physics-informed neural networks (PINNs) have emerged as a powerful deep learning approach for solving partial differential equations (PDEs) in the physical sciences, yet their behavior remains largely opaque and is typically understood through failure mode analyses rather than explicit interpretability.
To address this issue, we introduce \PINNfluence, a training data attribution framework for interpreting PINNs based on influence functions.
By extending influence functions to composite physics-informed training objectives, we enable fine-grained attribution between predictions, loss components, and training data points.
Through benchmark experiments across various PDEs, we demonstrate that influence patterns provide granular diagnostics that distinguish structural characteristics across well-trained and poorly-trained PINNs.
\PINNfluence thus opens a new avenue for understanding and improving the reliability of PINNs through the lens of their data.

\end{abstract}

\section{Introduction}

Partial differential equations (PDEs) are the central modeling language in the sciences, describing how quantities evolve across space and time under governing laws.
Solving these equations accurately and efficiently is essential for scientific prediction and control, yet often computationally demanding. In recent years, machine learning (ML) approaches have proven themselves exceptionally adept at approximating solutions to such problems, bundling fast inference and strong empirical performance once trained.
Physics-informed neural networks (PINNs) \cite{raissi_physics-informed_2019} integrate physical knowledge by embedding PDEs into neural network training objectives \cite{karniadakis_physics-informed_2021,kim_knowledge_2021,cuomo_scientific_2022}, enabling applications across fluid mechanics, electromagnetics, disease modeling, and optics, inter alia \cite{cai_physics-informed_2021,beltran-pulido_physics-informed_2022,berkhahn_physics-informed_2022,medvedev_modeling_2023}.
Despite their flexibility, understanding and improving PINN behavior when training fails remains challenging \cite{wang_understanding_2021,krishnapriyan_characterizing_2021,wang_when_2022,RathoreChallengesTrainingPINNs2024,liu_config_2025}.
\begin{figure*}[ht]
    \centering 
    \includegraphics[width=\textwidth]{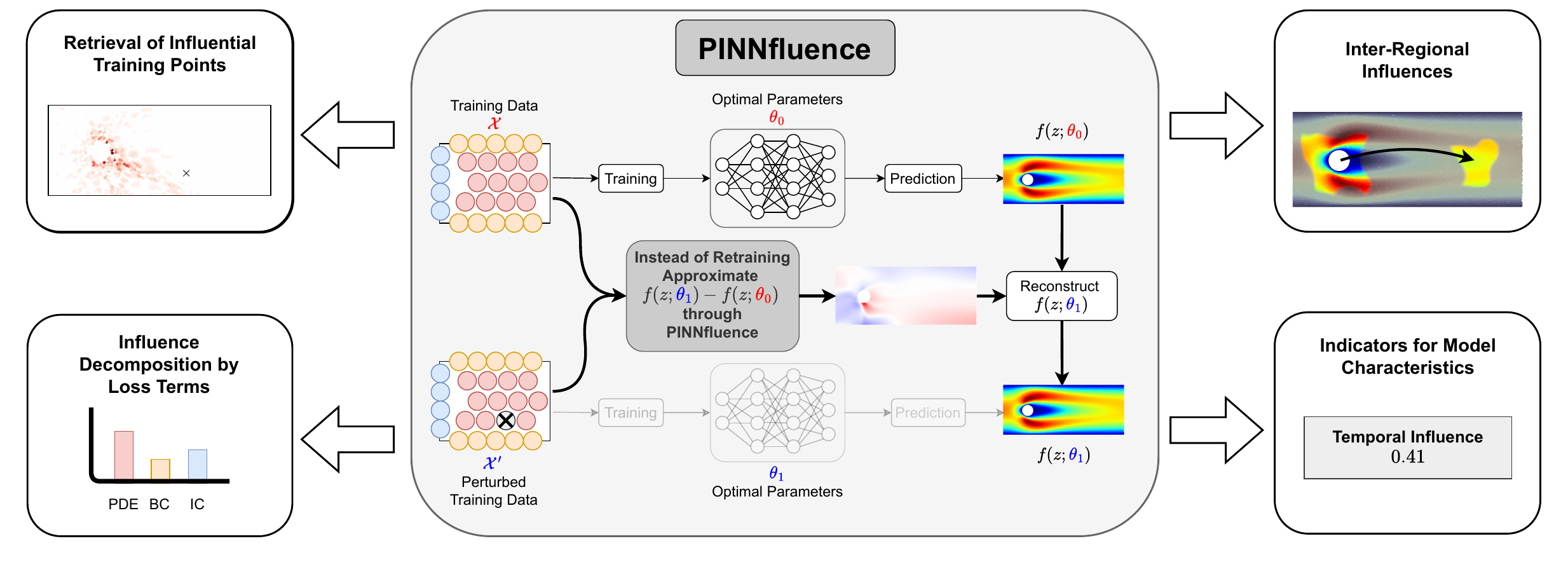}
    \caption{\PINNfluence approximates the effect of perturbing training data by estimating the difference $f(z, \textcolor{blue}{\theta_1}) - f(z, \textcolor{red}{\theta_0})$ \emph{without retraining}. Here, $\textcolor{red}{\theta_0}$ denotes the parameters obtained from the original training set $\textcolor{red}{\mathcal X}$, while $\textcolor{blue}{\theta_1}$ corresponds to a perturbed training set $\textcolor{blue}{\mathcal X'}$ that differs by a single training point.
    This enables attribution of model predictions at test points $\bz$ to individual training samples and loss components, supporting \textbf{(i)} retrieval of influential training points for the given prediction, \textbf{(ii)} inspection of how one region influences another, \textbf{(iii)} decomposition by  individual loss terms, and \textbf{(iv)} construction of diagnostic indicators that characterize model behavior.}
    \label{fig:visual_abstract}    
\end{figure*}

The existing understanding of PINNs is largely based on analyses of training dynamics, failure modes, and ill-conditioned loss landscapes. Prior work attempts to alleviate these issues through improvements in optimization, architecture, objective function design, and training data selection \cite{jagtap_adaptive_2020,mcclenny_self-adaptive_2023,wang_experts_2023,wu_propinn_2025,jeong_mitigating_2025}. Further, rigorous error guarantees and uncertainty quantification for PINNs are active fields of research and often difficult to obtain in practice \cite{EirasEfficientErrorCertification2024,hillebrecht_rigorous_2025}. Both of these strands of research aim to make PINNs reliable and performant, qualities that are crucial for wide-scale adoption. 
In parallel, a substantial body of work on explainable artificial intelligence (XAI) has developed tools to interpret and diagnose complex machine learning models, improving trust, scientific insights, and model validation \cite{ross_right_2017,rieger_interpretations_2020} in varied domains such as medical imaging \cite{hense_xmil_2024}, computer vision \cite{bach_pixel-wise_2015,bender_towards_2023} and natural language processing \cite{bricken2023monosemanticity}.

While interpretability\footnote{We use this term interchangeably with explainability.} has become a central theme in deep learning \cite{samek_explaining_2021,rudin_interpretable_2021} and scientific machine learning in particular \cite{wetzel_interpretable_2025}, PINNs still lack dedicated explainability methods that support transparency, systematic diagnosis and validation -- a precondition for their widespread application. In particular, although existing analyses of PINN failure modes provide valuable insights at the level of training dynamics and model classes, they fall short of providing diagnostic or attribution-based explanations that reveal how individual training points and physics constraints shape model predictions.
This leaves a gap for training-data-centric interpretability methods for PINNs, which we address in this work. This gap is of particular importance, as PINNs face a verification problem distinct from classical supervised learning: a low training loss does not imply a correct solution to the underlying PDE, since PINNs can converge to trivial solutions with low residuals yet high errors \cite{daw_mitigating_2023, rohrhoferRoleFixedPoints2023}. Trained models can thus exhibit pathological structure invisible to residual inspection alone. Such post-hoc methods have the potential to improve transparency and robustness by enabling fine-grained diagnosis and control of PINN behavior, thereby complementing existing approaches.

Specifically, we adapt influence functions, a training data attribution (TDA) method, to PINNs. TDA quantifies the influence of individual training points on model behavior and has proven effective for debugging models, identifying memorization and harmful data, and understanding model behavior through the lens of the training set in domains such as computer vision and large language models \cite{koh_understanding_2017,pruthi_estimating_2020,braun_influence_2022,hammoudeh_training_2024}.
An overview of our proposed \PINNfluence framework and the types of influence-based analyses it provides is shown in \cref{fig:visual_abstract}.

\paragraph{Our contributions can be summarized as follows:}
\begin{itemize}[itemsep=1pt, topsep=1pt]
    \item We generalize influence functions to PINNs, resulting in \PINNfluence, an interpretability framework that attributes predictions to individual training points and decomposes their influence by loss components.
    \item We provide a toolbox of diagnostic indicators, including temporal, directional, and region-to-region measures, that enable systematic analysis of trained PINNs.
    \item Across five time-dependent PDEs, we identify consistent structural differences between well-trained and poorly-trained models. 
    Well-trained models show diminishing influence of the initial condition (IC), while poorly-trained models exhibit persistent IC dependence over time.
\end{itemize}

\section{Theoretical Background}\label{sec:theory-background}

\subsection{Physics-Informed Neural Networks (PINNs)}

PDEs are ubiquitous modeling tools in the computational sciences: For example, they describe the motion of galaxies or the spread of infectious diseases. A PDE is typically posed together with initial and boundary conditions, such as the starting position of galaxies or the initial amount of infected specimens. A PDE with these constraints is called an initial-boundary value problem (IBVP).
PINNs \cite{raissi_physics-informed_2019} provide a widely-used ML-based framework for approximating solutions of IBVPs. Let $\Omega \subset \mathbb{R}^n$ be a connected, bounded, open set with piecewise smooth boundary $\partial \Omega$, and define its closure by $\overline{\Omega} := \Omega \cup \partial \Omega$. We study PDE constraints in the domain $\Omega$ together with $K$ boundary and/or initial constraints posed on subsets $\Gamma_k \subseteq \partial \Omega$:
\begin{align}
    \mathcal{N}[u](x) &= 0, \quad x \in \Omega \label{eq:IBVPpde}\\
    \mathcal{B}_k[u](x) &= 0, \quad x \in \Gamma_k\ \text{ for } k=1,\dots,K \label{eq:bc} \quad\text{,}
\end{align}
where $\mathcal{N}$ and $\mathcal{B}_k$ are differential operators acting on the solution  $u:\overline \Omega \to \mathbb R^d$ of \cref{eq:IBVPpde} and \cref{eq:bc} with $K$ initial and/or boundary conditions. Here, $x$ denotes the spatio-temporal coordinates, which may be purely spatial, purely temporal, both, or more generally any physical quantity on which the PDE depends. The above formulation \cref{eq:bc} encompasses both Dirichlet and Neumann boundary conditions, among others.
The aim of the PINN approach is to approximate a solution $u(x)$ of \cref{eq:IBVPpde}   and \cref{eq:bc} with a neural network $\phi(x; \theta)$ by optimizing its trainable parameters $\theta \in \Theta \subseteq \mathbb{R}^p$. Training proceeds by minimizing an empirical risk defined through the composite loss function
\begin{align}
    \mathcal L (\theta) &:= \frac{\lambda_{\mathrm{pde}}}{|\mathcal X_\mathrm{pde}|} \sum_{x\in\mathcal X_\mathrm{pde}} L_{\mathrm{pde}} (x;\theta) \nonumber\\
    &\quad + \sum_{k=1}^K \frac{\lambda_{\mathrm{bc},k}}{|\mathcal X_\mathrm{bc,k}|} \sum_{x\in\mathcal X_\mathrm{bc,k}} L_{\mathrm{bc},k}(x;\theta) \ , \label{eq:pinn-loss}
\end{align}
with per-sample losses $L_\text{pde} (\bx;\theta):= \norm{\mathcal N[\phi(\bx; \theta)]}_2^2$, $L_\text{bc,k} (\bx;\theta):= \norm{\mathcal B_k[\phi(\bx; \theta)]}_2^2$ and weights $\lambda_\text{pde},\lambda_{bc,k}>0$. We consider training points $\mathcal X_\text{pde}\subset\Omega$ and $\mathcal X_\text{bc,k}\subset \Gamma_k$, which construct $\mathcal X_\text{train} = \mathcal X_\text{pde}\cup \bigcup_k \mathcal X_\text{bc,k}$. Data-driven regression terms for incorporating measurements can be included but are omitted here for the sake of clarity. %

\subsection{Influence Functions (IFs)} \label{ssec:influence_functions}

Influence functions (IFs) \cite{ hampel_influence_1974,cook_residuals_1982, koh_understanding_2017} address the question of how a model’s behavior would change if specific training points were removed from or added to the training set.
Through a linear approximation of leave-one-out retraining, IFs enable the analysis of a model in terms of the contributions of individual training points. 

Compared to the original formulation of influence functions \cite{koh_understanding_2017}, we generalize the framework in several important ways:
First, we extend the analysis from strict minima to non-degenerate stationary points, which is critical in neural network training, where optimization typically converges to saddle points rather than global optima \cite{dauphin_identifying_2014}. This relaxation is especially relevant in PINNs, whose optimization landscapes are often ill-conditioned \cite{RathoreChallengesTrainingPINNs2024,RyckOperatorPreconditioningPerspective2024}.
Second, we refine the required assumptions on the loss landscape, noting that strong convexity, while necessary, is not sufficient. 
Finally, by allowing $f$ to be an arbitrary differentiable function, we generalize IFs from loss-to-loss attributions to influence \emph{from any quantity to any component of a composite objective}, which is crucial for analyzing PINN objectives, where quantities of interest extend beyond the loss to physical observables and individual constraint terms.

\begin{definition}[Influence Function] \label{def:influence_function}
    Let $L : \overline\Omega \times\Theta  \to \mathbb R_{\geq 0}$ be a twice differentiable, single-sample loss function that induces an empirical risk $\mathcal L(\theta)=\frac{1}{N}\sum_{\bx\in\mathcal X} L(\bx;\theta)$ over a training set $\mathcal X\subset\overline\Omega$ of size $N$, and $f:A\times\Theta\to\mathbb R^m$ be an arbitrary differentiable function. For two points $\bx\in\overline\Omega$, $z\in A$, we call
    \begin{align}
        \Inf_{\theta_0}^{L\to f} (\bx,z):= - \nabla_\theta f(z;\theta_0)\!^\top\ \mathcal H^{-1}_{\theta_0}\ \nabla_\theta L(\bx;\theta_0)
        \label{eq:influence-def}
    \end{align}
    the \emph{influence function} from $L$ at the point $\bx$ towards $f$ at the point $z$ with respect to $\theta_0$, where $\mathcal{H}_{\theta_0} = \nabla^2_{\theta} \mathcal L (\theta_0)$ denotes the Hessian at $\theta_0$, which is assumed to be invertible. 
\end{definition}

\begin{theorem}[Influence Functions Approximate Retraining] \label{thm:InfluenceFunction_mainText}
   Assume the setting of \cref{def:influence_function} with $L$ and $f$ being analytic. Further, assume that $\theta_0$ is a non-degenerate stationary point of $\mathcal L$ with respect to a training set $\mathcal X$ (i.e., $\nabla_\theta \mathcal L(\theta_0)=0$, $\det \mathcal{H}_{\theta_0} \neq 0$). Let $\theta_1$ denote the stationary point of $\mathcal L(\theta)$ with respect to $\mathcal X^\prime$, where we added or removed a training point $\bx^\pm$. For sufficiently large $N$, the resulting effect on the function $f$ is given by
    \begin{align*}
        f(z;\theta_1) - f(z;\theta_0) = \pm \Inf^{L\to f}_{\theta_0} (\bx^\pm,z) \frac{1}{N} + \mathcal O(1/N^2) \quad\text{.}
    \end{align*}
\end{theorem}
We provide the proof in \cref{app:theory-IFs}. Thus, the IF computes the effect of infinitesimally upweighting the contribution of a data point in the training objective. Downweighting a training point yields a sign-flipped influence expression.

In practice, IFs are most often evaluated for the influence on the test loss. This corresponds to the special case $f=L$, which reads $\Inf^{L\to L}_{\theta_0} (\bx,z)$. 
It is important to note that IFs constitute only a first-order approximation. In practical application, discrepancies may arise when the number of training points $N$ is not sufficiently large or when $\theta_0$ is only close to being a stationary point. Further, note that the accuracy of the approximation depends on the local geometry of the loss landscape.

\begin{corollary}[Properties of Influence Functions]\label{cor:linearity-additivity}
    Assume the setting of \cref{def:influence_function,thm:InfluenceFunction_mainText}.

    \textbf{(Linearity)}
    The influence function is bilinear in the loss $L$ and the quantity of interest $f$. For scalars $\alpha, \beta \in \mathbb{R}$, and suppressing the dependence on $(x,z)$ for brevity, we have
    \begin{align*}
    \Inf_{\theta_0}^{\alpha L_1+\beta L_2\to f} &= \alpha \Inf_{\theta_0}^{L_1\to f} + \beta \Inf_{\theta_0}^{L_2\to f}, \\
    \Inf_{\theta_0}^{L \to \alpha f_1+\beta f_2} &= \alpha \Inf_{\theta_0}^{L\to f_1}+ \beta \Inf_{\theta_0}^{L\to f_2} \quad .
    \end{align*}
    \textbf{(Additivity)}
    In addition, we can remove or add several points. Let $\mathcal X^+$ be a set of points added to the training set and $\mathcal X^-\subset\mathcal X$ a set of points removed. Then, for sufficiently large $N$, the resulting change in $f$ satisfies
    \begin{equation*}
        \begin{split}
            N \bigl(f(z;\theta_1)-f(z;\theta_0)\bigr)
            &\approx
            \sum_{\bx\in\mathcal X^+} \Inf_{\theta_0}^{L\to f}(\bx,z) \\
            &\quad
            - \sum_{\bx\in\mathcal X^-} \Inf_{\theta_0}^{L\to f}(\bx,z)\quad .
        \end{split}
    \end{equation*}
\end{corollary}

This corollary enables the analysis of influence between arbitrary regions of the domain and individual components of the PINN objective, forming the basis of the \PINNfluence framework developed in \cref{sec:methodology_pinnfluence}. Notably, the linearity in the loss argument implies that the loss weights directly scale corresponding influences, giving our approach a principled way to study loss-weighting effects.

Beyond point-wise effects, it is often practical to consider aggregated measures such as the total influence $\sum_{z \in R} \Inf_{\theta_0}^{L\to f}(\bx,z)$ or its absolute variant $\sum_{z \in R} \Big|\Inf_{\theta_0}^{L\to f}(\bx,z)\Big|$ over regions $R \subseteq \overline \Omega$. Together with \cref{cor:linearity-additivity}, this permits the study of influences from points to regions, regions to points, and region-to-region interactions, while simultaneously allowing for the analysis of individual loss components.
For a discussion on the practical considerations of the computation of influences, such as circumventing the costly explicit construction of the Hessian, we refer to \cref{app:IF-practical-considerations}.
\section{Related Work}\label{sec:related_work}
Interpretability and explainability have emerged as central themes in the machine learning community, driven by the increasing deployment of complex models in high-stakes applications. Naturally, this trend has also extended to scientific machine learning \cite{schnake_higher-order_2022,kitouni_neurons_2024,wetzel_interpretable_2025}, where model transparency and verification are closely tied to scientific validity.
Still, dedicated interpretability methods for PINNs remain scarce. 

\paragraph{Analyses and Improvements of PINN Training}
In the absence of dedicated interpretability frameworks, much of the existing understanding of PINNs has been developed through the systematic analysis of their training behavior.
There is a significant line of work investigating the mechanisms underlying observed pathologies. This includes gradient flow pathologies and ill-conditioned loss landscapes \cite{wang_when_2022,RyckOperatorPreconditioningPerspective2024}, information propagation failure \cite{krishnapriyan_characterizing_2021,daw_mitigating_2023}, failure of initial condition learning \cite{steger_how_2022}, and different phases of learning in PINNs \cite{anagnostopoulos_learning_2024}. Substantial efforts have focused on mitigating these failures, including loss reweighting \cite{wang_experts_2023,wang_respecting_2024}, adaptive sampling \cite{lu_deepxde_2021,liu_grad-rar_2022,wu_comprehensive_2023,LauPINNACLEPINNAdaptive2024}, refined network architectures \cite{jagtap_adaptive_2020,mcclenny_self-adaptive_2023}, and curvature-aware optimizers tailored to PINNs \cite{RathoreChallengesTrainingPINNs2024,vyasSOAPImprovingStabilizing2025a}, including energy natural gradients \cite{mullerAchievingHighAccuracy2023}, second-order approaches with gradient-alignment analysis \cite{wangGradientAlignmentPhysicsinformed2026}, and self-scaled quasi-Newton schemes \cite{kiyaniOptimizingOptimizerPhysicsinformed2025}. Furthermore, the neural tangent kernel (NTK) framework is a standard investigatory setting \cite{jacot_neural_2018,wang_when_2022}, although recent work has shown that this linearized regime breaks down for nonlinear PDEs \cite{bonfanti_challenges_2024}.
These studies provide valuable diagnostic insights into training dynamics and optimization pathologies but explicit post-hoc interpretability and data attribution mechanisms are missing. 
We provide a detailed methodological comparison of \PINNfluence with related training-time approaches in \cref{app:related_comparison}.

\paragraph{Interpretability in PINNs and Learning of the Solutions of PDEs}
Another line of research aims to increase interpretability in scientific machine learning for PDEs more explicitly. Here, some approaches include constraining the architecture to increase interpretability \cite{ramabathiran_spinn_2021} or developing novel architectures to be interpretable by design, such as Kolmogorov-Arnold Networks \cite{liu_kan_2024} and GINN-KANs \cite{ranasinghe_ginn-kan_2024}, but they do not address post-hoc analysis of already trained PINNs.

Although not aimed at PINNs, another emerging area of interpretability research is concerned with the post-hoc analysis of the learned representation of data-driven PDE solvers.
Recently, methods from the field of mechanistic interpretability \cite{bricken2023monosemanticity} such as sparse auto\-encoders \cite{cunningham2023sparseautoencodershighlyinterpretable} and linear probing \cite{AlainUnderstandingIntermediateLayers2018} have been used to investigate neural operators trained to solve PDEs and predict weather dynamics \cite{macmillan_towards_2025,fear_physics_2025}, discovering concept activation vectors and latent features that correspond to weather and flow phenomena such as vorticity and diffusivity. 
Although promising, these methods do not extend to PINNs.

Despite these advances, explicit attribution and diagnostic tools for PINNs remain largely unexplored. In particular, to the best of our knowledge, no training-data-centric framework for PINNs exists that enables attributing model behavior to individual collocation points or physics
constraints, which motivates the approach proposed in this work.

\section{Methodology of \PINNfluence} \label{sec:methodology_pinnfluence}

\begin{table}[t]
    \centering
    \caption{Overview of \PINNfluence-based diagnostics used in this work and the insights they provide.}
    \scriptsize
    \begin{tabular}{@{}p{.28\linewidth}p{.22\linewidth}p{.35\linewidth}@{}}
        \toprule
        \textbf{Diagnostic Question} 
        & \textbf{\PINNfluence tool} 
        & \textbf{Insight} \\
        \midrule[.5\lightrulewidth]

        Which training points shape a given \mbox{prediction}?
        & Point-wise \mbox{influence maps}
        & Localizes sensitivity to \mbox{individual} collocation points. \\
        \midrule[.5\lightrulewidth]

        How do individual physics constraints shape model behavior?
        & Loss-component influence \mbox{decomposition} \cref{eq:loss_fractions_decomposition}
        & Quantifies the relative contribution of PDE, boundary, and IC. Exposes IC-dominated failure patterns. \\
        \midrule[.5\lightrulewidth]

        Do learned solutions reflect temporally oriented influence?
        & Temporal \mbox{influence indicator} \cref{eq:causality}
        & Shows that well-trained PINNs exhibit near-symmetric temporal influence, while asymmetry signals failure. \\
        \midrule[.5\lightrulewidth]

        How does influence distribute across domain subsets?
        & Regional influence indicators
        & Probes problem-specific structure: downstream propagation, spatial symmetry, shock locality, see \cref{app:additional-investigations}. \\
        \midrule[.5\lightrulewidth]
        \bottomrule
    \end{tabular}
    \label{tab:diagnostic_questions}
\end{table}
In this section, we introduce our toolbox of \PINNfluence-based analysis primitives and indicators for diagnosing PINN behavior. Conceptually, \PINNfluence decomposes influence along two axes: \emph{what} contributes and \emph{where} contributions arise.
Building on the theoretical results of \cref{sec:theory-background}, we show how \PINNfluence can be utilized at different levels of granularity.
Starting from point-wise influences, we construct decompositions of loss components and regions, and define normalized indicators that summarize influence effects for systematic comparison across models.

These tools are designed to test concrete diagnostic hypotheses about the behavior of an optimized model, such as \emph{``Which constraints dominate a prediction?''} and \emph{``How does influence align with temporal or directional structure of the underlying PDE?''} In \cref{tab:diagnostic_questions}, we give a summary of the concrete questions together with the insights they yield.

\paragraph{Point-wise Influences}
As fundamental building blocks of our analysis, point-wise influences \cref{eq:influence-def} quantify how an individual training point affects a function $f$, such as the prediction, the loss or a physical observable, evaluated at a specific test location.
Influence values can be positive and negative, indicating whether a training point increases or decreases $f$ at the test location.
Empirically, we observe that the signs of influence effects can be noisy. Since we are primarily interested in the strength of the effect, we mainly consider absolute influence values $| \Inf_{\theta_0}^{L \to f} (\bx, \bz) |$, which provide a more robust estimate of sensitivity.
Further, points with large absolute influences can be interpreted as regions of high complexity for the model \cite{zhang_rethinking_2022}.
Thus, point-wise influences are particularly suited for attributing model behavior at individual points in the domain $\Omega$ to specific training points.
This enables diagnosis beyond localization: one can identify not only where a model fails, but also which training points dominate predictions across different regions, yielding a structural characterization invisible to residual analysis.

\paragraph{Point-to-Region Influences}
Building on the linearity and additivity properties of influence functions (see \cref{cor:linearity-additivity}), we aggregate point-wise influences to study region-level effects.
Throughout the section, we denote $R_{\text{tr}},R_{\text{te}} \subset \Omega$ as subsets of training and test samples, respectively.
Fixing a training point $\bx \in \mathcal{X}$ and summing its influence over test locations yields point-to-region influence maps, revealing where and how strongly this datum affects a quantity of interest across the domain.
\begin{align}
    \Inf_{\theta_0}^{L\to f}(x, R_{\text{te}}) = \sum_{z \in R_{\text{te}}} \Inf_{\theta_0}^{L\to f}(x,z)\quad.
    \label{eq:point_to_region}
\end{align}
Notably, this provides insights into how the quantity of interest $f$ would change at a macroscopic scale if the training point were perturbed or removed.
Analogously, one can fix a test point and aggregate the estimated effects of a set of training points $R_{\text{tr}}$, which reveals how each training point contributes to model behavior at that location.
 
\paragraph{Decomposition by Individual Loss Components}

The linearity of the influence in the loss argument (\cref{cor:linearity-additivity}) allows us to decompose it with respect to every individual loss term. 
This allows for fine-grained analysis of the effect of the PDE residual, or any of the boundary and initial conditions on a given test point's prediction. Notably, residual inspection alone does not offer the same level of granularity, as it does not permit the attribution of predictions or quantities of interest to specific loss components or constraints.
We obtain the \emph{fractional contribution} of each loss term by taking its ratio with respect to the sum of all absolute contributions, which yields
\begin{align}
    r_{L_i}(\bx,\bz) = \frac{\left| \Inf_{\theta_0}^{L_i \to f} (\bx,\bz) \right| }{\sum_{j} \left| \Inf_{\theta_0}^{L_j \to f} (\bx,\bz) \right| } \in [0,1] \quad,
    \label{eq:loss_fractions_decomposition}
\end{align}
satisfying $\sum_i r_{L_i}(x,z)=1$.
Since individual influences may have opposing signs, we additionally report a cancellation score
\begin{align}
    \kappa(\bx,\bz)  &=
    1- \frac{\left| \sum_j \Inf_{\theta_0}^{L_j \to f}(\bx,\bz)  \right|}{\sum_{j} \left| \Inf_{\theta_0}^{L_j \to f}(\bx,\bz)  \right| } \nonumber \\
    &= 1 - \frac{\left|\Inf_{\theta_0}^{L \to f}(\bx,\bz)  \right|}{\sum_{j} \left| \Inf_{\theta_0}^{L_j \to f}(\bx,\bz)  \right| }\in [0,1]
    \label{eq:loss_fractions_cancellation}
\end{align}
that quantifies the degree to which loss terms reinforce ($\kappa \approx 0$) or counteract ($\kappa \approx 1$) each other. 
When cancellation is low, the fractions reliably indicate each term's contribution to the total influence. Conversely, high cancellation corresponds to competing constraints whose individual effects largely negate each other at the prediction level.

\paragraph{Region-based Indicators}

In analogous fashion to the loss components decomposition, we can decompose influence with respect to training and test regions.
We first define relative regional influences, which capture the estimated effect on model behavior of certain regions relative to the whole domain.
The \emph{relative absolute regional influence} of $R_{\text{tr}}$ on $R_{\text{te}}$ is given by
\begin{align}
\rho_{\theta_0}^{L \to f}
\bigl(R_{\text{tr}}, R_{\text{te}}\bigr) = \frac{1}{|R_{\text{te}}|} \sum_{\bz \in R_{\text{te}}} 
    \frac{\displaystyle\sum_{\bx \in R_{\text{tr}}} \left| \Inf_{\theta_0}^{L \to f} (\bx, \bz) \right|}{\displaystyle\sum_{\bx \in \mathcal{X}_{\mathrm{train}}} \left| \Inf_{\theta_0}^{L \to f}(\bx, \bz) \right|}\quad,
    \label{eq:regional_influence}
\end{align}
with $\rho_{\theta_0}^{L \to f} \in [0,1]$. It allows quantifying how strongly a given training region contributes to model behavior over a specific test region.
For region-level analysis, point-wise loss decompositions \cref{eq:loss_fractions_decomposition} can also be aggregated over subsets of training and test points 
$
\bar r_{L_i}(R_{\text{tr}}, R_{\text{te}})
=\frac{1}{|R_{\text{te}}|}
\sum_{z \in R_{\text{te}}}\sum_{\bx \in R_{\text{tr}}}
r_{L_i}(x,z).$
An important variation of the aforementioned regional indicator arises in time-dependent problems. For such settings, we define a \emph{temporal influence indicator} that quantifies how strongly the model behavior at test points is influenced by the training points in the past
\begin{align}
    &\eta_{\theta_0}^{L \to f}(R_{\text{tr}}, R_{\text{te}}) = \nonumber \\
    &1-\frac{1}{|R_{\text{te}}|} \sum_{\bz \in R_{\text{te}}} 
    \frac{\displaystyle\sum_{\bx \in R_{\text{tr}}:\bx_t \leq \bz_t} \left| \Inf_{\theta_0}^{L \to f}(\bx, \bz) \right|}{\displaystyle\sum_{\bx \in \mathcal{X}_{\mathrm{train}}} \left| \Inf_{\theta_0}^{L \to f}(\bx, \bz) \right|}.
    \label{eq:causality}
\end{align}
Again, we have $\eta_{\theta_0}^{L \to f} \in [0,1]$. The subtraction from $1$ ensures that smaller values of $\eta$ correspond to stronger past-directed influence; larger values imply the opposite. Assuming that the distribution of training points is uniform, values of $\approx 0.5$ are expected if influence is temporally symmetric. Notably, this notion cannot be probed through residual analysis alone. 
These indicators can be specialized to various other diagnostic hypotheses, e.g., boundary dominance, spatial locality, or symmetry, through appropriate choices of the loss component $L_i$, the function $f$, and the sets of training and test regions.
In analogy to \cref{eq:causality}, directional indicators can be constructed for time-independent problems by defining ordering relations over spatial directions. For this and further instances, we refer to \cref{app:additional-investigations}.
\newcommand{\problemsfiglength}{0.15\textwidth}
\begin{figure}[t]
\centering
\begin{subfigure}[t]{\problemsfiglength}
    \includegraphics[width=\textwidth]{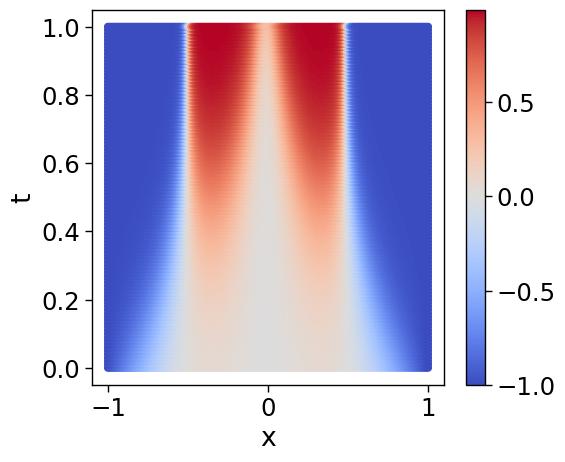}
    \caption{Allen-Cahn}
\end{subfigure}
\hfill 
\begin{subfigure}[t]{\problemsfiglength}
    \includegraphics[width=\textwidth]{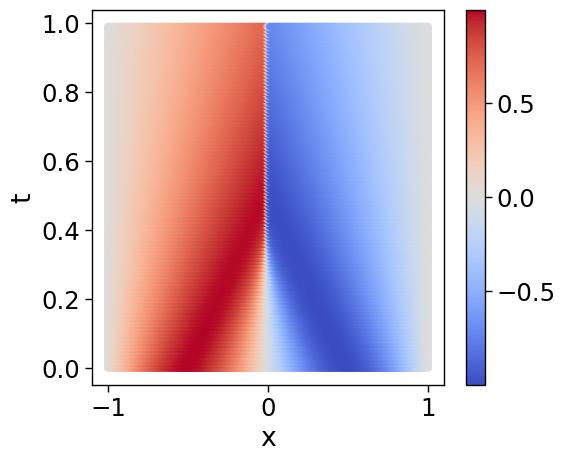}
    \caption{Burgers'}
\end{subfigure}
\hfill
\begin{subfigure}[t]{\problemsfiglength}
    \includegraphics[width=\textwidth]{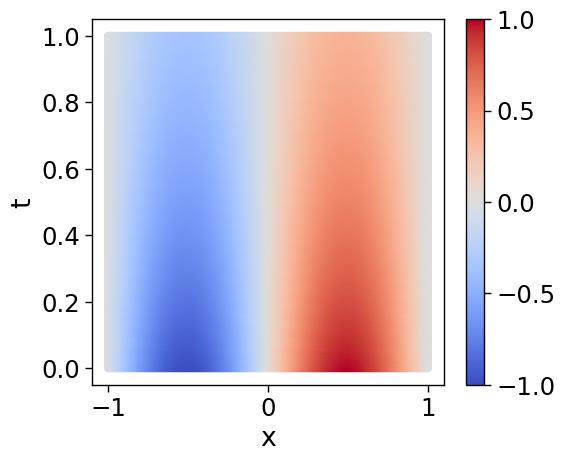}
    \caption{Heat}
\end{subfigure}
\hfill
\begin{subfigure}[t]{\problemsfiglength}
    \includegraphics[width=\textwidth]{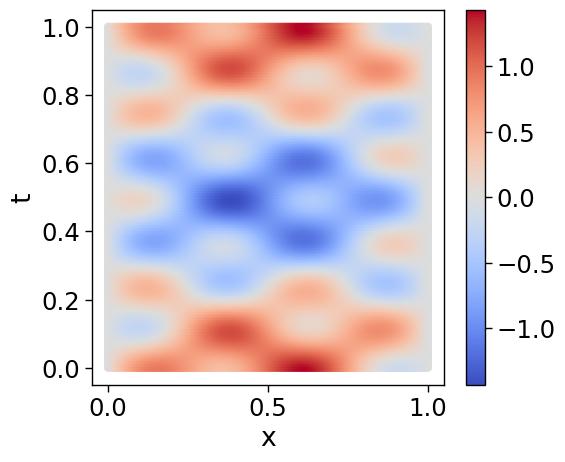}
    \caption{Wave}
\end{subfigure}
\begin{subfigure}[t]{\problemsfiglength}
    \includegraphics[width=\textwidth]{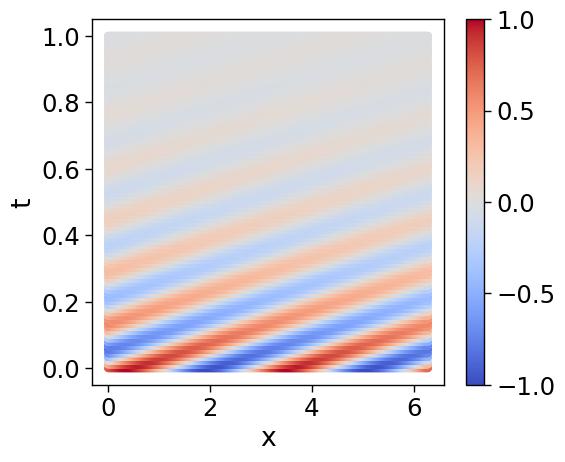}
    \caption{Drift-Diffusion}
\end{subfigure}
\hfill

\caption{Investigated problems.}
\label{fig:investigated_pdes}
\end{figure}

Indicators summarize influence pattern analysis and provide first-order, Hessian-based sensitivity diagnostics under the trained parameters of a PINN.
Notably, \PINNfluence does not provide formal error guarantees or interventional causal claims. It characterizes the converged model's sensitivity to how predictions would respond to perturbations in the training data, rather than the training dynamics that produced the structure. In this sense, \PINNfluence supports the comparative and hypothesis-driven diagnosis of trained PINNs, as demonstrated in the following section.

\section{Experiments}
We demonstrate how \PINNfluence helps to understand the behavior of PINNs across five time-dependent PDEs spanning diverse physics: the heat, Allen-Cahn, Burgers', wave, and drift-diffusion equations, shown in \cref{fig:investigated_pdes}.
Additionally, we provide an analysis of two steady-state problems, namely the Poisson and Navier-Stokes equations in \cref{app:directional_indicator}.
For every problem, we include two different settings: a \emph{well-trained} model that accurately solves the IBVP, and a \emph{poorly-trained} model that fails to accurately solve the IBVP due to insufficient data or suboptimal optimization.
Comparing influence patterns between these two configurations allows us to identify differences in how the two classes of models depend on their training data at convergence.
This enables us to provide diagnostic indicators that distinguish well-performing from pathological model states in a controlled way.
To assess consistency, we include $10$ random seeds for each configuration.
The complete problem specifications and training setup are provided in \cref{app:problems}.

Across experiments, we focus on three recurring research questions that exemplify how \PINNfluence yields insights into PINN behavior:
\begin{enumerate}[label=(\roman*)]
    \item Can \textsc{PINNfluence} give insights into differences between well-trained and poorly-trained models, beyond localizing where errors occur?
    \item Can \textsc{PINNfluence} attribute model sensitivity to specific loss components $L_i$?
    \item Can we reveal structural properties of trained PINNs using \textsc{PINNfluence}, such as temporal influence patterns that are not evident from residuals alone?
\end{enumerate}

\paragraph{Point-wise Attribution}

\begin{figure}[tb]
    \centering
    \begin{subfigure}[t]{0.45\linewidth}
          \includegraphics[width=\textwidth]{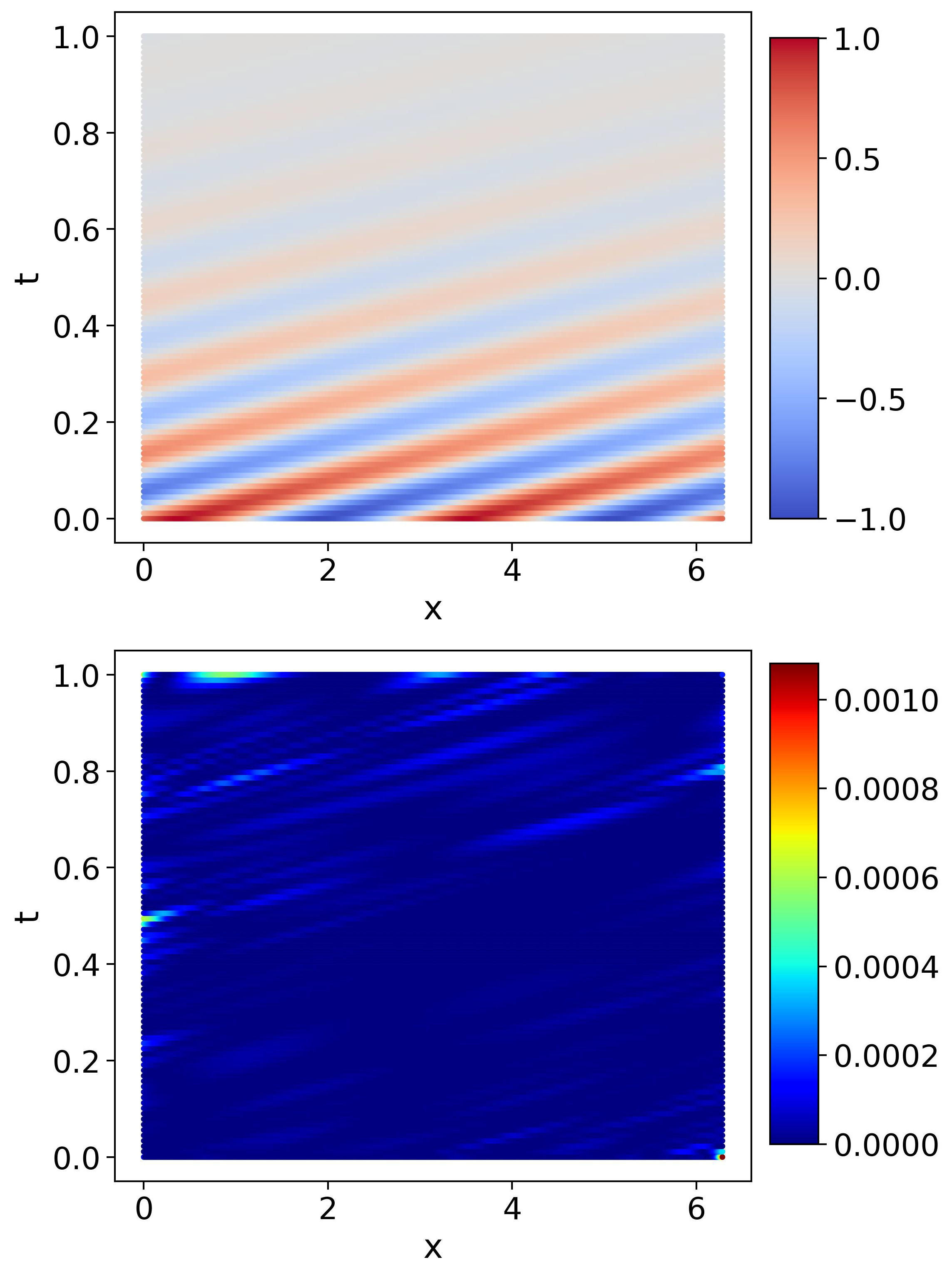}
          \caption{Well-trained}
    \end{subfigure}
    \begin{subfigure}[t]{0.45\linewidth}
        \includegraphics[width=0.97\textwidth]{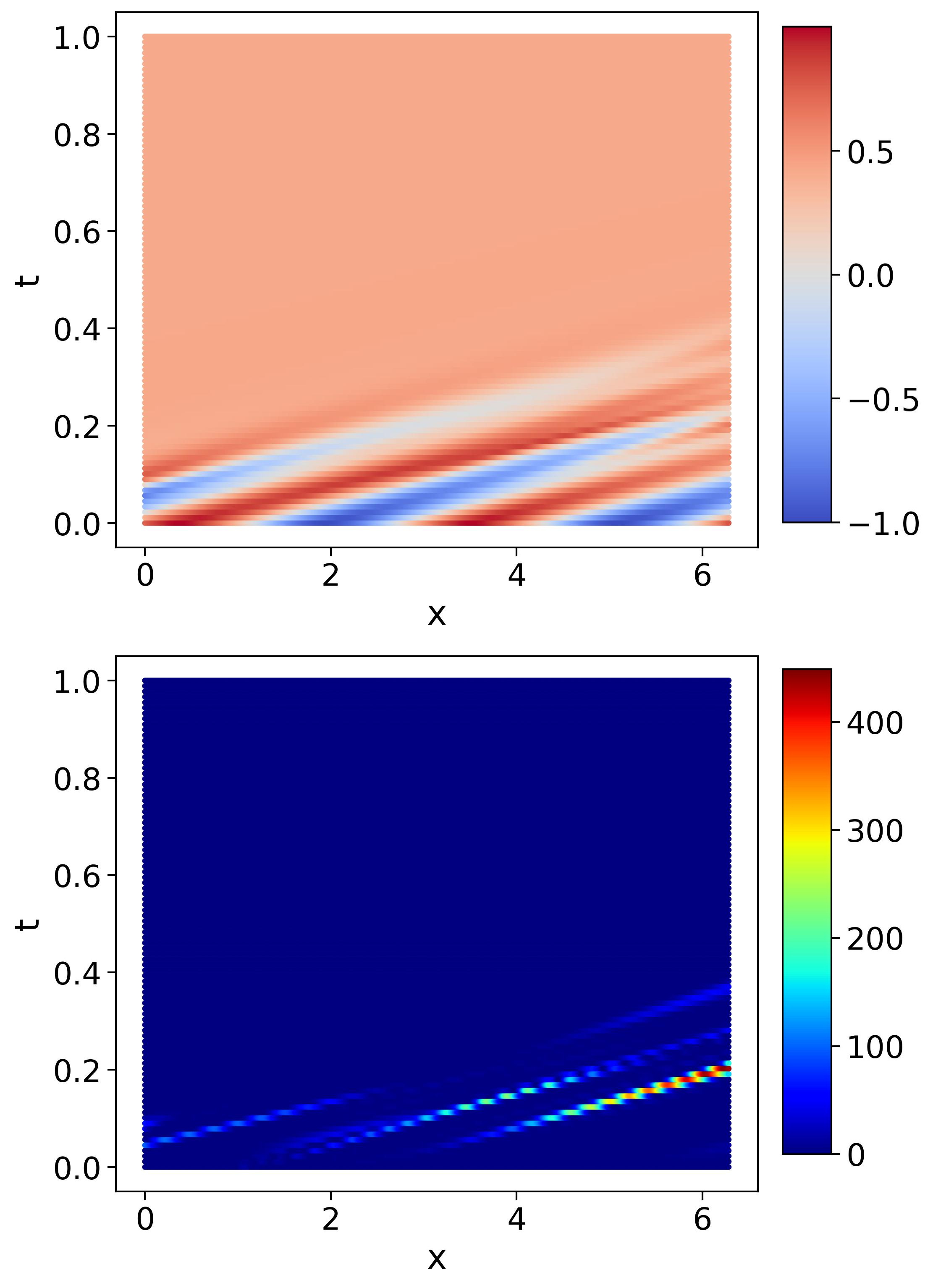}
          \caption{Poorly-trained}
    \end{subfigure}
    \caption{Prediction and losses for the drift-diffusion problem for a representative seed.}
    \label{fig:drift_diffusion_preds_losses}
\end{figure}

To demonstrate the capabilities of \PINNfluence, we focus our analysis on the drift-diffusion equation, a convection problem where PINNs are known to suffer from \emph{propagation failure}, i.e., collapsing to near-constant solutions forward in time \cite{krishnapriyan_characterizing_2021}.
This exact failure mode is present in our poorly-trained configuration and is straightforward to localize.
Predictions show the collapse to a constant output emerging at around $t=0.2$ and 
the losses indicate a large PDE violation at the transition region, as shown in \cref{fig:drift_diffusion_preds_losses}. 

\begin{figure}[t]
    \centering
    \begin{subfigure}[t]{0.45\linewidth}
        \includegraphics[width=\linewidth]{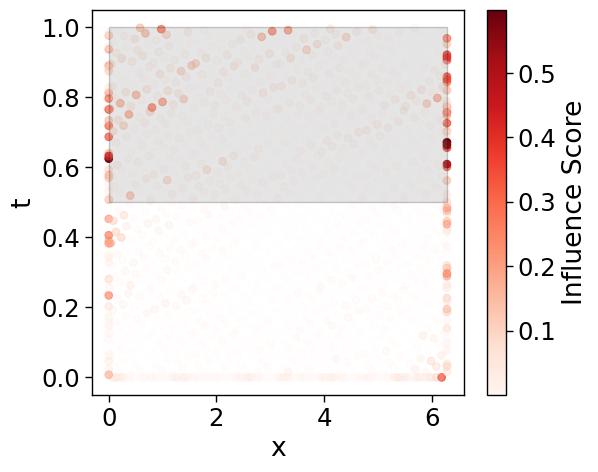}
        \caption{Well-trained}
    \end{subfigure}
    \hfill
    \begin{subfigure}[t]{0.45\linewidth}
        \includegraphics[width=\linewidth]{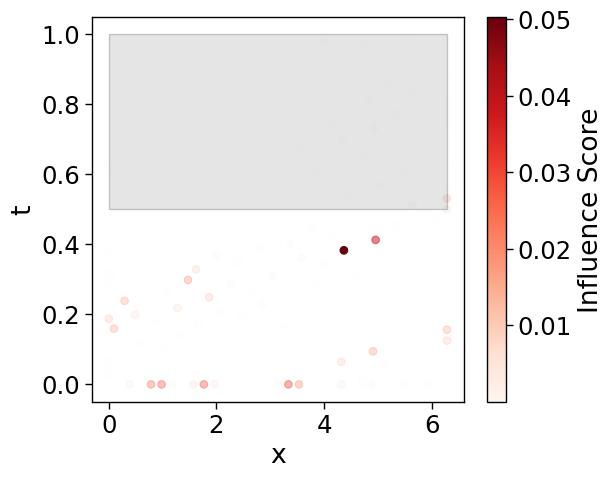}
        \caption{Poorly-trained}
    \end{subfigure}
    \caption{Point-to-region influence heatmaps showcasing the distribution of influences across the training set affecting the network \emph{prediction} at the top half ($t>0.5$, marked in gray) of the domain.}
    \label{fig:heatmap_test_point_drift_diffusion_output}
\end{figure}

\textsc{PINNfluence} aids us in understanding how the prediction was shaped, through the lens of the PINN's training data. 
\cref{fig:heatmap_test_point_drift_diffusion_output} shows point-to-region influence attribution maps for model predictions at times $t>0.5$, identifying training points to which these predictions are most sensitive.
For the well-trained model, the prediction in that region is primarily shaped by information from the data points on the periodic boundaries ($x=0$ and $x=2\pi$).
For the poorly-trained model, the influence patterns differ significantly: 
alongside the region of failure, the prediction is highly sensitive to training points along the initial condition ($t=0$), while influence from the boundary appears absent.
Thus, \PINNfluence reveals a key distinction invisible to residuals: which training data shapes predictions differs drastically between configurations.

\paragraph{Decomposition by Individual Loss Components}
Point-wise attribution reveals the training points to which the models are most sensitive. 
Using the linearity of influence functions in the loss argument, \PINNfluence also allows us to directly decompose and quantify influence contributions into individual loss terms \cref{eq:loss_fractions_decomposition}.
This enables studying the contributions of loss terms on a more global scale.

\newcommand{\ratioslength}{0.23\textwidth}
\begin{figure}[b]
\setlength{\fboxsep}{0pt}      %
\setlength{\fboxrule}{0.2pt}   %
\centering
\hspace*{0.82em}
\begin{subfigure}[t]{0.443\textwidth}
    \fbox{\includegraphics[width=\dimexpr\linewidth-2\fboxsep-2\fboxrule\relax]{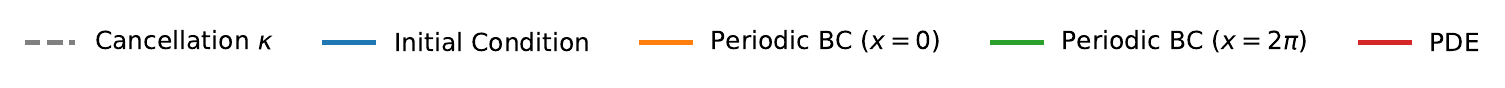}}
\end{subfigure}
\\[-0.2ex]
\begin{subfigure}[t]{\ratioslength}
    \includegraphics[width=\textwidth]{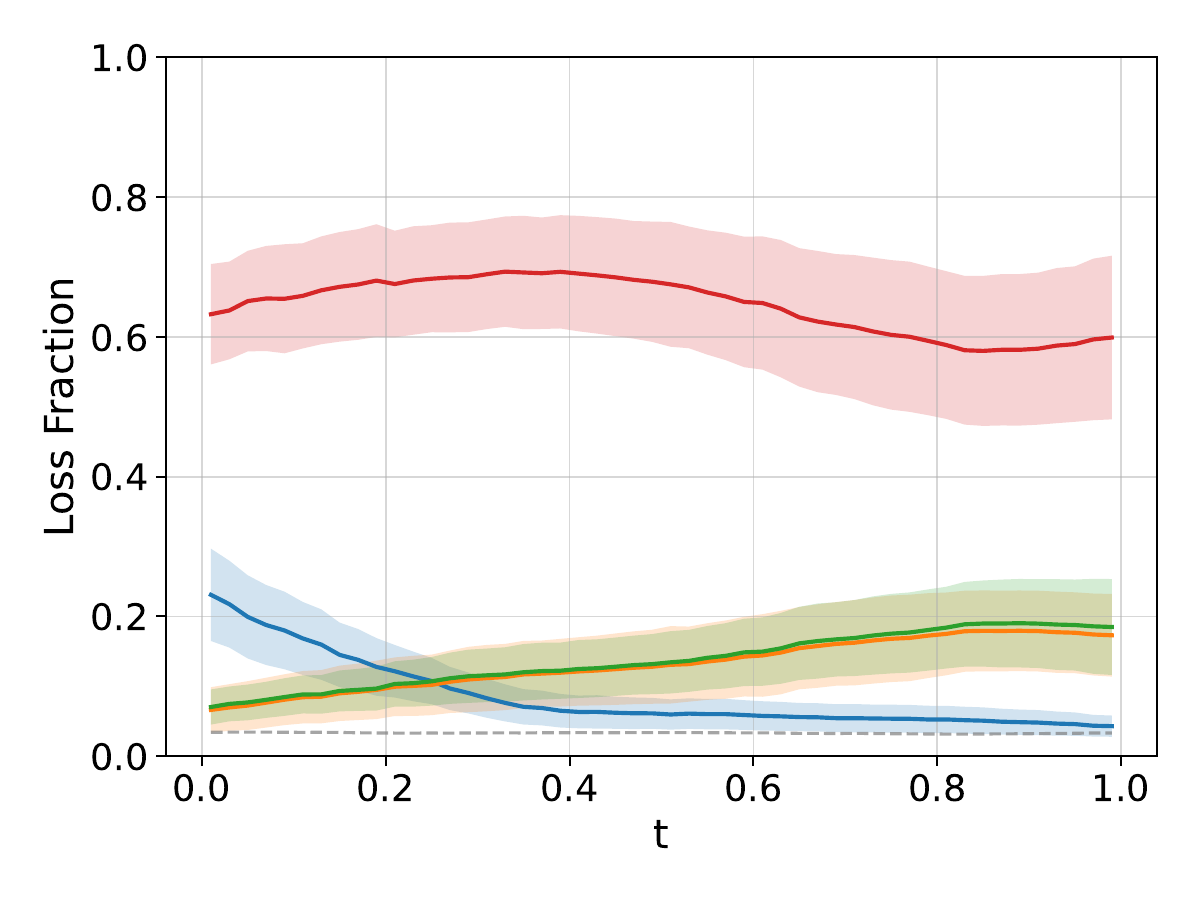}
    \caption{Well-trained}
\end{subfigure}
\hfill 
\begin{subfigure}[t]{\ratioslength}
    \includegraphics[width=\textwidth]{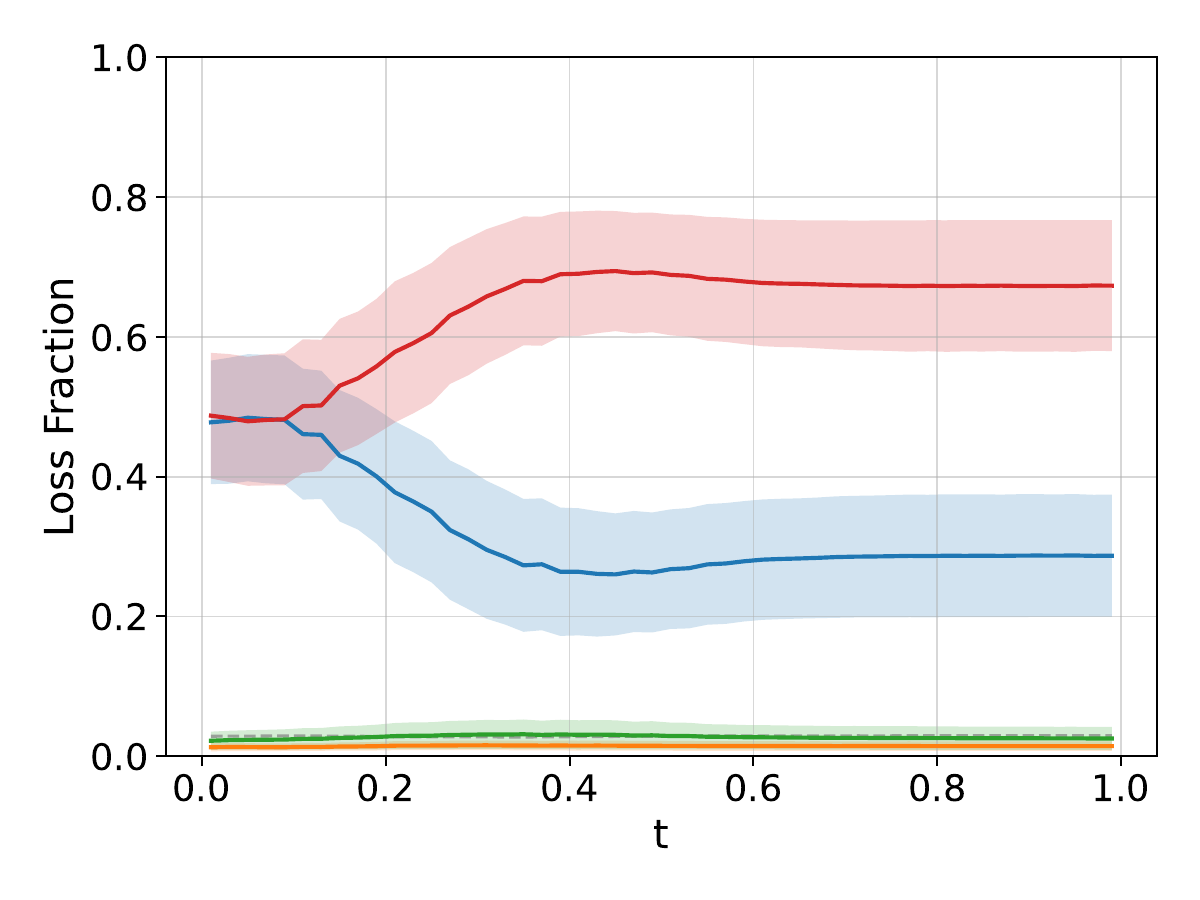}
    \caption{Poorly-trained}
\end{subfigure}
\caption{Fractions of influences $\bar r_{L_i}(\mathcal{X},R_{\mathrm{te}})$ for 50 time bins, averaged over seeds. Shaded regions indicate the standard deviation across seeds.}
\label{fig:drift_diffusion_loss_fractions}
\end{figure}

We aggregate the fractional influence contribution $\bar r_{L_i}$ of each loss component (the PDE residual, the initial and the two periodic boundary conditions), averaged over time bins in the test data, as shown in \cref{fig:drift_diffusion_loss_fractions}.

For the well-trained models, the IC contribution starts at around $\bar r_{L_\mathrm{ic}} \approx 0.25$ near $t=0$ and decreases asymptotically towards zero with increasing time. 
Simultaneously, the influence contributions of both periodic boundary conditions increase evenly over time, indicating that the model's predictions reflect the problem's periodic structure. The PDE term dominates throughout.

For the poorly-trained model, the IC displays a much greater share of influence. 
While it also decreases with time, the term's relative influence $\bar r_{L_\mathrm{ic}}$ remains elevated in comparison to the well-trained model. 
After $t\approx0.4$, the fractions plateau, coinciding with pathological, near-constant outputs and stagnating gradients.
Notably, the boundary conditions exhibit negligible contributions, i.e., the model's predictions are effectively insensitive to its boundary training data. The model has converged to a state where perturbations to boundary points have no effect, a clear structural misalignment for a problem driven by boundary propagation.

\begin{figure}[b]
\setlength{\fboxsep}{0pt}      %
\setlength{\fboxrule}{0.2pt}   %
\centering
\begin{subfigure}[t]{0.37\textwidth}
\fbox{\includegraphics[width=\dimexpr\linewidth-2\fboxsep-2\fboxrule\relax]{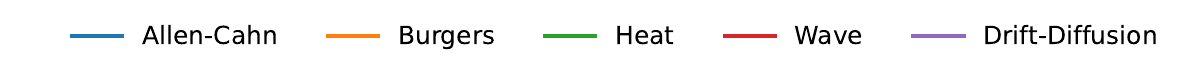}}
\end{subfigure}
\\[-0.2ex]
\begin{subfigure}[t]{\ratioslength}
    \includegraphics[width=\textwidth]{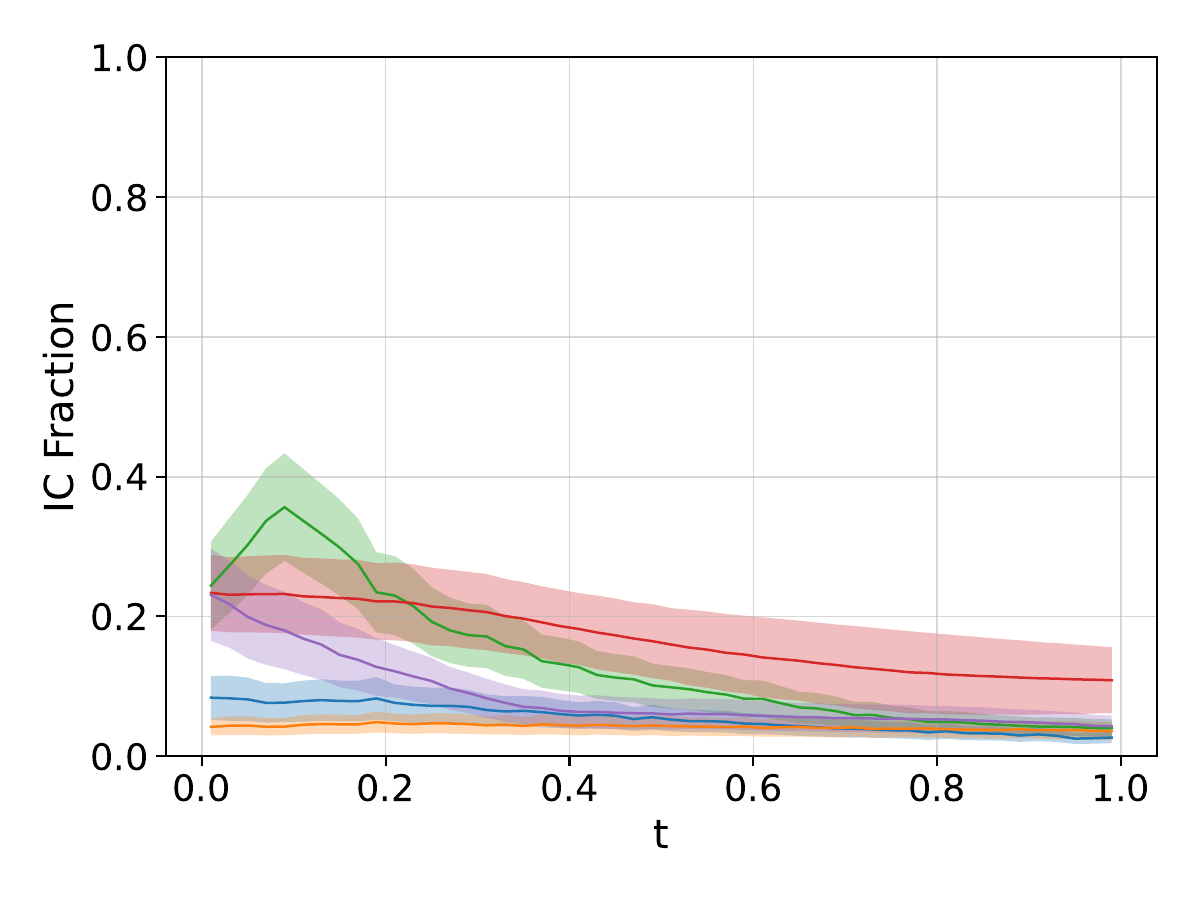}
    \caption{Well-trained}
\end{subfigure}
\hfill 
\begin{subfigure}[t]{\ratioslength}
        \includegraphics[width=\textwidth]{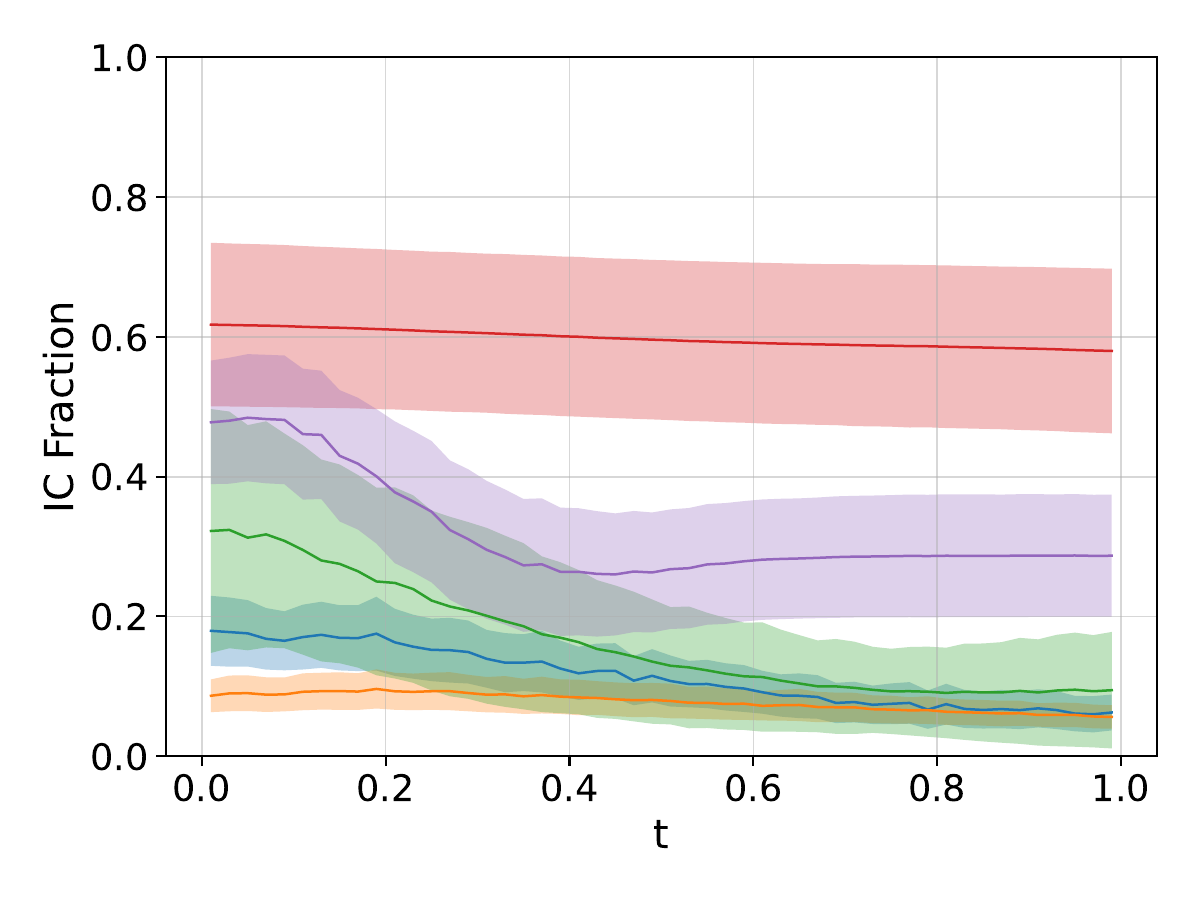}
        \caption{Poorly-trained}
\end{subfigure}
\caption{Fractions of IC influences $\bar r_{L_{\text{ic}}}(\mathcal{X},R_{\mathrm{te}})$ for 50 time bins, across all problems, averaged over seeds. Shaded regions indicate the standard deviation across seeds.}
    \label{fig:loss_fractions_time_dependent_probs}
\end{figure}

We further show that the elevated relative IC importance $\bar r_{L_\mathrm{ic}}$ can be identified across all investigated problems, as depicted in \cref{fig:loss_fractions_time_dependent_probs}.
Throughout all well-trained models, the IC influence share at $t=0$ is lower and decays more rapidly compared to their poorly-trained counterparts.
For the latter, the IC generally displays greater influence over time.
Note that while the rate of decay varies by problem, the general diagnostic insight holds: well-trained models show diminishing IC sensitivity with time, poorly-trained models do not.
This is striking, as not all suboptimally trained models perform equally \emph{poorly}. 
We refer the interested reader to \cref{app:model_preds_and_targets}, where all respective model predictions and targets are visualized.
\PINNfluence can thus attribute model sensitivity to individual loss components $L_i$, thereby reliably distinguishing well-trained from poorly-trained models.
We further present a complementary analysis in \cref{app:model-quality-sweep}, varying the number of training points on the time-dependent PDEs. There, we show that \PINNfluence indicators respond to the gradual transition between failure and success in PDE-specific ways, revealing each problem's data requirements.

\paragraph{Indicator-based Analysis}
The loss decomposition reveals that poorly-trained models are overly sensitive to changes in the initial condition. 
We now examine this pattern from a complementary perspective: the temporal distribution of influences across the domain. 
Time-dependent PDEs possess an inherent causal structure in which initial conditions determine future states. 
One might therefore hypothesize that a well-trained PINN that yields an accurate approximation of the solution would reflect this causality.
Under this proposition, predictions at time $t$ would be predominantly influenced by training points at times $\leq t$.
We probe this hypothesis using the temporal influence indicator \cref{eq:causality} and compare it against the baseline temporal average $\bar t =\frac{1}{|\mathcal{X}|} \sum_{x \in \mathcal{X}} x_t$ of the training set. Note that sampling on the initial boundary introduces a distribution skewed toward earlier times.

\begin{table}[t]
\caption{Mean temporal influence indicator for each problem, computed with respect to predictions $f=\hat u$. We report mean $\pm$ standard deviation over 10 seeds.}
\label{tab:mean_directionality}
\centering
\begin{threeparttable}
\begin{tabular}{lccccc}
\toprule
Problem & $\bar t$ & Well-trained & Poorly-trained \\
\midrule 
Heat & 0.46 & 0.33 $\pm$ 0.02 & 0.26 $\pm$ 0.06 \tnote{(*)}  \\
Allen-Cahn & 0.43 & 0.50 $\pm$ 0.02 & 0.32 $\pm$ 0.05 \\
Burgers' & 0.43 & 0.41 $\pm$ 0.02 & 0.28 $\pm$ 0.02 \\
Drift-Diffusion & 0.46 & 0.46 $\pm$ 0.04 & 0.21 $\pm$ 0.06 \\
Wave  & 0.43 & 0.41 $\pm$ 0.03 & 0.11 $\pm$ 0.02 \\
\bottomrule
\end{tabular}
\begin{tablenotes}
\item[(*)] \small The baseline is $\bar t=0.41$ for the poorly-trained model due to different sampling.
\end{tablenotes}

\end{threeparttable}
\end{table}
\cref{tab:mean_directionality} reports temporal influence indicators across all time-dependent problems, computed with respect to the prediction.
Contrary to the aforementioned hypothesis, well-trained models do not exhibit a temporal influence aligned with the causal structure of the underlying PDEs. Rather, they show an influence distribution that is close to the data distribution baseline $\bar t$, only deviating slightly from it. 
This shows that well-trained models are, on average, equally sensitive to changes in training data at both earlier and later times.
This is sensible given that PINNs are trained to solve a given problem globally, instead of sequentially.
Hence, this illustrates that a well-trained PINN has learned to leverage information from across the spatio-temporal domain.
Poorly-trained models, on the other hand, display skewed temporal influences. 
This finding aligns directly with the elevated IC influence ratio observed earlier:
these models show disproportionate sensitivity to early-time training points as exemplified in \cref{fig:loss_fractions_time_dependent_probs}.
Hence, strong causal alignment in terms of influences is a signal of training failure rather than of success. 

The previous results constitute only a selection of potential questions that one may study using \PINNfluence. 
In \cref{app:additional-investigations}, we present complementary use-cases, such as additional point-wise attribution heatmaps for other problems, the decomposition of loss terms for each setting. We also introduce further indicators, including a demonstration of region-to-region influence analysis. Notably, we also study steady-state problems with appropriately constructed influence-based indicators. Beyond these extensions to new problems, we also verify in \cref{app:soap_nncg} on two optimizers, NNCG \cite{RathoreChallengesTrainingPINNs2024} and SOAP \cite{vyasSOAPImprovingStabilizing2025a}, that \PINNfluence is applicable independently of optimizer choice.

To summarize, \PINNfluence offers a complementary perspective on PINNs by attributing predictions directly to the training data and losses.
It reliably distinguishes between well-trained and poorly-trained PINNs across diverse problems, enhancing the understanding of PINNs.

\paragraph{Limitations}
\PINNfluence inherits limitations from influence functions. 
PINN loss landscapes are often severely ill-conditioned, leading to near-singular Hessians, a setting under which raw leave-one-out retraining is known to be an unreliable validation target in practice \cite{basu_influence_2020}. In \cref{app:IF-practical-considerations}, we examine the impact of this fragility for our benchmark PDEs. The low-rank Arnoldi approximation recovers projected inverse Hessians faithfully and influence estimates align with local fine-tuning behavior that IFs actually approximate (PBRF, \citealp{baeIfInfluenceFunctions2022}), outperforming a gradient-only baseline \cite{charpiat_input_2019}. While these findings support reliability in our setting, the approach still remains a first-order approximation and may exhibit discrepancies under larger perturbations or more ill-conditioned settings.
While the explicit computation of the inverse Hessian can be avoided through inverse-Hessian vector products, computational cost is non-negligible.
The pairwise evaluation of influences scales linearly in both training and test set size, though it is typically faster than PINN training itself. In \cref{app:computational-costs}, we provide an overview of the practical scaling of our approach.

Our methodological choices introduce further caveats:
Aggregating absolute influences discards sign information, which is partially addressed by the cancellation score $\kappa$.
Furthermore, indicator design is problem-dependent and requires domain knowledge for appropriate construction; we provide a practitioner's guide in \cref{app:practitioners-guide}.
Interpretation also requires care: influences quantify sensitivity to perturbations in training data, not causal importance. 
High attribution does not imply necessity or sufficiency for correct behavior. In particular, the temporal indicator should be read as a diagnostic signal rather than a statement about physical causality.
\paragraph{Future Work}
Beyond diagnosis, influence scores naturally rank training points by importance, suggesting avenues for actively improving PINNs. For example, targeted resampling of collocation points from influential regions and informed adjustment of loss weights that rescales components according to their attributed influence both present themselves as sensible extensions of the \PINNfluence framework. We outline both future directions schematically in \cref{app:future-work}.

\section{Conclusion}
Existing studies of PINNs focus primarily on training dynamics and failure modes, but do not provide attribution-based explanations of how training data and physics constraints shape predictions. To address this, we propose \PINNfluence, a novel, training-data-centric interpretability method that adapts influence functions to PINNs and their composite losses. 
By enabling attribution between training data, loss components, and model predictions, \PINNfluence supports fine-grained diagnostics of PINN behavior that go beyond residual-based analysis.
Our experiments show that loss-component and temporal influence patterns can distinguish well-trained from poorly-trained PINNs, localize pathological patterns in the trained model's data dependence, and reveal structural properties of learned solutions.
Overall, while with limitations, our method presents a principled diagnostic lens for studying PINNs with the aim of imbuing scientific machine learning with interpretability methods.
In future work, exploring the use of these methods for intervening in training and informed model design emerges as a promising next step.

\section*{Acknowledgments}
This work was supported by the Fraunhofer Internal Programs (Fraunhofer) as grant ESPINN (PREPARE 40-08394).
\section*{Code Availability}

The implementation of \textsc{PINNfluence} is available at \url{https://github.com/aleks-krasowski/pinnfluence}.

\section*{Impact Statement}

This work introduces a training-data-centric interpretability framework for physics-informed neural networks. 
By enabling practitioners to attribute model behavior to specific training points and loss components, \PINNfluence provides diagnostic capabilities that complement existing methods. 
We hope this opens up new directions for understanding PINN behavior and, in future work, 
for developing principled interventions based on this method. 

As an interpretability method for scientific machine learning, we do not anticipate negative societal impacts from this work.

\bibliography{references_clean}
\bibliographystyle{icml2026}

\newpage
\appendix
\crefalias{section}{appendix}
\crefalias{subsection}{appendix}
\crefalias{subsubsection}{appendix}
\onecolumn

\section*{Appendix}
The Appendix is structured as follows: 
\cref{app:IFs_supplementary_info} focuses on influence functions and provides an extended theoretical background in \cref{app:theory-IFs} and detailed practical considerations in \cref{app:IF-practical-considerations}, including validation of the Hessian approximation quality, empirical confirmation of influence estimates, computational cost analysis, and a discussion of alternative training data attribution methods.
\Cref{app:related_comparison} contrasts \PINNfluence with training-time diagnostics, \cref{app:practitioners-guide} gives a short recipe for constructing diagnostic indicators, and \cref{app:future-work} outlines resampling and loss-weighting extensions.

\cref{app:problems} contains descriptions of the investigated problems, including two steady-state PDEs (\cref{app:pdes}).
\cref{app:training} specifies the training environment and hyperparameters for model configurations across problems and showcases predictions in \cref{app:model_preds_and_targets}.

Finally, \cref{app:additional-investigations} features supplemental experiments utilizing \PINNfluence. 
Concretely, we adapt the temporal indicator for steady-state problems in \cref{app:directional_indicator}, and showcase additional region-to-region indicators in \cref{app:additional_indicators_region}. We further sweep the training-data quantity continuously in \cref{app:model-quality-sweep} to confirm that the regional indicators respond to gradual variation in training quality, and verify that the diagnostics depend on the converged model rather than the optimizer used to reach it by evaluating curvature-aware optimizers in \cref{app:soap_nncg}. Finally, we provide loss-term decompositions for the remaining problems in \cref{app:decomposition_plots} and additional point-wise heatmaps in \cref{app:pointwise_plots}.

\section{Supplementary Information for Influence Functions}
\label{app:IFs_supplementary_info}
\subsection{Theoretical Background}
\label{app:theory-IFs}

In this part, we formulate the key steps for deriving influence functions, initially introduced in \cite{hampel_influence_1974} to study robust estimators, and which were recently applied to neural networks \cite{koh_understanding_2017,koh_accuracy_2019,hammoudeh_training_2024}. However, as already observed in \cite{schioppa_theoretical_2023}, the functions one typically encounters in neural networks require a more exhaustive analysis. Therefore, we carefully redraw the results proposed in \cite{koh_understanding_2017} and generalize them slightly for the application to PINNs. In particular, we formulate influence functions in a manner that enables the study of the effect of various modifications of the problem (i.e., changes in the training set, in the loss weighting, or in the PDE) on the prediction and any derived quantity. Furthermore, we provide rigorous proofs for the statements made in \cref{ssec:influence_functions}. 

We start with a crucial lemma for influence functions, which was already noted in \cite{cook_residuals_1982}. As remarked in \cite{schioppa_theoretical_2023}, influence functions are usually stated for strictly convex functions\footnote{Note that for the application of influence functions with global minima \citep[e.g.][]{koh_understanding_2017}, strict convexity is a necessary, but not a sufficient criterion.} in the vicinity of global minima. However, in practice, we typically only reach saddle points \cite{dauphin_identifying_2014}, which is the reason why the following lemma is extended to stationary points.

\begin{lemma}[$\operatorname{arg min}$ Trick] \label{lem:argmin_trick}
    Consider an open set of parameters $\Theta \subseteq \mathbb{R}^p$ and an open set $U \subseteq \mathbb{R}$. Let $g: \Theta \times U \to \mathbb{R}$ be a twice continuously differentiable function and $(\theta_0, \epsilon_0)\in\Theta\times U$, such that $\theta_0$ is a non-degenerate stationary point of $g(\, \cdot \, , \epsilon_0)$, i.e., $\nabla_\theta g(\theta_0, \epsilon_0) = 0$ and the Hessian $\mathcal{H}_\theta(\theta_0, \epsilon_0) \coloneq \nabla_{\theta}^2 g(\theta_0, \epsilon_0) \in \mathbb{R}^{p \times p}$ is invertible. 
    
    Then, there exists a non-empty, open neighborhood $U_0 \subseteq U$ around $\epsilon_0$, such that $g(\,\cdot\, , \epsilon)$ has a unique stationary point in the neighborhood of $\theta_0$ for all $\epsilon \in U_0$. Denote by $h: U_0 \to \Theta$ the function which describes such stationary points. Then $h$ is continuously differentiable and
    \begin{align}
        \frac{\partial h(\epsilon)}{\partial \epsilon} = \left. - \mathcal{H}_{\theta_0}(h(\epsilon), \epsilon)^{-1} \cdot \nabla_{\theta}\frac{\partial g(\theta, \epsilon)}{\partial \epsilon} \right|_{\theta=h(\epsilon)}\quad. \label{eq:argmin_trick}
    \end{align}
    Furthermore, if $g$ is analytic in a neighborhood of $(\theta_0,\epsilon_0)$, then we can choose $U_0$ such that $h$ is analytic as well. Moreover, if $B\subseteq U$ is an open, connected set, such that for every $\epsilon\in B$, there exists a non-degenerate stationary point of $g(\,\cdot\, ,\epsilon)$, then $h$ is extendable to $B$ and $h: B \to \Theta$ is also continuously differentiable with \cref{eq:argmin_trick}.
\end{lemma}
\begin{proof}
    Let us denote the gradient of $g$ as $s(\theta, \epsilon) \coloneq \nabla_\theta g(\theta, \epsilon)$. By our initial assumption, $g$ is twice continuously differentiable, implying that $s$ is continuously differentiable. Since $(\theta_0, \epsilon_0)$ is a stationary point, $s(\theta_0, \epsilon_0) = 0$, the implicit function theorem \citep[see][Theorem 2]{oliveira_implicit_2012} guarantees the existence of an open set $U_0 \subseteq U$ around $\epsilon_0$ and a unique, continuously differentiable function $h: U_0 \rightarrow \Theta$ such that $h(\epsilon_0) = \theta_0$ and $s(h(\epsilon), \epsilon) = 0$ for all $\epsilon \in U_0$. This establishes that for any $\epsilon \in U_0$, a unique stationary point exists in the neighborhood of $\theta_0$. And $h$ is continuously differentiable with \cref{eq:argmin_trick}.

    Now if $g$ is analytic, then so is its gradient $s$. Therefore also the function $h$ is analytic, following the holomorphic implicit function theorem \citep[see][Section 7.6]{fritzsche_holomorphic_2002}.

    If for all $\epsilon\in B$ the above assumptions are fulfilled, then we can apply \cref{lem:argmin_trick} for all of these points $\epsilon$. The function $h:B\to\Theta$ is continuously differentiable, since it is continuously differentiable in every point and \cref{eq:argmin_trick} extends to all points in $B$. Moreover, $h(B)$ is connected, since $h$ is continuous and $B$ is connected by assumption.
\end{proof}

By means of this theorem, we can now prove \cref{thm:InfluenceFunction_mainText} from \cref{ssec:influence_functions}. For better readability, we restate that theorem.

\begin{theorem}[Influence Functions Approximate Retraining, {analogous to \cref{thm:InfluenceFunction_mainText}}] \label{thm:InfluenceFunction_app}
    Assume the setting of \cref{def:influence_function} with $L$ and $f$ being analytic, and that $\theta_0$ is a non-degenerate stationary point of $\mathcal L$ with respect to a training set $\mathcal X$. Let $\theta_1$ denote the stationary point of $\mathcal L(\theta)$ (i.e., $\nabla_\theta \mathcal L(\theta_0)=0$, $\det \nabla_\theta^2\mathcal L(\theta_0) \neq 0$) with respect to $\mathcal X'$, where we added or removed a training point $\bx^\pm$ (with $\bx^-\in\mathcal X$). For sufficiently large $N$, the resulting effect on the function $f$ is given by
    \begin{align}
        f(z,\theta_1) - f(z,\theta_0) = \pm \Inf^{L\to f}_{\theta_0} (\bx^\pm,z) \frac{1}{N} + \mathcal O(1/N^2) \quad\text{.}
    \end{align}
\end{theorem}
\begin{proof}
    Let $\tilde {\mathcal L} (\theta,\epsilon):= \mathcal L(\theta) + \epsilon\ \kappa(\theta)$ be a small perturbation of the original loss with some differentiable function $\kappa:\Theta\to\mathbb R$ and $\epsilon\in\mathbb R$. Since $\theta_0$ is a non-degenerate stationary point of $\mathcal L(\theta)$, $(\theta_0,0)$ is a non-degenerate stationary point of $\tilde {\mathcal L}(\theta,\epsilon)$ in the sense of \cref{lem:argmin_trick}. Hence, there is a neighborhood $U_0$ around the origin $\epsilon_0=0$, such that the stationary points $\theta_\epsilon$ of $\tilde{\mathcal L}(\theta,\epsilon)$ are uniquely defined for all $\epsilon\in U_0$ and the derivative of $\theta_\epsilon$ is given by \cref{eq:argmin_trick}.

    Since $\theta_\epsilon$ is also analytic, due to the assumptions and \cref{lem:argmin_trick}, we can approximate it with a Taylor expansion as
    \begin{align}
        \theta_\epsilon - \theta_0 = \frac{\partial\theta_\epsilon}{\partial\epsilon}\Bigg|_{\epsilon=0} \epsilon + \mathcal O(\epsilon^2) = - \mathcal H^{-1}_{\theta_0} \cdot \nabla_\theta \kappa(\theta)\big|_{\theta=\theta_0} \cdot \epsilon + \mathcal O(\epsilon^2) \label{eq:theta-epsilon}
    \end{align}
    with $\mathcal H_{\theta_0} = \nabla_\theta^2 \mathcal L(\theta)\big|_{\theta=\theta_0}$. When setting $\kappa(\theta)=L(x^+;\theta)$, we have $\tilde{\mathcal L}(\theta,1/N) = \frac{1}{N}\sum_{x\in\mathcal X\cup \{x^+\}} L(x;\theta)$, which has the same stationary points as the empirical risk over $\mathcal X\cup \{x^+\}$, which is $\frac{1}{N+1}\sum_{x\in\mathcal X\cup \{x^+\}} L(x;\theta)$. In other words, we have $\theta_{1/N} = \theta_1$ providing that $1/N\in U_0$, which is always satisfied for large $N$. We obtain by chain rule
    \begin{align}
        f(z,\theta_1) - f(z,\theta_0) = \nabla_\theta f(z,\theta_0) \cdot \frac{\partial\theta_\epsilon}{\partial\epsilon}\Bigg|_{\epsilon=0} \cdot \frac{1}{N} + \mathcal O(1/N^{2}) \label{eq:proof-statement}
    \end{align}
    which concludes the proof for $x^+$ by application of \cref{lem:argmin_trick}. Similarly, we obtain the result for $x^-$ by considering $\kappa(\theta)=-L(x^-;\theta)$ with an additional sign.

    The \cref{cor:linearity-additivity} follows immediately, by considering the linearity in equations \cref{eq:theta-epsilon} and \cref{eq:proof-statement}.
\end{proof}

From the details of the proof we also immediately see the effect of adding several or removing several points. Hence, we find that a change from a training set $\mathcal X$ to a set $\mathcal X^\prime=\mathcal X \cup\mathcal X^+ \setminus \mathcal X^-$, where we assume $\mathcal X^-\subset\mathcal X$ will affect a function $f:A \times \Theta$ according to
\begin{align}
    f(z;\theta_1)-f(z;\theta_0) =  \frac{1}{N} \Bigg(\sum_{\bx\in\mathcal X^+} \Inf_{\theta_0}^{L\to f}(\bx,z) - \sum_{\bx\in\mathcal X^-} \Inf_{\theta_0}^{L\to f}(\bx,z)\Bigg) + \mathcal O(1/N^2) \label{eq:app-groupinf}
\end{align}
for large $N>N_\text{min}$. There is always a value $N_\text{min}$, where \cref{eq:app-groupinf} is valid, but the precise number depends on the set $U_0$, and for large changes of the training set $\mathcal X \to \mathcal X^\prime$ this number might increase.

One can easily extend \cref{thm:InfluenceFunction_app} to study also the influence of other parts in the optimization task. E.g.\ one could study the influence of certain parameters in the PDE, such as the diffusivity constant $\alpha$ in the heat equation. By computing the Taylor expansion with respect to that parameter $\alpha$, one approximates the effect of increasing or lowering that parameter. The proof is completely analogous to the proof of \cref{thm:InfluenceFunction_app} and is left to the reader.

\subsection{Practical Considerations On The Computation Of Influences} \label{app:IF-practical-considerations}

Influence functions are typically derived under the assumption of a well-conditioned Hessian at a strict local minimum \cite{koh_understanding_2017}, which in practice is very unlikely to hold for neural networks in general.
For PINNs, in particular, the loss landscape is known to be non-convex and ill-conditioned \cite{RathoreChallengesTrainingPINNs2024}, and converged models reach non-degenerate stationary points rather than strict minima. 
Whether the resulting influence estimates remain numerically meaningful is therefore a legitimate concern.

We address this in two ways. First, we check whether the low-rank Arnoldi approximations that we utilize faithfully invert the Hessian on the gradients that enter the influence computation. 
Second, we evaluate whether the resulting influence scores predict actual data-removal effects, validated using the Proximal Bregman Response Function \cite{baeIfInfluenceFunctions2022}, which captures the local fine-tuning behavior that IFs approximate.
We further quantify computational costs and situate \textsc{PINNfluence} among alternative training data attribution methods.

\subsubsection{Approximating the inverse Hessian}

Explicit computation of the inverse Hessian $\mathcal{H}_{\theta_0}^{-1}$ is costly, but can be circumvented by utilizing inverse-Hessian vector products (IHVPs).
Several efficient approximations of IHVPs have been proposed in the literature, including the linear time stochastic second-order algorithm (LiSSA) \cite{agarwal_second-order-2017, koh_understanding_2017}, Arnoldi iterations \cite{schioppa_scaling_2021} and Eigenvalue-corrected Kronecker-Factored Approximate Curvature (EK-FAC) \cite{grosse_studying_2023}. 
We utilize a modified version of the \texttt{ArnoldiInfluenceFunction} from the Captum library \cite{kokhlikyan_captum_2020} with the following hyperparameters:
$200$ Arnoldi iterations, top-$50$ retained eigenvectors, and damping $\mathcal{H}_{\theta_0} + \lambda I$ with $\lambda = 10^{-3}$. 

Concretely, the approximation recovers top eigenvectors and eigenvalues of the damped Hessian using only Hessian-vector products, from which a low-rank surrogate of $\mathcal{H}^{-1}_{\theta_0}$ is assembled as follows.
Let $V \in \mathbb{R}^{p \times k}$ contain the top-$k$ eigenvectors of  the damped Hessian with the corresponding eigenvalues on the diagonal of $S \in \mathbb{R}^{k \times k}$, and let $R = V S^{-\frac{1}{2}}$, so that $RR^\top = VS^{-1}V^\top \approx \mathcal{H}^{-1}_{\theta_0}$. 
With $R$ at hand, the influence calculation in \cref{eq:influence-def} becomes 
\begin{align}
    \label{eq:arnoldi_if}
    \Inf^{L\to f}_{\theta_0}(x, z) &= -\bigl(\nabla_\theta f(z; \theta_0)^\top R\bigr)\bigl(R^\top \nabla_\theta L(x; \theta_0)\bigr) \quad.
\end{align}

\paragraph{Hessian approximation quality}
To validate the IHVP approximation, we compute the full dense Hessian for each problem configuration.
The exact spectra confirm severe ill-conditioning with condition numbers from $\kappa \approx 10^{20}$ to $10^{33}$ across all problems (\cref{tab:hessian_condition_numbers}). 
Damping $\mathcal{H}_{\theta_0} + \lambda I$ with $\lambda = 10^{-3}$ reduces $\kappa$ to between $10^{8}$ and $10^{15}$.
Across all configurations, the smallest retained eigenvalue exceeds $\lambda$ by at least three orders of magnitude, 
so in practice, damping has negligible effect on the top-$k$ subspace and primarily acts as a safeguard against the ill-conditioned tail.
We note that small or even negative eigenvalues are admissible in this tail since $\theta_0$ is a non-degenerate stationary point rather than a strict minimum.

What remains is to confirm that the discarded tail is in fact irrelevant for influence computation.
We probe how well $R R^\top$ acts as an inverse to $\mathcal{H}_{\theta_0}$ on the gradients that enter the influence estimation by evaluating the cosine similarity between $\mathcal{H}_{\theta_0}RR^\top g$ and $g$, where $g$ is either a per-sample loss gradient $\nabla_\theta L(x)$ or a prediction gradient $\nabla_\theta f(x)$ at a training point $x \in \mathcal{X}_{\text{train}}$. 
\cref{tab:cosine_similarity_hessian_recon} reports the mean cosine similarities over $10$ seeds.
Recovery is near-perfect for Allen-Cahn, Burgers', Heat, and the well-trained Drift-Diffusion, and remains strong for the remaining configurations. 
The gradients therefore concentrate in the top-$k$ subspace and the truncation to top-$k$ eigendirections acts as a regularization of a numerically ill-conditioned operator rather than purely as a computational shortcut (\cref{tab:hessian_condition_numbers}).
In fact, it has been observed that low-rank Arnoldi approximations can outperform exact inversion of the Hessian in practice \cite{schioppa_scaling_2021}.

\begin{table}[ht]
\centering
\caption{Cosine similarities between reconstructed $\mathcal{H}_{\theta_0}RR^\top g$ and true gradients $g$ averaged over the whole training set $\mathcal{X}_{\text{train}}$ and over 10 runs (mean $\pm$ std).}
\label{tab:cosine_similarity_hessian_recon}
\begin{threeparttable}
    \begin{tabular}{lcccc}
\toprule
 & \multicolumn{2}{c}{$\nabla_\theta L$} & \multicolumn{2}{c}{$\nabla_\theta f$} \\
 & Well-trained & Poorly-trained & Well-trained & Poorly-trained \\
Problem &  &  &  &  \\
\midrule
Heat & $1.00 \pm 0.00$\tnote{(*)} & $1.00 \pm 0.00$\tnote{(*)} & $1.00 \pm 0.00$\tnote{(*)} & $1.00 \pm 0.00$\tnote{(*)} \\
Allen-Cahn & $1.00 \pm 0.00$\tnote{(*)} & $1.00 \pm 0.00$\tnote{(*)} & $0.99 \pm 0.03$ & $0.99 \pm 0.02$ \\
Burgers' & $0.99 \pm 0.01$ & $0.99 \pm 0.01$ & $1.00 \pm 0.01$ & $0.99 \pm 0.00$ \\
Drift-Diffusion & $1.00 \pm 0.01$ & $0.90 \pm 0.10$ & $0.99 \pm 0.01$ & $0.85 \pm 0.08$ \\
Wave & $0.99 \pm 0.02$ & $0.99 \pm 0.02$ & $0.88 \pm 0.05$ & $0.92 \pm 0.04$ \\
Poisson & $0.96 \pm 0.04$ & $0.98 \pm 0.02$ & $0.93 \pm 0.01$ & $0.93 \pm 0.02$ \\
Navier-Stokes\tnote{(**)} & $0.87 \pm 0.14$ & $0.80 \pm 0.16$ & $0.87 \pm 0.06$ & $0.94 \pm 0.02$ \\
\bottomrule
\end{tabular}
\begin{tablenotes}
\item[(*)] \small For Heat and for Allen-Cahn, the stds are below $7 \cdot 10^{-4}$ and $10^{-3}$, respectively.
\item[(**)] \small For Navier-Stokes $f$ is $u$, i.e., the predicted velocity in $x$ direction.
\end{tablenotes}
\end{threeparttable}
\end{table}

\begin{table}[]
    \caption{Mean condition numbers of the Hessian $\mathcal{H}_{\theta_0}$ at three stages of the inverse approximation: the exact (dense) Hessian, the damped Hessian $\mathcal{H}_{\theta_0}+\lambda I$ with $\lambda=10^{-3}$, and the top-$k$ subspace retained by the Arnoldi iteration.}
    \label{tab:hessian_condition_numbers}
    \centering
    \begin{tabular}{lcccccc}
    \toprule 
     & \multicolumn{2}{c}{Exact $\mathcal{H}_{\theta_0}$} & \multicolumn{2}{c}{Damped $\mathcal{H}_{\theta_0}+\lambda I$} & \multicolumn{2}{c}{Arnoldi top-$k$} \\
Problem & Poorly-tr. & Well-tr. & Poorly-tr. & Well-tr. & Poorly-tr. & Well-tr. \\
\midrule
Heat & $3.0\times 10^{21}$ & $8.5\times 10^{20}$ & $9.2\times 10^{8}$ & $2.4\times 10^{11}$ & $4.5\times 10^{6}$ & $8.4\times 10^{5}$ \\
Allen-Cahn & $9.7\times 10^{25}$ & $9.9\times 10^{24}$ & $1.8\times 10^{12}$ & $9.2\times 10^{12}$ & $1.4\times 10^{4}$ & $2.1\times 10^{4}$ \\
Burgers' & $3.5\times 10^{21}$ & $4.7\times 10^{21}$ & $1.3\times 10^{12}$ & $8.5\times 10^{11}$ & $1.0\times 10^{4}$ & $5.1\times 10^{3}$ \\
Drift-Diffusion & $2.6\times 10^{23}$ & $8.0\times 10^{23}$ & $3.4\times 10^{12}$ & $2.1\times 10^{12}$ & $3.1\times 10^{3}$ & $1.2\times 10^{3}$ \\
Wave & $1.3\times 10^{32}$ & $1.3\times 10^{33}$ & $2.0\times 10^{14}$ & $8.6\times 10^{15}$ & $1.8\times 10^{3}$ & $8.2\times 10^{2}$ \\
Navier-Stokes & $5.0\times 10^{24}$ & $6.4\times 10^{27}$ & $3.7\times 10^{11}$ & $1.2\times 10^{13}$ & $5.2\times 10^{2}$ & $4.0\times 10^{2}$ \\
Poisson & $2.2\times 10^{22}$ & $1.3\times 10^{22}$ & $4.9\times 10^{12}$ & $3.5\times 10^{13}$ & $1.9\times 10^{4}$ & $3.1\times 10^{2}$ \\
    \bottomrule 
    \end{tabular}
\end{table}

\subsubsection{Validation of Influence Functions}

The first-order approximation underlying influence functions assumes convexity and is found to be fragile as a true leave-one-out estimator \cite{basu_influence_2020}, particularly for very deep neural networks. 
While our model configurations are fairly shallow, especially in comparison to modern computer vision and LLM architectures, PINNs exhibit non-convex and ill-conditioned loss landscapes \cite{RathoreChallengesTrainingPINNs2024},
so the question of whether influence estimates correspond to actual data-removal effects warrants empirical validation.
Despite the theoretical limitations, influence functions have been found to reliably recover influential samples in practice \cite{koh_accuracy_2019,BasuSecondOrderGroupInfluence2020}.
Recent work suggests that they are better understood as approximators of local fine-tuning behavior rather than exact leave-one-out retraining \cite{baeIfInfluenceFunctions2022,wei_final-model-only_2024}.
We confirm this distinction directly: correlation with exact leave-one-out retraining is near zero (\cref{tab:loo}), while correlation with the PBRF, \textit{the object influence functions are designed to approximate}, is substantially higher.

\paragraph{Influence estimates do not approximate leave-one-out retraining}

We first verify that influence functions do not recover exact leave-one-out effects in our setting. 
For each problem, we retrain from scratch with 30 individual training points removed and correlate the observed prediction change with the IF estimate. 
\cref{tab:loo} confirms near-zero Spearman correlations across all PDEs and training regimes, consistent with findings in IF literature \cite{basu_influence_2020}.

\begin{table}[h]
    \caption{Leave-one-out validation. Spearman rank correlation between IF-predicted and LOO-observed prediction differences for
30 randomly removed training points, retrained from scratch, averaged over 10 seeds (mean $\pm$ std).}
    \label{tab:loo}
    \centering
\begin{tabular}{lllll}
\toprule
 & \multicolumn{2}{c}{Influence Function} & \multicolumn{2}{c}{Grad-Dot} \\
 Problem & Poorly-trained & Well-trained & Poorly-trained & Well-trained  \\
\midrule
Heat & $-0.02 \pm 0.13$ & $0.00 \pm 0.03$ & $0.01 \pm 0.06$ & $0.00 \pm 0.02$ \\
Allen-Cahn & $-0.01 \pm 0.04$ & $0.00 \pm 0.02$ & $-0.05 \pm 0.05$ & $0.00 \pm 0.01$ \\
Burgers' & $0.00 \pm 0.02$ & $-0.00 \pm 0.01$ & $0.00 \pm 0.02$ & $-0.00 \pm 0.02$ \\
Drift-Diffusion & $0.01 \pm 0.03$ & $-0.00 \pm 0.02$ & $0.01 \pm 0.02$ & $-0.00 \pm 0.01$ \\
Wave & $0.02 \pm 0.07$ & $0.00 \pm 0.03$ & $0.01 \pm 0.02$ & $-0.00 \pm 0.03$ \\
Navier-Stokes & $0.02 \pm 0.14$ & $-0.05 \pm 0.09$ & $0.03 \pm 0.07$ & $-0.01 \pm 0.05$ \\
Poisson & $-0.01 \pm 0.04$ & $-0.01 \pm 0.04$ & $-0.00 \pm 0.02$ & $0.00 \pm 0.02$ \\
\bottomrule
\end{tabular}
\end{table}

\paragraph{Proximal Bregman Response Function}

Recent work shows that influence functions are better understood as approximators of the Proximal Bregman Response Function (PBRF) \cite{baeIfInfluenceFunctions2022}, which captures the regularized local effect of data removal via fine-tuning rather than retraining from scratch.

For each problem, we sample $30$ training points randomly. 
For each sampled point individually, we fine-tune the converged model for $1\,000$ L-BFGS steps with that point removed and compute the Spearman rank correlation between IF-predicted and PBRF-observed prediction differences across test points (averaged over $10$ seeds). 
As a baseline, we additionally report Grad-Dot \cite{charpiat_input_2019}, a gradient-based attribution method which corresponds to setting $\mathcal{H}^{-1}_{\theta_0} = I$, to confirm the utility of the inverse Hessian. 
\cref{tab:pbrf} reports the resulting Spearman rank correlations for both well-trained and poorly-trained configurations.

Influence functions achieve Spearman rank correlations of $0.08$ to $0.77$ across the evaluated PDEs and both training regimes, substantially exceeding the gradient-only baseline.
The inverse-Hessian factor therefore carries meaningful information about data-removal effects beyond what is available from raw gradients.

\begin{table}[ht]
\centering
\caption{PBRF validation. Spearman rank correlation between IF-predicted and PBRF-observed prediction differences for $30$ randomly removed training points, averaged over $10$ seeds.
}
\label{tab:pbrf}
\begin{threeparttable}
\begin{tabular}{lcccc}
\toprule
 & \multicolumn{2}{c}{Influence Function} & \multicolumn{2}{c}{Grad-Dot} \\
 Problem & Poorly-trained & Well-trained & Poorly-trained & Well-trained \\
\midrule
Heat & $0.45 \pm 0.26$ & $0.28 \pm 0.06$ & $0.09 \pm 0.06$ & $0.12 \pm 0.05$ \\
Allen-Cahn\tnote{(*)} & $0.69 \pm 0.12$ & $0.45 \pm 0.06$ & $-0.01 \pm 0.06$ & $0.00 \pm 0.03$ \\
Burgers' & $0.17 \pm 0.21$ & $0.77 \pm 0.04$ & $0.09 \pm 0.07$ & $-0.08 \pm 0.09$ \\
Drift-Diffusion & $0.08 \pm 0.23$ & $0.54 \pm 0.20$ & $0.07 \pm 0.13$ & $0.15 \pm 0.06$ \\
Wave\tnote{(*)} & $0.15 \pm 0.25$ & $0.47 \pm 0.11$ & $-0.00 \pm 0.04$ & $0.02 \pm 0.05$ \\
Navier-Stokes\tnote{(\dag)} & $0.36 \pm 0.14$ & $0.37 \pm 0.15$ & $-0.04 \pm 0.15$ & $-0.01 \pm 0.06$ \\
Poisson & $0.54 \pm 0.16$ & $0.47 \pm 0.12$ & $0.02 \pm 0.09$ & $0.12 \pm 0.05$ \\
\bottomrule
\end{tabular}
\begin{tablenotes}
\item[(*)] \small For the poorly-trained Allen-Cahn and Wave configurations, training 
uses Adam (cf. \cref{tab:app_hyperparams}); PBRF is computed with $1\,000$ L-BFGS iterations, 
matching the protocol used for the other PDEs.
\item[(\dag)] \small For Navier-Stokes $f$ is $u$, i.e., the predicted velocity in $x$ direction.
\end{tablenotes}
\end{threeparttable}
\end{table}

\subsubsection{Computational Costs}\label{app:computational-costs}

We characterize the cost of \textsc{PINNfluence} both asymptotically and empirically.
Let $n \in \mathbb{N}$ denote the number of training points, $m \in \mathbb{N}$ the number of test points, $p \in \mathbb{N}$ the number of network parameters, $k \in \mathbb{N}$ the number of retained top eigenvectors, and $d \in \mathbb{N}$ the number of Arnoldi iterations.

\paragraph{One-time Costs (Approximation of the IHVP)}
The Arnoldi approximation produces a low-rank surrogate $R \in \mathbb{R}^{p \times k}$ with $R R^\top \approx \mathcal{H}_{\theta_0}^{-1}$, where $\theta_0 \in \mathbb{R}^p$ are the parameters of the converged neural network.
We compute Hessian-vector products exactly by accumulating per-batch contributions across all mini-batches.
This stage requires $\mathcal{O}(n \cdot p \cdot d)$ time and $\mathcal{O}(d \cdot p)$ peak space.
The subsequent distillation to the top-$k$ eigenpairs reduces the stored representation to $\mathcal{O}(k \cdot p)$, with $k < d \ll p$ in practice.

\paragraph{Per-pair Costs (Influence Scoring)}
For a training-test pair $(x,z)$ using Arnoldi iterations, the influence score decomposes as in \cref{eq:arnoldi_if},
yielding a per-pair time complexity of $\mathcal{O}(p \cdot k)$.
Note that both projected gradients $\nabla_\theta f(z;\theta_0)^\top R$ and $R^\top \nabla_\theta L(x;\theta_0)$ can be cached, such that the full $n \times m$ influence matrix requires $\mathcal{O}\bigl( (n+m) \cdot p \cdot k \bigr)$ time for projection and $\mathcal{O}(n \cdot  m  \cdot  k)$ for the pairwise dot products, 
with an additional $\mathcal{O}\bigl((n + m)  \cdot  k\bigr)$ space for the cached projections and $\mathcal{O}(n \cdot m)$ for storing all results.

\paragraph{Empirical Costs}
In \cref{tab:comp_costs_model_size,tab:comp_costs_train_size,tab:comp_costs_test_size}, we report the wall-clock time and peak memory footprints across model size, training set size and test set size, all measured on the Drift-Diffusion PDE as a representative configuration. 
Training points $n$ are split into $1\,000$ domain points, $100$ initial and $100$ boundary points. 
The training-size sweep scales these ratios proportionally.
All measurements use $6$ cores of Intel Xeon Gold 6448H CPUs and are reported as mean $\pm$ standard deviation over $10$ seeds.
We note that GPU acceleration and implementation-level improvements may reduce wall-clock time significantly.

In practice, an $\sim 18\times$ increase in model size (from $2\,241$ to $40\,801$ parameters) only roughly doubles wall-clock time, and the empirical scaling in $n$ and $m$ matches the asymptotic behavior outlined above.
For reference, training a PINN for the same problem with $3\times 64$ hidden neurons and $1\,200$ training points for $15\,000$ Adam and $5\,000$ L-BFGS iterations takes $200.7$s on average on the same hardware (across 10 runs). 

\begin{table}[]
    \centering
    \caption{Computational costs by model size for $n=1\,200$ training and $m=1\,000$ test points. We report mean $\pm$ standard deviation across 10 runs.}
    \label{tab:comp_costs_model_size}
    \begin{tabular}{cccccc}
    \toprule 
         Layers & Parameters $p$ & Hessian Approx. (s) & Influence Scoring (s) & Total (s) & Memory (MB) \\ 
         \midrule 
  $3 \times 16$ & $609$ & $12.6 \pm 1.6$ & $4.1 \pm 0.44$ & $16.7 \pm 2.0$ & $260 \pm 48$ \\
  $3 \times 32$ & $2\,241$ & $14.2 \pm 1.5$ & $5.5 \pm 0.56$ & $19.7 \pm 2.1$ & $534 \pm 56$ \\
  $3 \times 64$ & $8\,577$ & $17.4 \pm 2.5$ & $7.6 \pm 1.8$ & $25.0 \pm 4.1$ & $775 \pm 164$ \\
  $3 \times 100$ & $20\,601$ & $19.9 \pm 2.5$ & $11.1 \pm 1.9$ & $31.0 \pm 4.4$ & $1\,236 \pm 259$ \\
  $5 \times 100$ & $40\,801$ & $28.0 \pm 3.8$ & $17.2 \pm 2.8$ & $45.2 \pm 6.5$ & $1\,959 \pm 66$ \\
         \bottomrule
    \end{tabular}
\end{table}

\begin{table}[]
    \centering
    \caption{Computational costs by training set size for $m=1\,000$ test points and an MLP with $3\times 64$ hidden layers. We report mean $\pm$ standard deviation across 10 runs.}
    \label{tab:comp_costs_train_size}
    \begin{tabular}{ccccc}
    \toprule 
         Number of Train points $n$ & Hessian Approx. (s) & Influence Scoring (s) & Total (s) & Memory (MB) \\ 
         \midrule 
  $100$ & $6.2 \pm 0.31$ & $3.6 \pm 0.29$ & $9.8 \pm 0.54$ & $846 \pm 244$ \\
  $1\,000$ & $12.3 \pm 1.6$ & $7.1 \pm 1.3$ & $19.4 \pm 2.9$ & $662 \pm 240$ \\
  $10\,000$ & $99.7 \pm 11.5$ & $37.7 \pm 7.6$ & $137 \pm 18.2$ & $739 \pm 167$ \\
  $100\,000$ & $978 \pm 105$ & $332 \pm 61.7$ & $1310 \pm 160$ & $1\,131 \pm 134$ \\
         \bottomrule
    \end{tabular}
\end{table}

\begin{table}[]
    \centering
    \caption{Computational costs by test set size for $n=1\,200$ training points and an MLP with $3\times 64$ hidden layers. We report mean $\pm$ standard deviation across 10 runs.}
    \label{tab:comp_costs_test_size}
    \begin{tabular}{ccccc}
    \toprule 
         Number of Test points $m$ & Hessian Approx. (s) & Influence Scoring (s) & Total (s) & Memory (MB) \\ 
         \midrule 
  $100$ & $18.3 \pm 2.5$ & $4.3 \pm 0.64$ & $22.6 \pm 3.1$ & $684 \pm 341$ \\
  $1\,000$ & $18.2 \pm 2.2$ & $7.9 \pm 1.1$ & $26.1 \pm 3.3$ & $758 \pm 188$ \\
  $10\,000$ & $18.0 \pm 2.1$ & $43.3 \pm 8.7$ & $61.3 \pm 10.7$ & $925 \pm 112$ \\
  $100\,000$ & $18.7 \pm 2.0$ & $406 \pm 60.0$ & $425 \pm 61.8$ & $1\,271 \pm 178$ \\
         \bottomrule
    \end{tabular}
\end{table}

\subsubsection{Alternative Data Attribution Methods}

Beyond influence functions, alternative training data attribution methods exist with different trade-offs. 
Retraining-based approaches such as naive leave-one-out retraining and Data Shapley \cite{ghorbani_data_2019} provide gold-standard estimates but are computationally prohibitive.
Simpler approximations include similarity-based methods that compare loss gradients (e.g., Grad-Dot \cite{charpiat_input_2019}), or penultimate-layer representations \cite{pezeshkpour_empirical_2021}. 
Alternative approaches that do not rely on estimating leave-one-out retraining effects include the Representer Points method \cite{yeh_representer_2018}, which avoids Hessian computation via $L^2$ regularization and application of the representer theorem.
TracIn \cite{pruthi_estimating_2020} accumulates gradient dot products throughout training to trace influence over the optimization trajectory. 
DualXDA \cite{yolcu_sparse_2025} performs attribution in the penultimate layer using SVM dual coefficients for sparse and efficient estimates, further employing relevance back-propagation \cite{bach_pixel-wise_2015} to pinpoint influential features in the given train and test samples, explaining the interaction of these data points.

\subsection{Comparison to Training-Time Diagnostics}
\label{app:related_comparison}

To highlight the core methodological distinction, we compare against three representative works that too examine how loss components, collocation points, and temporal structure shape PINN behavior: \citet{wangGradientAlignmentPhysicsinformed2026} diagnose gradient conflicts between loss terms via cosine-similarity metrics, PINNACLE \cite{LauPINNACLEPINNAdaptive2024} selects collocation points based on alignment with NTK eigenfunctions, and \citet{wang_respecting_2024} enforce temporal causality during training. These approaches operate during training at batch or trajectory granularity. \PINNfluence instead operates post-hoc -- that is, after training -- and per-point: it attributes individual predictions to individual training points and individual loss components on a converged model, addressing complementary questions rather than competing with training-time interventions.
\cref{tab:related_comparison_app} summarizes the methodological distinctions between \PINNfluence and the closest training-time approaches discussed in \cref{sec:related_work}.

\begin{table}[h]
\centering
\caption{Comparison of related training-time diagnostic methods and \PINNfluence.}
\label{tab:related_comparison_app}
\small
\setlength{\tabcolsep}{4pt}
\begin{tabular}{@{}lp{0.18\linewidth}p{0.15\linewidth}p{0.18\linewidth}p{0.22\linewidth}@{}}
\toprule
 & Gradient Alignment \cite{wangGradientAlignmentPhysicsinformed2026} & PINNACLE \cite{LauPINNACLEPINNAdaptive2024} & Causal PINN \cite{wang_respecting_2024} & \PINNfluence \\
\midrule
When & Training & Training & Training & Post-hoc \\
Goal & Fix gradient conflicts & Select collocation pts & Enforce temporal causality & Attribute predictions to training data \\
Granularity & 1 Scalar/loss-pair/step & Batch-level & Temporal slice & Per point $\times$ per prediction $\times$ per loss \\
Per-prediction attribution? & No & No & No & Yes \\
\bottomrule
\end{tabular}
\end{table}

\newpage

Beyond the aforementioned differences, the methods differ in the kind of question they answer. Gradient alignment analysis \cite{wangGradientAlignmentPhysicsinformed2026} reveals \emph{that} gradients of different loss terms conflict during training and proposes a condensed scalar cosine similarity per loss pair. \PINNfluence on the other hand identifies \emph{where} in the domain and \emph{through which} loss component this interplay shapes a given model prediction. As a result, it provides a spatiotemporal resolution that aggregate gradient statistics do not provide. PINNACLE \cite{LauPINNACLEPINNAdaptive2024} ranks candidate collocation points \emph{prospectively} based on their alignment with eNTK eigenfunctions, whereas our approach operates \emph{retrospectively} by quantifying an already converged model's sensitivity to training data perturbations. Therefore, these methods capture different notions of influence: contribution to convergence versus counterfactual perturbation.
A consequence of the per-point post-hoc design is that \PINNfluence can surface phenomena inaccessible to training-time analyses. Causality-enforcing training schemes \cite{wang_respecting_2024} impose temporal ordering during training; our experiments reveal that well-trained PINNs exhibit temporally symmetric rather than causally ordered influence (\cref{tab:mean_directionality}), a finding inaccessible to training-dynamics analyses alone.

\subsection{Practitioner's Guide}\label{app:practitioners-guide}
In order to exemplify the application of \PINNfluence, we give a short recipe on how to use and develop diagnostic indicators in this subsection.
The general-purpose influence heatmaps and loss-component decompositions are always computable and already constitute the principal diagnostic signal throughout this work (e.g., \cref{fig:heatmap_test_point_drift_diffusion_output,fig:drift_diffusion_loss_fractions}).
Generating new indicators for hypothesis-driven testing does require some domain knowledge, but enables domain-specific questions beyond generic diagnostics. This reliance on human judgment is inherent to most interpretability methods, where practitioners must decide which patterns are meaningful. This is the case for example with image saliency maps in computer vision or token attributions in natural language processing.

We suggest the following recipe. First, (a) compute and inspect the generic influence patterns to identify anomalies. This step does not require domain knowledge. 
Second, (b) examine whether the observed patterns align with some expected physical behavior of the problem. This can include notions of symmetries, spatial locality, or temporal causality.
Third, (c) formalize these hypotheses into indicators for quantitative verification, using constructions as shown in \cref{eq:regional_influence} and \cref{eq:causality}. The prominent example in our work is the temporal influence indicator \cref{eq:causality}. Further instantiations are provided in \cref{app:additional-investigations} and can be adapted as templates to new PDEs.
In summary, step (a) is problem-agnostic and applicable out of the box. Steps (b) and (c) presuppose some domain knowledge to know what physical behavior to expect and to formalize the hypothesis into an indicator. In the main text, \cref{tab:diagnostic_questions} summarizes which tool answers which diagnostic question.

\subsection{Resampling and Loss-Weighting Based on \PINNfluence}\label{app:future-work}
In this subsection, we succinctly outline two possible extensions of our framework for active training interventions in PINNs.
The diagnostics developed in this work are post-hoc, but the influences they are based upon can in principle also be used to rank training points and loss components by their effect on model predictions or physical observables. This suggests two possible directions for utilizing \PINNfluence diagnostics as interventions: resampling and loss-reweighting.

\paragraph{Resampling}
Adaptive collocation sampling resamples or adds points based on some scoring function of potential training points, conventionally the magnitude of the PDE residual \cite{lu_deepxde_2021,liu_grad-rar_2022,wu_comprehensive_2023,LauPINNACLEPINNAdaptive2024}. The influence scores of \PINNfluence constitute another possible scoring criterion. Using \cref{eq:influence-def}, one can score current as well as hypothetical training points at any domain location, since it is defined on the whole domain. The most straightforward such criterion is a variation of an absolute point-to-region indicator (see \cref{eq:point_to_region} and \cref{eq:regional_influence}).
Such an influence-guided train–score–resample scheme would complement residual-based sampling.

\paragraph{Loss-weighting}
The imbalance of loss components is a known PINN failure mode \cite{wang_when_2022,wangGradientAlignmentPhysicsinformed2026}. As stated in the main text, the influence is linear in the loss argument (see \cref{cor:linearity-additivity}). 
Scaling a loss weight $\lambda_i$ thus scales the corresponding influence contribution by the same factor. The fractional loss contributions $r_{L_i}$ \cref{eq:loss_fractions_decomposition} can therefore act as both a diagnostic and a control signal.
For example, the persistent IC dominance observed in poorly-trained models (see \cref{fig:loss_fractions_time_dependent_probs}) could be addressed by iteratively rebalancing the loss weights during training: downweighting the overly attributed loss term (or equivalently upweighting the remaining terms), turning the diagnostic signal into an actionable reweighting criterion.

\section{Problem Specifications}
\label{app:problems}
\subsection{Investigated Partial Differential Equations}
\label{app:pdes}

\paragraph{Allen-Cahn Equation:} A nonlinear reaction-diffusion equation modeling phase separation in materials which is given by
\begin{align}
    \label{problem:allen_cahn}
    \frac{\partial u}{\partial t} - \epsilon^2 \frac{\partial^2u}{\partial x^2} + f(u) &= 0 \quad. 
\end{align}
We take $\Omega = (-1,1) \times (0,1)$, select $f(u) = -5(u^3 - u)$ and $\epsilon = \sqrt{10^{-4}}$, with the following boundary conditions $u(x, 0) = x^2 \cos (\pi x)$, $u(-1, t) = -1$ and $u(1, t) = -1$. 

\begin{figure}
\centering
\begin{subfigure}{0.3\textwidth}
    \includegraphics[width=\textwidth]{figures/icml_preds/allen_cahn_true.jpg}
    \caption{Targets}
\end{subfigure}
\hfill 
\begin{subfigure}{0.3\textwidth}
    \includegraphics[width=\textwidth]{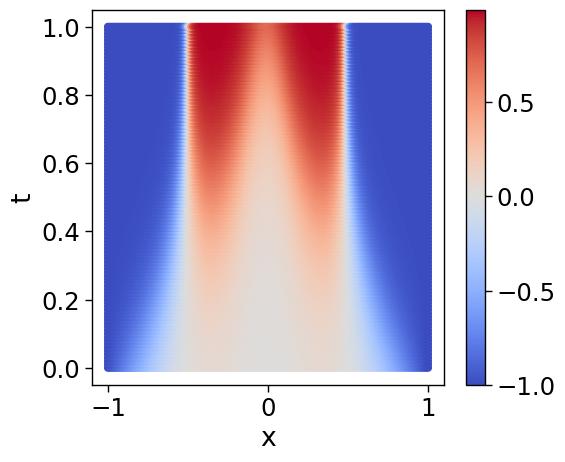}
    \caption{Well-trained}
\end{subfigure}
\hfill
\begin{subfigure}{0.3\textwidth}
    \includegraphics[width=\textwidth]{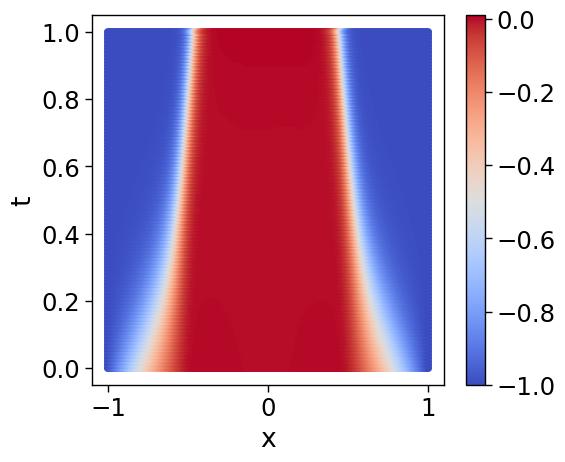}
    \caption{Poorly-trained}
\end{subfigure}
\caption{Allen-Cahn equation: target values and predictions for well and poorly-trained configurations averaged over 10 seeds.}
\label{app:preds_allen_cahn}
\end{figure}

In the specified setting the reaction term $-5(u^3-u)=-5u(u^2 -1)$ has two stable wells at $u = \pm 1$ for regions with negligible diffusion. The setup drives negative $u$ to $-1$ and positive to $1$, respectively. Note however that the reaction term may also collapse to $0$ for $u=0$, allowing for low PDE residuals with an incorrect solution. We obtained reference values from \cite{wu_comprehensive_2023} shown in \cref{app:preds_allen_cahn}.

\begin{figure}
\centering
\begin{subfigure}{0.3\textwidth}
    \includegraphics[width=\textwidth]{figures/icml_preds/burgers_true.jpg}
    \caption{Targets}
\end{subfigure}
\hfill 
\begin{subfigure}{0.3\textwidth}
    \includegraphics[width=\textwidth]{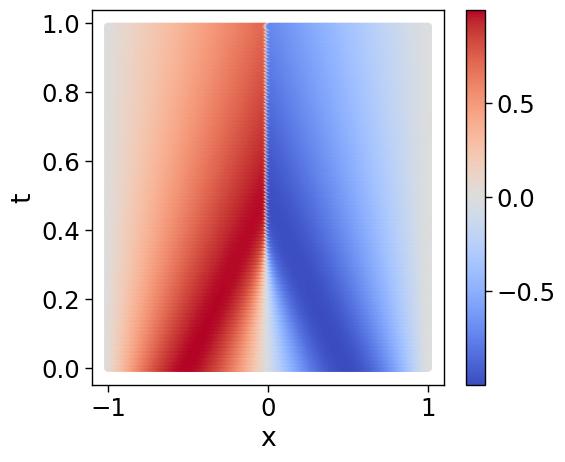}
    \caption{Well-trained}
\end{subfigure}
\hfill
\begin{subfigure}{0.3\textwidth}
    \includegraphics[width=\textwidth]{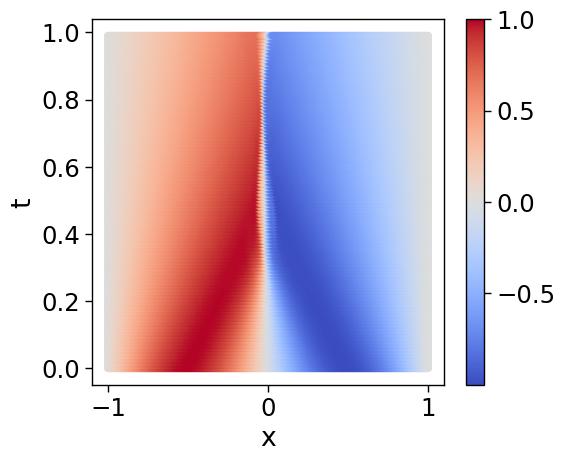}
    \caption{Poorly-trained}
\end{subfigure}
\caption{Burgers' equation: target values and predictions for well and poorly-trained configurations averaged over 10 seeds.}
\label{app:preds_burgers}
\end{figure}

\paragraph{Burgers' Equation:} A second-order nonlinear PDE that models phenomena combining convection and diffusion, often used as a simplified model for fluid dynamics or traffic flow. It is given by 
\begin{align}
    \frac{\partial u}{\partial t} + u \frac{\partial u}{\partial x} -\nu \frac{\partial^2u}{\partial x^2}&= 0\quad. 
\end{align}
We take $\Omega = (-1,1) \times (0,1)$ and set the viscosity to $\nu = \frac{10^{-3}}{\pi}$, with the following boundary conditions $u(x, 0) = -\sin(\pi x)$, $ u(-1, t) = 0$ and $u(1, t) = 0$. In this setup it develops a sharp shock around $x=0$. We obtained reference values from \cite{wu_comprehensive_2023} shown in \cref{app:preds_burgers}.

\begin{figure}
\centering
\begin{subfigure}{0.3\textwidth}
    \includegraphics[width=\textwidth]{figures/icml_preds/drift_diffusion_true.jpg}
    \caption{Targets}
\end{subfigure}
\hfill 
\begin{subfigure}{0.3\textwidth}
    \includegraphics[width=\textwidth]{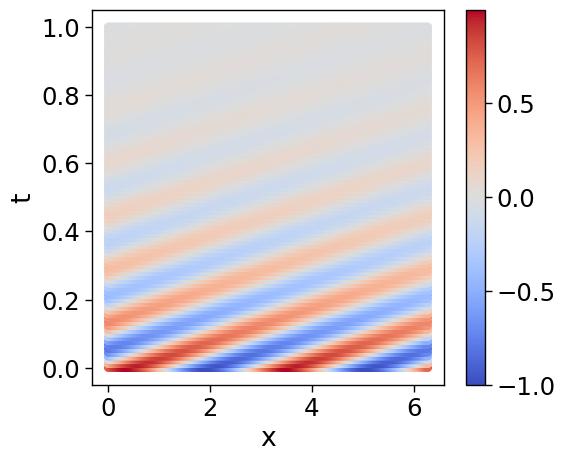}
    \caption{Well-trained}
\end{subfigure}
\hfill
\begin{subfigure}{0.3\textwidth}
    \includegraphics[width=\textwidth]{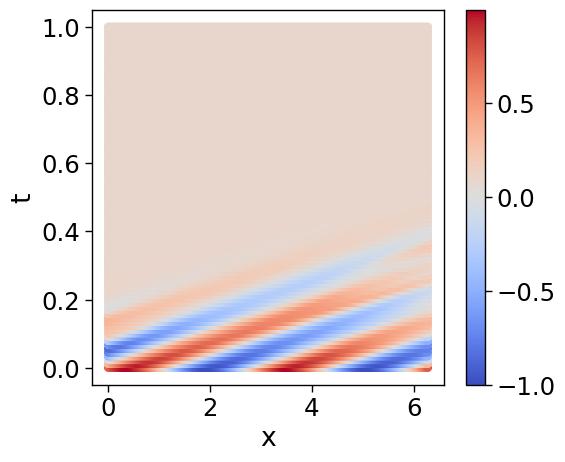}
    \caption{Poorly-trained}
\end{subfigure}
\caption{Drift-diffusion equation: target values and predictions for well and poorly-trained configurations averaged over 10 seeds.}
\label{app:preds_drift_diff}
\end{figure}

\paragraph{Drift-Diffusion Equation:} A second-order linear PDE describing diffusion with coefficient $\alpha$ with given drift $\beta$
\begin{align}
    \frac{\partial u}{\partial t} - \alpha \frac{\partial^2 u}{\partial x^2} + \beta \frac{\partial u}{\partial x} &= 0
\end{align}
We take $\Omega = (0, 2\pi) \times (0,1)$ and set $\alpha=1, \beta=20$, with the following initial and boundary conditions $u(x, 0) = \sin\left(2x + \frac{\pi}{4}\right)$, $u(0,t) = \sin\left(\frac{\pi}{4} - 2 \cdot \beta  t\right)\ e^{-4 \alpha t}$ and $u(2\pi,t) = \sin\left(\frac{17\pi}{4} - 2 \beta t\right)\  e^{-4 \alpha t}$. This yields the closed form solution
\begin{align}
    u(x,t) = \sin\left(2x-2\beta t+\frac{\pi}{4}\right)\ e^{-4\alpha t}\ .
\end{align}
A similar problem was found to cause propagation failures in PINNs \cite{krishnapriyan_characterizing_2021}, where the PINN fails to generalize over the domain, which motivated the inclusion of this example as a prominent problem in the main text. We provide a visualization in \cref{app:preds_drift_diff}.

\paragraph{Heat Equation:} A second-order linear PDE for modeling diffusion
\begin{align}
    \frac{\partial u}{\partial t} - \frac{\partial^2u}{\partial x^2} - 
    s(\bx) = 0\quad,
\end{align}
where $s:\Omega\to\mathbb{R}$ denotes a source term, which we set to $s(x,t) \coloneq (\pi^2 - 1) \sin(\pi x) e^{-t}$. 

\begin{figure}
\centering
\begin{subfigure}{0.3\textwidth}
    \includegraphics[width=\textwidth]{figures/icml_preds/diffusion_true.jpg}
    \caption{Targets}
\end{subfigure}
\hfill 
\begin{subfigure}{0.3\textwidth}
    \includegraphics[width=\textwidth]{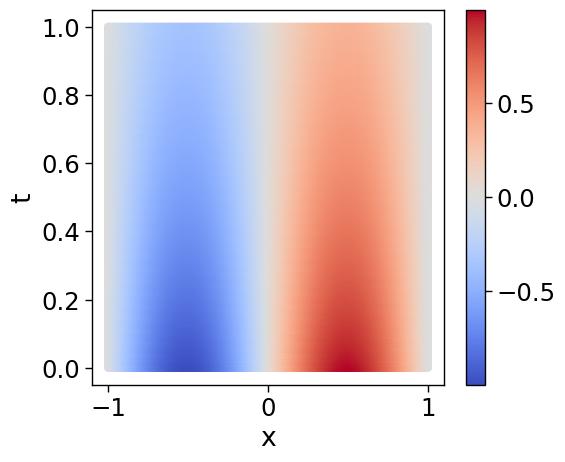}
    \caption{Well-trained}
\end{subfigure}
\hfill
\begin{subfigure}{0.3\textwidth}
    \includegraphics[width=\textwidth]{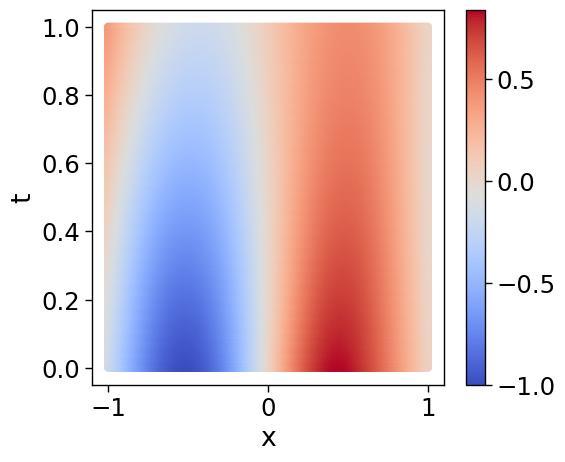}
    \caption{Poorly-trained}
\end{subfigure}
\caption{Heat equation: target values and predictions for well and poorly-trained configurations averaged over 10 seeds.}
\label{app:preds_heat}
\end{figure}

We chose $\Omega = (-1,1) \times (0,1)$, with the following boundary conditions $u(x, 0) = \sin(\pi x)$, $u(-1, t ) = 0$ and $u(1, t) = 0$. This yields the closed form solution \cite{wu_comprehensive_2023}.
\begin{align}
    u(x,t) = \sin(\pi x) e^{-t}
\end{align}
The heat equation is a considerably simple PDE for basic benchmarks. See \cref{app:preds_heat} for visualization.

\begin{figure}
\centering
\begin{subfigure}{0.3\textwidth}
    \includegraphics[width=\textwidth]{figures/icml_preds/wave_true.jpg}
    \caption{Targets}
\end{subfigure}
\hfill 
\begin{subfigure}{0.3\textwidth}
    \includegraphics[width=\textwidth]{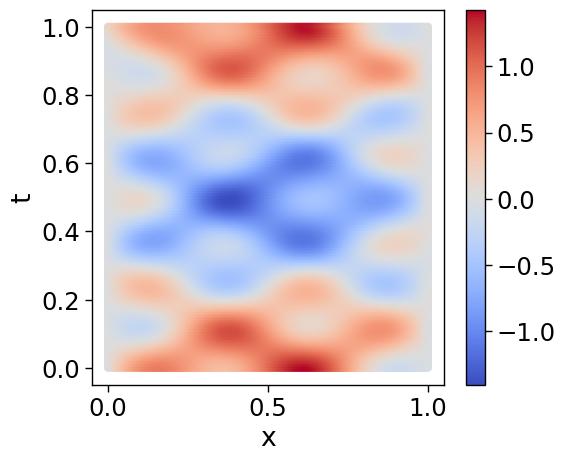}
    \caption{Well-trained}
\end{subfigure}
\hfill
\begin{subfigure}{0.3\textwidth}
    \includegraphics[width=\textwidth]{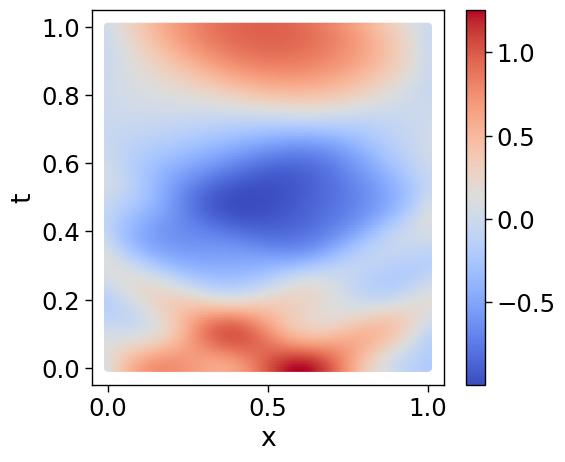}
    \caption{Poorly-trained}
\end{subfigure}
\caption{Wave equation: target values and predictions for well and poorly-trained configurations averaged over 10 seeds.}
\label{app:preds_wave}
\end{figure}

\paragraph{Wave Equation:} A second-order linear PDE describing the propagation of waves (e.g., vibrations, acoustics), given by 
\begin{align}
    \frac{\partial^2u}{\partial t^2} - c^2 \frac{\partial^2 u}{\partial x^2} &= 0
\end{align}
We take $\Omega = (0,1) \times (0,1)$, set $c=2$, with the following boundary conditions $u(x, 0) = \sin (\pi x) + \frac{1}{2} \sin (4 \pi x)$, $\frac{\partial u}{\partial t}(x, 0) = 0$, $u(0, t) = 0$ and $u(1, t) = 0$.
This yields the following closed form solution
\begin{align}
    u(x,t) = \sin(\pi x)\cos(\pi c t) + \frac{1}{2}\sin(4\pi x)\cos(4\pi c t).
\end{align}
This IBVP yields antisymmetric patterns as well as high and low frequency changes in the solution shown in \cref{app:preds_wave}.

\paragraph{Navier-Stokes Equations:} A system of second-order nonlinear PDEs describing the motion of viscous fluids, such as water or air. We follow the formulation from the DFG flow around cylinder benchmark in the laminar case with the Reynolds number being set to $20$ \cite{hirschel_benchmark_1996}. This results in an incompressible and steady instance of the Navier-Stokes equations, which are given in this case as
\begin{itemize}
    \item $x$-Velocity: $u \frac{\partial u}{\partial x} + v \frac{\partial u}{\partial y} + \frac{1}{\rho} \frac{\partial p}{\partial x}- \nu \left( \frac{\partial^2 u}{\partial x^2} + \frac{\partial^ 2 u}{\partial y^ 2} \right) = 0$
    \item $y$-Velocity: $u \frac{\partial v}{\partial x} + v \frac{\partial v}{\partial y} + \frac{1}{\rho} \frac{\partial p}{\partial y} - \nu \left( \frac{\partial^2 v}{\partial x^2} + \frac{\partial^ 2 v}{\partial y^ 2} \right) = 0$
    \item Continuity: $\frac{\partial u}{\partial x} + \frac{\partial v}{\partial y} = 0$
\end{itemize}
where $(u,v) \in \mathbb{R}^2$ denotes the velocity field, $p \in \mathbb{R}$ the pressure, $\rho$ the fluid density and $\nu$ the kinematic viscosity. 
A Reynolds number of $20$ induces $\rho=1$ and $\nu=\frac{1}{1000}$. 
We solve the non-dimensionalized form of these equations, scaling coordinates by characteristic length $L^* = 0.1$ and velocities by $U^* =0.2$. 
The domain is set to $\Omega = (0, 2.2) \times (0, 0.41) \setminus C_r(0.2, 0.2)$, where $C_r$ denotes a cylinder with center at $(0.2, 0.2)$ and radius $r=0.05$.
The boundary conditions are given as follows, the inflow is characterized by a parabola $u(0, y) = \left( \frac{4 U y(0.41 - y)}{0.41^2}, 0\right)$ where $U=0.3$ denotes the maximum inflow velocity.
As boundary conditions for the lower and upper walls $\Gamma_{\text{bottom}} = (0, 2.2) \times \{0\}$ and $\Gamma_{\text{top}} = (0, 2.2) \times \{0.41\}$ as well as on the cylinder boundary $\Gamma_{\text{cylinder}} = \partial C_r(0.2,0.2)$, the predicted $x$ and $y$-velocity should be 0.
    
Furthermore, for the outflow on the right edge, i.e., $\Gamma_{\text{outflow}} = \{2.2\} \times (0, 0.41)$ do-nothing boundary conditions are defined, i.e., $\nu \frac{\partial u}{\partial x} - p$ and $\nu \frac{\partial v}{\partial y}$ should be $0$.
    
\begin{figure}
\centering
\includegraphics[width=0.3\textwidth]{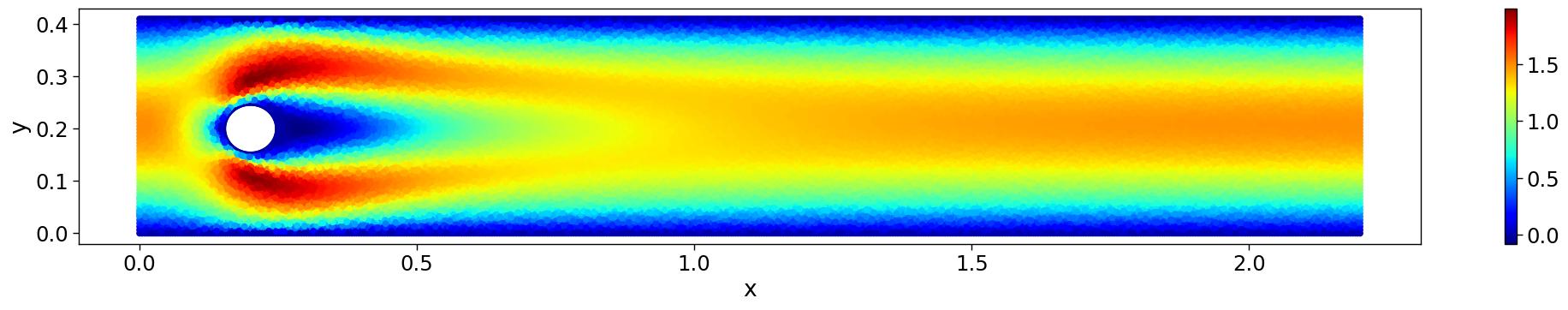}
\hfill
\includegraphics[width=0.3\textwidth]{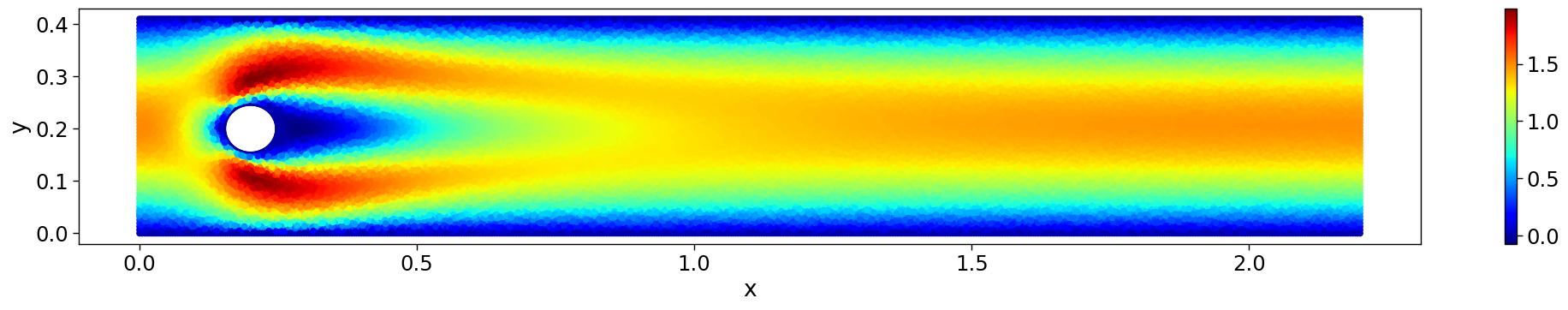}
\hfill
\includegraphics[width=0.3\textwidth]{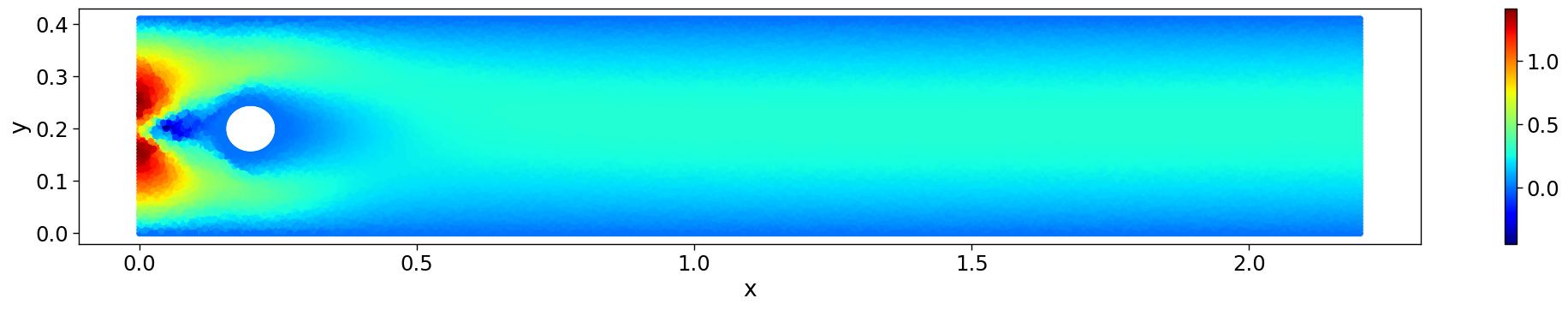}

\vspace{0.3em}

\includegraphics[width=0.3\textwidth]{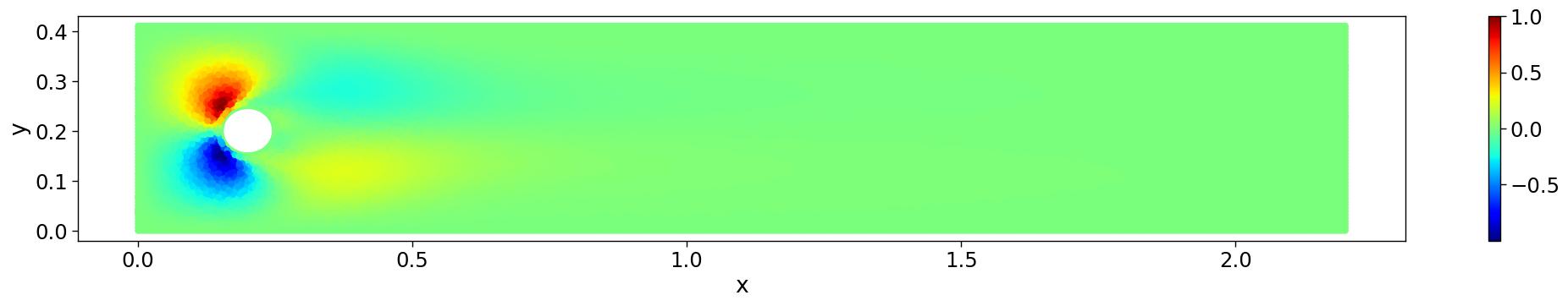}
\hfill
\includegraphics[width=0.3\textwidth]{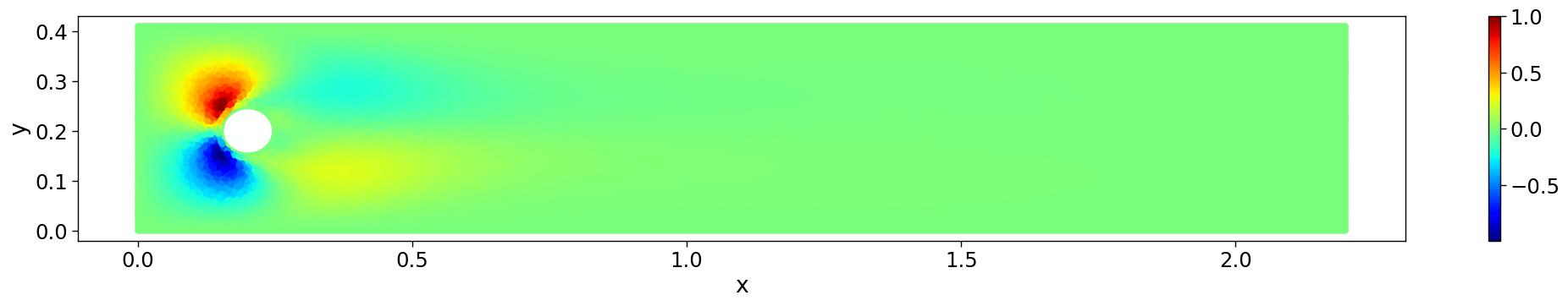}
\hfill
\includegraphics[width=0.3\textwidth]{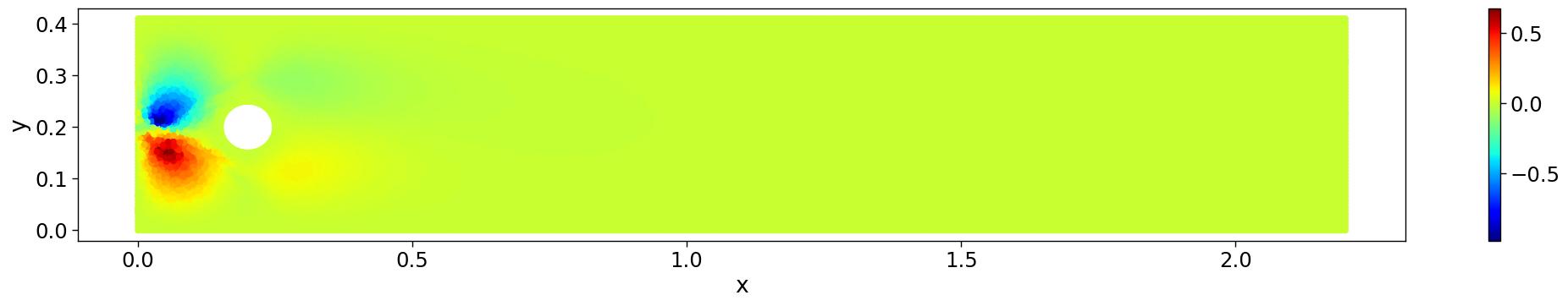}

\vspace{0.3em}

\begin{subfigure}{0.3\textwidth}
    \includegraphics[width=\textwidth]{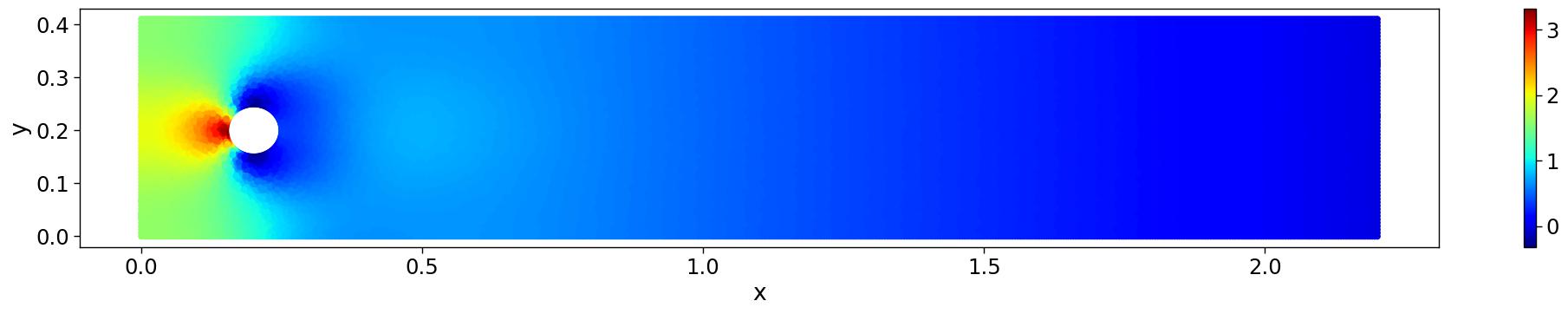}
    \caption{Target}
\end{subfigure}
\hfill
\begin{subfigure}{0.3\textwidth}
    \includegraphics[width=\textwidth]{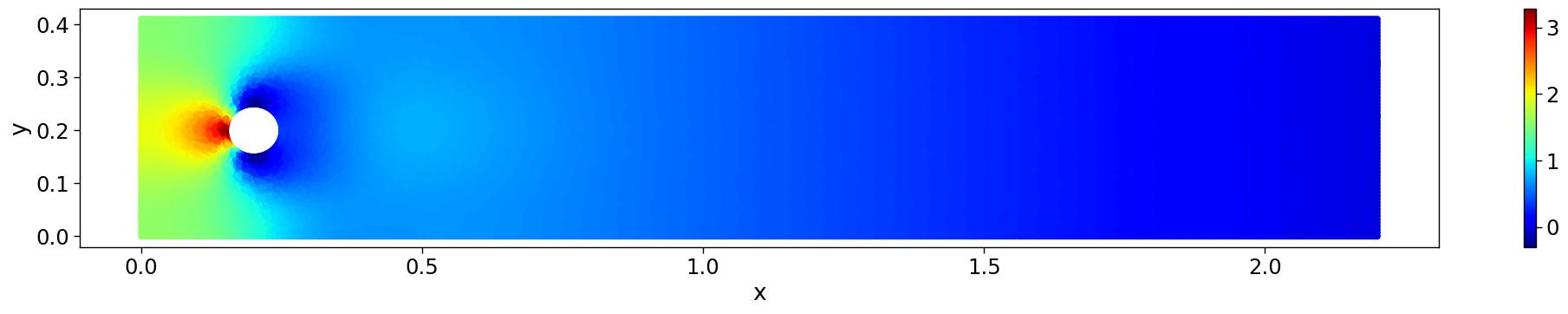}
    \caption{Well-trained}
\end{subfigure}
\hfill
\begin{subfigure}{0.3\textwidth}
    \includegraphics[width=\textwidth]{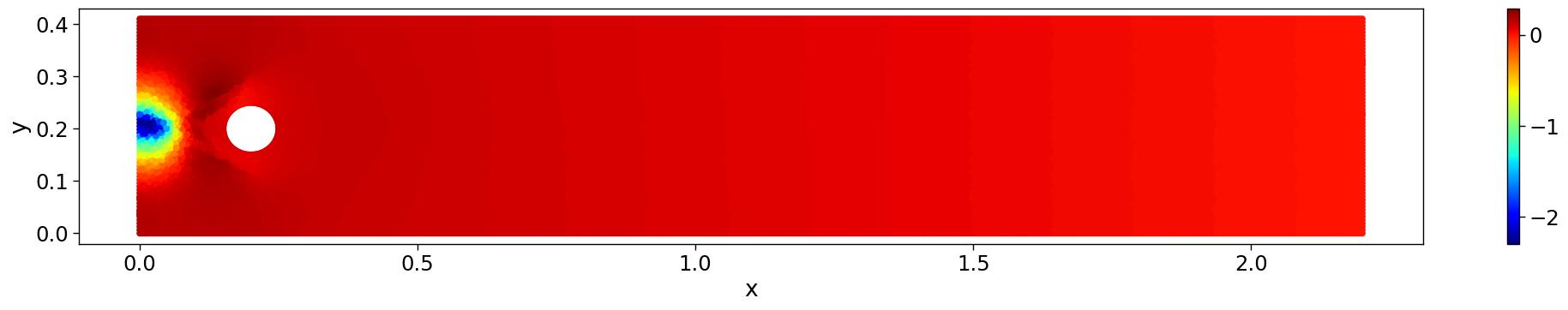}
    \caption{Poorly-trained}
\end{subfigure}

\caption{Navier-Stokes equations: target values and predictions for $x$-velocity $u$ (top row), $y$-velocity $v$ (middle row), and pressure $p$ (bottom row). Predictions averaged over 10 seeds.}
\label{app:preds_ns}
\end{figure}

Target values are obtained from \cite{wang_experts_2023} and are visualized in \cref{app:preds_ns}.

\paragraph{Poisson Equation with a point source over a unit disk:} The Poisson equation models the electrostatic potential induced by an electric point charge and is given by
\begin{align}
    \frac{\partial^2 u}{\partial x^2} + \frac{\partial^2u}{\partial y^2} + \delta_\sigma(\bx) &= 0 \quad,
    \label{eq:poisson_disk_pde}
\end{align}
where $\delta_{\sigma}$ denotes a Gaussian approximation of the Dirac delta with bandwidth $\sigma = 0.02$. The domain we consider is the unit disk $\Omega = \{ x \in \mathbb{R}^2:\| x \|_2 < 1 \}$ with a unit point source at the origin $c=(0,0)$. The analytical solution corresponding to the Dirac source (recovered in the limit $\sigma \to 0$) is radially symmetric and is given as 
\begin{align}
    u(r) = -\frac{1}{2\pi} \log r\quad ,
\end{align}
where $r=\|\bx\|_2$ with $0<r\leq1$. We specify the following boundary condition $u(\bx) = 0$ for $\|\bx\| = 1$.

\begin{figure}
\centering
\begin{subfigure}{0.3\textwidth}
    \includegraphics[width=\textwidth]{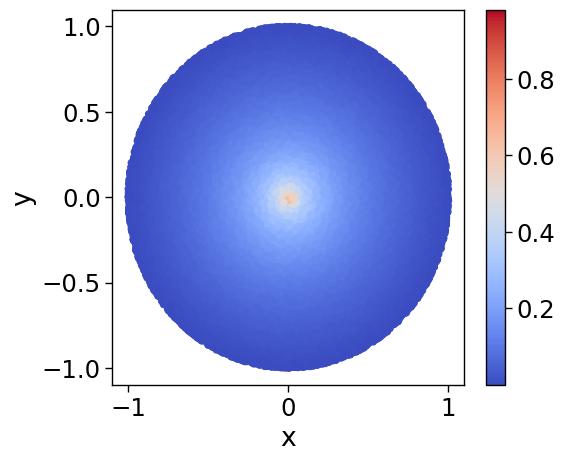}
    \caption{Targets}
\end{subfigure}
\hfill 
\begin{subfigure}{0.3\textwidth}
    \includegraphics[width=\textwidth]{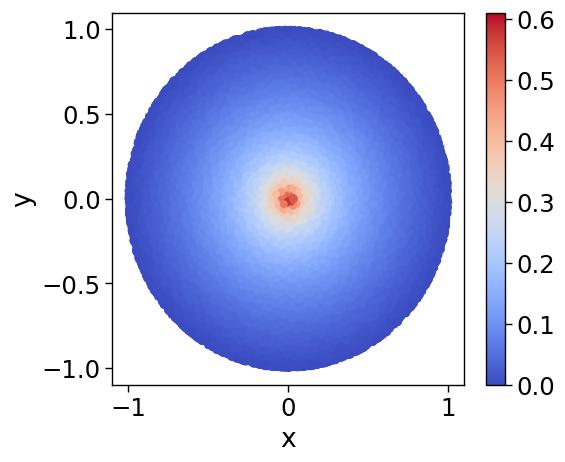}
    \caption{Well-trained}
\end{subfigure}
\hfill
\begin{subfigure}{0.3\textwidth}
    \includegraphics[width=\textwidth]{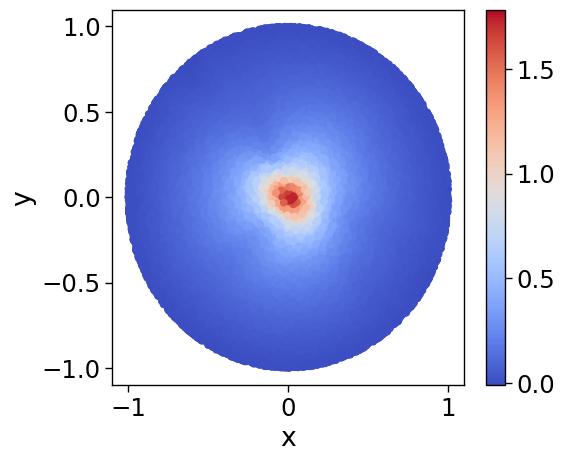}
    \caption{Poorly-trained}
\end{subfigure}
\caption{Poisson equation: target values and predictions for well and poorly-trained configurations averaged over 10 seeds.}
\label{app:preds_poisson}
\end{figure}

This problem yields a difficult to learn singularity and exhibits radial symmetries, which result in an interesting benchmark problem. We provide a visualization in \cref{app:preds_poisson}.

\subsection{Training Setup}
\label{app:training}

All models have been trained with the DeepXDE library \cite{lu_deepxde_2021}, including the geometry generation and sampling, with PyTorch \cite{paszke_pytorch_2019} used as its backend. 
Optimization was carried out in two stages, first with the Adam optimizer \cite{kingma_adam_2017} followed by fine-tuning with the second-order L-BFGS optimizer \cite{liu_limited_1989}. 
Note that to construct some of the poorly-trained models, we deliberately skipped the L-BFGS optimization step. Further, we note that for the Allen-Cahn well-trained configuration a minority of seeds collapse to
the spurious constant solution $u\equiv 0$ in the domain interior despite the well-trained
protocol (see the reaction term discussion in \cref{app:pdes}). We therefore report the
well-trained Allen-Cahn statistics over seeds that converged to the correct solution. 

The hyperparameters for each configuration are presented in \cref{tab:app_hyperparams}. All models were trained as simple feed-forward neural networks with equal number of neurons per hidden layer and $\tanh$ activations.

\begin{table}[h]
\centering
\caption{Hyperparameters for well-trained and poorly-trained configurations. Layers and neurons specify depth and width of the feed-forward neural networks; $N_{\text{domain}}$, $N_{\text{bc}}$, $N_{\text{ic}}$ specify numbers of sampled points inside the domain, on boundaries and on the initial time slice; $N_{\text{iter}}^{(\text{Adam})}$ and $N_{\text{iter}}^{(\text{L-BFGS})}$ specify the number of optimization steps using the respective optimizer.}
\label{tab:app_hyperparams}
\begin{tabular}{llccccccc}
\toprule
Problem & Config & Layers & Neurons & $N_{\text{domain}}$ & $N_{\text{bc}}$ & $N_{\text{ic}}$ & $N_{\text{iter}}^{(\text{Adam})}$ & $N_{\text{iter}}^{(\text{L-BFGS})}$ \\
\midrule
\multirow{2}{*}{Allen-Cahn} 
    & Well-trained & 3 & 64 & 2\,500 & 500 & 500 & 100\,000 & 25\,000 \\
    & Poorly-trained  & 3 & 64 & 2\,500 & 500 & 500 & 100\,000 & ---   \\
\midrule
\multirow{2}{*}{Burgers'} 
    & Well-trained & 3 & 32 & 2\,500 & 500 & 500 & 50\,000 & 12\,000 \\
    & Poorly-trained  & 3 & 32 & 500    & 100 & 100 & 50\,000 & 12\,000 \\
\midrule 
\multirow{2}{*}{Drift-Diffusion} 
    & Well-trained & 3 & 64 & 1\,000 & 100 & 100 & 15\,000 & 5\,000 \\
    & Poorly-trained & 3 & 64 & 200    & 20  & 20  & 15\,000 & 5\,000 \\
\midrule
\multirow{2}{*}{Heat} 
    & Well-trained & 3 & 32 & 1\,000 & 100 & 100 & 15\,000 & 5\,000 \\
    & Poorly-trained & 3 & 32 & 10 & 2  & 2  & 15\,000 & 5\,000 \\
\midrule
\multirow{2}{*}{Wave} 
    & Well-trained & 5 & 100 & 2\,500 & 500 & 500 & 100\,000 & 25\,000 \\
    & Poorly-trained & 5 & 100 & 2\,500 & 500 & 500 & 100\,000 & --- \\
\midrule
\multirow{2}{*}{Navier-Stokes} 
    & Well-trained & 3 & 64 & 7\,500  & 2\,500 & --- & 100\,000 & 25\,000 \\
    & Poorly-trained & 3 & 64 & 1\,500  & 500    & --- & 100\,000 & 25\,000 \\
\midrule
\multirow{2}{*}{Poisson} 
    & Well-trained & 3 & 32 & 2\,500 & 500 & --- & 50\,000 & 12\,000 \\
    & Poorly-trained & 3 & 32 & 100    & 20  & --- & 15\,000 & 5\,000 \\
\bottomrule
\end{tabular}
\end{table}

\subsection{Model Predictions and Targets}
\label{app:model_preds_and_targets}

\begin{table}[]
    \caption{$L_2$ relative errors across all configurations. Reported as mean $\pm$ standard deviation $\times$ order of magnitude across all 10 seeds.}
    \label{tab:l2_errors_well_poorly_trained}
    \centering
    \begin{tabular}{lcc}
        \toprule 
        Problem & Well-trained & Poorly-trained \\ 
        \midrule 
        Heat & $(2.8 \pm 1.3) \times 10^{-4}$ & $(2.0 \pm 0.4 )\times 10^{-1}$ \\ 
        Allen-Cahn & $(6.3 \pm 3.4) \times 10^{-2}$ & $(5.3 \pm 0.01) \times 10^{-1}$ \\
        Burgers' & $(5.6 \pm 3.4) \times 10^{-3}$ & $(1.8 \pm 0.7) \times 10^{-1}$\\
        Drift-Diffusion & $(8.6 \pm 9.3) \times 10^{-3}$ & $(8.6 \pm 4.3) \times 10^{-1}$ \\ 
        Wave & $(6.8 \pm 4.7) \times 10^{-2}$ & $(3.8 \pm 0.1) \times 10^{-1}$ \\ 
        Poisson & $(2.9 \pm 0.2) \times 10^{-2}$ & $(1.9 \pm 0.7 )\times 10^0$ \\ 
        Navier-Stokes ($x$-Velocity) & $(3.4 \pm 0.8) \times 10^{-3}$ & $(7.8 \pm 1.2) \times 10^{-1}$ \\
        Navier-Stokes ($y$-Velocity) & $(1.2 \pm 0.2 )\times 10^{-2}$ & $(1.4 \pm 0.3) \times 10^{0}$ \\
        Navier-Stokes (Pressure) & $(8.3 \pm 1.3) \times 10^{-3}$ & $(9.7 \pm 1.7 )\times 10^{-1}$ \\
        \bottomrule
    \end{tabular}
\end{table}

\clearpage

\section{Additional Investigations}\label{app:additional-investigations}

\subsection{Reframing the Temporal Indicator for Steady-State Problems as a Measure of Directionality}
\label{app:directional_indicator}

The causality indicator can be extended to the featured steady-state problems, replacing temporal influence with directional influence, appropriate to the problem structure. 

For the Navier-Stokes equation, we consider the inflow boundary condition analogous to an initial condition. 
Here, the flow propagates along the $x$-axis instead. 
This results in the analogous indicator to \cref{eq:causality} with a different spatial coordinate
\begin{align}
    \eta_{\theta_0}^{L \to f}(R_{\text{tr}}, R_{\text{te}}) = 
    &1-\frac{1}{|R_{\text{te}}|} \sum_{\bz \in R_{\text{te}}} 
    \frac{\displaystyle\sum_{\bx \in R_{\text{tr}:\bx_x \leq \bz_x}} \left| \Inf_{\theta_0}^{L \to f}(\bx, \bz) \right|}{\displaystyle\sum_{\bx \in \mathcal{X}} \left| \Inf_{\theta_0}^{L \to f}(\bx, \bz) \right|}.
    \label{eq:causality_navier_stokes}
\end{align}
For the Poisson equation, we consider the problem stated at a disk domain, featuring a singularity in the center acting as a source. The directional indicator measures influence from training points closer to the center, relative to the total influence
\begin{align}
    \eta_{\theta_0}^{L \to f}(R_{\text{tr}}, R_{\text{te}}) = 
    &1-\frac{1}{|R_{\text{te}}|} \sum_{\bz \in R_{\text{te}}} 
    \frac{\displaystyle\sum_{\bx \in R_{\text{tr}:\norm{\bx} \leq \norm{\bz}}} \left| \Inf_{\theta_0}^{L \to f}(\bx, \bz) \right|}{\displaystyle\sum_{\bx \in \mathcal{X}} \left| \Inf_{\theta_0}^{L \to f}(\bx, \bz) \right|}.
    \label{eq:causality_poisson}
\end{align}

\begin{table}[h]
\caption{Spatial directionality indicator for steady-state problems, evaluated for predictions $f=\hat u$. We report mean $\pm$ standard deviation over 10 seeds.}
\label{tab:mean_directionality_steady}
\centering
\begin{tabular}{lccccc}
\toprule
Problem & $f$ & $\bar x$ & Well-trained & Poorly-trained \\
\toprule 
Poisson & $\hat u$ & 0.28 & 0.29 $\pm$ 0.03  & 0.69 $\pm$ 0.06 \\
\midrule
\multirow{3}{*}{Navier-Stokes} 
 & $x$-velocity & 0.48 & 0.15 $\pm$ 0.01 & 0.15 $\pm$ 0.03 \\
 & $y$-velocity & 0.48 & 0.25 $\pm$ 0.01 & 0.2 $\pm$ 0.05 \\
 & Pressure  & 0.48 & 0.14 $\pm$ 0.02 & 0.11 $\pm$ 0.02 \\
\bottomrule
\end{tabular}
\end{table}

\cref{tab:mean_directionality_steady} reports directionality indicators. 
For the well-trained Poisson model, the indicator matches the expected baseline. For the poorly-trained model, the indicator is far above the baseline. 
Contrary to the time-dependent problems, the model is heavily influenced by points near the boundary.
This represents the inverse of the time-dependent pathology: rather than over-relying on the source term the poorly-trained model completely neglects it. Note that it does not feature a separate loss term and is instead modeled solely through the PDE residual. 

For the Navier-Stokes problem, we consider steady flow around an obstacle close to the inflow. 
We find that the directionality indicator shows no separation between the model configurations. 
This is because influences are concentrated around the obstacle, which is the defining structure of the flow, rather than distributed along the flow direction. 
The measured indicator values confirm this, as they closely match the ratio of training points downstream versus upstream of the obstacle.

\clearpage

\subsection{Additional Indicators}
\label{app:additional_indicators_region}

In this section we demonstrate how one may construct problem-specific indicators using \PINNfluence to specifically probe individual model characteristics. 

\paragraph{Navier-Stokes: Downstream Influence on Individual Regions}

The region-based influence indicator $\rho_{\theta_0}^{L\to f}$ \cref{eq:regional_influence} allows us to quantify the influence of one region in the training set $R_{\text{tr}}$ onto another region in the test set $R_{\text{test}}$.
We will exemplify this on the Navier-Stokes problem.

\begin{figure}[ht]
    \centering
    \includegraphics[width=0.8\linewidth]{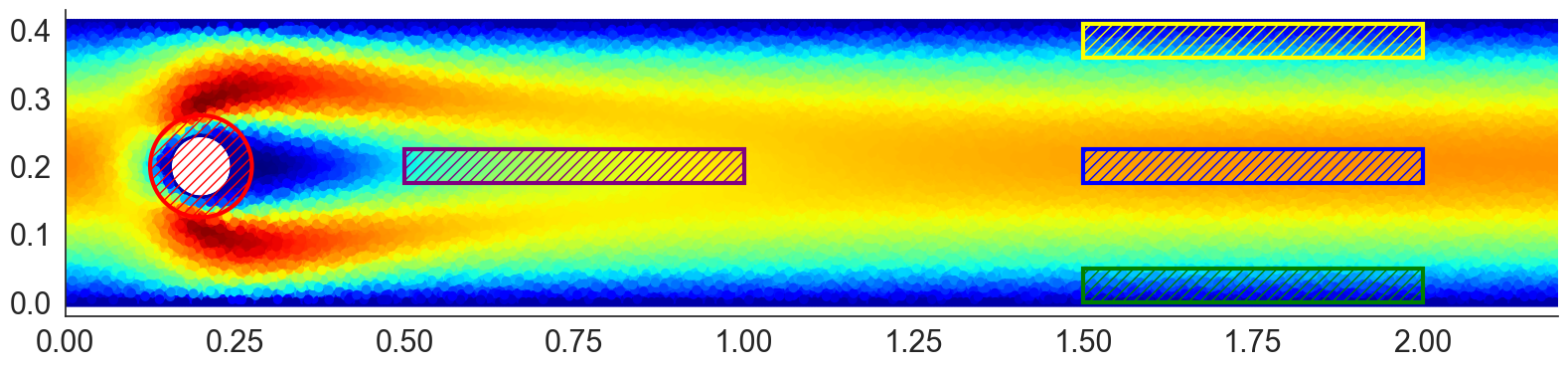}
    \caption{Regions of interest to quantify the influence of the region surrounding the cylinder on downstream locations: (i) closely behind the cylinder; (ii) close to the outflow, centered; (iii) close to the outflow, top; (iv) close to the outflow, bottom.}
    \label{fig:ns_regions_of_interest}
\end{figure}

\begin{table}[h]
\caption{Influence scores from cylinder region for different loss terms and spatial regions. We report mean $\pm$ standard deviation over 10 seeds.}
\label{tab:cylinder_influences}
\centering
\begin{tabular}{llcccc}
   \toprule 
   $f$ & Configuration & Center-Rectangle & Outflow-Rectangle & Top-Rectangle & Bottom-Rectangle \\ 
   \midrule
   \multirow[t]{2}{*}{$x$-Velocity} & Well-trained & 0.43 $\pm$ 0.05 & 0.38 $\pm$ 0.05 & 0.33 $\pm$ 0.05 & 0.36 $\pm$ 0.05 \\
   & Poorly-trained & 0.31 $\pm$ 0.09 & 0.27 $\pm$ 0.09 & 0.26 $\pm$ 0.09 & 0.25 $\pm$ 0.09 \\
   \midrule
   \multirow[t]{2}{*}{$y$-Velocity} & Well-trained & 0.27 $\pm$ 0.04 & 0.25 $\pm$ 0.03 & 0.22 $\pm$ 0.03 & 0.23 $\pm$ 0.02 \\
   & Poorly-trained & 0.24$\pm$ 0.08 & 0.23 $\pm$ 0.08 & 0.23 $\pm$ 0.09 & 0.21 $\pm$ 0.09 \\
   \midrule
   \multirow[t]{2}{*}{Pressure} & Well-trained & 0.48 $\pm$ 0.07 & 0.47 $\pm$ 0.07 & 0.47 $\pm$ 0.07 & 0.46 $\pm$ 0.07 \\
   & Poorly-trained & 0.25 $\pm$ 0.06& 0.25$\pm$ 0.06 & 0.24 $\pm$ 0.06& 0.26$\pm$ 0.06 \\
   \midrule
   \multirow[t]{2}{*}{Loss} & Well-trained & 0.32 $\pm$ 0.03 & 0.28 $\pm$ 0.04 & 0.24 $\pm$ 0.03 & 0.25 $\pm$ 0.02 \\
   & Poorly-trained & 0.26$\pm$ 0.08 & 0.23$\pm$ 0.08 & 0.23$\pm$ 0.09 & 0.22 $\pm$ 0.06\\
   \bottomrule
\end{tabular}
\end{table}

The results presented in \cref{tab:cylinder_influences} confirm physical expectations: excluding the predictions for pressure, the influence of the cylinder is strongest in the rectangle directly downstream and diminishes toward the outflow. 
Top and bottom regions show similar values, consistent with the flow's approximate symmetry. 
Note, though, that the cylinder is not perfectly centered, hence slight asymmetries are expected. 
Well-trained models exhibit consistently higher influences than poorly-trained ones, most strikingly for pressure, and lower variance across seeds.

\clearpage

\paragraph{Wave Equation: Symmetry Indicator}

We test for symmetry in influences for the wave equation by comparing the influence of the left spatial half ($x<0.5$) onto the right ($x\geq 0.5$) and vice versa.

\begin{table}[h]
\caption{Spatial symmetry indicators for the Wave equation. Values represent the fraction of influence from one spatial half to the opposite half. The ratio quantifies symmetry (values near 1.0 indicate symmetric influence patterns). We report mean $\pm$ standard deviation over 10 seeds.}
\label{tab:wave_spatial_symmetry}
\centering
\begin{tabular}{lcccc}
\toprule
$f$ & Configuration & $\rho_{\theta_0}^{L\to f}(R_{\text{left}}, R_{\text{right}})$ & $\rho_{\theta_0}^{L\to f}(R_{\text{right}}, R_{\text{left}})$ & $\frac{\rho_{\theta_0}^{L\to f}(R_{\text{right}}, R_{\text{left}})}{\rho_{\theta_0}^{L\to f}(R_{\text{left}}, R_{\text{right}})}$  \\
\midrule
\multirow{2}{*}{$\hat{u}$} 
 & Well-trained & 0.48 $\pm$ 0.04 & 0.49 $\pm$ 0.05 & 1.01 $\pm$ 0.17 \\
 & Poorly-trained & 0.48 $\pm$ 0.04 & 0.51 $\pm$ 0.04 & 0.96 $\pm$ 0.14 \\
\midrule 
\multirow{2}{*}{Loss} 
 & Well-trained & 0.35 $\pm$ 0.03 & 0.36 $\pm$ 0.04 & 0.98 $\pm$ 0.18 \\
 & Poorly-trained & 0.38  $\pm$ 0.02& 0.40 $\pm$ 0.02 & 0.95 $\pm$ 0.08 \\
\bottomrule
\end{tabular}
\end{table}

As shown in \cref{tab:wave_spatial_symmetry}, both configurations display symmetric influence patterns with ratios close to $1$.
Notably, the influence fractions of $0.48$ to $0.51$ indicate that predictions in each half are nearly equally shaped by training points from the opposite half, consistent with perfect symmetry. 
Although the poorly-trained model fails to capture high-frequency features in the solution, these results suggest that it still accurately learns the problem's spatial symmetry.
The loss decomposition, however, clearly separates well-trained from poorly-trained models, see \cref{app:decomposition_plots}.

\paragraph{Burgers' Equation: Locality of the Shock}
\label{app:burgers_shock_locality_indicator}

Burgers' equation develops a sharp shock around $x=0$. 
We test how locally distributed influences are around the shock region by computing the region-to-region influence from training points within $x \in [-0.2,0.2]$ onto test points within the same region.

\begin{table}[ht!]
\caption{Shock locality indicator for Burgers' equation. Values represent the fraction of influence on shock region predictions ($x$ coordinates in $[-0.2,0.2]$) originating from training points within the same region. 
We report mean $\pm$ standard deviation over 10 seeds.
}
\label{tab:burgers_shock_proximity}
\centering
\begin{tabular}{lcc}
\toprule
$f$ & Configuration & $\rho_{\theta_0}^{L\to f}(R_{\text{shock}}, R_{\text{shock}})$ \\
\midrule
\multirow{2}{*}{$\hat{u}$} 
 & Well-trained & 0.66 $\pm$ 0.03 \\
 & Poorly-trained & 0.38 $\pm$ 0.06 \\
\midrule 
\multirow{2}{*}{Loss} 
 & Well-trained & 0.71 $\pm$ 0.02 \\
 & Poorly-trained & 0.47 $\pm$ 0.05 \\
\bottomrule
\end{tabular}
\end{table}

\cref{tab:burgers_shock_proximity} indicates that well-trained models show strong locality, while poorly-trained models exhibit weaker locality,
indicating over-sensitivity to regions irrelevant to the shock. 

\clearpage
\subsection{Continuous Sweep over Training-Data Quantity}\label{app:model-quality-sweep}

The evaluation of \textsc{PINNfluence} on controlled well- and poorly-trained models shows that the method distinguishes accurately between them. 
To show that \PINNfluence's regional indicators respond to \emph{gradual} variation in training quality rather than as a binary artifact of two hand-picked configurations, we additionally sweep the fraction of domain training points continuously, holding the well-trained architecture and optimizer fixed (\cref{tab:app_hyperparams}). 
IC/BC points scale proportionally with the domain count to preserve the 1:5 or 1:10 ratio used throughout each configuration. 
We aggregate indicator values by the mean $\pm$ standard deviation over 10 seeds.

\paragraph{Shock proximity on Burgers'}
The shock-proximity indicator
$\rho^{L \to \hat{u}}(R_s, R_s)$ with $R_s = [-0.2, 0.2]$
(\cref{app:burgers_shock_locality_indicator}) rises near-monotonically with domain points
(\cref{tab:burgers-sweep}), recovering the gap between the well-
and poorly-trained configurations of \cref{tab:burgers_shock_proximity} without reference to a held-out ground truth.

\begin{table}[h]
\caption{Burgers': Shock-proximity indicator across domain-point counts
(mean $\pm$ std over 10 seeds).}
\label{tab:burgers-sweep}
\centering
\small
\begin{tabular}{lccccc}
\toprule
Domain pts & 250 & 750 & 1250 & 1750 &  2500 \\
\midrule
$\rho^{L \to \hat{u}}(R_s, R_s)$ & 0.24 $\pm$ 0.04 & 0.50 $\pm$ 0.03 & 0.60 $\pm$ 0.03 & 0.65 $\pm$ 0.03 & 0.66 $\pm$ 0.03  \\
\bottomrule
\end{tabular}
\end{table}

\paragraph{Temporal indicator across PDEs}
We additionally sweep over the temporal influence
indicator (see \cref{eq:causality}) for cross-PDE comparison over all time-dependent problems, with the results given in \cref{tab:eta-sweep}.
This indicator quantifies the extent to which predictions are influenced by training points at earlier times; low values flag over-reliance on early-time data. 
The pattern shared by Burgers', Allen-Cahn, Heat, and Drift-Diffusion is a recovery of the indicator values from depressed low-fraction values toward the well-trained reference (\cref{tab:mean_directionality}), with the transition occurring at a PDE-specific data scale that is itself diagnostic of each problem's data requirements. Wave is the exception and is discussed separately below.

\begin{table}[h]
\caption{Temporal influence indicator across PDEs and
training-data fractions (mean $\pm$ std over 10 seeds). Full sets contain
$2500$ domain points (Allen-Cahn, Burgers', Wave) or $1000$ (Heat,
Drift-Diffusion) and IC/BC points sampled at respective ratios matching \cref{tab:app_hyperparams}.}
\centering
\small
\begin{tabular}{lccccc}
\toprule
Fraction & Allen-Cahn & Burgers' & Heat & Drift-Diff. & Wave \\
\midrule
$0.1$ & $0.41 \pm 0.05$ & $0.25 \pm 0.02$ & $0.29 \pm 0.03$ & $0.28 \pm 0.09$ & $0.40 \pm 0.05$ \\
$0.3$ & $0.47 \pm 0.02$ & $0.32 \pm 0.02$ & $0.31 \pm 0.02$ & $0.45 \pm 0.08$ & $0.41 \pm 0.06$ \\
$0.5$ & $0.48 \pm 0.03$ & $0.35 \pm 0.01$ & $0.31 \pm 0.02$ & $0.52 \pm 0.03$ & $0.41 \pm 0.07$ \\
$0.7$ & $0.48 \pm 0.04$ & $0.38 \pm 0.03$ & $0.31 \pm 0.03$ & $0.50 \pm 0.02$ & $0.40 \pm 0.07$ \\
$1.0$ & $0.50 \pm 0.02$ & $0.40 \pm 0.02$ & $0.32 \pm 0.03$ & $0.47 \pm 0.02$ & $0.40 \pm 0.07$ \\
\bottomrule
\end{tabular}
\label{tab:eta-sweep}
\end{table}

\textbf{Burgers'} mirrors the shock-proximity sweep
($0.25$--$0.40$).
\textbf{Allen-Cahn} saturates already near fraction $0.3$, and even the
lowest fraction exceeds the poorly-trained
value of $0.32$ (\cref{tab:mean_directionality}), since that configuration differs in optimizer choice
rather than data quantity (\cref{tab:app_hyperparams}).
\textbf{Heat} varies only mildly across the swept range
($0.29$--$0.32$): the equation needs substantially fewer points to be
learned and only breaks down at very low data.
The poorly-trained Heat row in \cref{tab:app_hyperparams} corresponds to fraction $0.01$
with an indicator value of $0.26$ (\cref{tab:mean_directionality}), matching the depressed regime seen at low fractions in the other PDEs.
\textbf{Drift-Diffusion} is non-monotone, peaking near fraction $0.5$
before mildly declining at full data: low values at low fractions are
consistent with the information-propagation failure discussed in
\cref{sec:related_work}, while at high fractions influence distributes
uniformly across the temporal domain and saturates the indicator.
\textbf{Wave} departs from this pattern: the indicator stays essentially flat ($\approx 0.40$) across the swept range and shows none of the low-fraction depression seen elsewhere. 
It therefore does not track training quality monotonically, and the monotone recovery reading should not be carried over to it from the other PDEs.

Taken together, the two sweeps show that \PINNfluence's regional indicators respond to training quality in a PDE-dependent fashion: monotonically for Burgers', with early saturation, non-monotonicity, or (for Wave) no monotone trend in the others. Additionally, we observe that the form of this response itself encodes PDE-specific information about each problem's data requirements. 

\clearpage

\subsection{Alternative Optimizers}
\label{app:soap_nncg}

A natural concern is whether the diagnostics produced by \textsc{PINNfluence} reflect genuine properties of the converged model or are an artifact of the optimizer used to reach it. 
Because \textsc{PINNfluence} operates post-hoc on a stationary point, its outputs should depend on where training converged rather than on how it got there. 
To show this, we additionally provide evaluations of models trained with curvature-aware optimizers. 

We consider NysNewton-CG (NNCG) \cite{RathoreChallengesTrainingPINNs2024} and SOAP \cite{vyasSOAPImprovingStabilizing2025a}.
NNCG is a second-order method tailored to ill-conditioned PINN landscapes.
SOAP instead utilizes Adam \cite{kingma_adam_2017} with the Shampoo preconditioner \cite{guptaShampooPreconditionedStochastic2018}. 
Recent work \cite{wangGradientAlignmentPhysicsinformed2026} has shown that this optimizer excels at resolving gradient-conflicts between individual loss terms (PDE, IC, BCs) present in PINN training, hence achieving more stable training in comparison to Adam.

\paragraph{Setup}
For the alternative optimizers, we adopt the well-trained configurations' training points and network setups of \cref{tab:app_hyperparams}.
The SOAP experiments replace Adam+L-BFGS completely and are trained from scratch. We use $N^{(\mathrm{SOAP})}_{\mathrm{iter}} = N^{(\mathrm{Adam})}_{\mathrm{iter}}$. For the NNCG experiments we take the well-trained models and fine-tune them additionally with $1\,000$ steps of NNCG\footnote{As NNCG runs extremely slowly with default hyperparameters, Wave and Navier-Stokes use lower tolerances for the Hessian-approximation in NNCG optimization.}.

\paragraph{Results}

\begin{figure}[ht]
\centering
\begin{minipage}[t]{0.48\textwidth}
    \centering
    \includegraphics[width=0.48\textwidth]{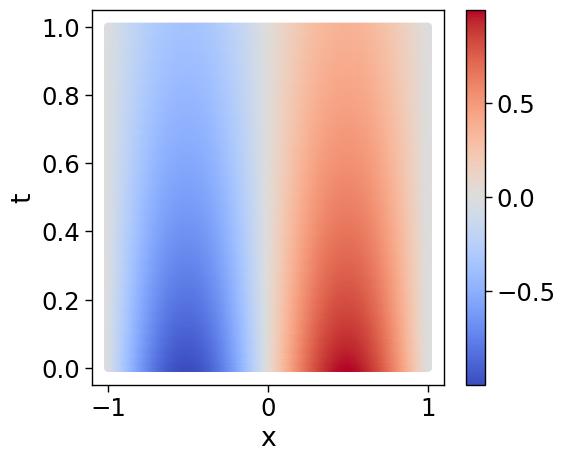}
    \includegraphics[width=0.48\textwidth]{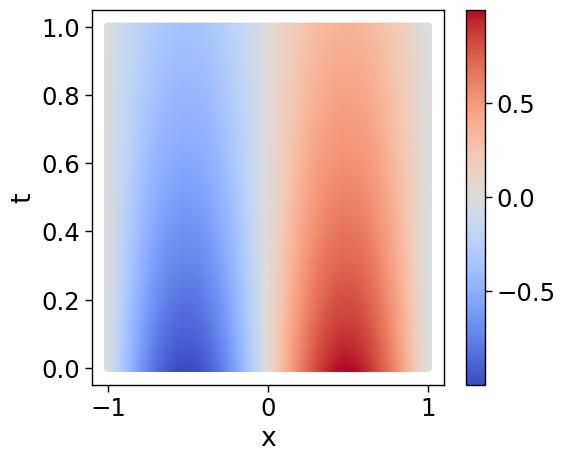}
    \subcaption{Heat}
    \label{fig:heat_preds_soap_nncg}
\end{minipage}
\hfill
\begin{minipage}[t]{0.48\textwidth}
    \centering
    \includegraphics[width=0.48\textwidth]{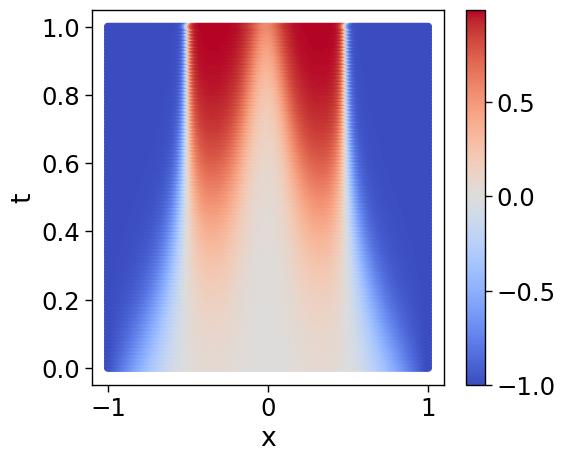}
    \includegraphics[width=0.48\textwidth]{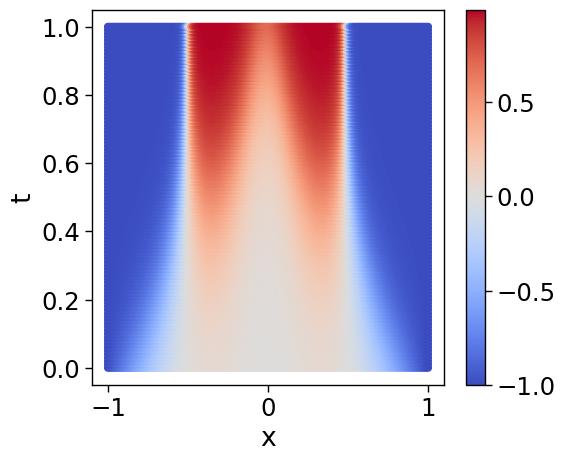}
    \subcaption{Allen-Cahn}
    \label{fig:allen_cahn_preds_soap_nncg}
\end{minipage}

\vspace{1em}

\begin{minipage}[t]{0.48\textwidth}
    \centering
    \includegraphics[width=0.48\textwidth]{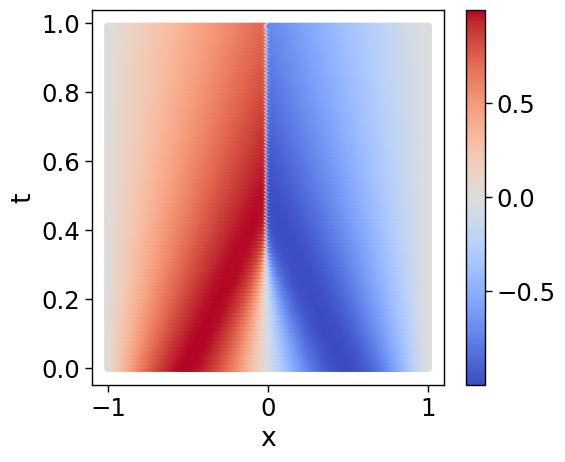}
    \includegraphics[width=0.48\textwidth]{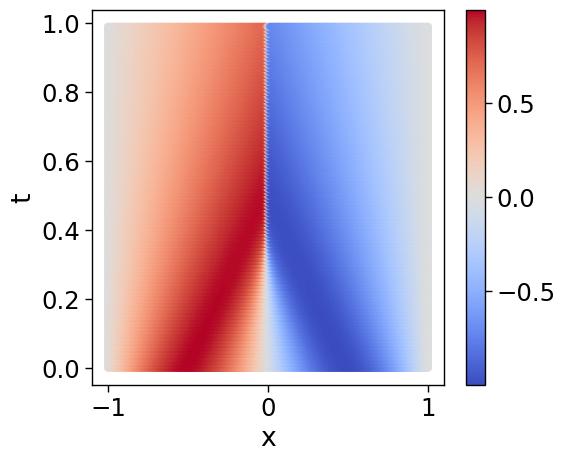}
    \subcaption{Burgers'}
    \label{fig:burgers_preds_soap_nncg}
\end{minipage}
\hfill
\begin{minipage}[t]{0.48\textwidth}
    \centering
    \includegraphics[width=0.48\textwidth]{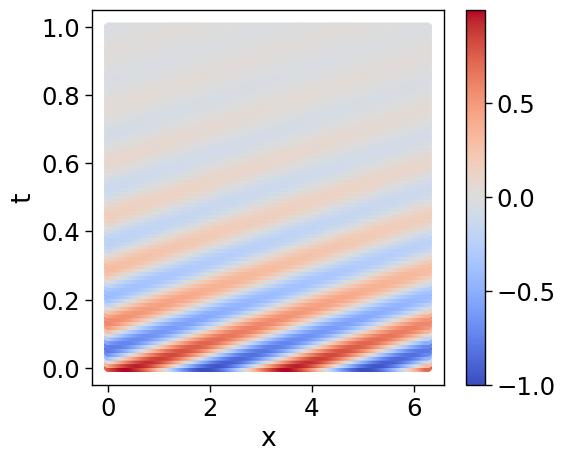}
    \includegraphics[width=0.48\textwidth]{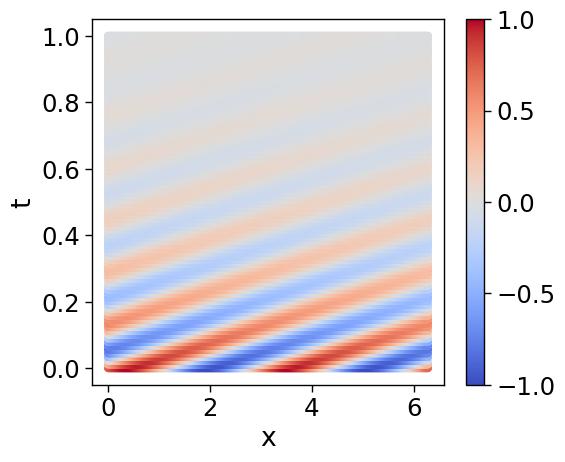}
    \subcaption{Drift-Diffusion}
    \label{fig:drift_diff_preds_soap_nncg}
\end{minipage}

\vspace{1em}

\begin{minipage}[t]{0.48\textwidth}
    \centering
    \includegraphics[width=0.48\textwidth]{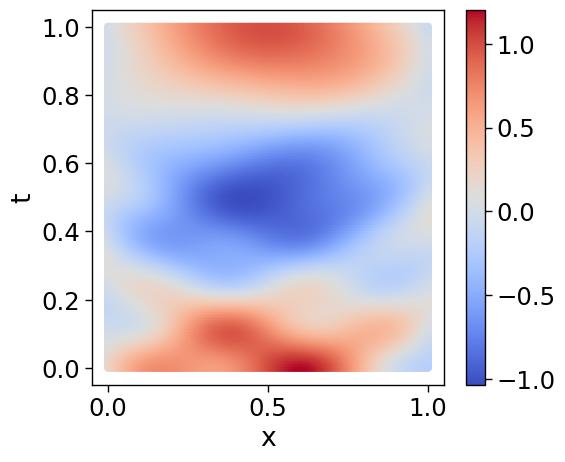}
    \includegraphics[width=0.48\textwidth]{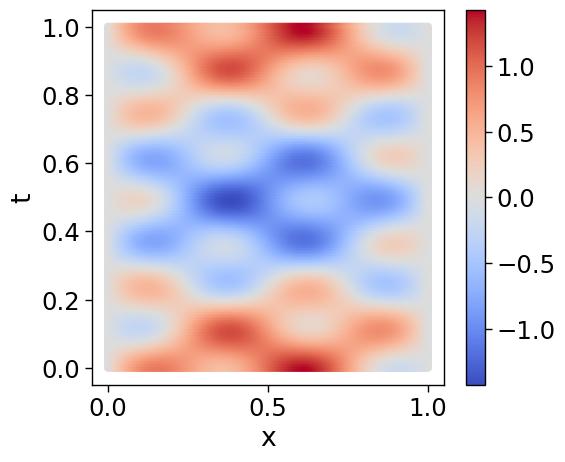}
    \subcaption{Wave}
    \label{fig:wave_preds_soap_nncg}
\end{minipage}
\hfill
\begin{minipage}[t]{0.48\textwidth}
    \centering
    \includegraphics[width=0.48\textwidth]{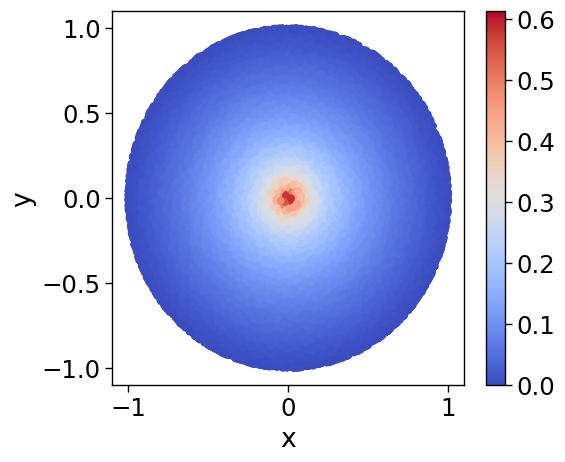}
    \includegraphics[width=0.48\textwidth]{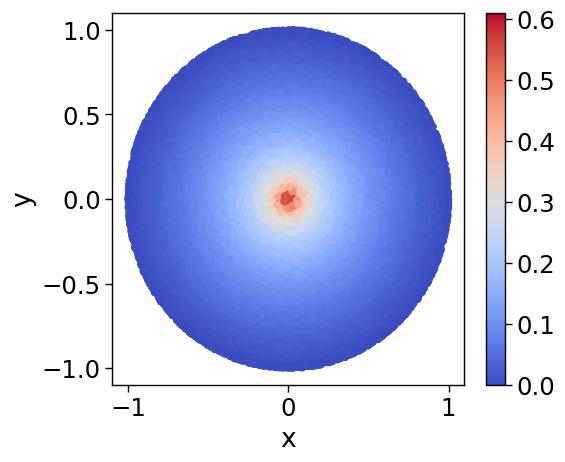}
    \subcaption{Poisson}
    \label{fig:poisson_preds_soap_nncg}
\end{minipage}

\vspace{1em}

\begin{minipage}[t]{0.8\textwidth}
    \centering
    \includegraphics[width=0.48\textwidth]{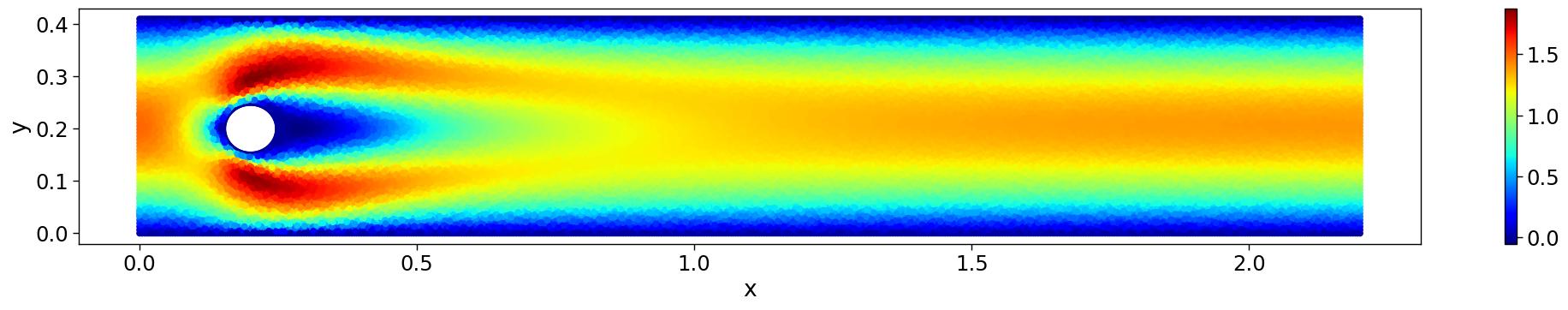}
    \includegraphics[width=0.48\textwidth]{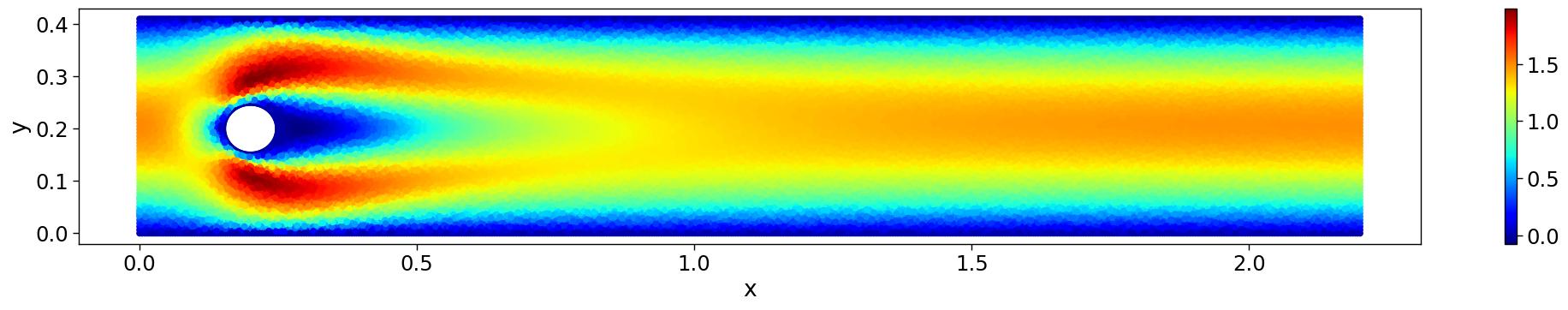}\\[0.3em]
    \includegraphics[width=0.48\textwidth]{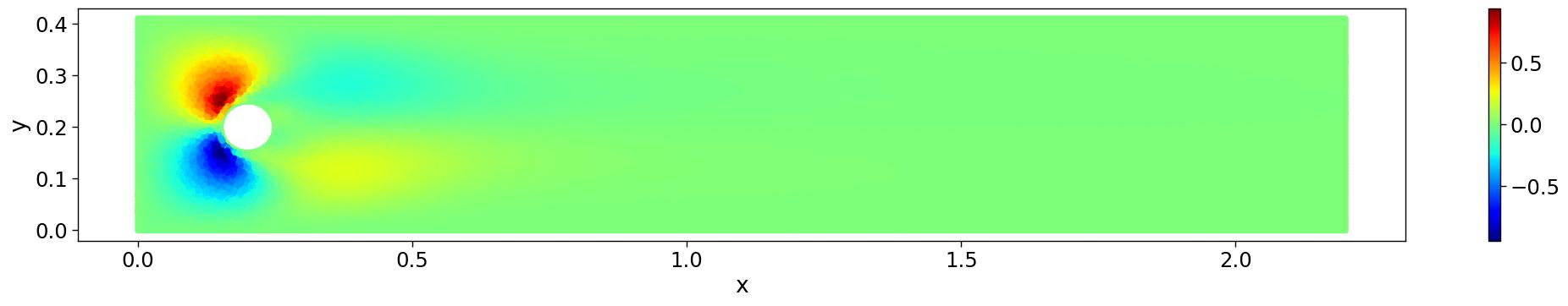}
    \includegraphics[width=0.48\textwidth]{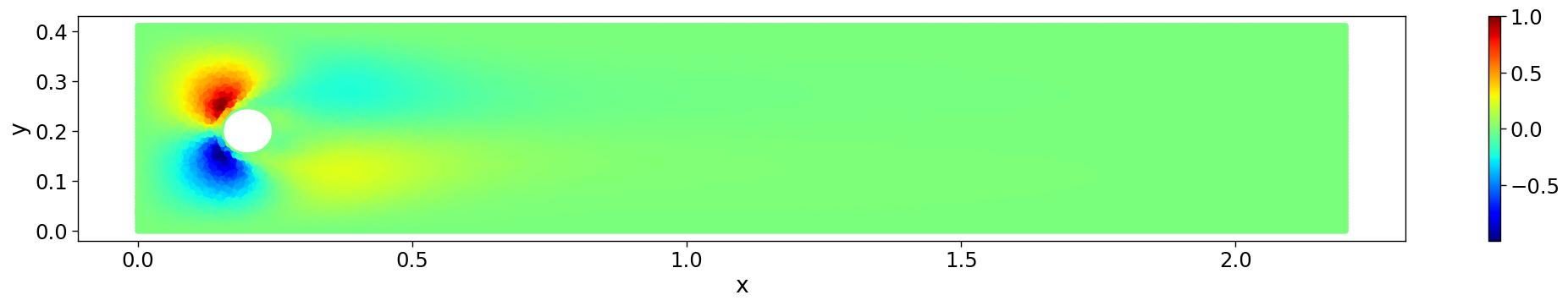}\\[0.3em]
    \includegraphics[width=0.48\textwidth]{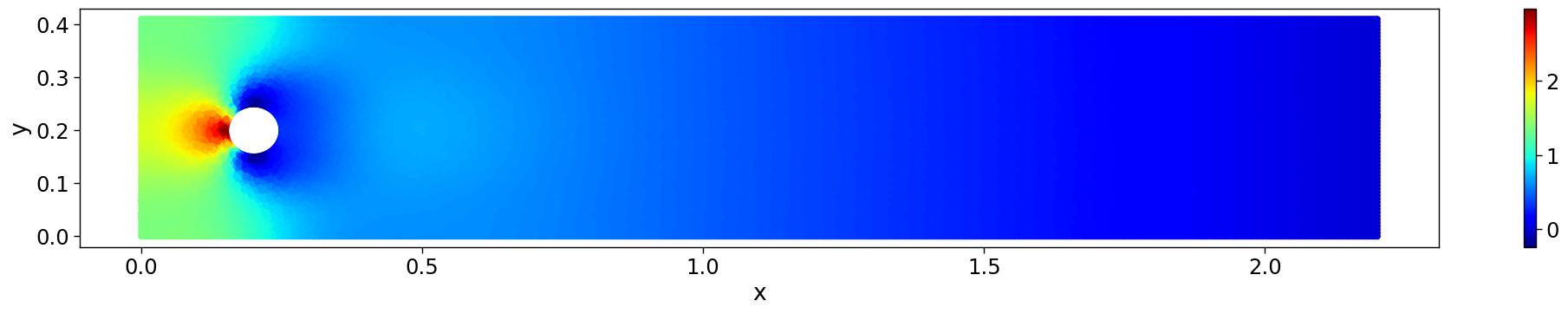}
    \includegraphics[width=0.48\textwidth]{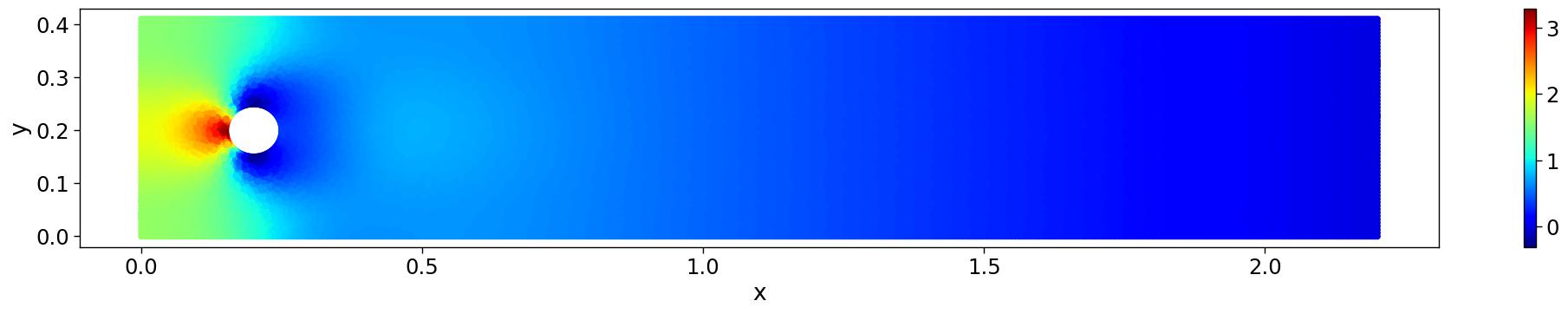}
    \subcaption{Navier-Stokes ($u$, $v$, $p$ top to bottom)}
    \label{fig:ns_preds_soap_nncg}
\end{minipage}

\caption{Predictions for SOAP (left) and NNCG (right) across all PDEs.}
\label{fig:all_soap_nncg_preds}
\end{figure}

\cref{fig:all_soap_nncg_preds} shows prediction plots across all problems for both SOAP and NNCG trainings. Almost all configurations achieve good results, the one exception being the Wave equation with the SOAP optimizer, where similar patterns arise as in the poorly-trained configuration.

\cref{tab:mean_directionality_soap_nncg} reports the resulting temporal (and spatial) influence indicator values (see \cref{tab:mean_directionality,tab:mean_directionality_steady} for well- and poorly-trained values). 
NNCG closely reproduces the well-trained Adam+L-BFGS reference across every evaluated problem. This is expected as NNCG fine-tunes already trained models.
SOAP, however, is trained from scratch and thus, we assume, converges to meaningfully different optima. It matches the well-trained configuration only for Allen-Cahn and Burgers' equations. For the Heat, Drift-Diffusion, and Poisson equations, it produces accurate solutions whose indicator values nonetheless fall well below the well-trained reference. For the Wave equation, SOAP fails to solve the problem, exhibiting both high error and low indicator values, mirroring the poorly-trained configuration.

\begin{figure}[ht]
\setlength{\fboxsep}{0pt}      %
\setlength{\fboxrule}{0.2pt}
\centering

\begin{minipage}[t]{0.48\textwidth}
    \centering
    \fbox{\includegraphics[width=\dimexpr\linewidth-2\fboxsep-2\fboxrule\relax]{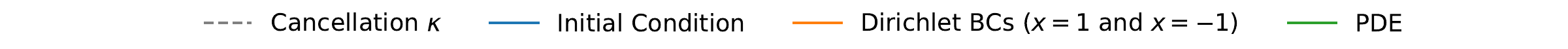}}\\[0.3ex]
    \includegraphics[width=0.48\textwidth]{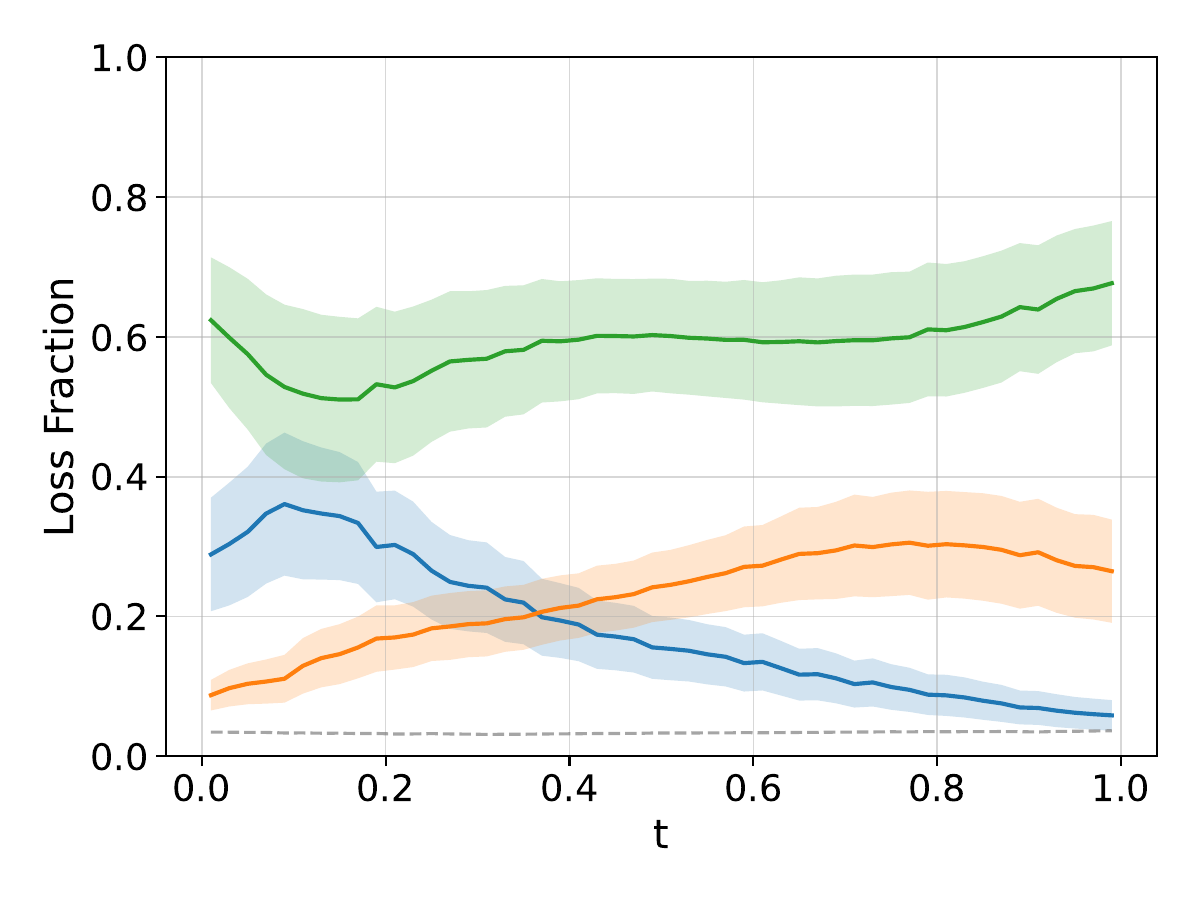}
    \includegraphics[width=0.48\textwidth]{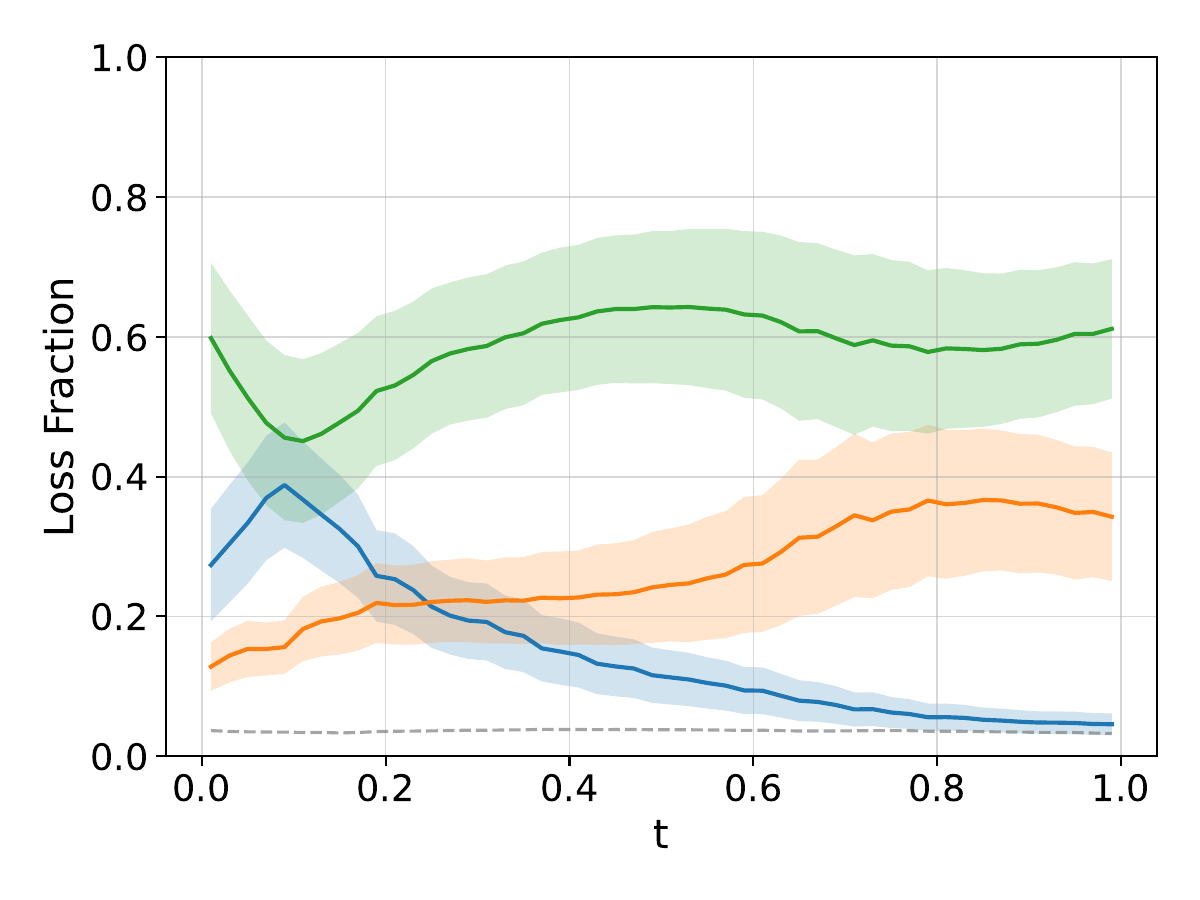}
    \subcaption{Heat}
    \label{fig:heat_loss_fracs_soap_nncg}
\end{minipage}
\hfill
\begin{minipage}[t]{0.48\textwidth}
    \centering
    \fbox{\includegraphics[width=\dimexpr\linewidth-2\fboxsep-2\fboxrule\relax]{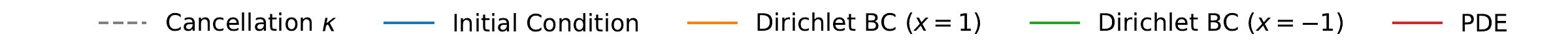}}\\[0.3ex]
    \includegraphics[width=0.48\textwidth]{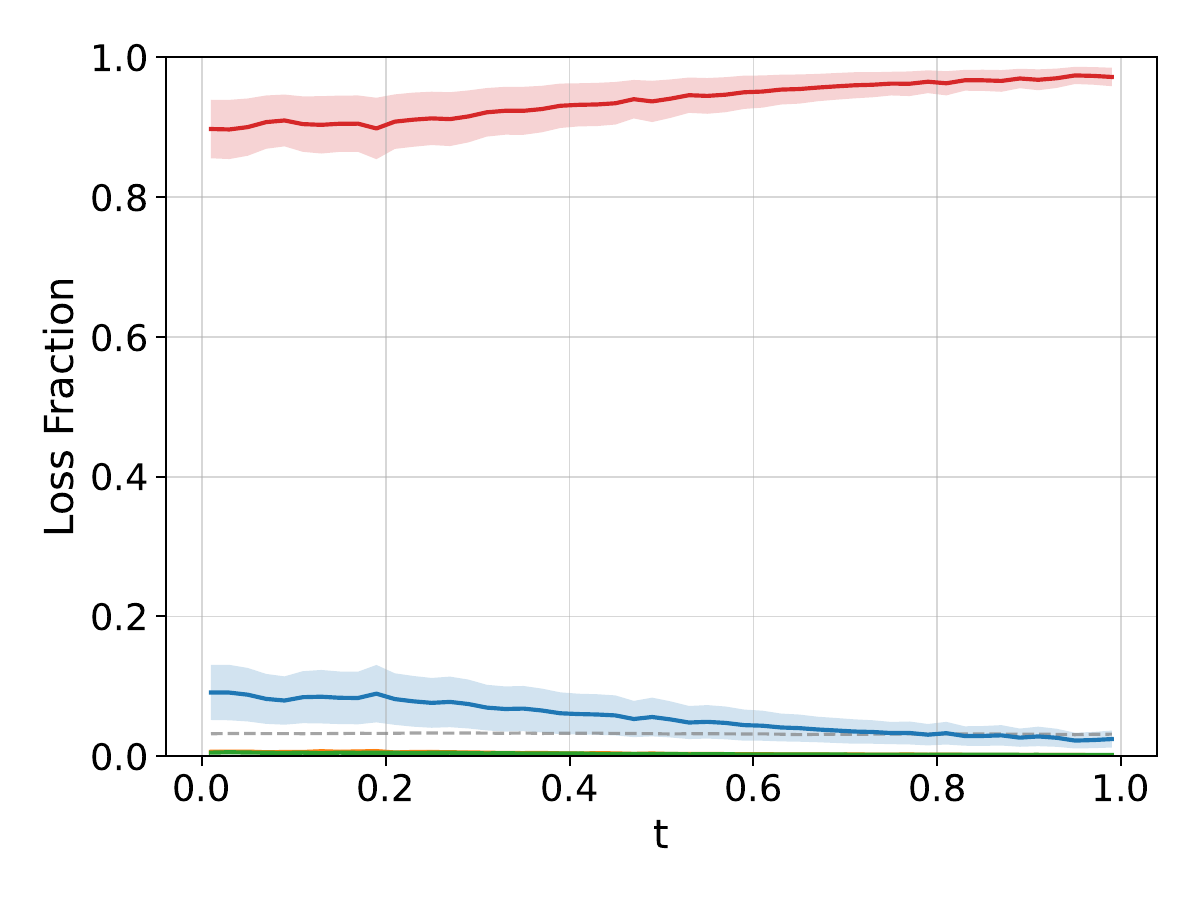}
    \includegraphics[width=0.48\textwidth]{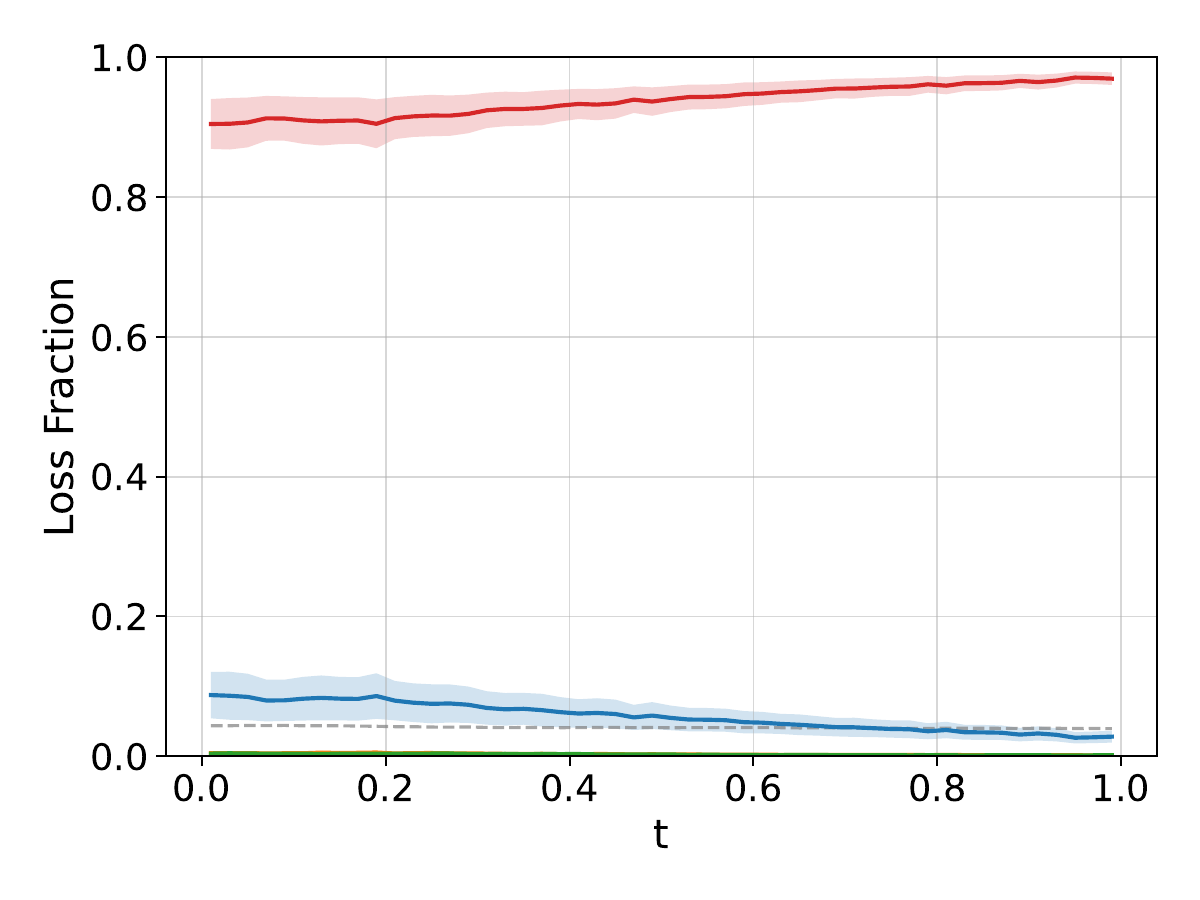}
    \subcaption{Allen-Cahn}
    \label{fig:allen_cahn_loss_fracs_soap_nncg}
\end{minipage}
\vspace{1em}

\begin{minipage}[t]{0.48\textwidth}
    \centering
    \fbox{\includegraphics[width=\dimexpr\linewidth-2\fboxsep-2\fboxrule\relax]{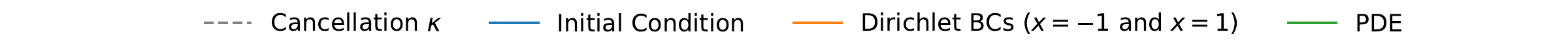}}\\[0.3ex]
    \includegraphics[width=0.48\textwidth]{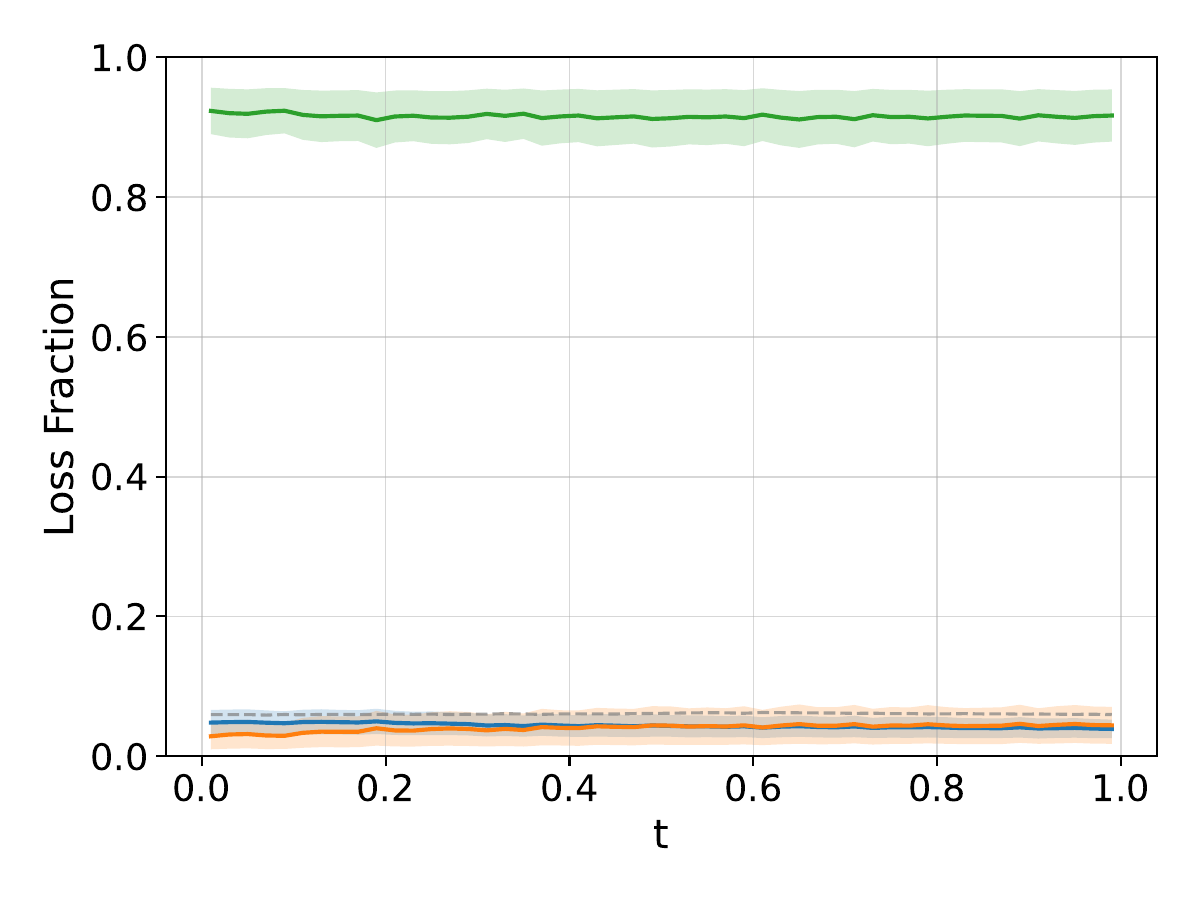}
    \includegraphics[width=0.48\textwidth]{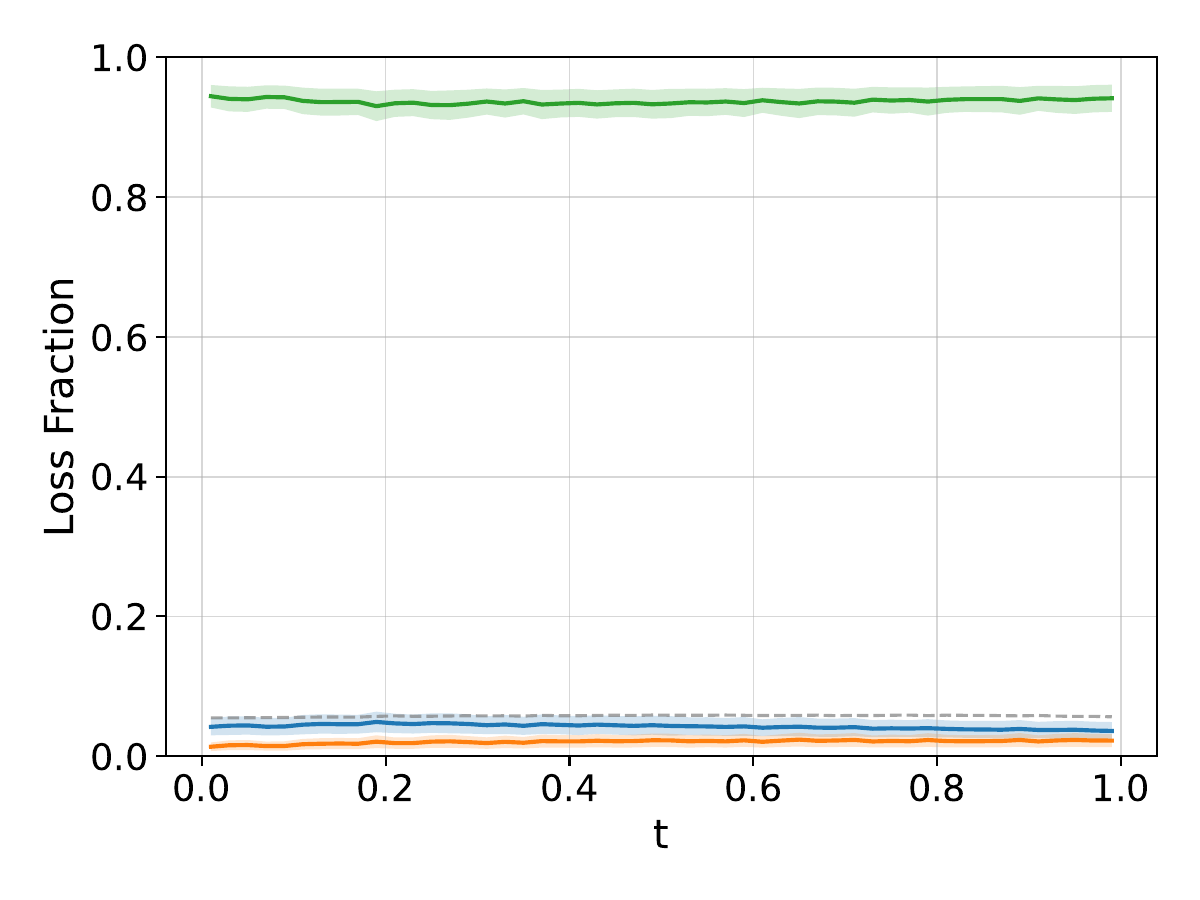}
    \subcaption{Burgers'}
    \label{fig:burgers_loss_fracs_soap_nncg}
\end{minipage}
\hfill
\begin{minipage}[t]{0.48\textwidth}
    \centering
    \fbox{\includegraphics[width=\dimexpr\linewidth-2\fboxsep-2\fboxrule\relax]{figures/icml_ratios/drift_diffusion_legend.pdf}}\\[0.3ex]
    \includegraphics[width=0.48\textwidth]{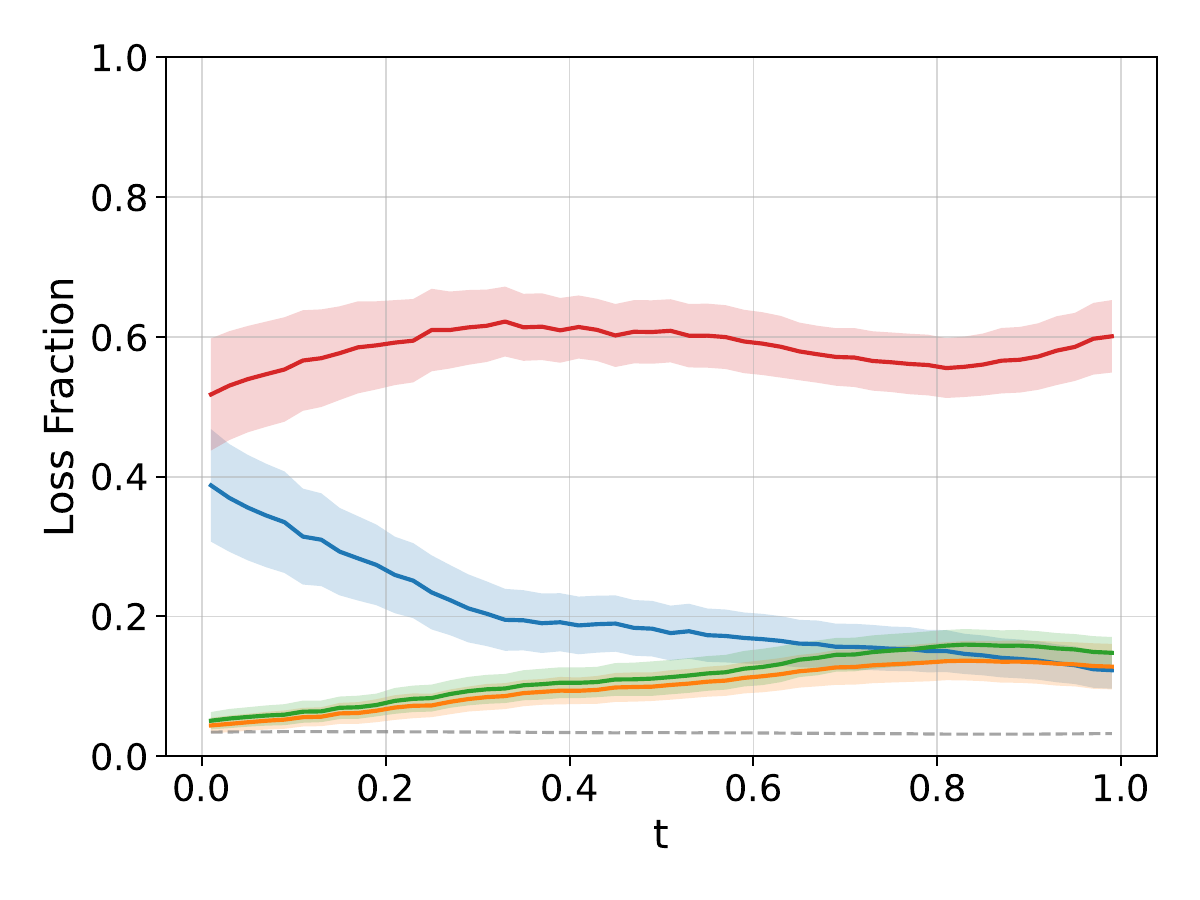}
    \includegraphics[width=0.48\textwidth]{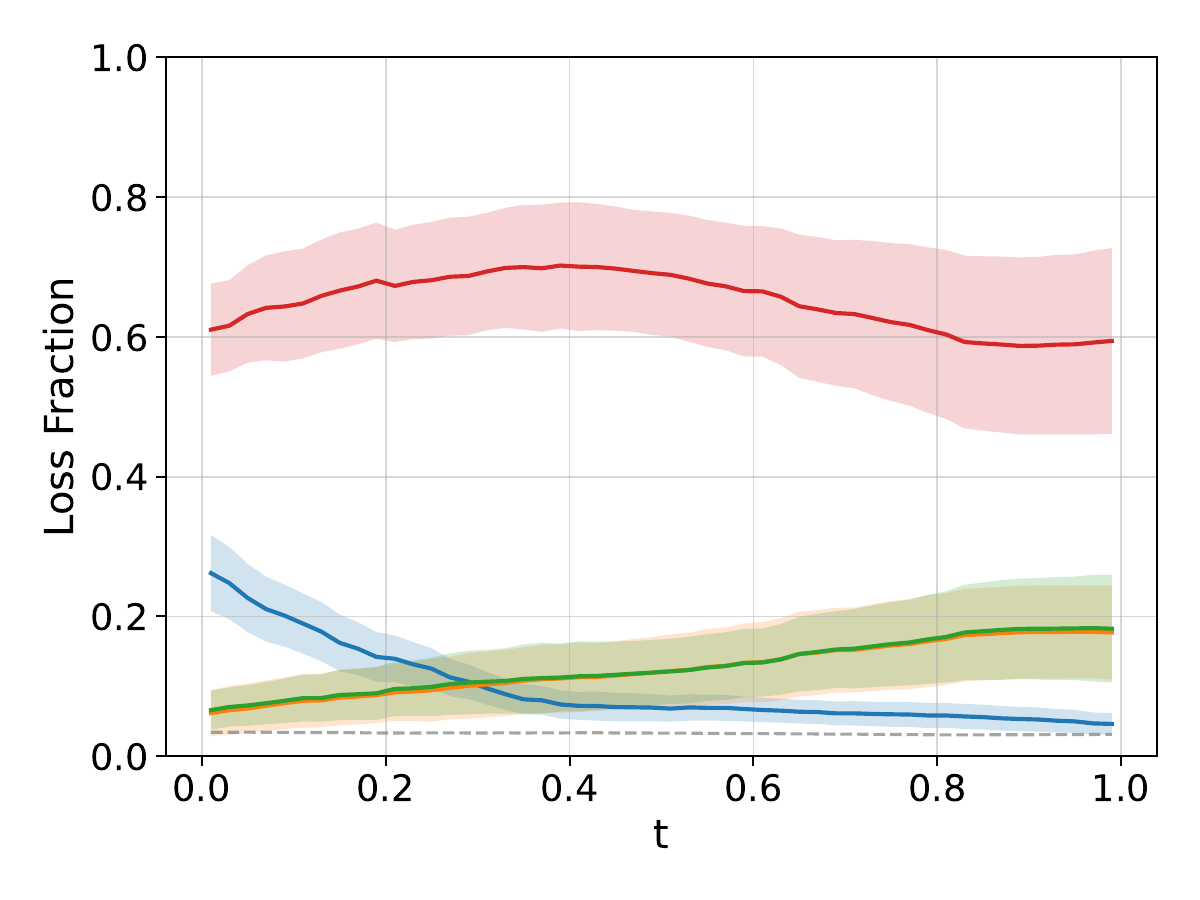}
    \subcaption{Drift-Diffusion}
    \label{fig:drift_diff_loss_fracs_soap_nncg}
\end{minipage}
\vspace{1em}

\begin{minipage}[t]{0.48\textwidth}
    \centering
\begin{tikzpicture}
    \node[inner sep=0pt] (topL)
        {\includegraphics[width=\linewidth]{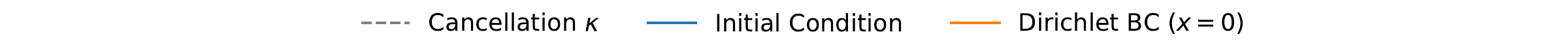}};
    \node[inner sep=0pt, anchor=north] (botL) at (topL.south)
        {\includegraphics[width=\linewidth]{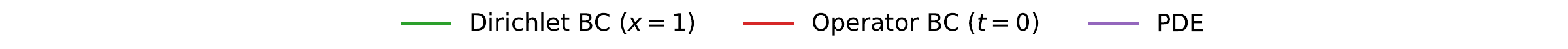}};
    \draw[line width=0.2pt] (topL.north west) rectangle (botL.south east);
\end{tikzpicture}\\[0.3ex]
    \includegraphics[width=0.48\textwidth]{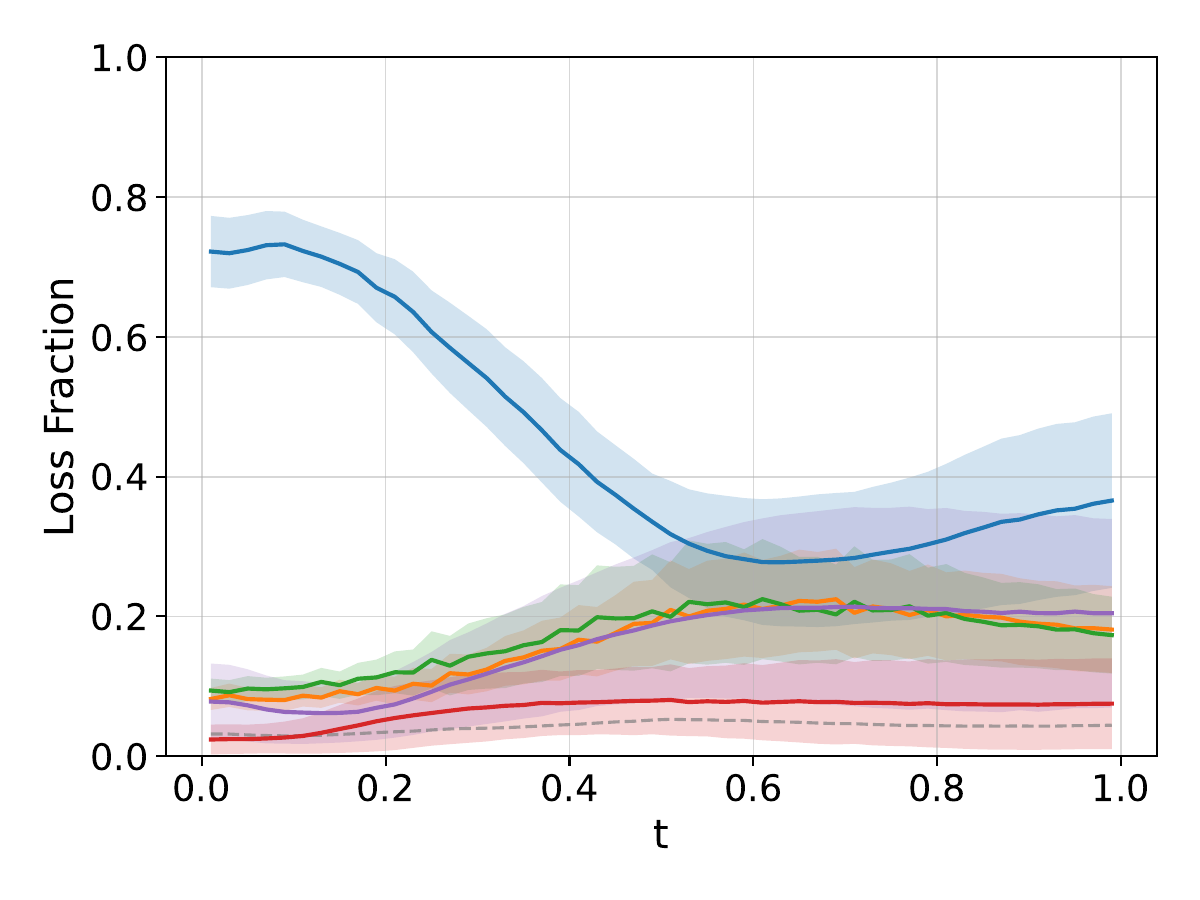}
    \includegraphics[width=0.48\textwidth]{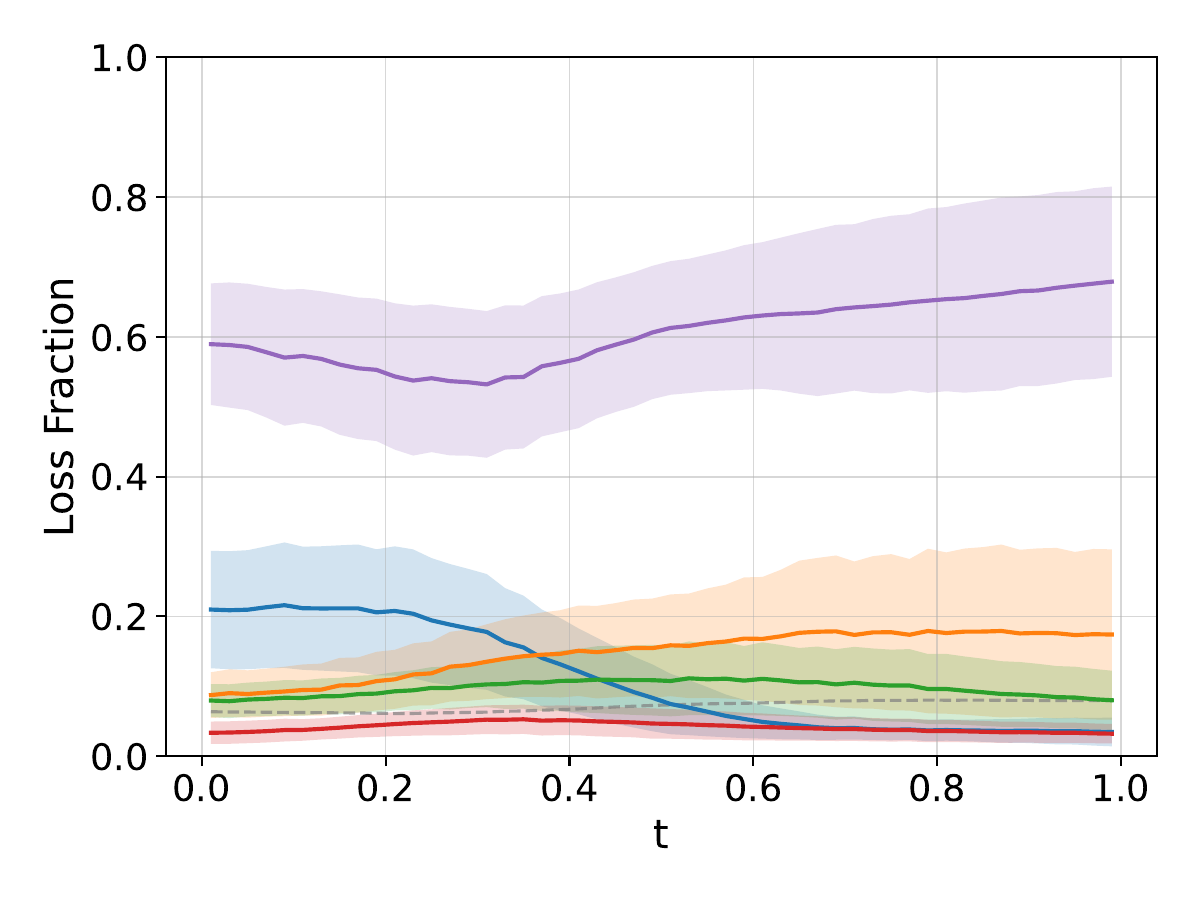}
    \subcaption{Wave}
    \label{fig:wave_loss_fracs_soap_nncg}
\end{minipage}
\hfill
\begin{minipage}[t]{0.48\textwidth}
    \centering
    \fbox{\includegraphics[width=\dimexpr\linewidth-2\fboxsep-2\fboxrule\relax]{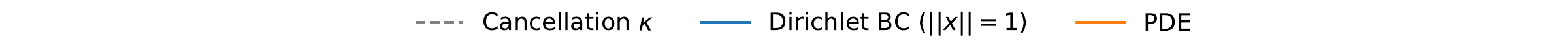}}\\[0.3ex]
    \includegraphics[width=0.48\textwidth]{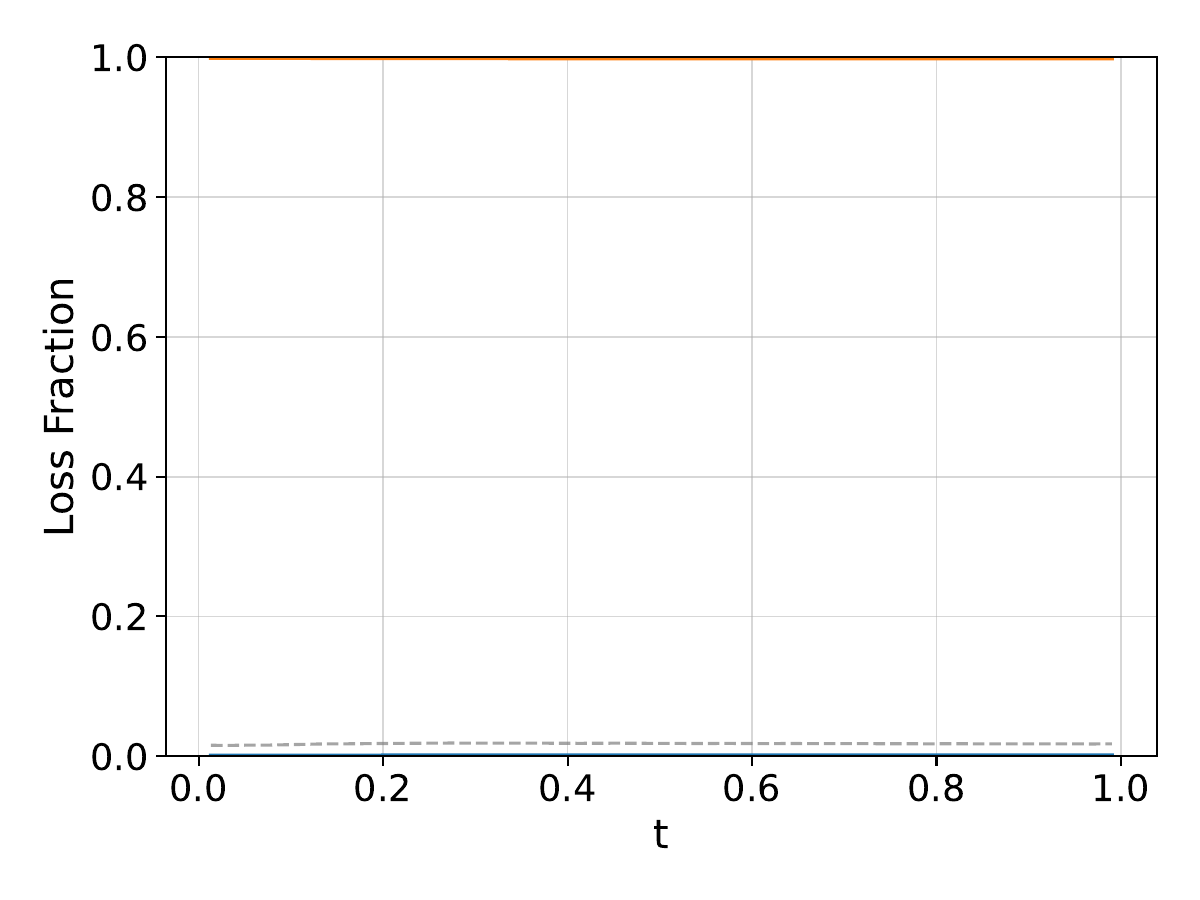}
    \includegraphics[width=0.48\textwidth]{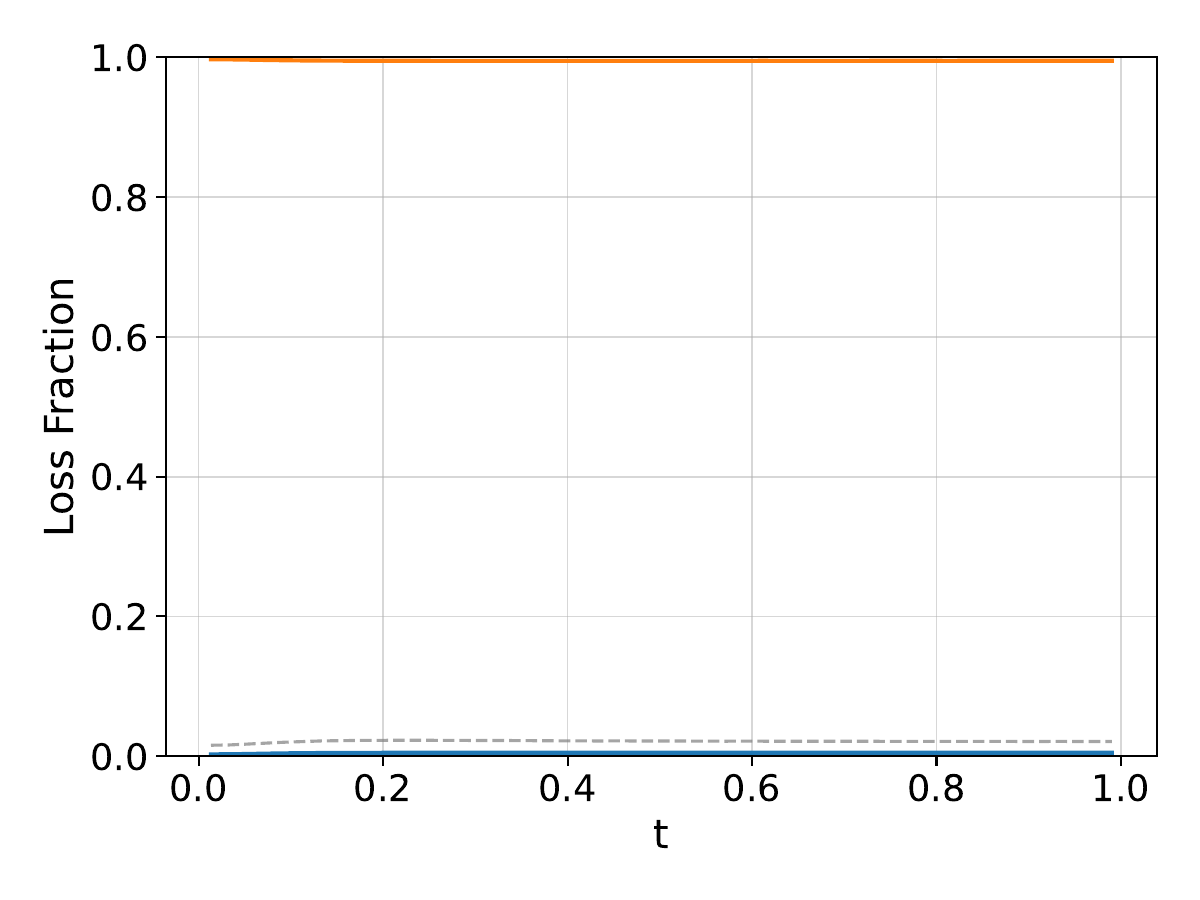}
    \subcaption{Poisson}
    \label{fig:poisson_loss_fracs_soap_nncg}
\end{minipage}
\vspace{1em}

\begin{minipage}[t]{0.48\textwidth}
    \centering
\begin{tikzpicture}
    \node[inner sep=0pt] (topN)
        {\includegraphics[width=\linewidth]{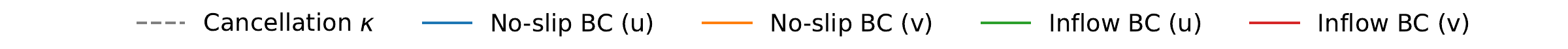}};
    \node[inner sep=0pt, anchor=north] (botN) at (topN.south)
        {\includegraphics[width=\linewidth]{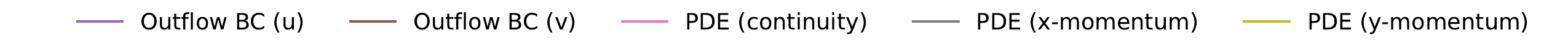}};
    \draw[line width=0.2pt] (topN.north west) rectangle (botN.south east);
\end{tikzpicture}\\[0.3ex]
    \includegraphics[width=0.48\textwidth]{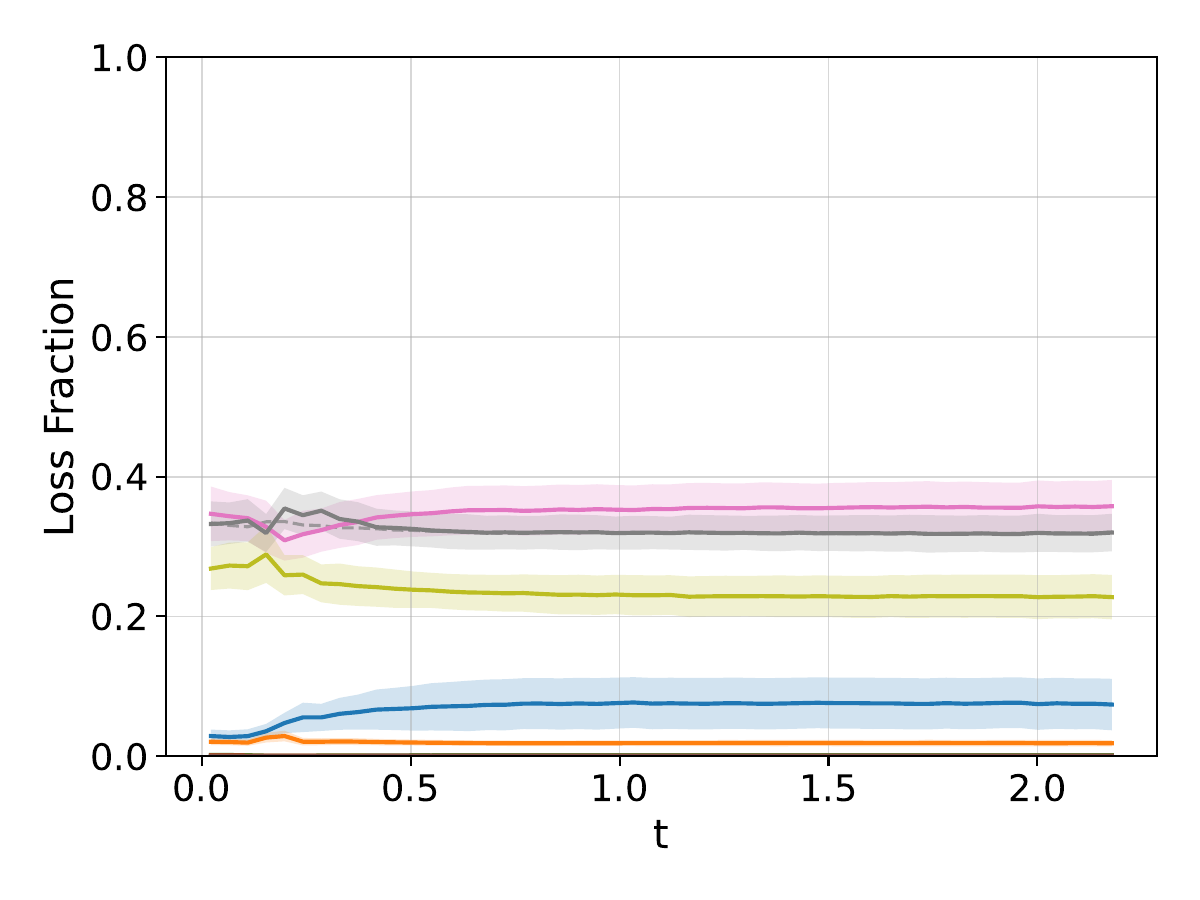}
    \includegraphics[width=0.48\textwidth]{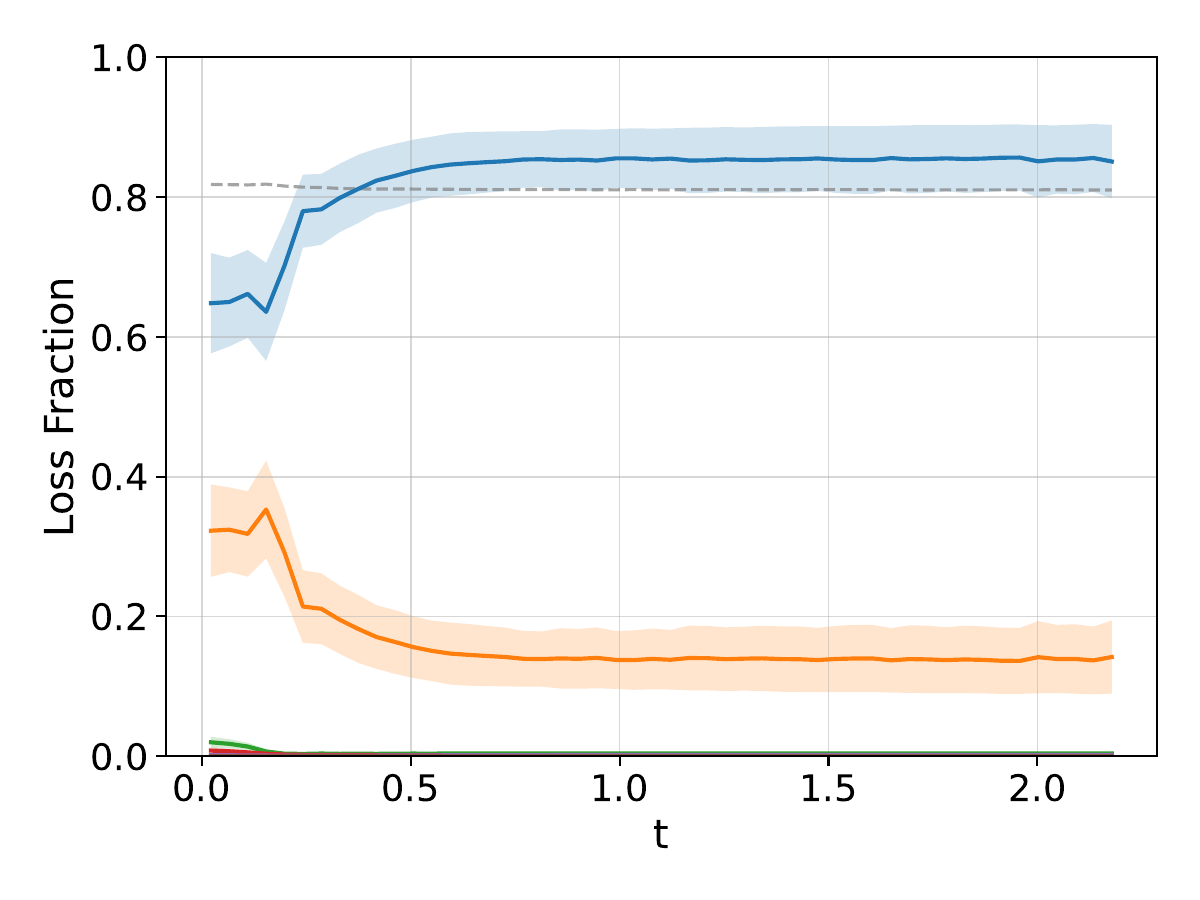}
    \subcaption{Navier-Stokes}
    \label{fig:ns_loss_fracs_soap_nncg}
\end{minipage}

\caption{Loss fractions $\bar r_{L_i}(\mathcal{X}, R_{\text{test}})$ for the curvature-aware optimizers SOAP (left) and NNCG (right) across all PDEs. Influences are binned along the relevant coordinate (time for the time-dependent PDEs, distance from the center for Poisson, and the $x$-coordinate for Navier-Stokes) and averaged over seeds; shaded regions indicate the standard deviation.}
\label{fig:all_soap_nncg_loss_fracs}
\end{figure}

These accurate-but-low-indicator SOAP solutions isolate what the indicator captures: the data dependence of the converged model,
not the accuracy of its solution. The Drift-Diffusion case illustrates this. SOAP reaches an accurate solution whose influence profile
nonetheless leans more heavily on the initial condition than the Adam+L-BFGS well-trained
model (\cref{fig:drift_diff_loss_fracs_soap_nncg}), plausibly reflecting SOAP's IC-weighted
resolution of inter-component gradient conflicts \cite{wangGradientAlignmentPhysicsinformed2026}. This underscores how \PINNfluence characterizes the converged model: the indicator describes the relationship of model behavior with respect to its data and does not constitute an optimizer-independent measure of solution quality.

Finally, for Navier-Stokes the loss-component cancellations are large under both optimizers
(\cref{fig:ns_loss_fracs_soap_nncg}), so the corresponding fractions should be read with care.

The Hessian condition numbers (\cref{tab:hessian_condition_numbers_soap_nncg}) and the
inverse-Hessian gradient cosine similarities (\cref{tab:hessian_gradient_recon_soap_nncg}) are
comparable to those of the Adam+L-BFGS runs, indicating that the approximation underlying
\PINNfluence is equally reliable for these optimizers.

\begin{table}[t]
\caption{Temporal (and spatial, for the steady-state problems) influence indicator
for models trained with the SOAP and NNCG optimizers, computed with respect to
predictions $f=\hat u$, alongside the relative $L_2$ error of each configuration.
For the steady-state Poisson and Navier-Stokes problems the indicator and baseline
are spatial rather than temporal (cf.\ \cref{tab:mean_directionality_steady}); the
baseline column reports $\bar t$ for the time-dependent problems and $\bar x$ for
the steady-state ones. Well- and poorly-trained reference values are given in
\cref{tab:mean_directionality,tab:mean_directionality_steady}, and the corresponding
reference $L_2$ errors in \cref{tab:l2_errors_well_poorly_trained}. We report
mean $\pm$ standard deviation over 10 seeds.}
\label{tab:mean_directionality_soap_nncg}
\centering
\begin{tabular}{lccccc}
\toprule
 & & \multicolumn{2}{c}{SOAP} & \multicolumn{2}{c}{NNCG} \\
Problem & Baseline & Indicator & $L_2$ Rel.\ Error & Indicator & $L_2$ Rel.\ Error \\
\midrule
Heat & $0.46$ & $0.25 \pm 0.02$ & $(1.4 \pm 0.7) \times 10^{-3}$ & $0.32 \pm 0.02$ & $(2.6 \pm 1.0) \times 10^{-4}$\\
Allen-Cahn & $0.43$ & $0.52 \pm 0.05$ & $(3.3 \pm 0.3) \times 10^{-2}$ & $0.50 \pm 0.02$ & $(6.2 \pm 3.3) \times 10^{-2}$ \\
Burgers' & $0.43$ & $0.45 \pm 0.04$ & $(1.7 \pm 1.1) \times 10^{-2}$ & $0.40 \pm 0.02$ & $(5.5 \pm 3.3) \times 10^{-3}$ \\
Drift-Diffusion & $0.46$ & $0.25 \pm 0.04$ & $(2.5 \pm 1.1) \times 10^{-3}$ & $0.47 \pm 0.04$ & $(5.6 \pm 6.8) \times 10^{-3}$ \\
Wave & $0.43$ & $0.16 \pm 0.02$ & $(3.8 \pm 0.1) \times 10^{-1}$ & $0.45 \pm 0.03$ & $(1.3 \pm 0.5) \times 10^{-2}$ \\
Poisson & $0.28$ & $0.16 \pm 0.02$ & $(3.1 \pm 0.4) \times 10^{-2}$ & $0.29 \pm 0.03$ & $(3.0 \pm 0.2) \times 10^{-2}$ \\
Navier-Stokes ($x$-Velocity) & $0.48$ & $0.16 \pm 0.03$ & $(6.2 \pm 0.6) \times 10^{-2}$ & $0.15 \pm 0.01$ & $(2.7 \pm 1.3) \times 10^{-3}$ \\
\bottomrule
\end{tabular}
\end{table}

\begin{table}[]
    \caption{Mean condition numbers of the Hessian $\mathcal{H}_{\theta_0}$ at three stages of the inverse approximation: the exact (dense) Hessian, the damped Hessian $\mathcal{H}_{\theta_0}+\lambda I$ with $\lambda=10^{-3}$, and the top-$k$ subspace retained by the Arnoldi iteration. Wave is omitted due to computational costs.}
    \label{tab:hessian_condition_numbers_soap_nncg}
    \centering
    \begin{tabular}{lcccccc}
    \toprule 
     & \multicolumn{2}{c}{Exact $\mathcal{H}_{\theta_0}$} & \multicolumn{2}{c}{Damped $\mathcal{H}_{\theta_0}+\lambda I$} & \multicolumn{2}{c}{Arnoldi top-$k$} \\
Problem & SOAP & NNCG & SOAP & NNCG & SOAP & NNCG \\
\midrule
Heat & $3.1\times 10^{21}$ & $6.4\times 10^{20}$ & $9.8\times 10^{10}$ & $1.4\times 10^{11}$ & $8.3\times 10^{4}$ & $8.8\times 10^{5}$ \\
Allen-Cahn & $2.5\times 10^{24}$ & $1.7\times 10^{25}$ & $1.4\times 10^{13}$ & $2.8\times 10^{13}$ & $4.6\times 10^{3}$ & $2.1\times 10^{4}$ \\
Burgers' & $1.4\times 10^{22}$ & $4.4\times 10^{21}$ & $5.6\times 10^{11}$ & $8.0\times 10^{11}$ & $2.2\times 10^{3}$ & $5.1\times 10^{3}$ \\
Drift-Diffusion & $8.8\times 10^{23}$ & $2.8\times 10^{23}$ & $3.0\times 10^{11}$ & $4.8\times 10^{12}$ & $1.2\times 10^{3}$ & $1.2\times 10^{3}$ \\
Poisson & $7.7\times 10^{20}$ & $1.9\times 10^{21}$ & $4.5\times 10^{12}$ & $4.2\times 10^{13}$ & $2.0\times 10^{2}$ & $3.1\times 10^{2}$ \\
Navier-Stokes & $3.3\times 10^{22}$ & $3.0\times 10^{27}$ & $4.4\times 10^{13}$ & $6.5\times 10^{13}$ & $3.5\times 10^{2}$ & $5.7\times 10^{2}$ \\
    \bottomrule 
    \end{tabular}
\end{table}

\begin{table}[]
    \caption{Cosine similarities between reconstructed $\mathcal{H}_{\theta_0}RR^\top g$ and true gradients $g$ averaged over the whole training set $\mathcal{X}_{\text{train}}$ and over 10 runs (mean $\pm$ std). NNCG Wave is omitted due to computational costs.}
    \label{tab:hessian_gradient_recon_soap_nncg}
    \centering
\begin{threeparttable}
\begin{tabular}{lcccc}
\toprule
 & \multicolumn{2}{c}{$\nabla_\theta L$} & \multicolumn{2}{c}{$\nabla_\theta f$} \\
 & SOAP & NNCG & SOAP & NNCG \\
Problem &  &  &  &  \\
\midrule
Heat & $1.00 \pm 0.00$ & $1.00 \pm 0.00$ & $1.00 \pm 0.00$ & $1.00 \pm 0.00$ \\
Allen-Cahn & $1.00 \pm 0.00$ & $1.00 \pm 0.00$ & $0.97 \pm 0.05$ & $0.99 \pm 0.03$ \\
Burgers' & $0.99 \pm 0.01$ & $0.99 \pm 0.01$ & $0.99 \pm 0.00$ & $1.00 \pm 0.01$ \\
Drift-Diffusion & $0.99 \pm 0.01$ & $1.00 \pm 0.01$ & $0.98 \pm 0.02$ & $0.99 \pm 0.01$ \\
Wave & $1.00 \pm 0.00$ & --- & $0.97 \pm 0.02$ & --- \\
Poisson & $0.95 \pm 0.05$ & $0.96 \pm 0.04$ & $0.92 \pm 0.01$ & $0.93 \pm 0.01$ \\
Navier-Stokes\tnote{(*)} & $0.92 \pm 0.08$ & $0.86 \pm 0.14$ & $0.87 \pm 0.06$ & $0.88 \pm 0.06$ \\
\bottomrule
\end{tabular}
\begin{tablenotes}
\item[(*)] \small For Navier-Stokes $f$ is $u$, i.e., the predicted velocity in $x$ direction.
\end{tablenotes}
\end{threeparttable}
\end{table}

\clearpage

\subsection{Decompositions by Loss-Components for other Problems}
\label{app:decomposition_plots}

Here we present the loss decompositions for all problems not featured in the main text.

\newcommand{\appratioslength}{0.4\textwidth}
\begin{figure}[ht]
\setlength{\fboxsep}{0pt}      %
\setlength{\fboxrule}{0.2pt}   %
\centering
\hspace*{3.5em}
\begin{subfigure}[t]{0.8\textwidth}
    \fbox{\includegraphics[width=\dimexpr\linewidth-2\fboxsep-2\fboxrule\relax]{figures/icml_ratios/diffusion_legend.pdf}}
\end{subfigure}
\\[-0.2ex]
\hfill 
\begin{subfigure}[t]{\appratioslength}
    \includegraphics[width=\textwidth]{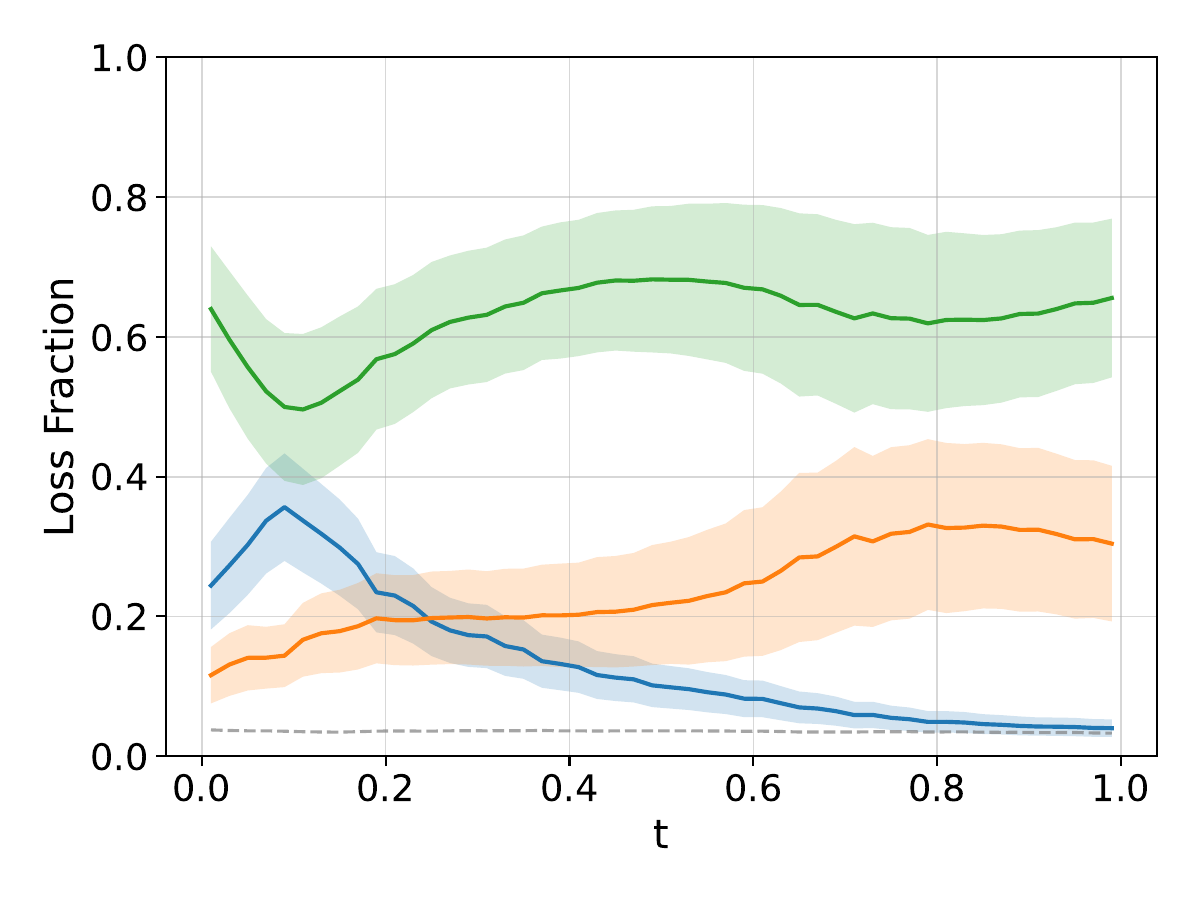}
    \caption{Well-trained}
\end{subfigure}
\hfill 
\begin{subfigure}[t]{\appratioslength}
    \includegraphics[width=\textwidth]{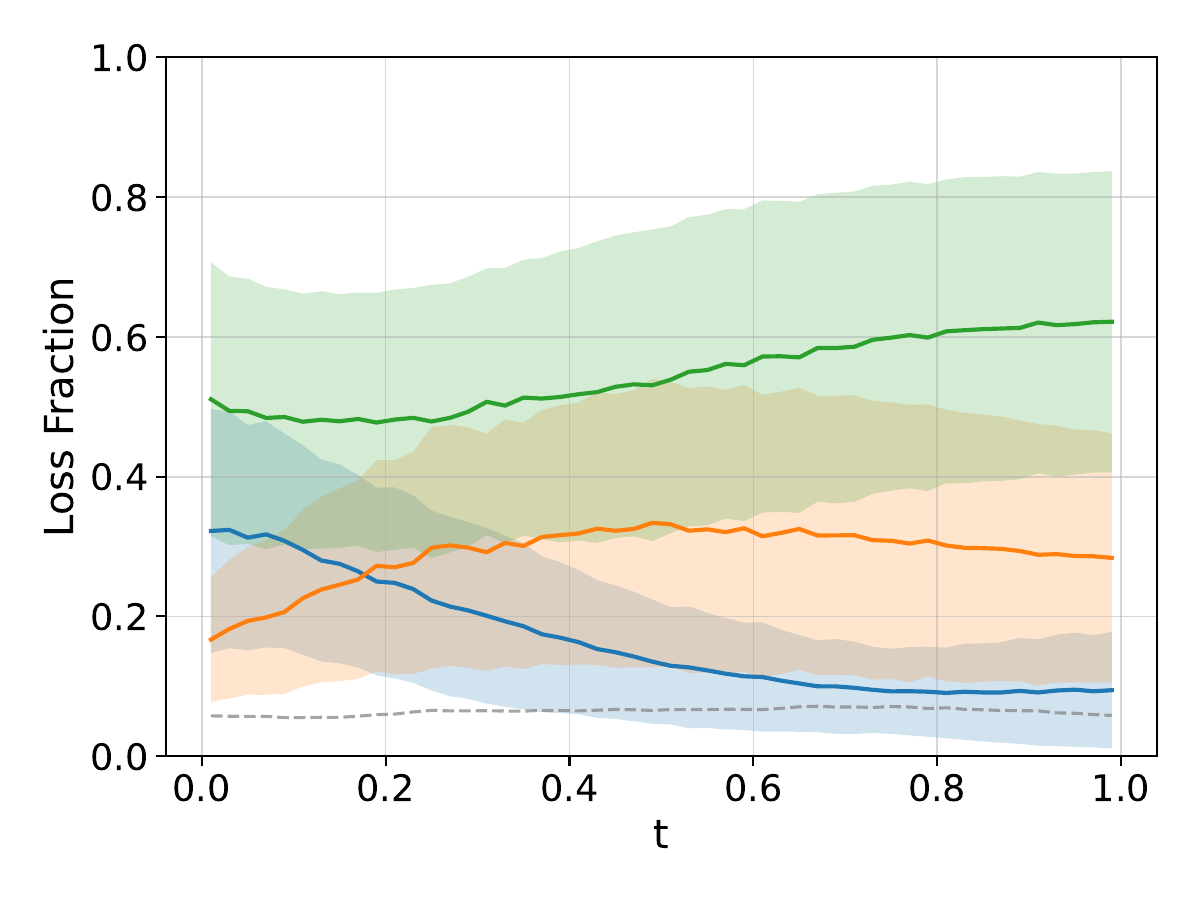}
    \caption{Poorly-trained}
\end{subfigure}
\hfill 
\caption{Heat equation: Fractions of influences $\bar r_{L_i}(\mathcal{X}, R_{\text{test}})$ for 50 time bins, averaged over seeds. Shaded regions indicate the standard deviation.}
\label{fig:app_diffusion_loss_ratios}
\end{figure}

\begin{figure}[ht]
\setlength{\fboxsep}{0pt}      %
\setlength{\fboxrule}{0.2pt}   %
\centering
\hspace*{3.5em}
\begin{subfigure}[t]{0.8\textwidth}
    \fbox{\includegraphics[width=\dimexpr\linewidth-2\fboxsep-2\fboxrule\relax]{figures/icml_ratios/allen_cahn_legend.pdf}}
\end{subfigure}
\\[-0.2ex]
\hfill 
\begin{subfigure}[t]{\appratioslength}
    \includegraphics[width=\textwidth]{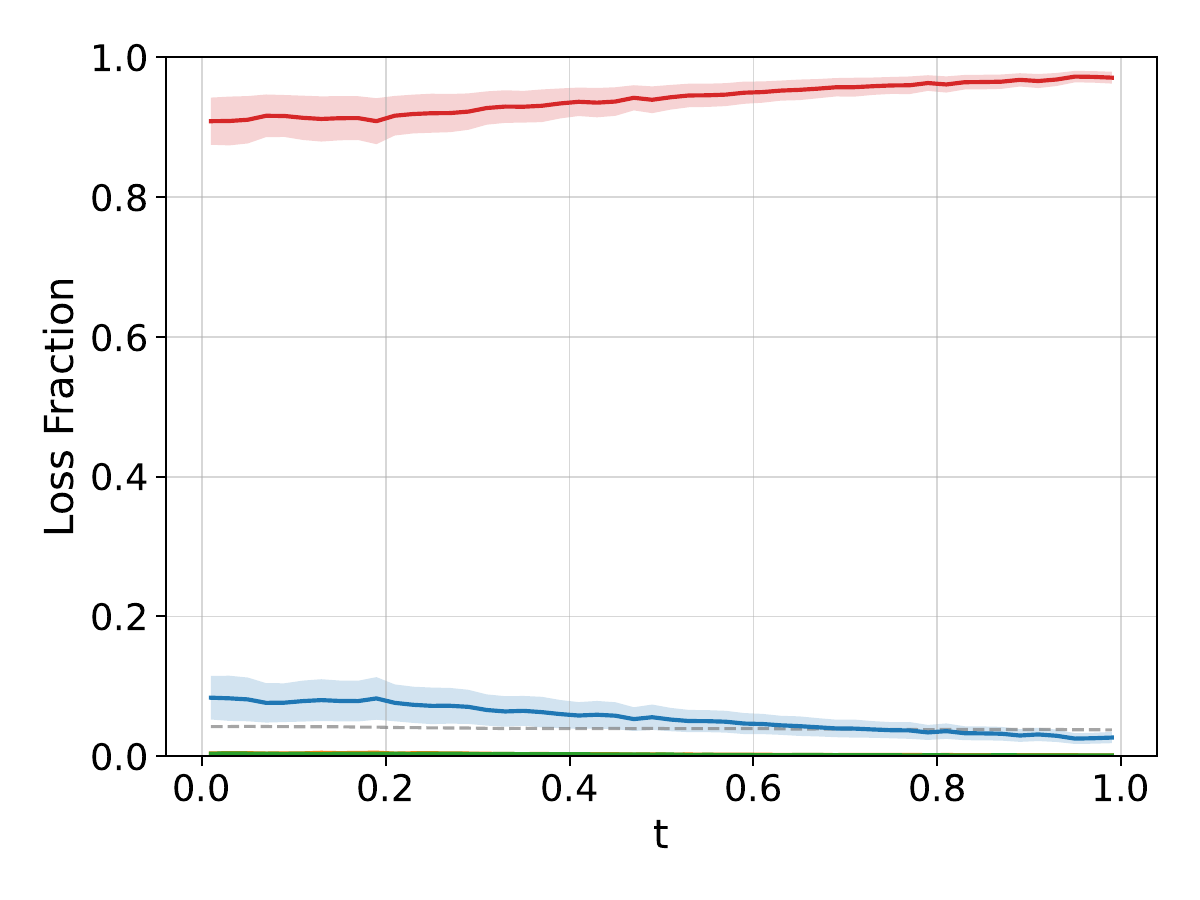}
    \caption{Well-trained}
\end{subfigure}
\hfill 
\begin{subfigure}[t]{\appratioslength}
    \includegraphics[width=\textwidth]{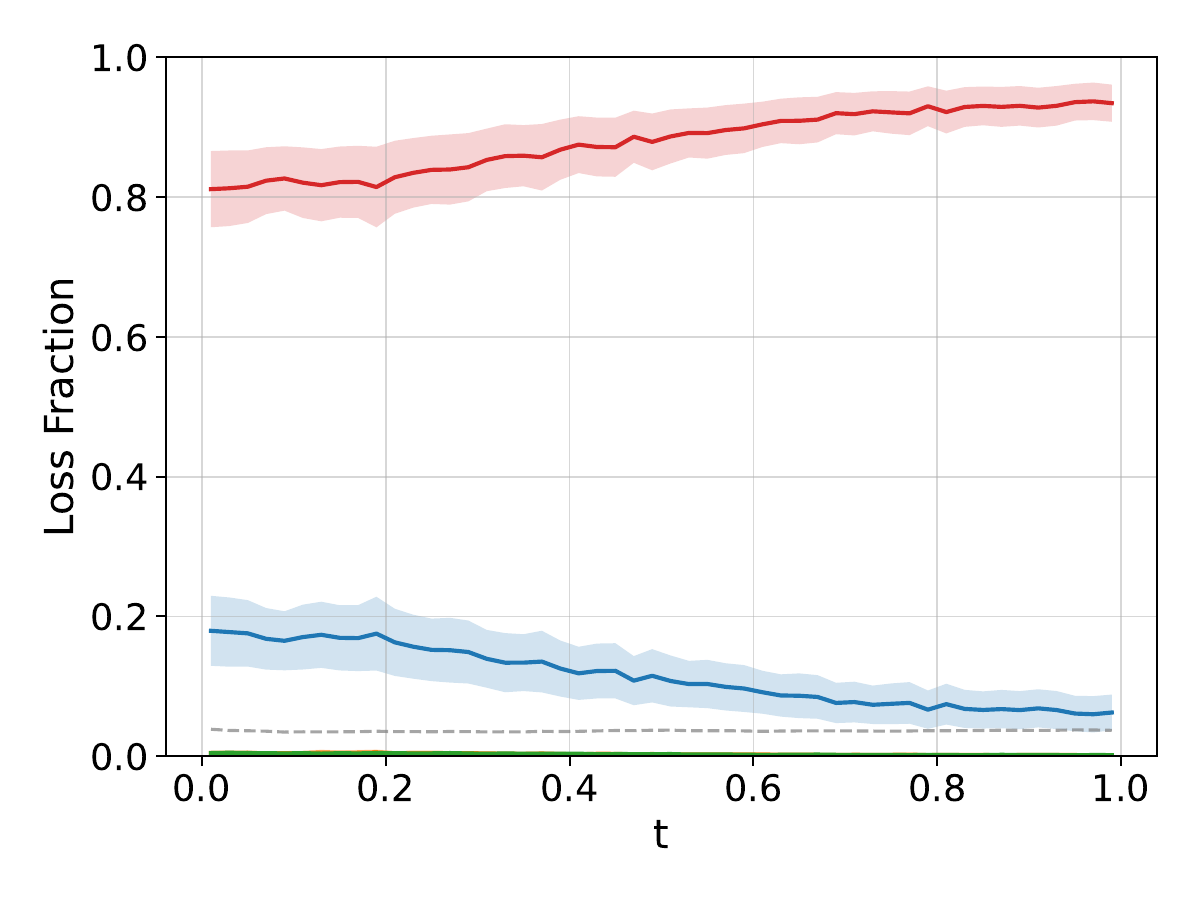}
    \caption{Poorly-trained}
\end{subfigure}
\hfill 
\caption{Allen-Cahn equation: Fractions of influences $\bar r_{L_i}(\mathcal{X}, R_{\text{test}})$ for 50 time bins, averaged over seeds. Shaded regions indicate the standard deviation.}
\label{fig:app_allen_cahn_loss_ratios}
\end{figure}

\begin{figure}[ht]
\setlength{\fboxsep}{0pt}      %
\setlength{\fboxrule}{0.2pt}   %
\centering
\hspace*{3.5em}
\begin{subfigure}[t]{0.8\textwidth}
    \fbox{\includegraphics[width=\dimexpr\linewidth-2\fboxsep-2\fboxrule\relax]{figures/icml_ratios/burgers_legend.pdf}}
\end{subfigure}
\\[-0.2ex]
\hfill 
\begin{subfigure}[t]{\appratioslength}
    \includegraphics[width=\textwidth]{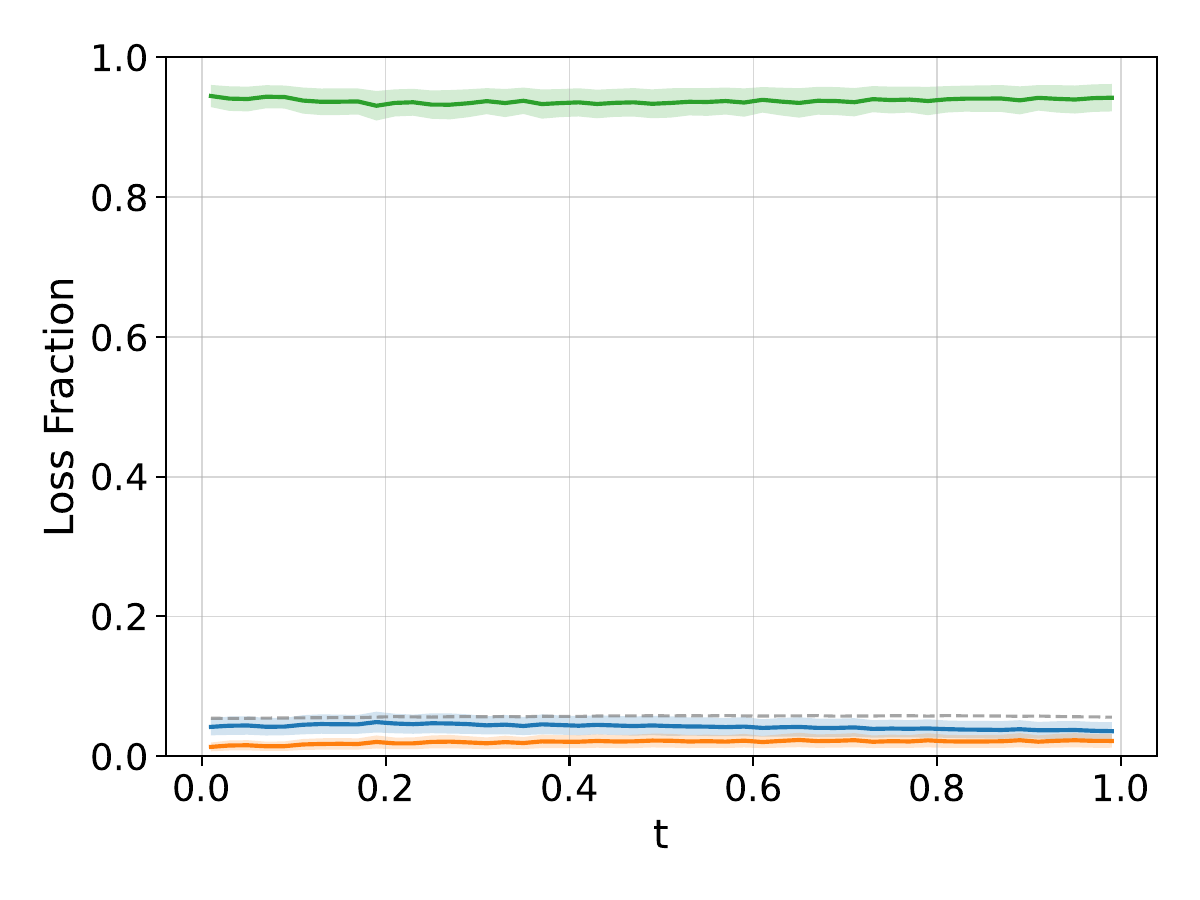}
    \caption{Well-trained}
\end{subfigure}
\hfill 
\begin{subfigure}[t]{\appratioslength}
    \includegraphics[width=\textwidth]{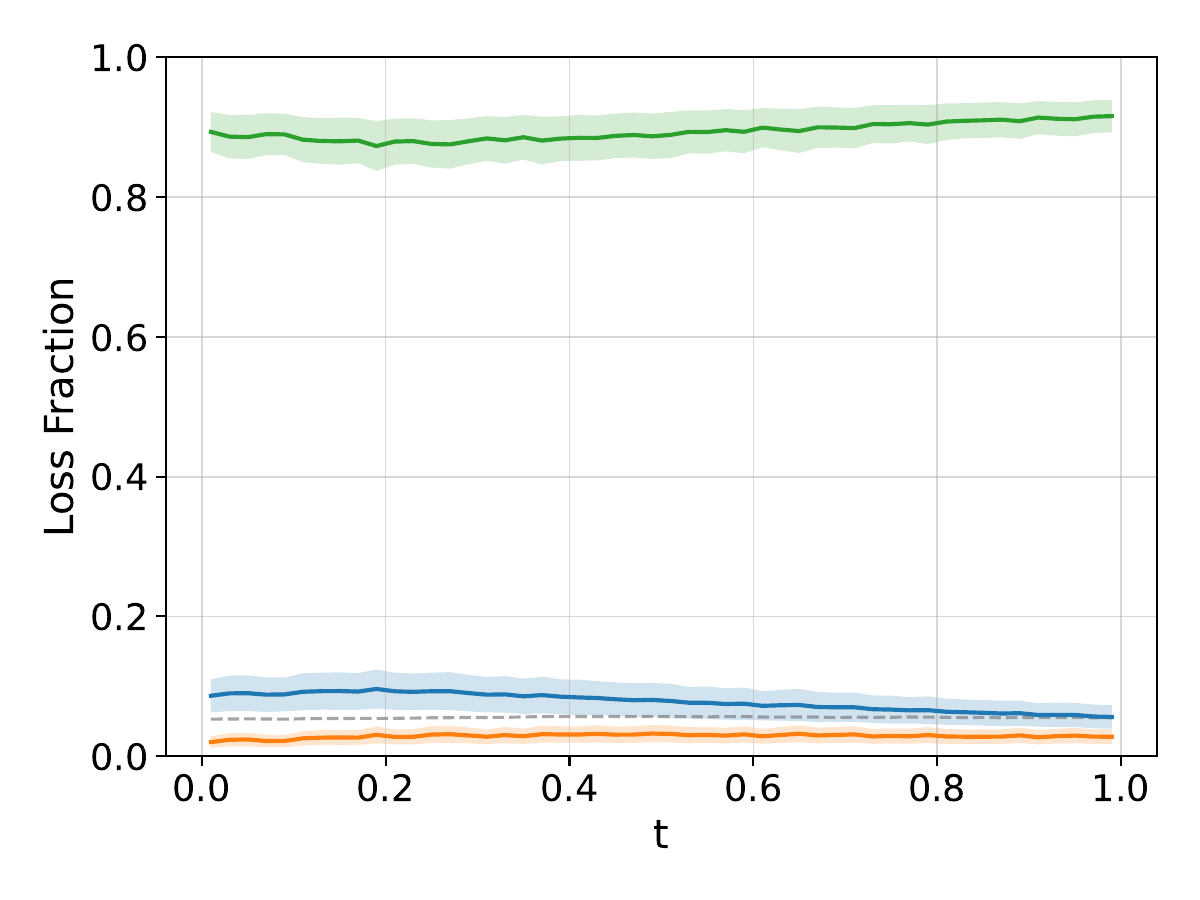}
    \caption{Poorly-trained}
\end{subfigure}
\hfill 
\caption{Burgers' equation: Fractions of influences $\bar r_{L_i}(\mathcal{X}, R_{\text{test}})$ for 50 time bins, averaged over seeds. Shaded regions indicate the standard deviation.}
\label{fig:app_burgers_loss_ratios}
\end{figure}

\begin{figure}[ht]
\setlength{\fboxsep}{0pt}      %
\setlength{\fboxrule}{0.2pt}   %
\centering
\hspace*{3.5em}
\begin{subfigure}[t]{0.8\textwidth}
    \begin{tikzpicture}
        \node[inner sep=0pt] (img) {\includegraphics[width=\linewidth]{figures/icml_ratios/wave_legend.pdf}};
        \draw[line width=0.2pt] (img.north west) -- (img.north east) -- (img.south east);
        \draw[line width=0.2pt] (img.north west) -- (img.south west);
    \end{tikzpicture}
\end{subfigure}
\\[-0.3ex]
\hspace*{3.5em}
\begin{subfigure}[t]{0.8\textwidth}
    \begin{tikzpicture}
        \node[inner sep=0pt] (img) {\includegraphics[width=\linewidth]{figures/icml_ratios/wave_legend_2.pdf}};
        \draw[line width=0.2pt] (img.south west) -- (img.south east) -- (img.north east);
        \draw[line width=0.2pt] (img.south west) -- (img.north west);
    \end{tikzpicture}
\end{subfigure}
\\[-0.2ex]
\hfill 
\begin{subfigure}[t]{\appratioslength}
    \includegraphics[width=\textwidth]{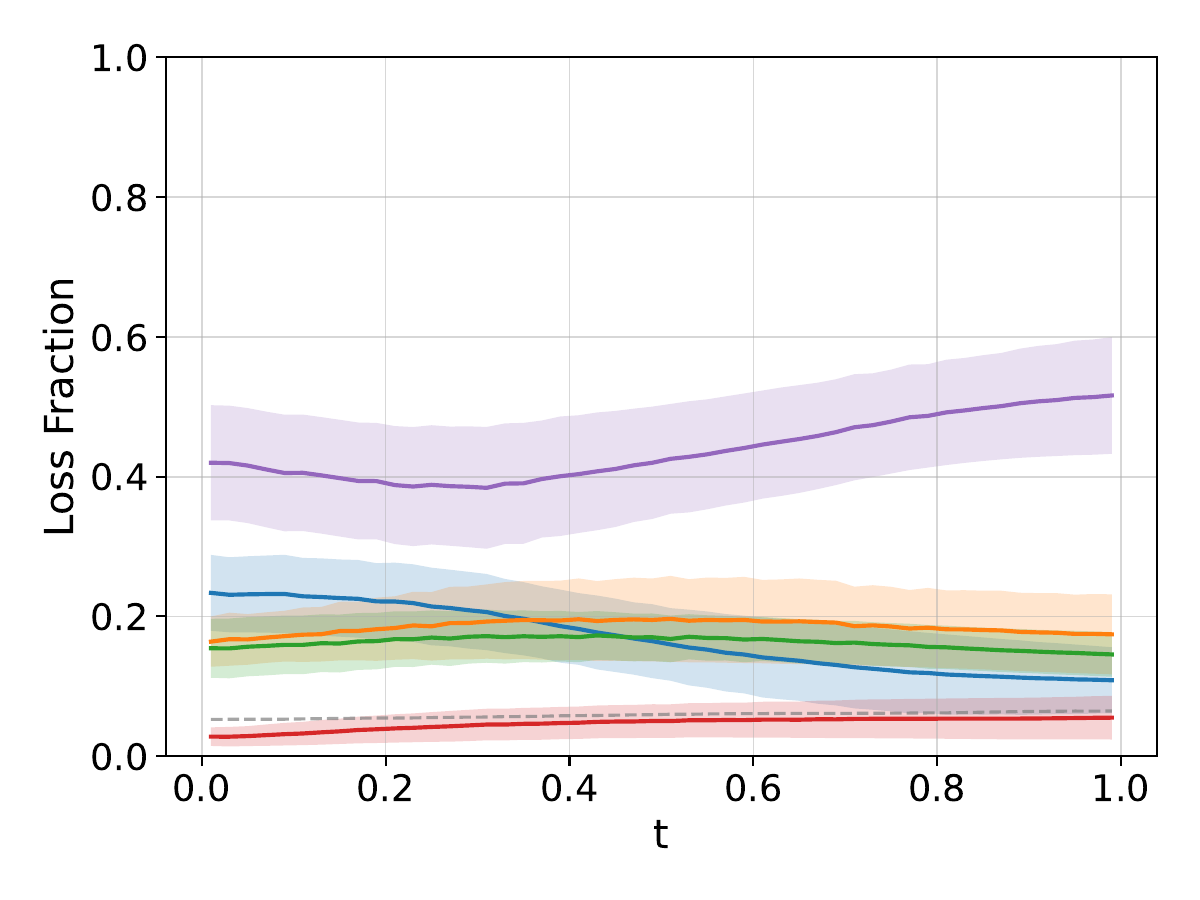}
    \caption{Well-trained}
\end{subfigure}
\hfill 
\begin{subfigure}[t]{\appratioslength}
    \includegraphics[width=\textwidth]{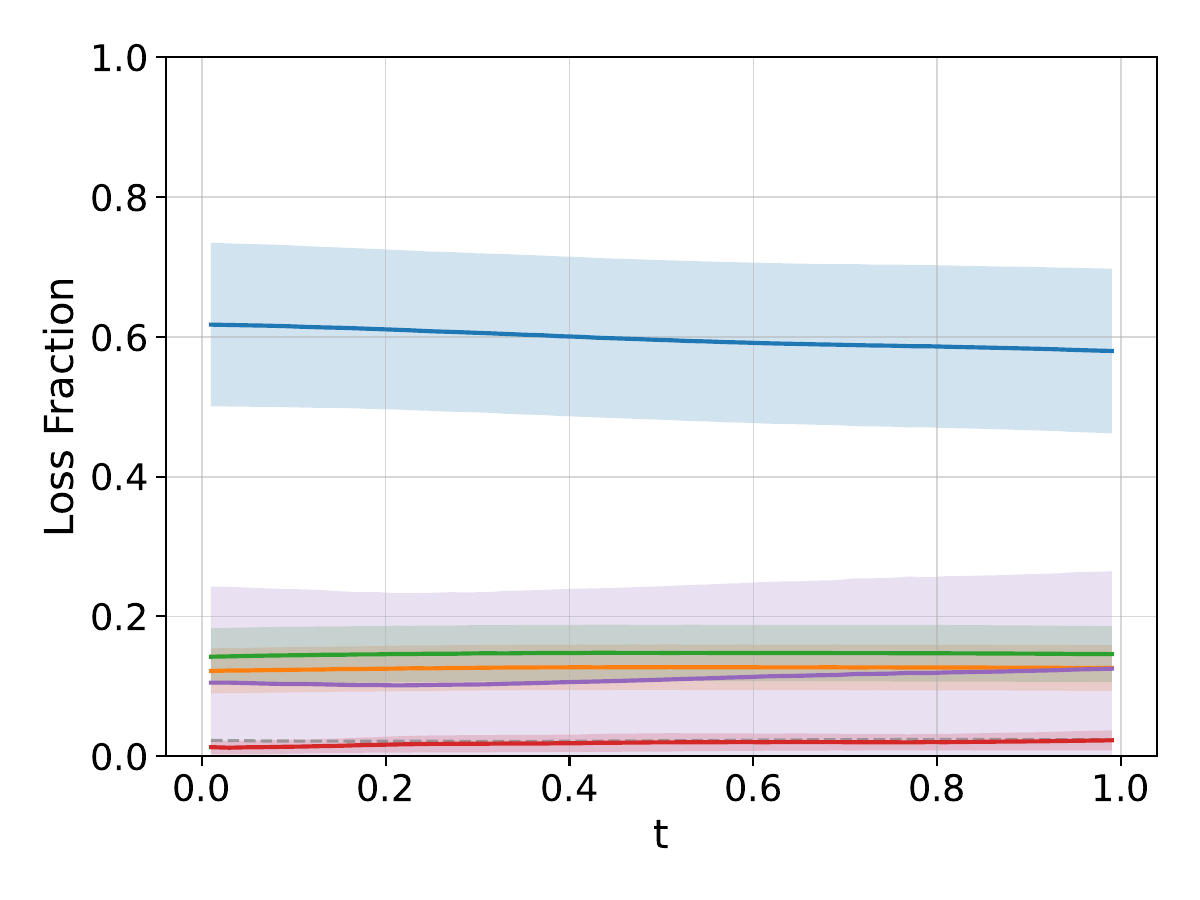}
    \caption{Poorly-trained}
\end{subfigure}
\hfill 
\caption{Wave equation: Fractions of influences $\bar r_{L_i}(\mathcal{X}, R_{\text{test}})$ for 50 time bins, averaged over seeds. Shaded regions indicate the standard deviation.}
\label{fig:app_wave_loss_ratios}
\end{figure}

\begin{figure}[ht]
\setlength{\fboxsep}{0pt}      %
\setlength{\fboxrule}{0.2pt}   %
\centering
\hspace*{3.5em}
\begin{subfigure}[t]{0.8\textwidth}
    \fbox{\includegraphics[width=\dimexpr\linewidth-2\fboxsep-2\fboxrule\relax]{figures/icml_ratios/poisson_legend.pdf}}
\end{subfigure}
\\[-0.2ex]
\hfill 
\begin{subfigure}[t]{\appratioslength}
    \includegraphics[width=\textwidth]{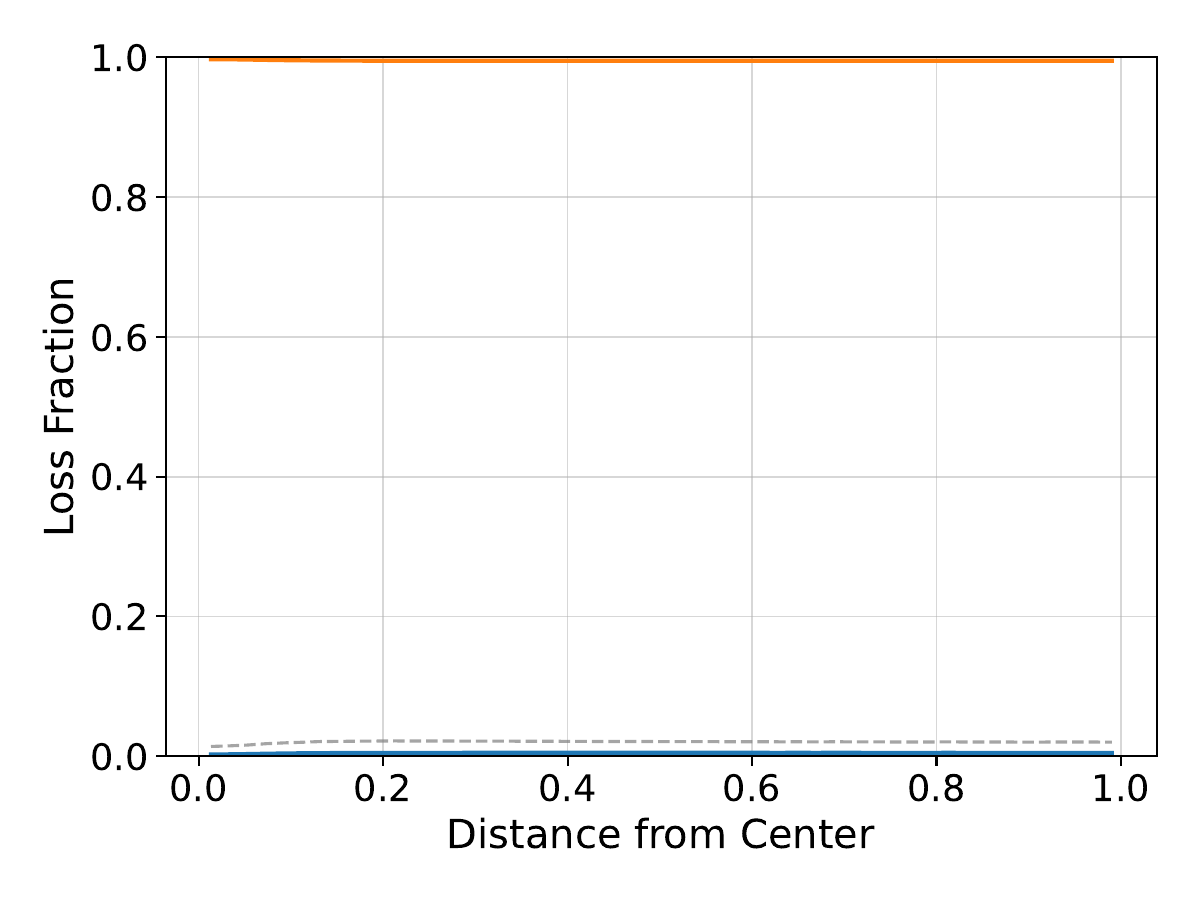}
    \caption{Well-trained}
\end{subfigure}
\hfill 
\begin{subfigure}[t]{\appratioslength}
    \includegraphics[width=\textwidth]{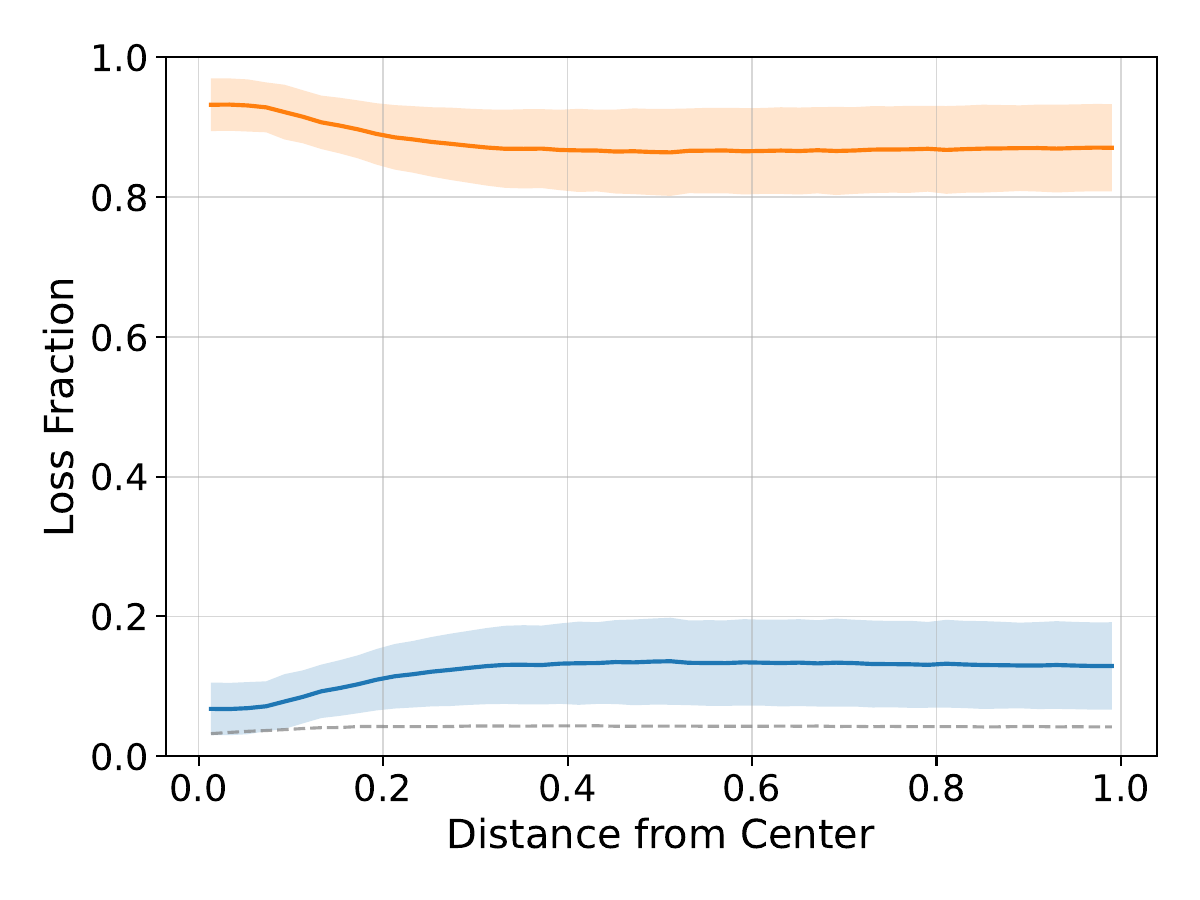}
    \caption{Poorly-trained}
\end{subfigure}
\hfill 
\caption{Poisson equation: Fractions of influences $\bar r_{L_i}(\mathcal{X}, R_{\text{test}})$ for 50 bins based on the distance from the center, averaged over seeds. Shaded regions indicate the standard deviation.}
\label{fig:app_poisson_loss_ratios}
\end{figure}

\newcommand{\nsratioslength}{0.25\textwidth}

\begin{figure}[ht]
\setlength{\fboxsep}{0pt}
\setlength{\fboxrule}{0.2pt}
\centering

\begin{subfigure}[t]{0.95\textwidth}
    \centering
    \begin{tikzpicture}
        \node[inner sep=0pt] (img) {\includegraphics[width=0.85\linewidth]{figures/icml_ratios/navier_stokes_legend.pdf}};
        \draw[line width=0.2pt] (img.north west) -- (img.north east) -- (img.south east);
        \draw[line width=0.2pt] (img.north west) -- (img.south west);
    \end{tikzpicture}
\end{subfigure}
\\[-0.3ex]
\begin{subfigure}[t]{0.95\textwidth}
    \centering
    \begin{tikzpicture}
        \node[inner sep=0pt] (img) {\includegraphics[width=0.85\linewidth]{figures/icml_ratios/navier_stokes_legend_1.pdf}};
        \draw[line width=0.2pt] (img.south west) -- (img.south east) -- (img.north east);
        \draw[line width=0.2pt] (img.south west) -- (img.north west);
    \end{tikzpicture}
\end{subfigure}
\\[1ex]

\hfill 
\begin{subfigure}[t]{\nsratioslength}
    \centering
    \includegraphics[width=\textwidth]{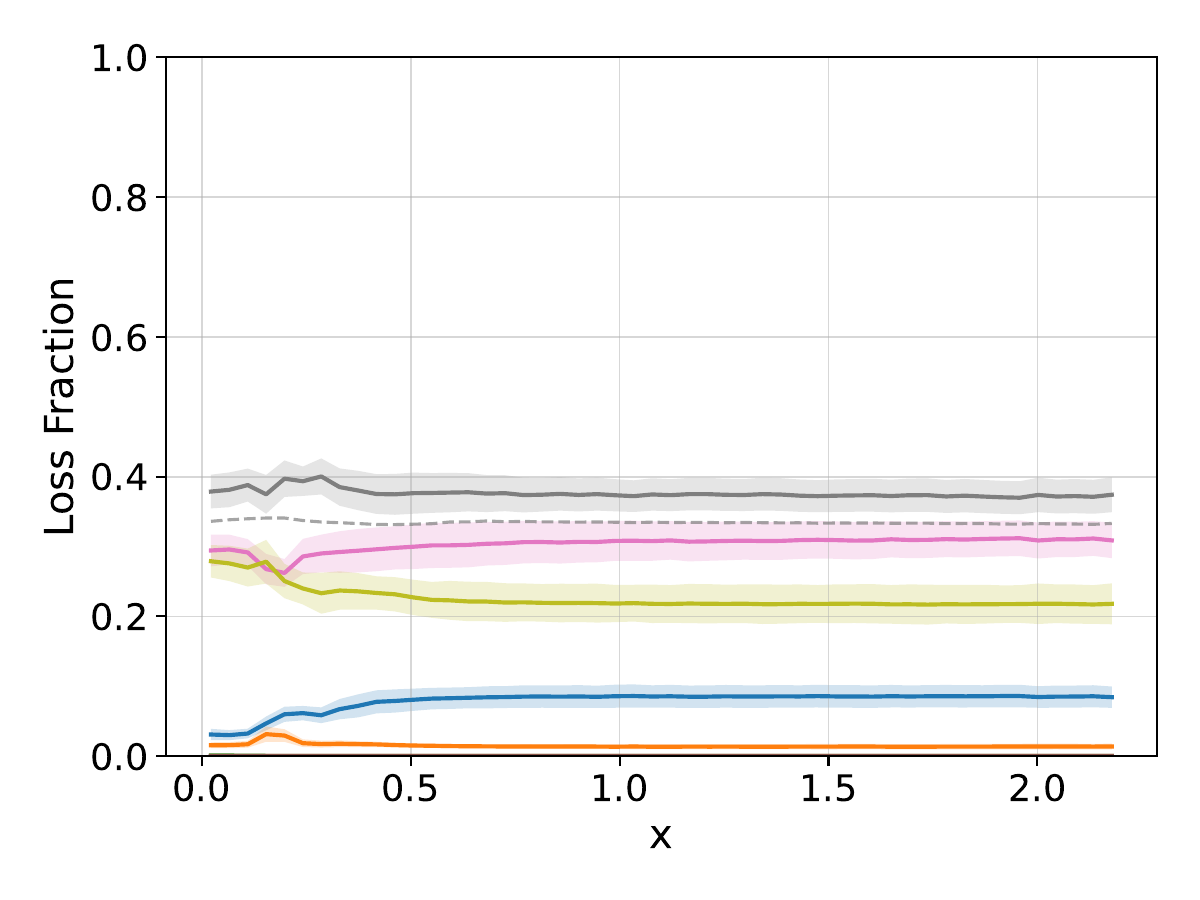}
    \caption{Well-trained ($x$-Velocity)}
    \label{fig:ns_good_0}
\end{subfigure}
\hfill
\begin{subfigure}[t]{\nsratioslength}
    \centering
    \includegraphics[width=\textwidth]{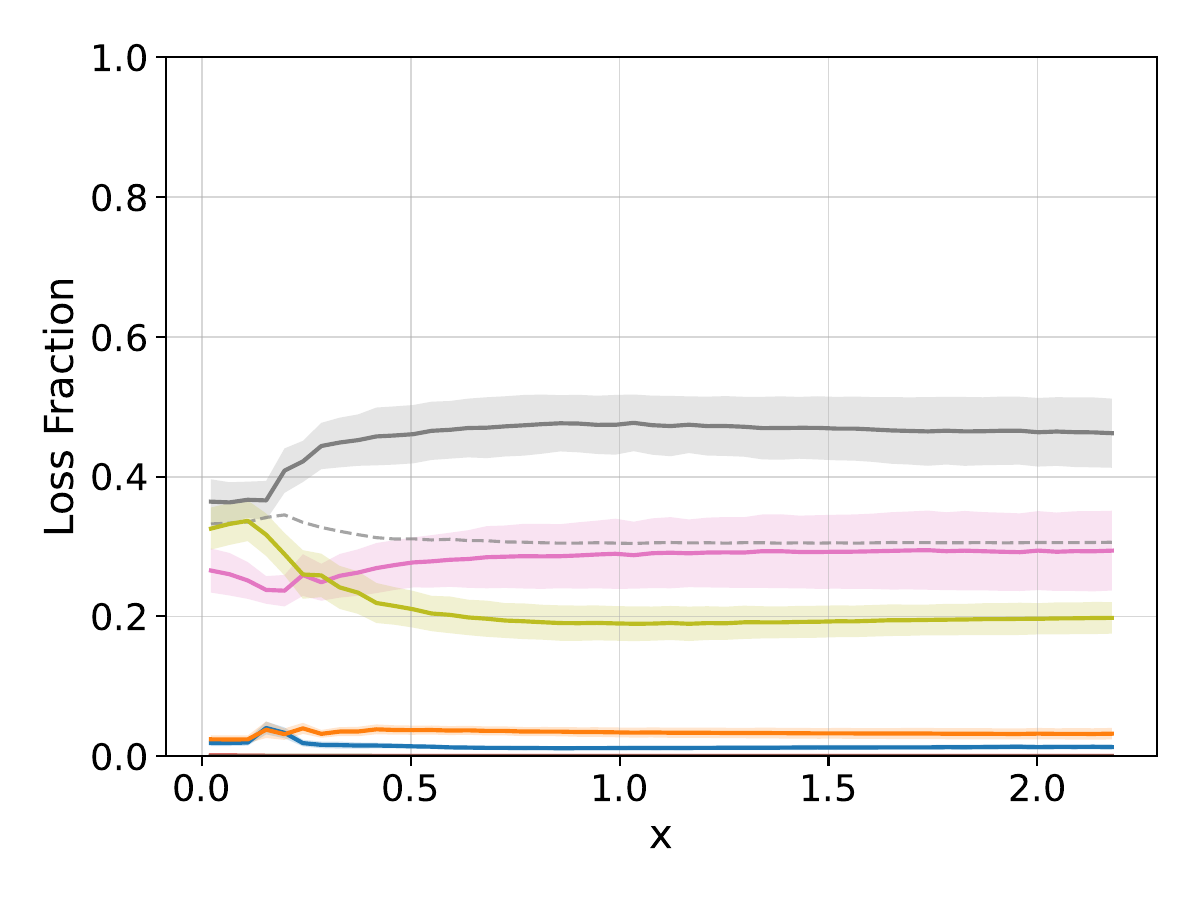}
    \caption{Well-trained ($y$-Velocity)}
    \label{fig:ns_good_1}
\end{subfigure}
\hfill
\begin{subfigure}[t]{\nsratioslength}
    \centering
    \includegraphics[width=\textwidth]{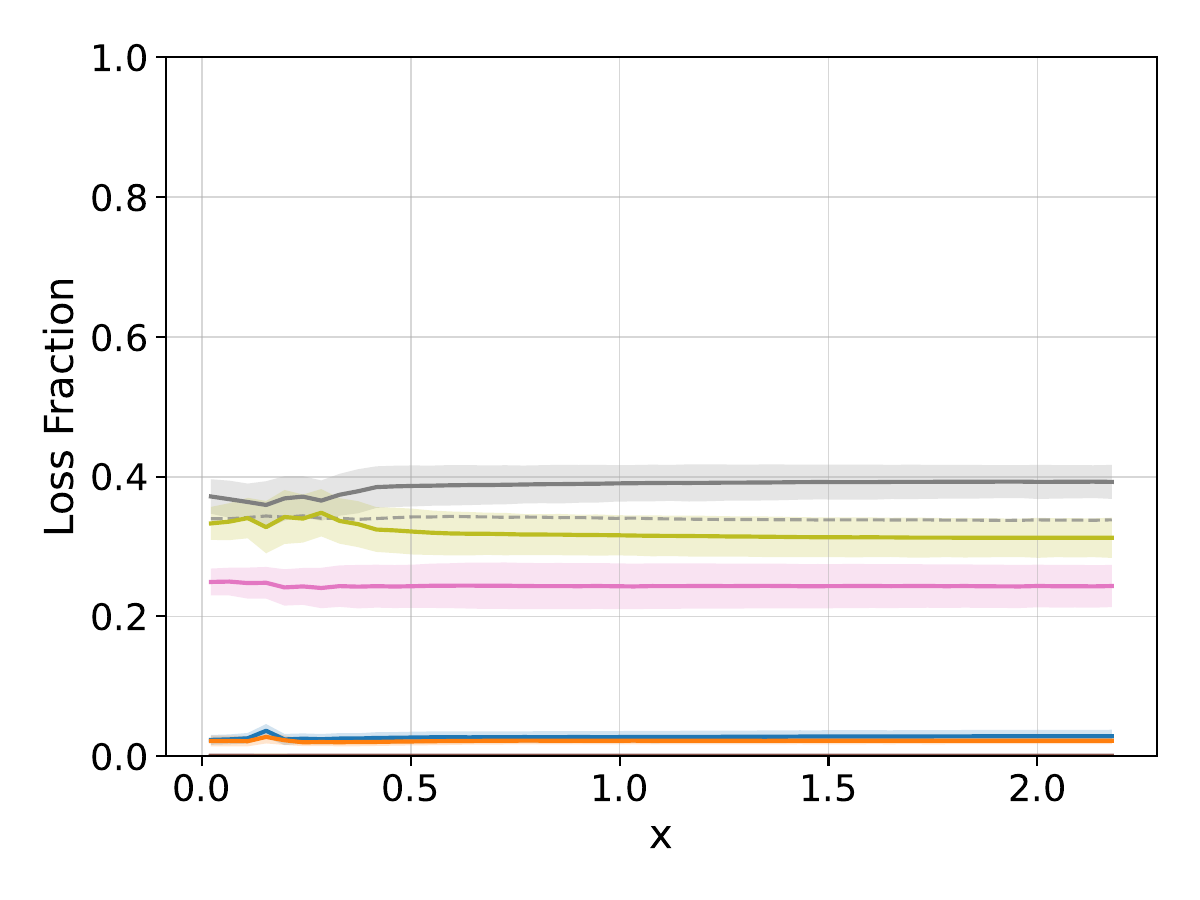}
    \caption{Well-trained (Pressure)}
    \label{fig:ns_good_2}
\end{subfigure}
\hfill
\\ 
\hfill
\begin{subfigure}[t]{\nsratioslength}
    \centering
    \includegraphics[width=\textwidth]{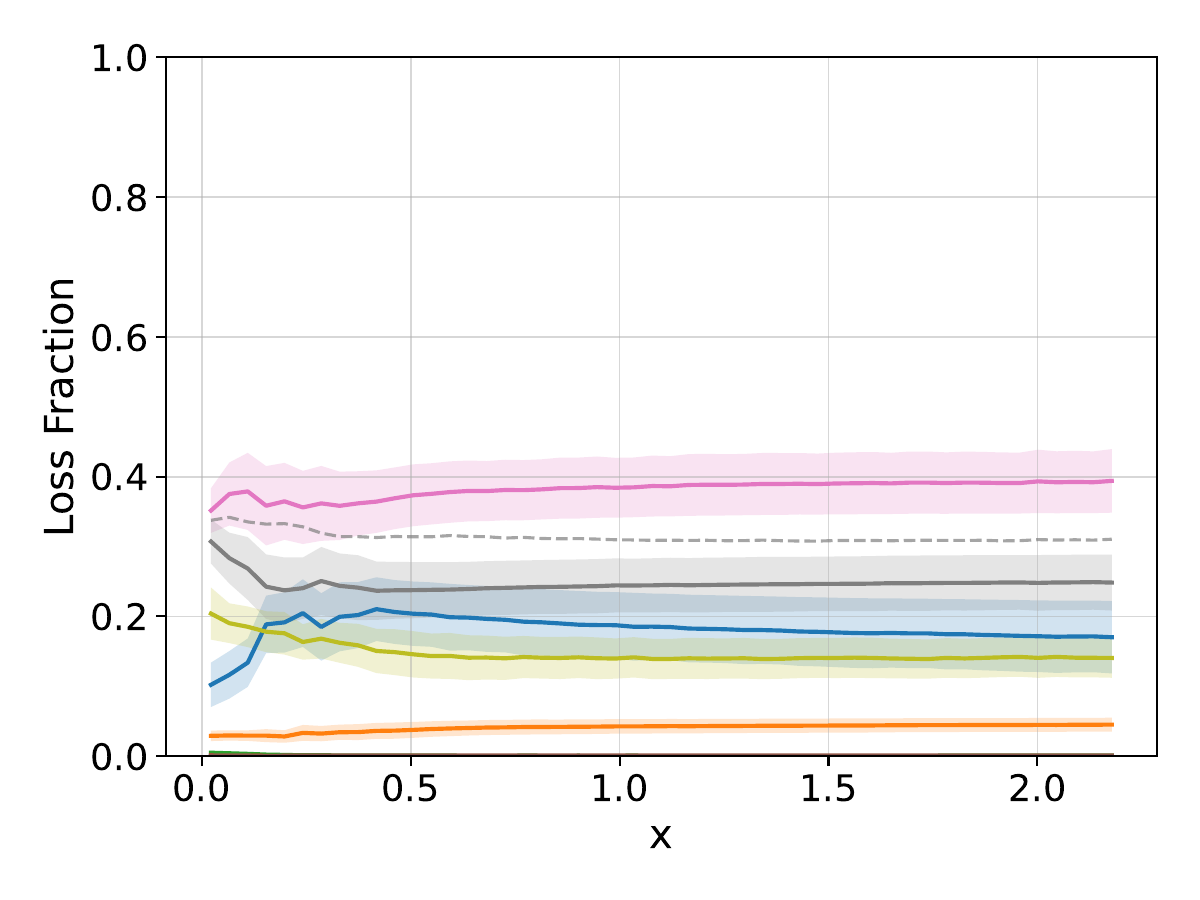}
    \caption{Poorly-trained ($x$-Velocity)}
    \label{fig:ns_bad_0}
\end{subfigure}
\hfill
\begin{subfigure}[t]{\nsratioslength}
    \centering
    \includegraphics[width=\textwidth]{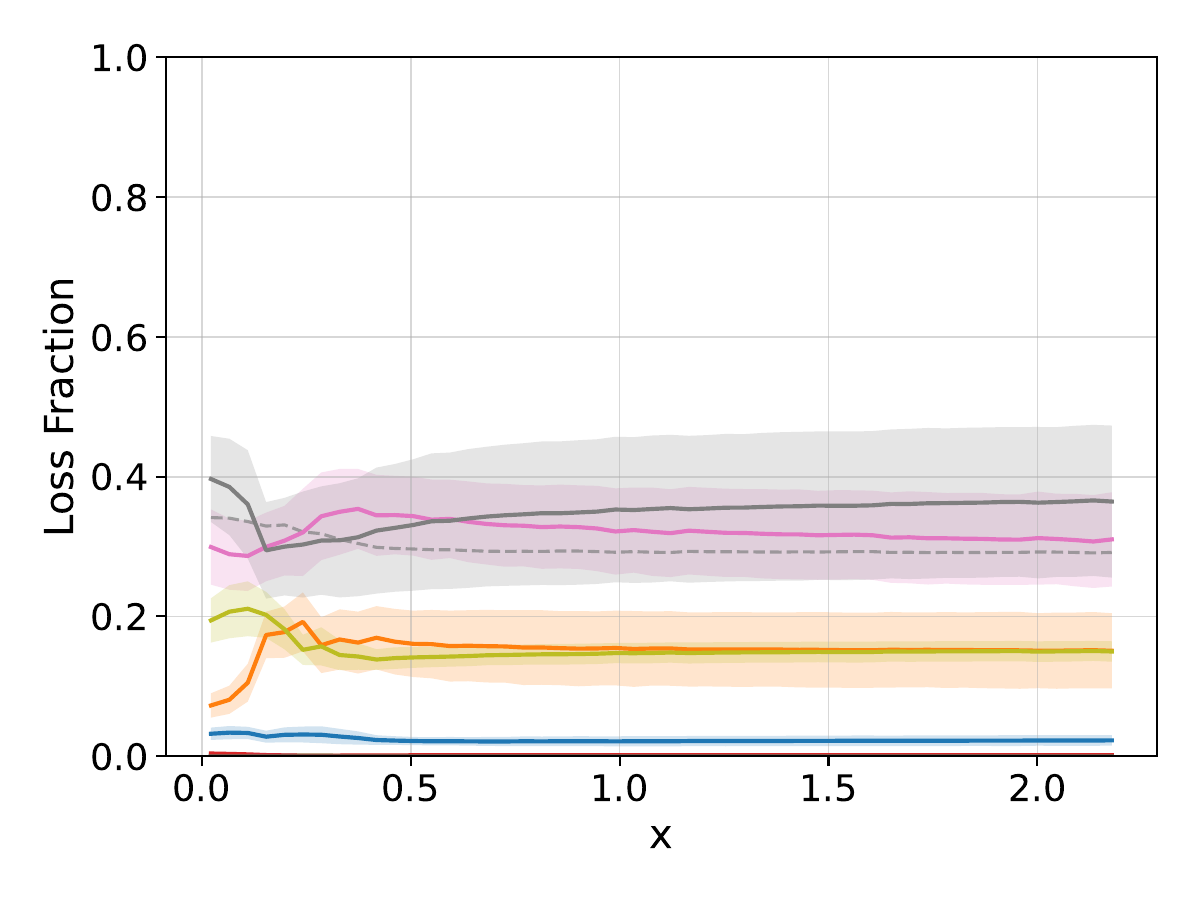}
    \caption{Poorly-trained ($y$-Velocity)}
    \label{fig:ns_bad_1}
\end{subfigure}
\hfill
\vspace{0.5ex}
\begin{subfigure}[t]{\nsratioslength}
    \centering
    \includegraphics[width=\textwidth]{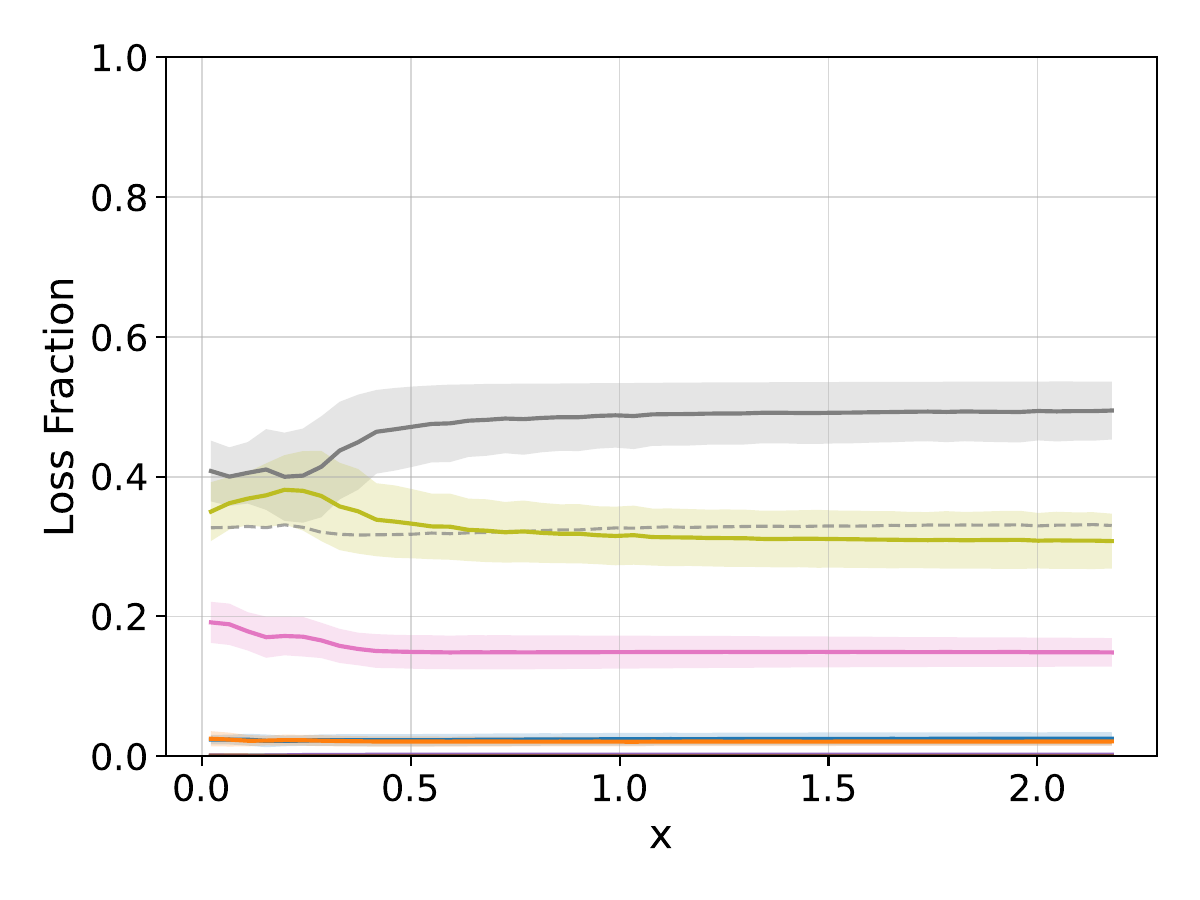}
    \caption{Poorly-trained (Pressure)}
    \label{fig:ns_bad_2}
\end{subfigure}
\hfill

\caption{Navier-Stokes equations: Fractions of influences $\bar r_{L_i}(\mathcal{X}, R_{\text{test}})$ for 50 bins of x coordinates, averaged over seeds. Shaded regions indicate the standard deviation. Evaluated for all three outputs respectively.}
\label{fig:app_navier_stokes_loss_ratios}
\end{figure}

\clearpage

\subsection{Supplementary Point-wise Influence Plots}
\label{app:pointwise_plots}

We visualize point-wise influence scores for each benchmark problem. 
For two representative points per problem, we compare well-trained and poorly-trained models side by side.
Red heatmaps show which training points influence a fixed test point, 
whereas blue heatmaps show how a fixed training point influences the whole domain.

\newcommand{\pointwisewidth}{0.22\textwidth}

\begin{figure}[h]
    \centering
    \begin{minipage}[t]{0.48\textwidth}
        \centering
        \begin{subfigure}[t]{0.48\textwidth}
            \centering
            \includegraphics[width=\textwidth]{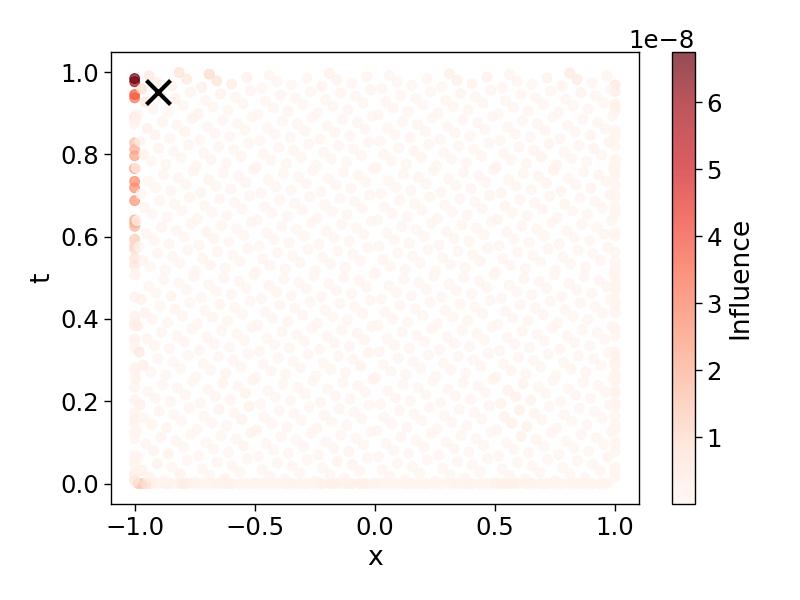}
        \end{subfigure}
        \hfill
        \begin{subfigure}[t]{0.48\textwidth}
            \centering
            \includegraphics[width=\textwidth]{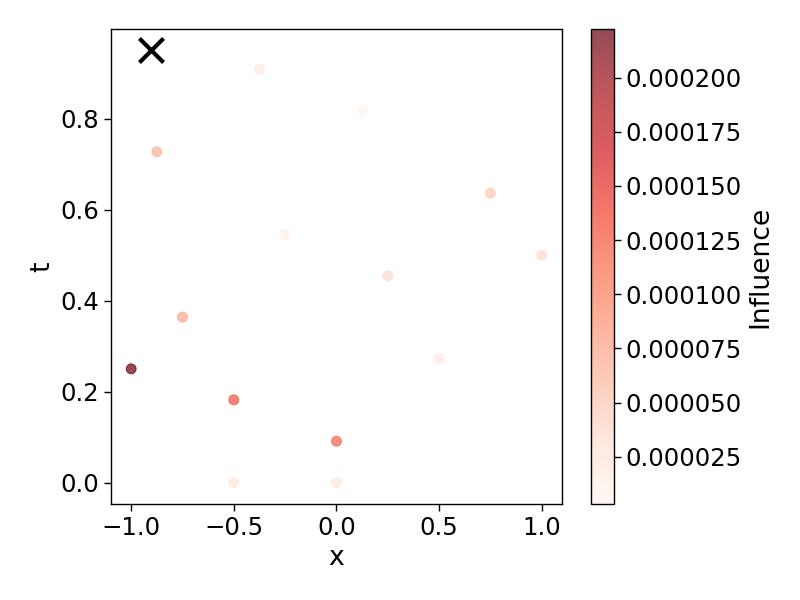}
        \end{subfigure}
        \\[0.5ex]
        \begin{subfigure}[t]{0.48\textwidth}
            \centering
            \includegraphics[width=\textwidth]{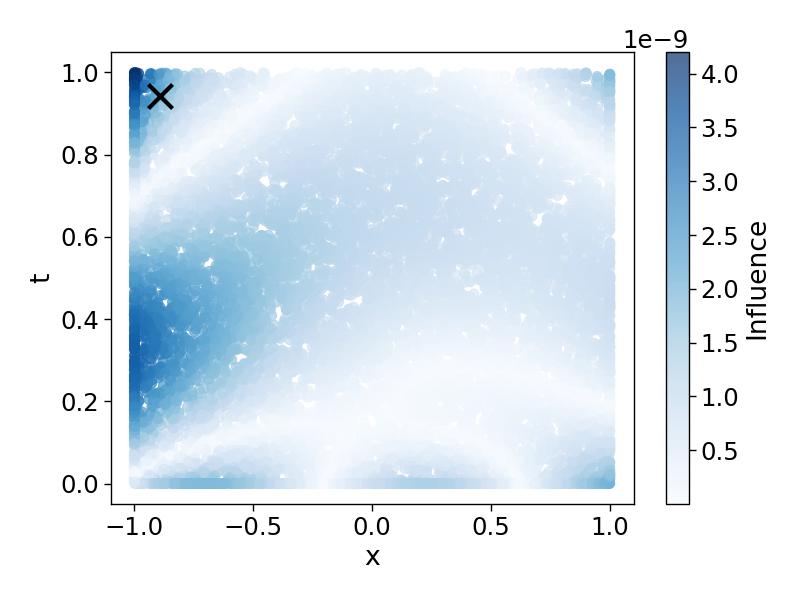}
        \end{subfigure}
        \hfill
        \begin{subfigure}[t]{0.48\textwidth}
            \centering
            \includegraphics[width=\textwidth]{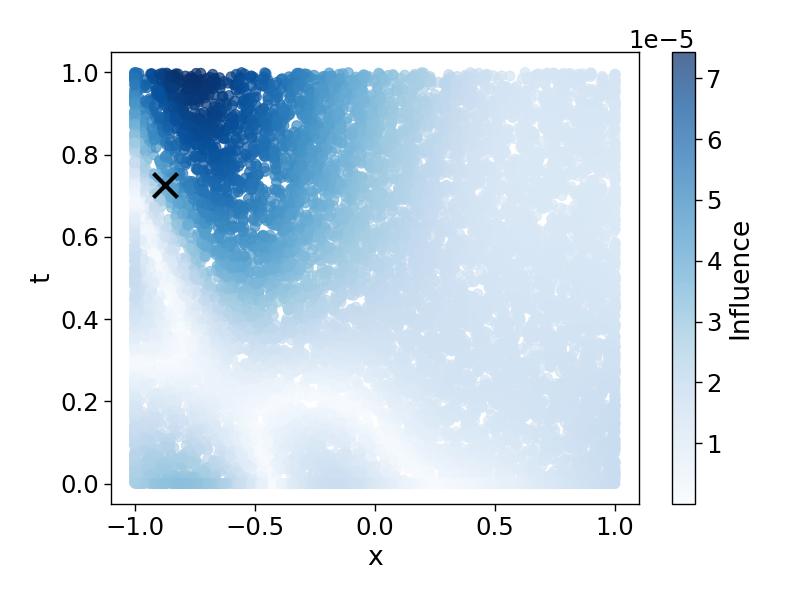}
        \end{subfigure}
    \end{minipage}
    \hfill
    \vrule 
    \hfill
    \begin{minipage}[t]{0.48\textwidth}
        \centering
        \begin{subfigure}[t]{0.48\textwidth}
            \centering
            \includegraphics[width=\textwidth]{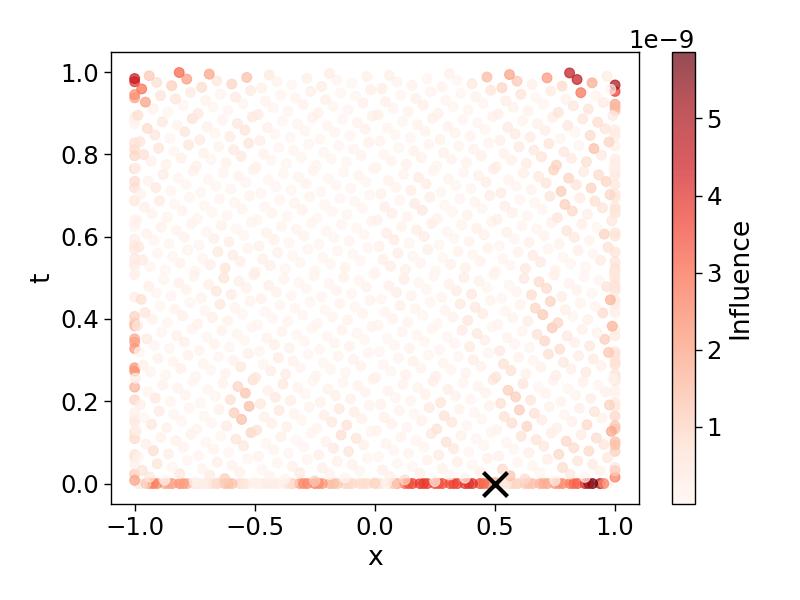}
        \end{subfigure}
        \hfill
        \begin{subfigure}[t]{0.48\textwidth}
            \centering
            \includegraphics[width=\textwidth]{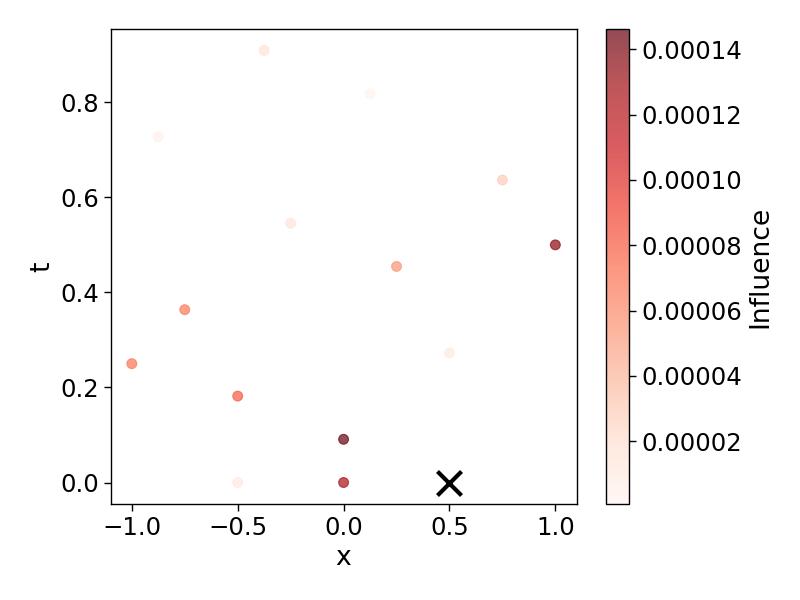}
        \end{subfigure}
        \\[0.5ex]
        \begin{subfigure}[t]{0.48\textwidth}
            \centering
            \includegraphics[width=\textwidth]{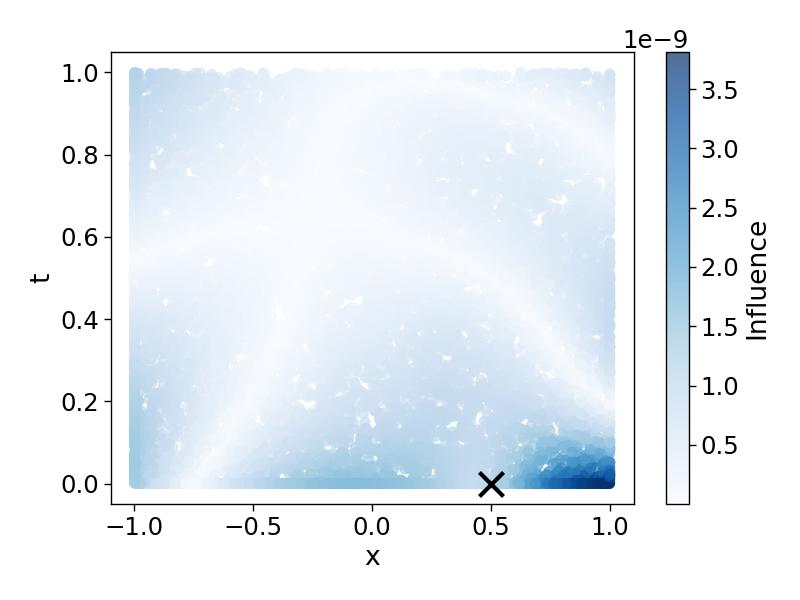}
        \end{subfigure}
        \hfill
        \begin{subfigure}[t]{0.48\textwidth}
            \centering
            \includegraphics[width=\textwidth]{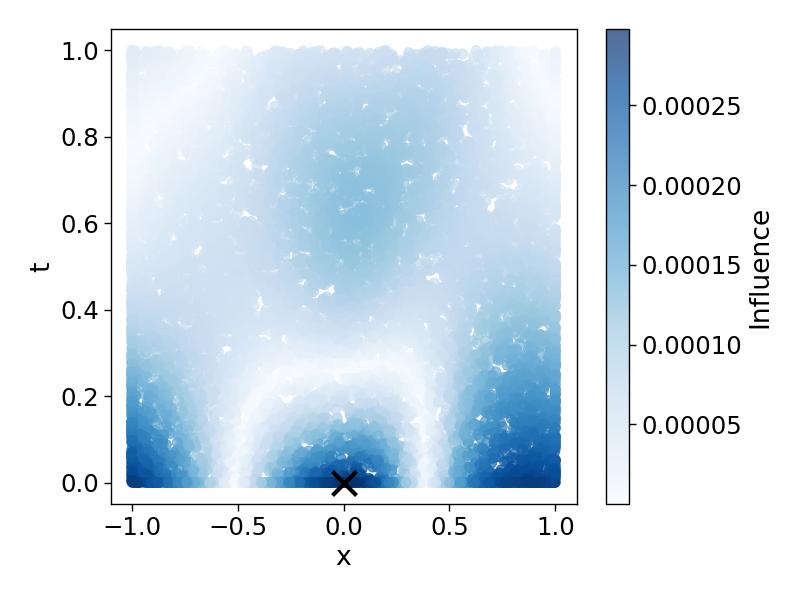}
        \end{subfigure}
    \end{minipage}
    \caption{Heat equation: Influence scores for well-trained (left) and poorly-trained (right) models. Top: training point influences for fixed test point. Bottom: influences for a fixed training point onto the domain.}
    \label{fig:diffusion_influences}
\end{figure}

\begin{figure}[h]
    \centering
    \begin{minipage}[t]{0.48\textwidth}
        \centering
        \begin{subfigure}[t]{0.48\textwidth}
            \centering
            \includegraphics[width=\textwidth]{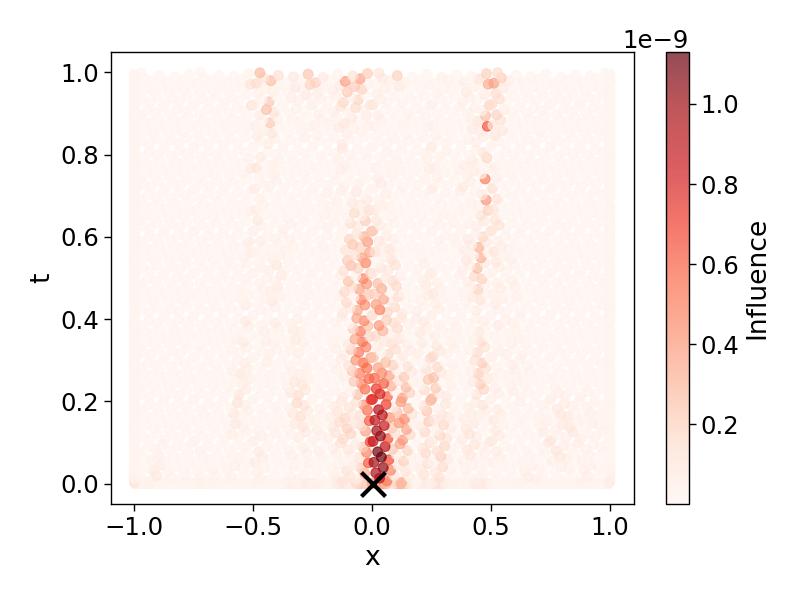}
        \end{subfigure}
        \hfill
        \begin{subfigure}[t]{0.48\textwidth}
            \centering
            \includegraphics[width=\textwidth]{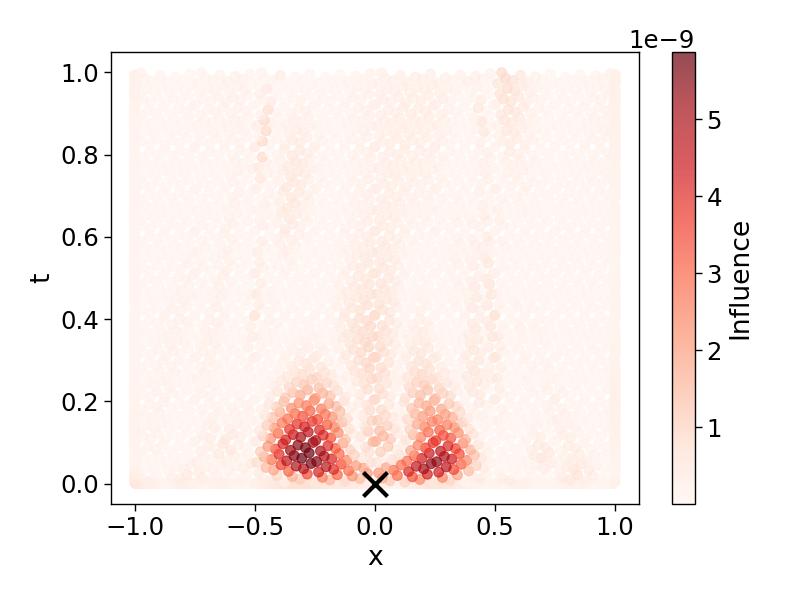}
        \end{subfigure}
        \\[0.5ex]
        \begin{subfigure}[t]{0.48\textwidth}
            \centering
            \includegraphics[width=\textwidth]{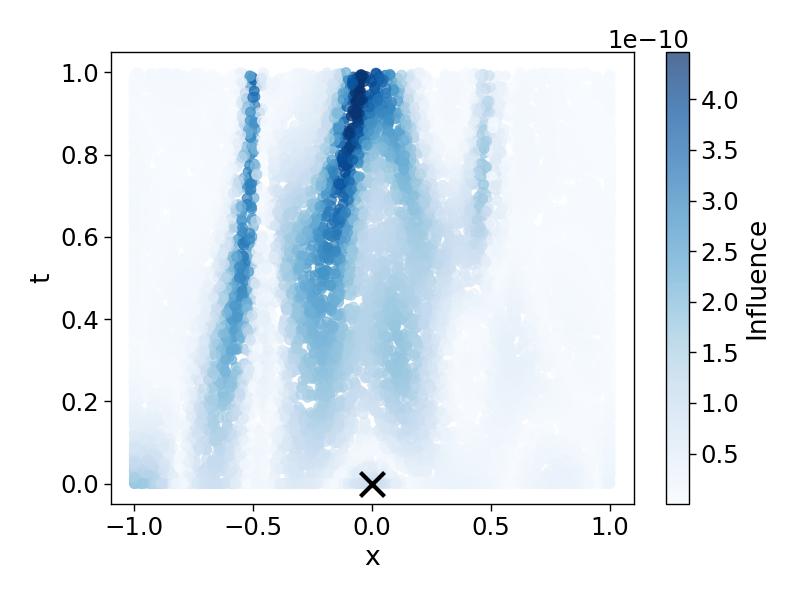}
        \end{subfigure}
        \hfill
        \begin{subfigure}[t]{0.48\textwidth}
            \centering
            \includegraphics[width=\textwidth]{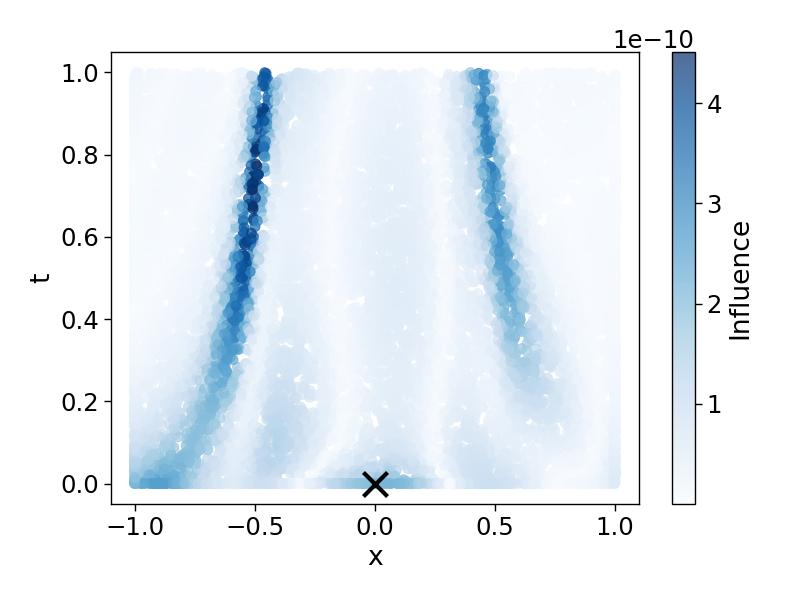}
        \end{subfigure}
    \end{minipage}
    \hfill
    \vrule 
    \hfill
    \begin{minipage}[t]{0.48\textwidth}
        \centering
        \begin{subfigure}[t]{0.48\textwidth}
            \centering
            \includegraphics[width=\textwidth]{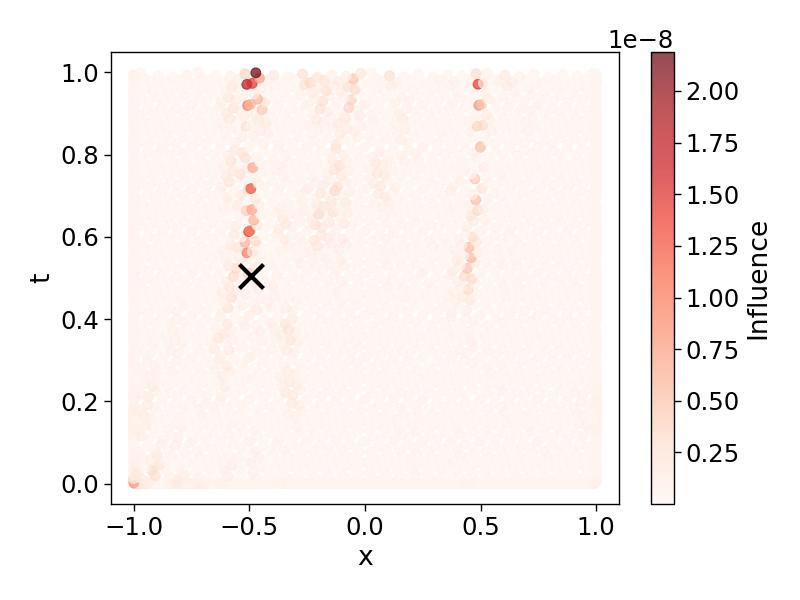}
        \end{subfigure}
        \hfill
        \begin{subfigure}[t]{0.48\textwidth}
            \centering
            \includegraphics[width=\textwidth]{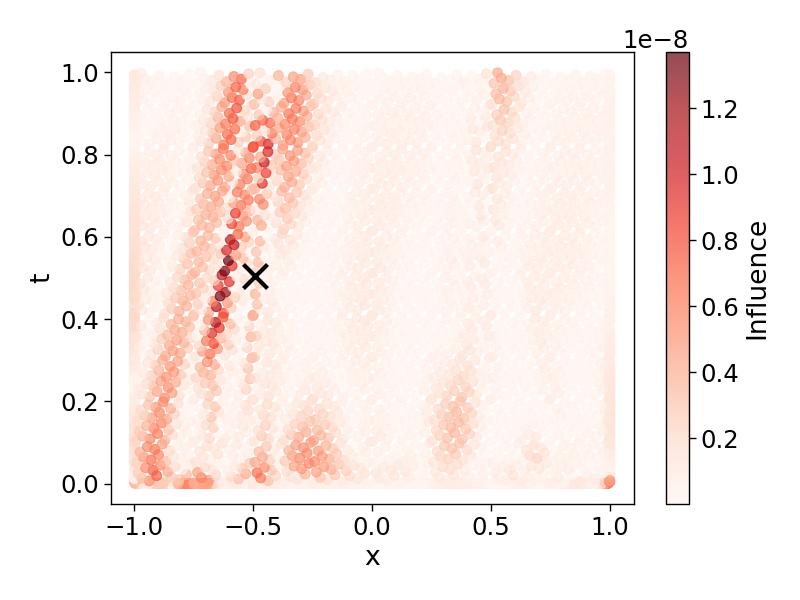}
        \end{subfigure}
        \\[0.5ex]
        \begin{subfigure}[t]{0.48\textwidth}
            \centering
            \includegraphics[width=\textwidth]{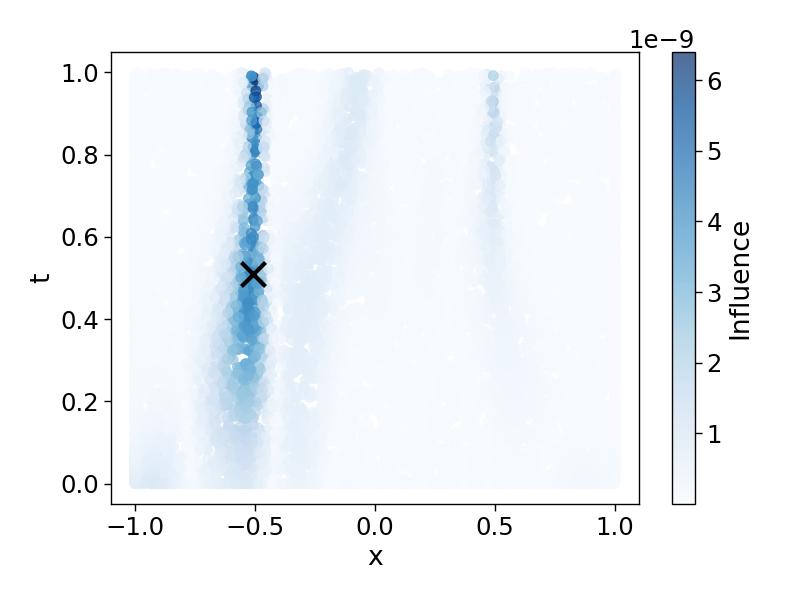}
        \end{subfigure}
        \hfill
        \begin{subfigure}[t]{0.48\textwidth}
            \centering
            \includegraphics[width=\textwidth]{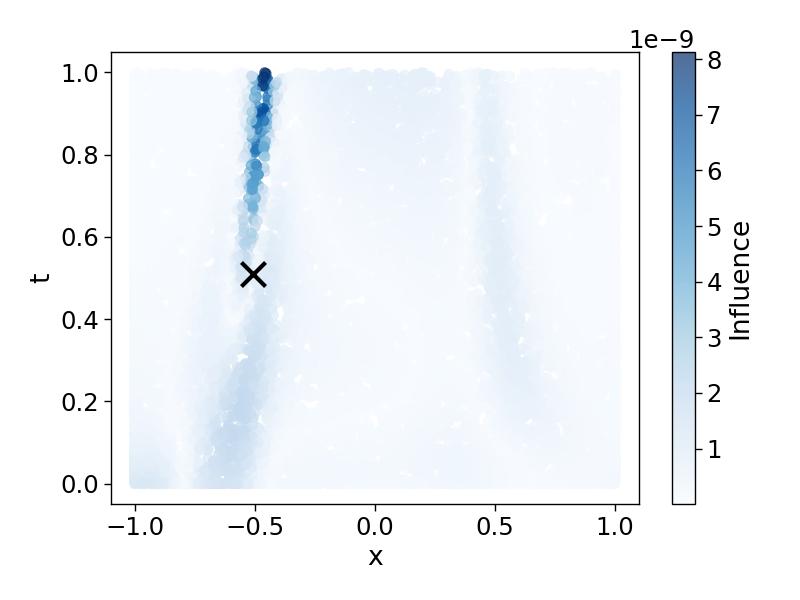}
        \end{subfigure}
    \end{minipage}
    \caption{Allen-Cahn equation: Influence scores for well-trained (left) and poorly-trained (right) models. Top: training point influences for fixed test point. Bottom: influences for a fixed training point onto the domain.}
    \label{fig:allen_cahn_influences}
\end{figure}

\begin{figure}
    \centering
    \begin{minipage}[t]{0.48\textwidth}
        \centering
        \begin{subfigure}[t]{0.48\textwidth}
            \centering
            \includegraphics[width=\textwidth]{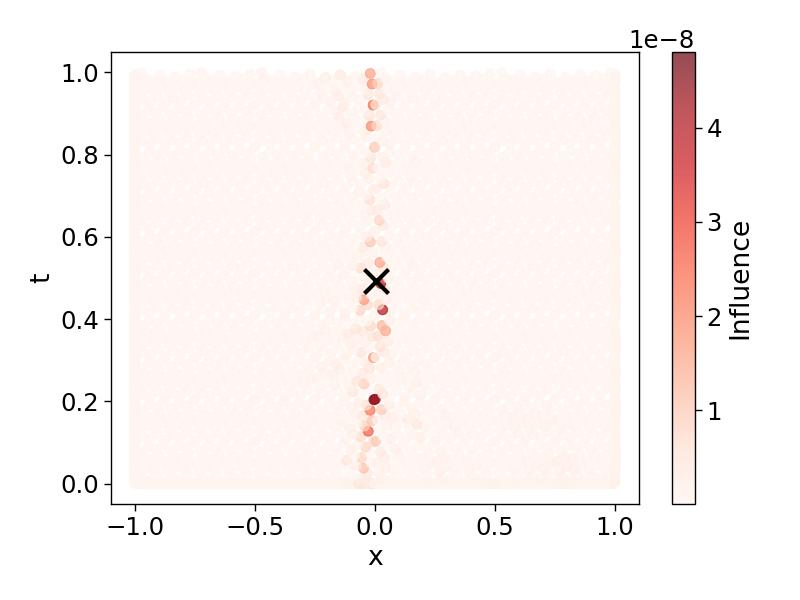}
        \end{subfigure}
        \hfill
        \begin{subfigure}[t]{0.48\textwidth}
            \centering
            \includegraphics[width=\textwidth]{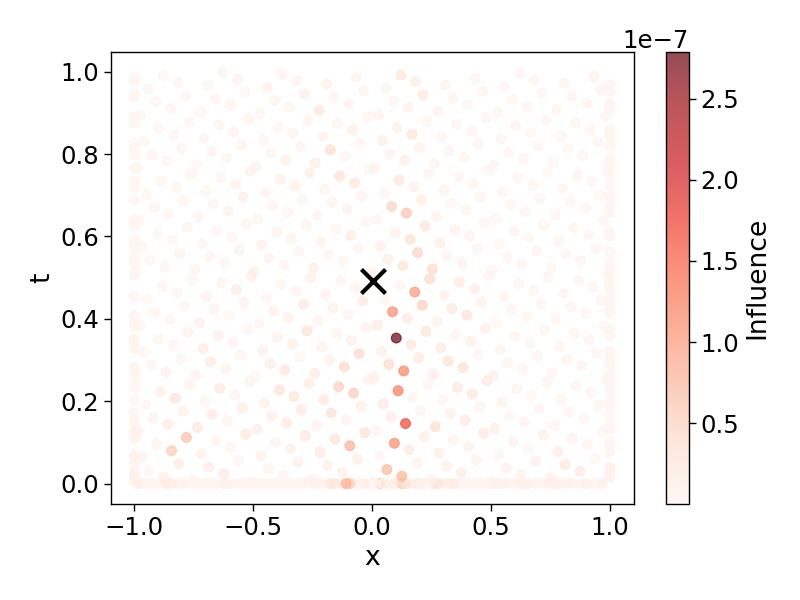}
        \end{subfigure}
        \\[0.5ex]
        \begin{subfigure}[t]{0.48\textwidth}
            \centering
            \includegraphics[width=\textwidth]{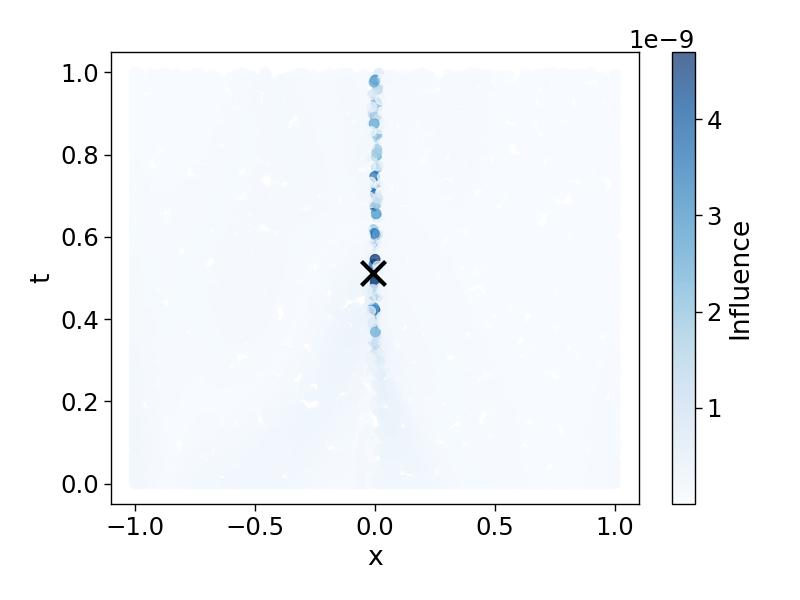}
        \end{subfigure}
        \hfill
        \begin{subfigure}[t]{0.48\textwidth}
            \centering
            \includegraphics[width=\textwidth]{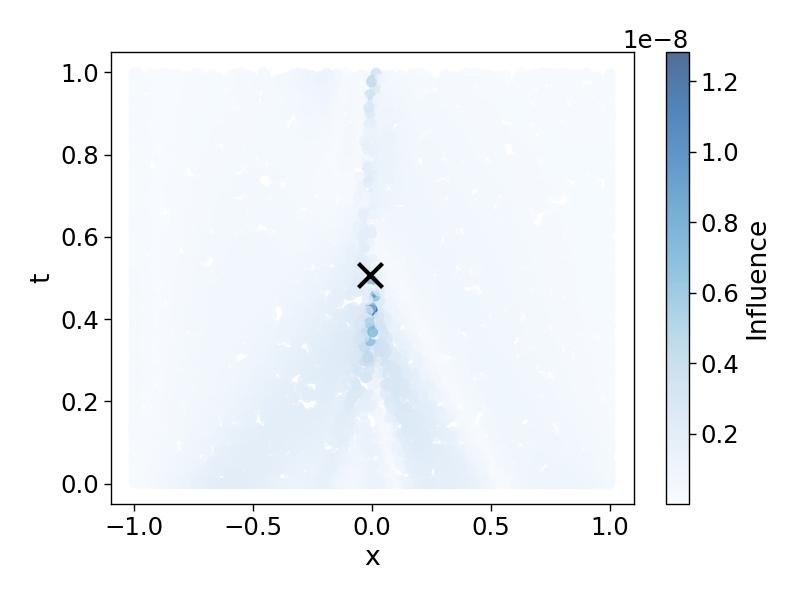}
        \end{subfigure}
    \end{minipage}
    \hfill
    \vrule 
    \hfill
    \begin{minipage}[t]{0.48\textwidth}
        \centering
        \begin{subfigure}[t]{0.48\textwidth}
            \centering
            \includegraphics[width=\textwidth]{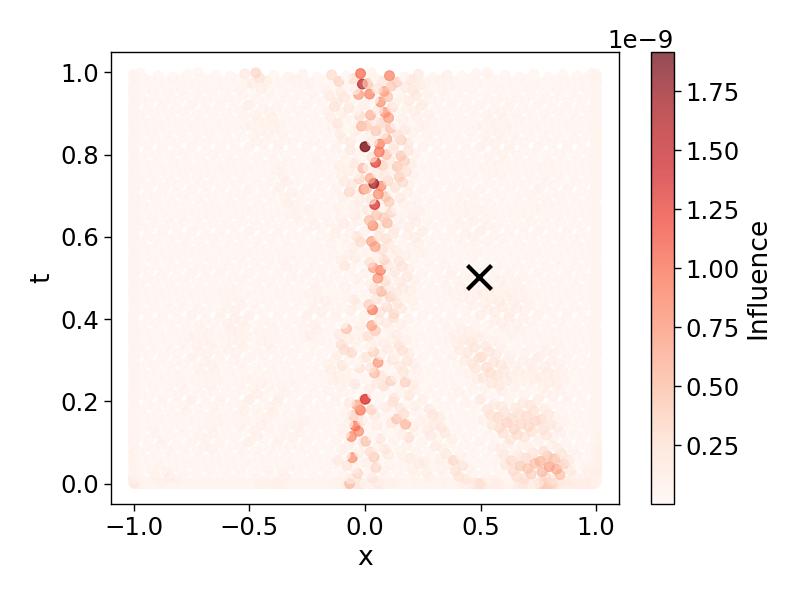}
        \end{subfigure}
        \hfill
        \begin{subfigure}[t]{0.48\textwidth}
            \centering
            \includegraphics[width=\textwidth]{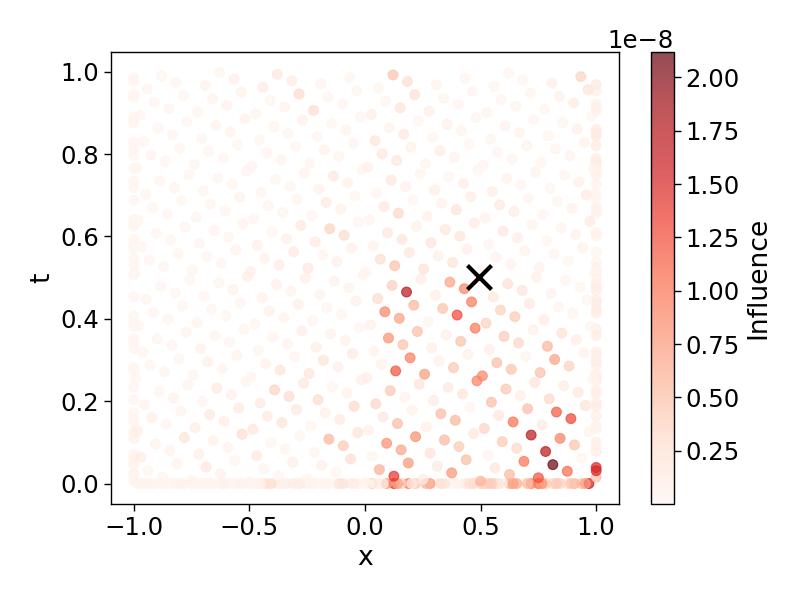}
        \end{subfigure}
        \\[0.5ex]
        \begin{subfigure}[t]{0.48\textwidth}
            \centering
            \includegraphics[width=\textwidth]{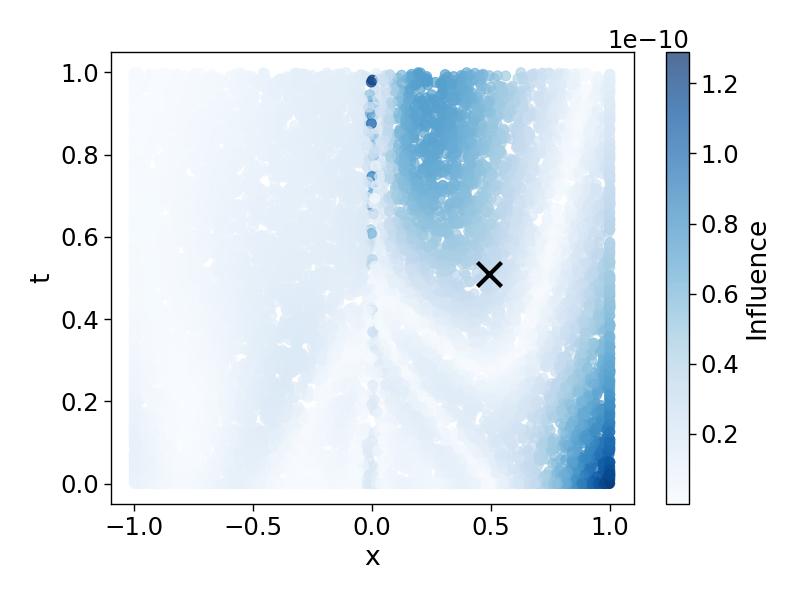}
        \end{subfigure}
        \hfill
        \begin{subfigure}[t]{0.48\textwidth}
            \centering
            \includegraphics[width=\textwidth]{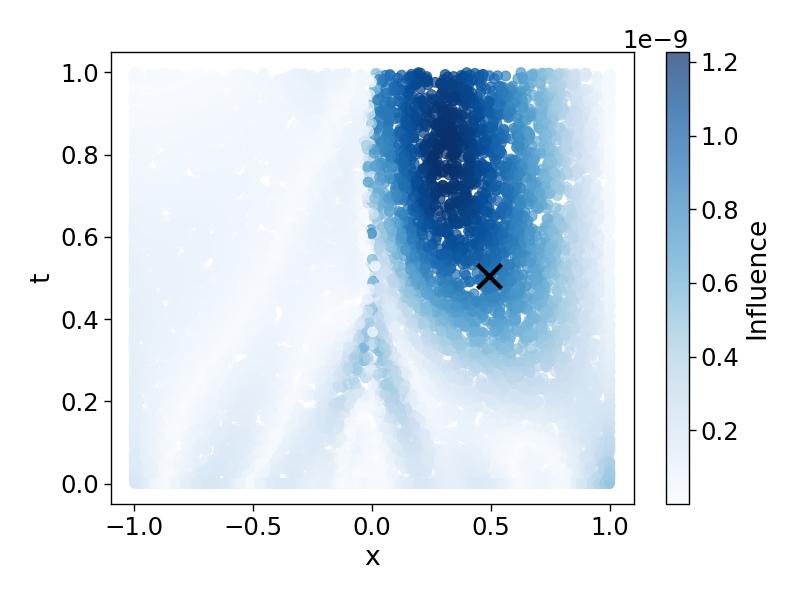}
        \end{subfigure}
    \end{minipage}
    \caption{Burgers equation: Influence scores for well-trained (left) and poorly-trained (right) models. Top: training point influences for fixed test point. Bottom: influences for a fixed training point onto the domain.}
    \label{fig:burgers_influences}
\end{figure}

\begin{figure}
    \centering
    \begin{minipage}[t]{0.48\textwidth}
        \centering
        \begin{subfigure}[t]{0.48\textwidth}
            \centering
            \includegraphics[width=\textwidth]{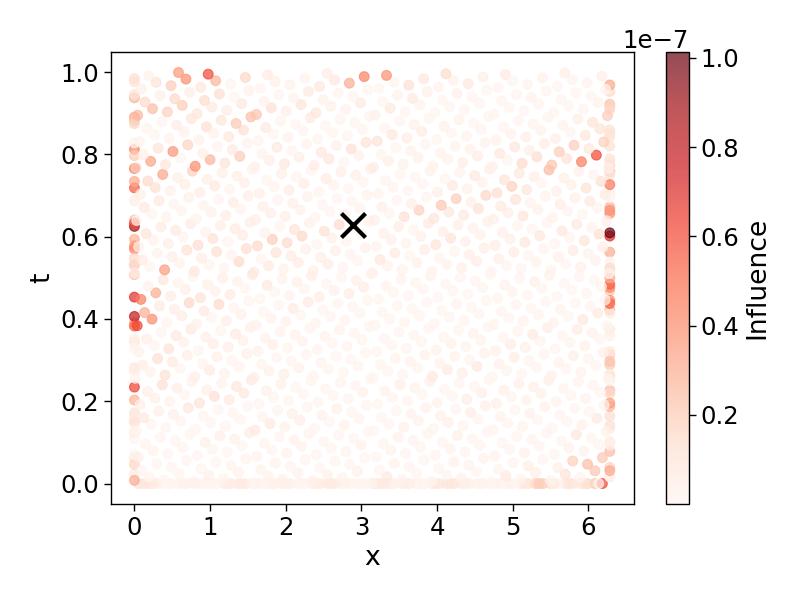}
        \end{subfigure}
        \hfill
        \begin{subfigure}[t]{0.48\textwidth}
            \centering
            \includegraphics[width=\textwidth]{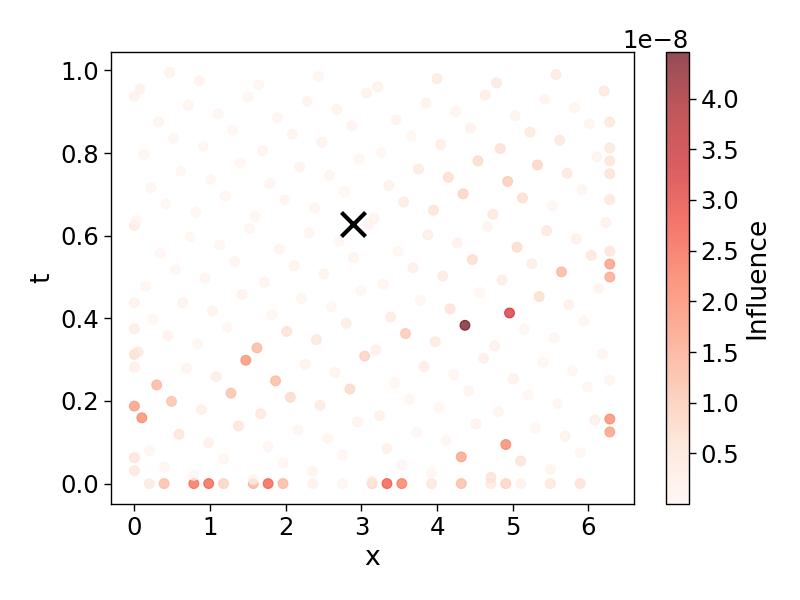}
        \end{subfigure}
        \\[0.5ex]
        \begin{subfigure}[t]{0.48\textwidth}
            \centering
            \includegraphics[width=\textwidth]{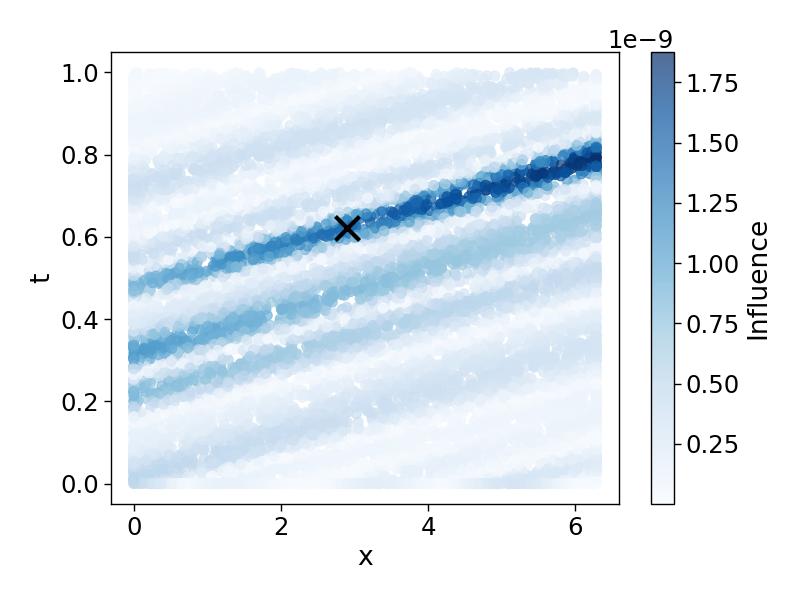}
        \end{subfigure}
        \hfill
        \begin{subfigure}[t]{0.48\textwidth}
            \centering
            \includegraphics[width=\textwidth]{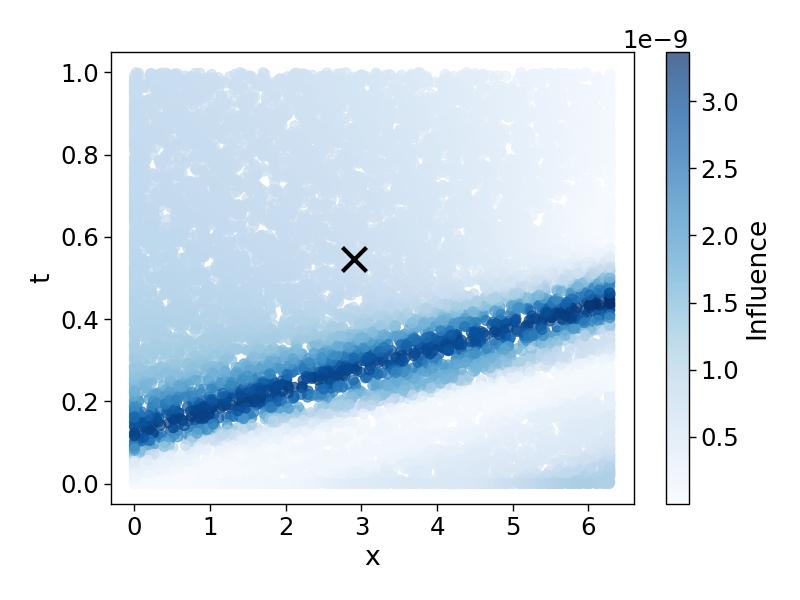}
        \end{subfigure}
    \end{minipage}
    \hfill
    \vrule 
    \hfill
    \begin{minipage}[t]{0.48\textwidth}
        \centering
        \begin{subfigure}[t]{0.48\textwidth}
            \centering
            \includegraphics[width=\textwidth]{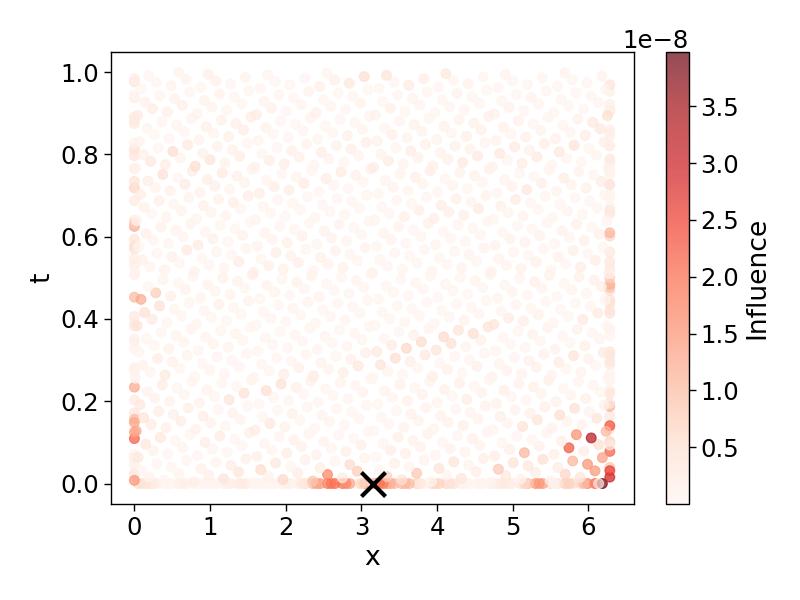}
        \end{subfigure}
        \hfill
        \begin{subfigure}[t]{0.48\textwidth}
            \centering
            \includegraphics[width=\textwidth]{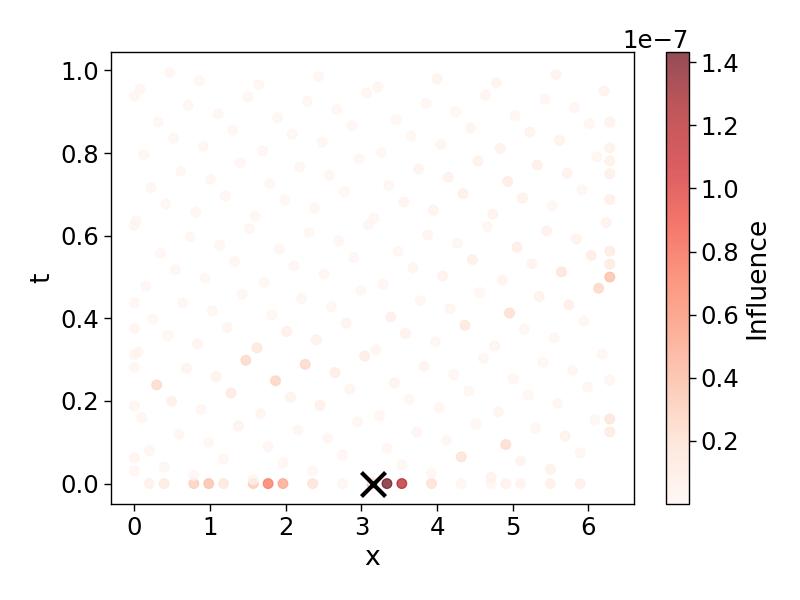}
        \end{subfigure}
        \\[0.5ex]
        \begin{subfigure}[t]{0.48\textwidth}
            \centering
            \includegraphics[width=\textwidth]{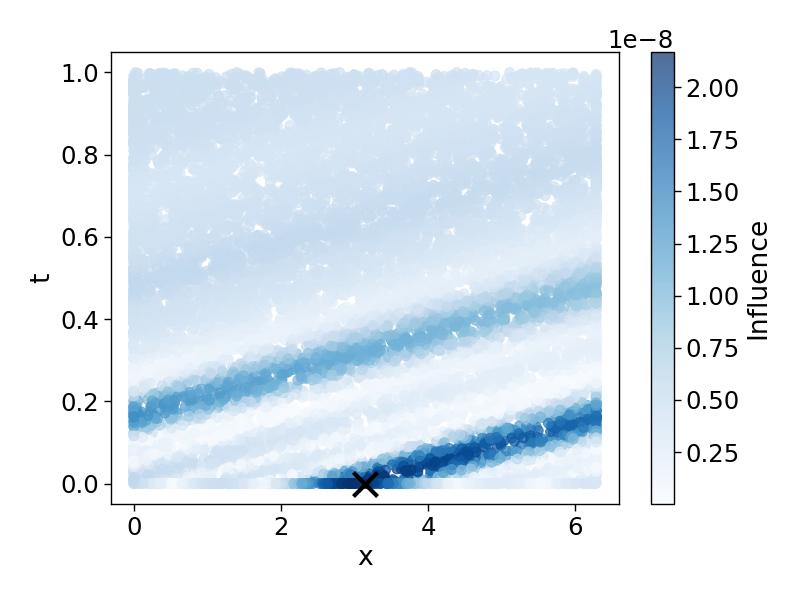}
        \end{subfigure}
        \hfill
        \begin{subfigure}[t]{0.48\textwidth}
            \centering
            \includegraphics[width=\textwidth]{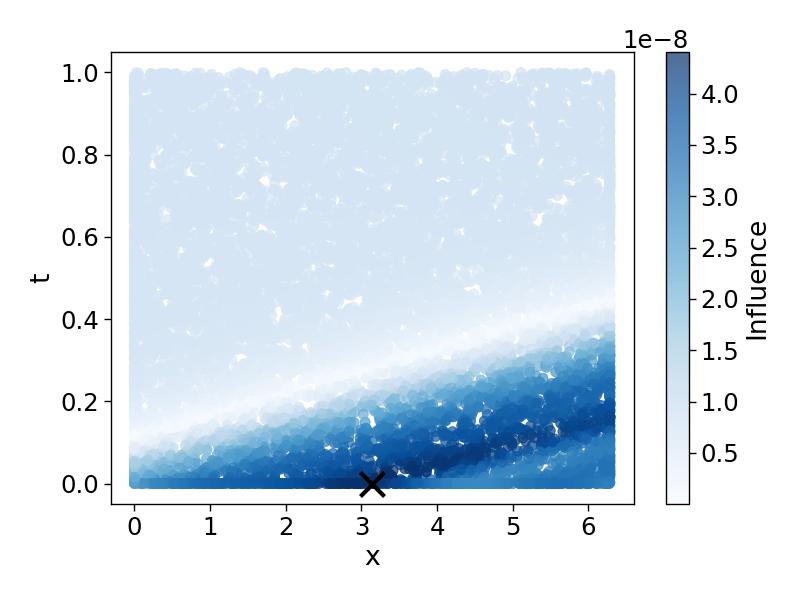}
        \end{subfigure}
    \end{minipage}
    \caption{Drift-diffusion equation: Influence scores for well-trained (left) and poorly-trained (right) models. Top: training point influences for fixed test point. Bottom: influences for a fixed training point onto the domain.}
    \label{fig:drift_diffusion_influences}
\end{figure}

\begin{figure}
    \centering
    \begin{minipage}[t]{0.48\textwidth}
        \centering
        \begin{subfigure}[t]{0.48\textwidth}
            \centering
            \includegraphics[width=\textwidth]{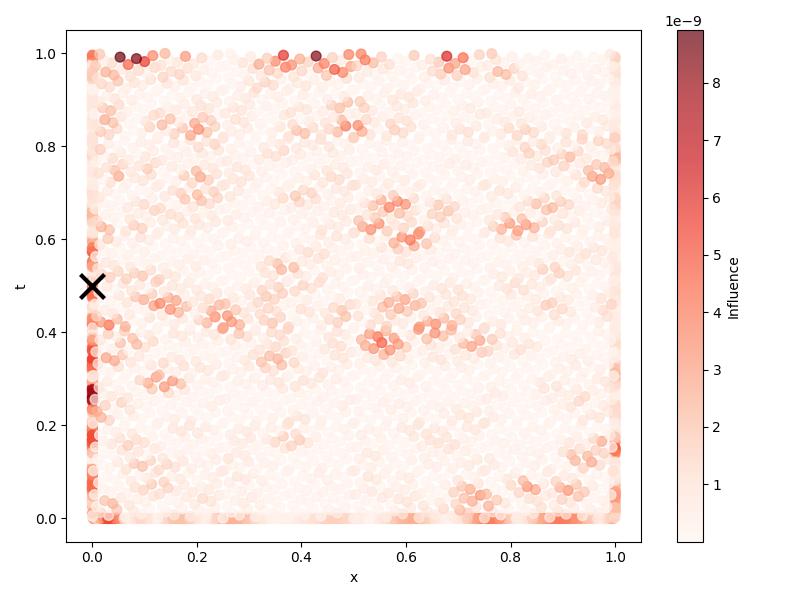}
        \end{subfigure}
        \hfill
        \begin{subfigure}[t]{0.48\textwidth}
            \centering
            \includegraphics[width=\textwidth]{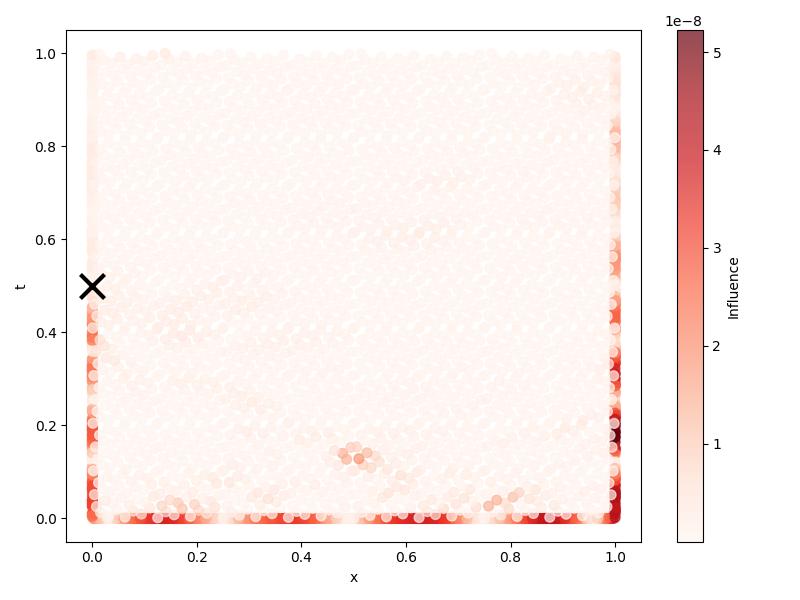}
        \end{subfigure}
        \\[0.5ex]
        \begin{subfigure}[t]{0.48\textwidth}
            \centering
            \includegraphics[width=\textwidth]{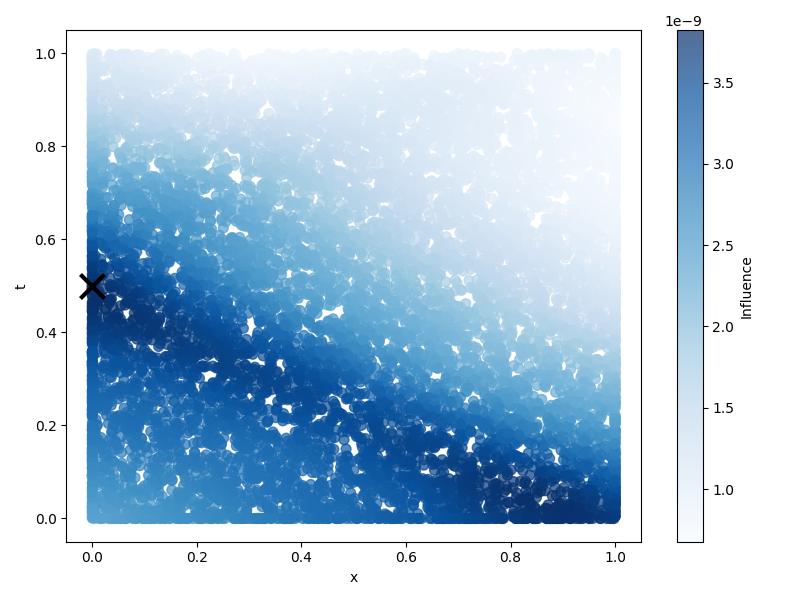}
        \end{subfigure}
        \hfill
        \begin{subfigure}[t]{0.48\textwidth}
            \centering
            \includegraphics[width=\textwidth]{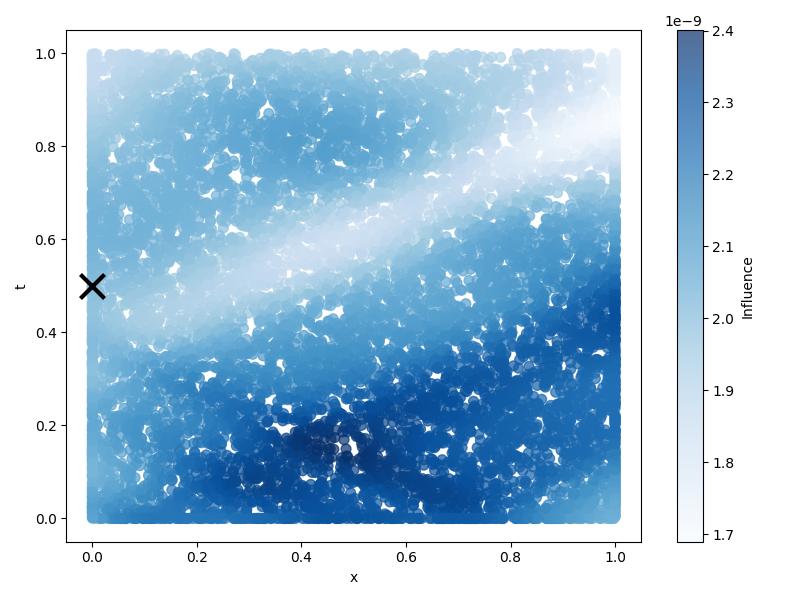}
        \end{subfigure}
    \end{minipage}
    \hfill
    \vrule 
    \hfill
    \begin{minipage}[t]{0.48\textwidth}
        \centering
        \begin{subfigure}[t]{0.48\textwidth}
            \centering
            \includegraphics[width=\textwidth]{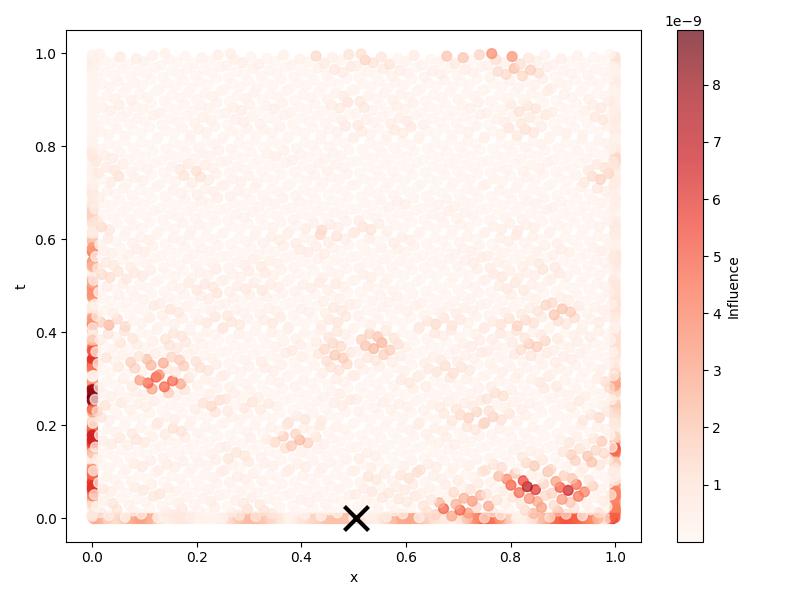}
        \end{subfigure}
        \hfill
        \begin{subfigure}[t]{0.48\textwidth}
            \centering
            \includegraphics[width=\textwidth]{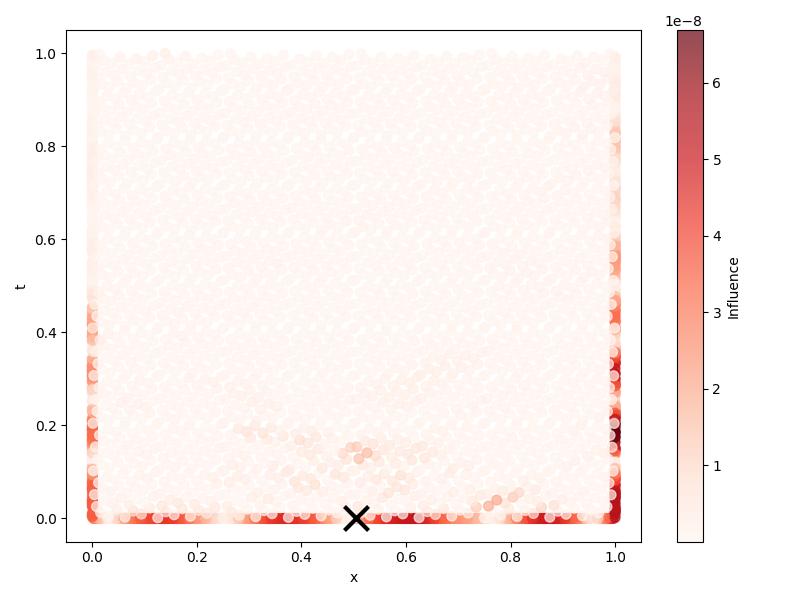}
        \end{subfigure}
        \\[0.5ex]
        \begin{subfigure}[t]{0.48\textwidth}
            \centering
            \includegraphics[width=\textwidth]{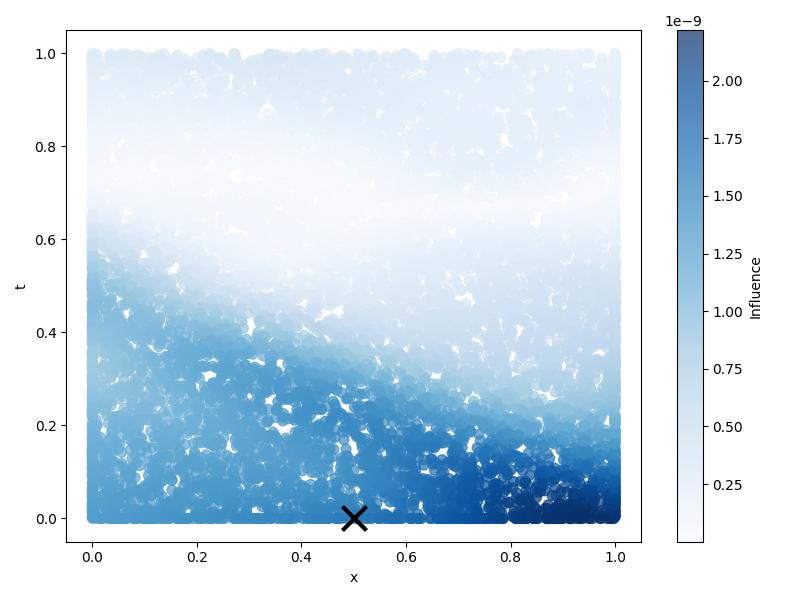}
        \end{subfigure}
        \hfill
        \begin{subfigure}[t]{0.48\textwidth}
            \centering
            \includegraphics[width=\textwidth]{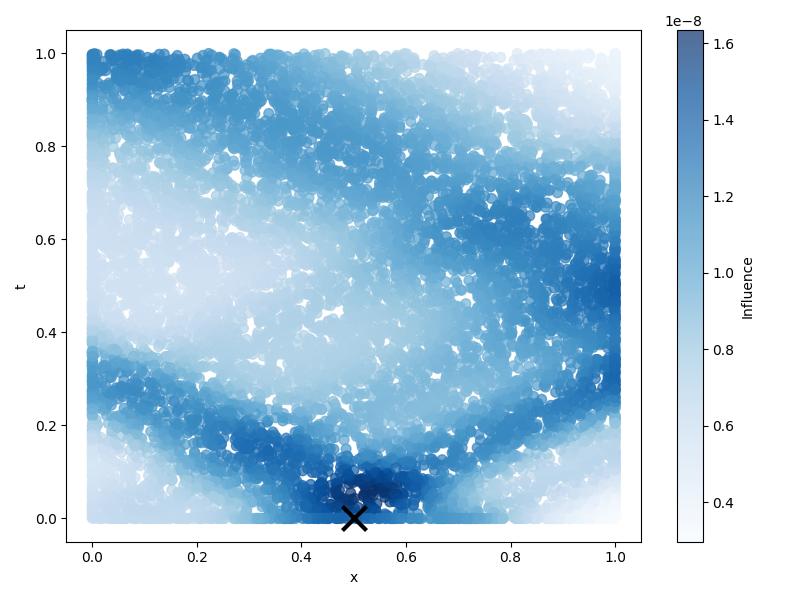}
        \end{subfigure}
    \end{minipage}
    \caption{Wave equation: Influence scores for well-trained (left) and poorly-trained (right) models. Top: training point influences for fixed test point. Bottom: influences for a fixed training point onto the domain.}
    \label{fig:wave_influences}
\end{figure}

\begin{figure}
    \centering
    \begin{minipage}[t]{0.48\textwidth}
        \centering
        \begin{subfigure}[t]{0.48\textwidth}
            \centering
            \includegraphics[width=\textwidth]{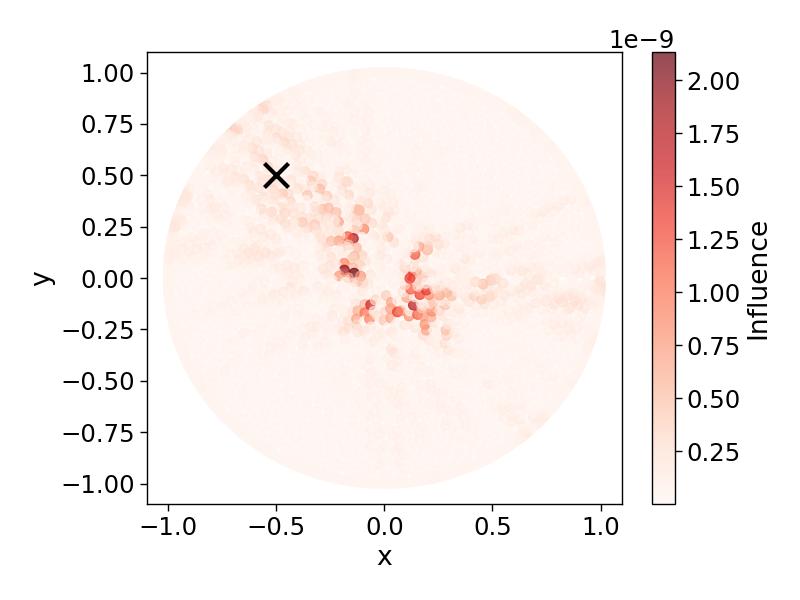}
        \end{subfigure}
        \hfill
        \begin{subfigure}[t]{0.48\textwidth}
            \centering
            \includegraphics[width=\textwidth]{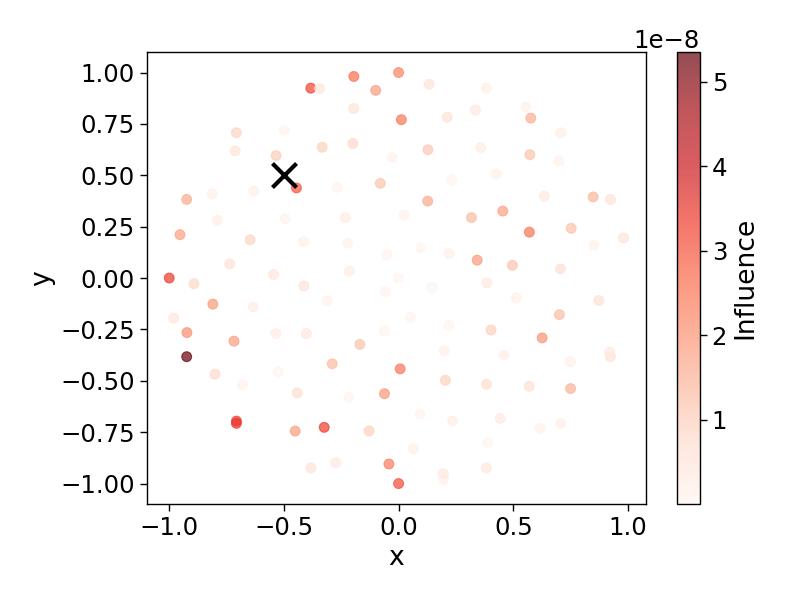}
        \end{subfigure}
        \\[0.5ex]
        \begin{subfigure}[t]{0.48\textwidth}
            \centering
            \includegraphics[width=\textwidth]{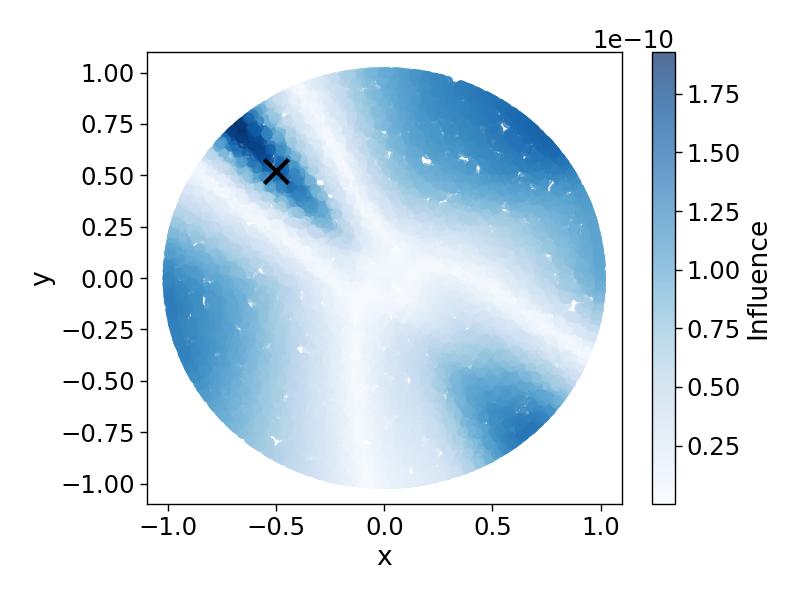}
        \end{subfigure}
        \hfill
        \begin{subfigure}[t]{0.48\textwidth}
            \centering
            \includegraphics[width=\textwidth]{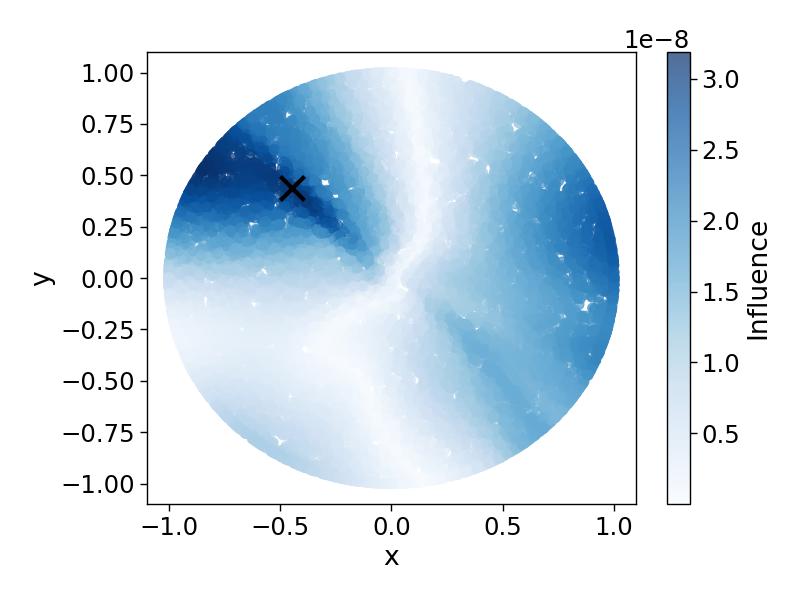}
        \end{subfigure}
    \end{minipage}
    \hfill
    \vrule 
    \hfill
    \begin{minipage}[t]{0.48\textwidth}
        \centering
        \begin{subfigure}[t]{0.48\textwidth}
            \centering
            \includegraphics[width=\textwidth]{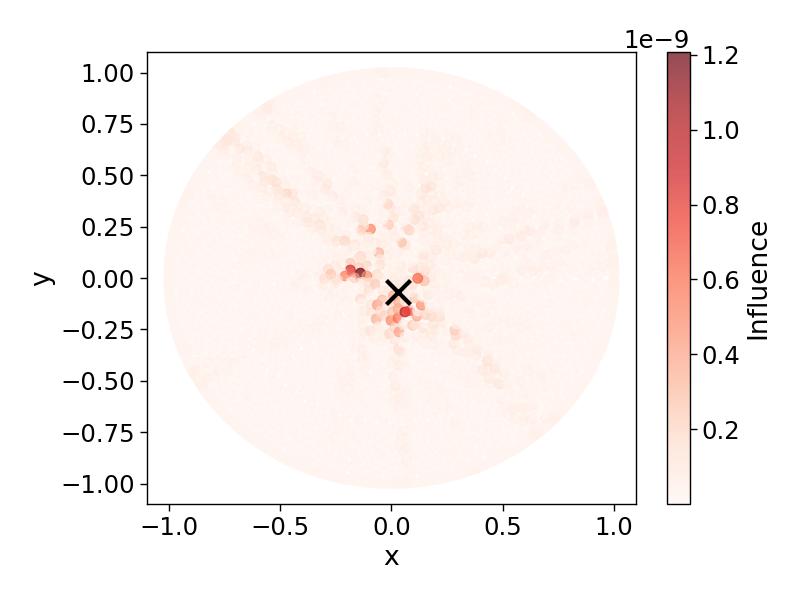}
        \end{subfigure}
        \hfill
        \begin{subfigure}[t]{0.48\textwidth}
            \centering
            \includegraphics[width=\textwidth]{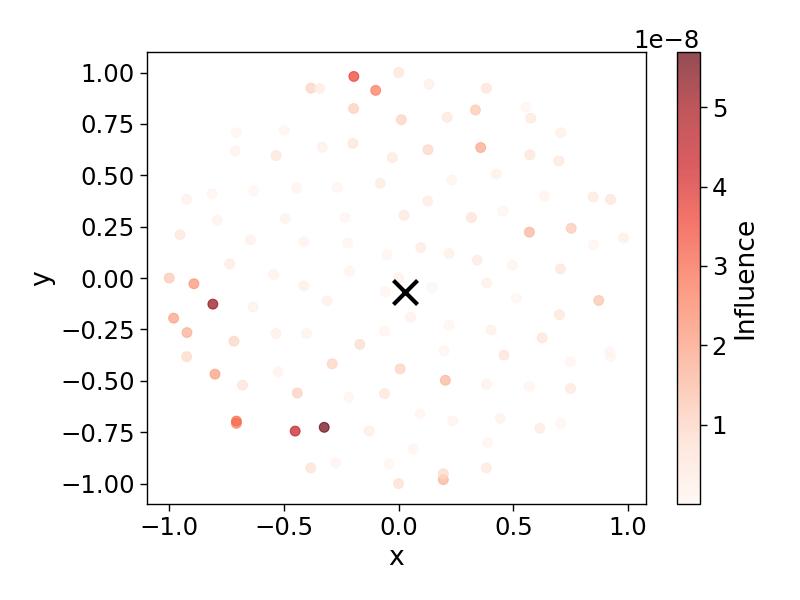}
        \end{subfigure}
        \\[0.5ex]
        \begin{subfigure}[t]{0.48\textwidth}
            \centering
            \includegraphics[width=\textwidth]{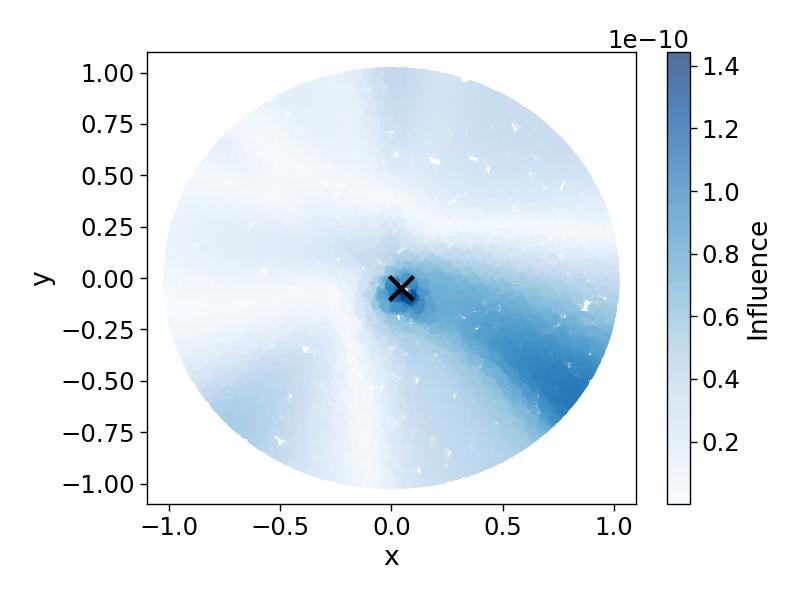}
        \end{subfigure}
        \hfill
        \begin{subfigure}[t]{0.48\textwidth}
            \centering
            \includegraphics[width=\textwidth]{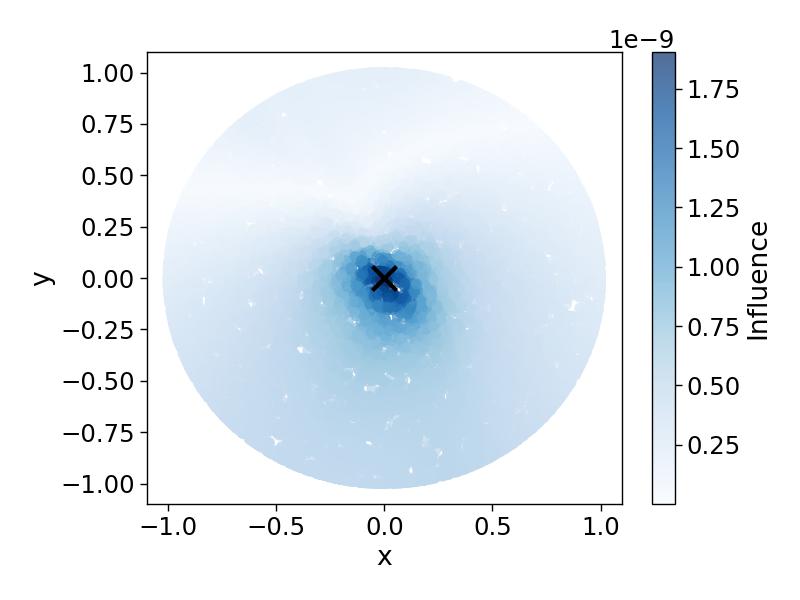}
        \end{subfigure}
    \end{minipage}
    \caption{Poisson equation: Influence scores for well-trained (left) and poorly-trained (right) models. Top: training point influences for fixed test point. Bottom: influences for a fixed training point onto the domain.}
    \label{fig:poisson_disk_influences}
\end{figure}

\begin{figure}
    \centering
    \begin{minipage}[t]{0.48\textwidth}
        \centering
        \begin{subfigure}[t]{0.48\textwidth}
            \centering
            \includegraphics[width=\textwidth]{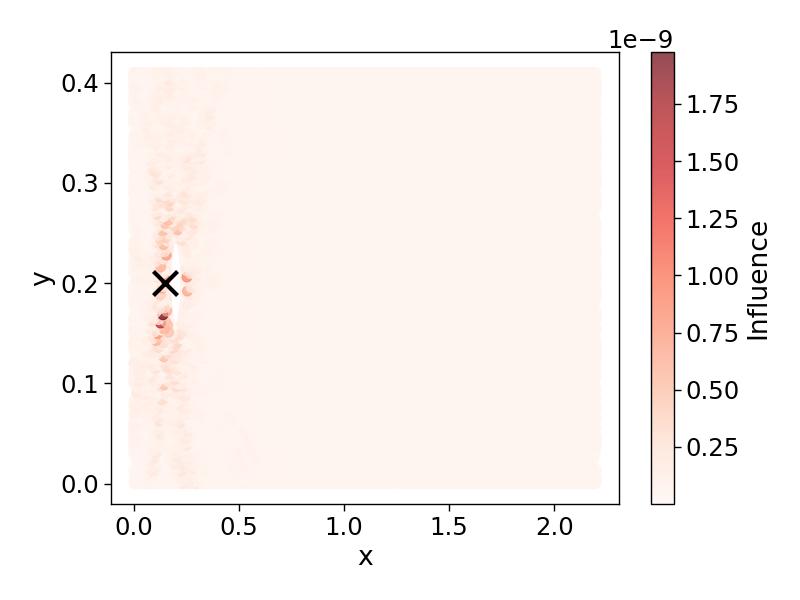}
        \end{subfigure}
        \hfill
        \begin{subfigure}[t]{0.48\textwidth}
            \centering
            \includegraphics[width=\textwidth]{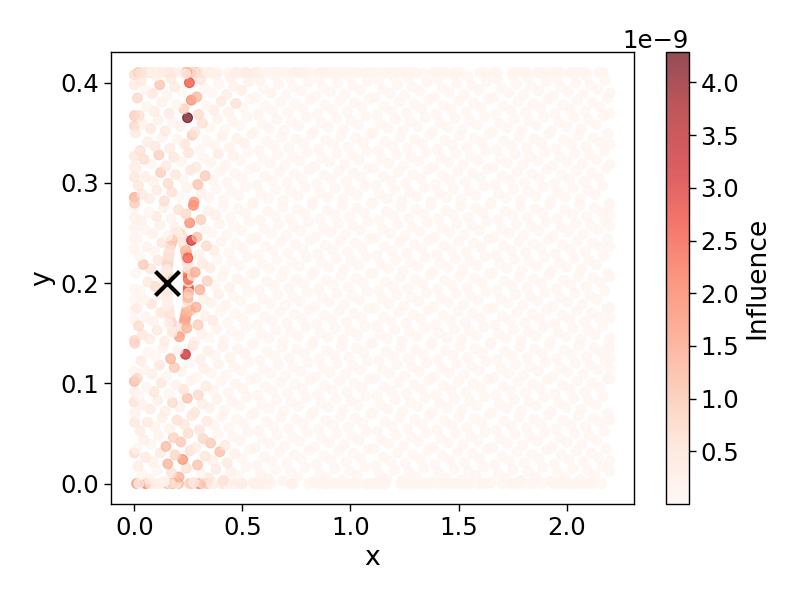}
        \end{subfigure}
        \\[0.5ex]
        \begin{subfigure}[t]{0.48\textwidth}
            \centering
            \includegraphics[width=\textwidth]{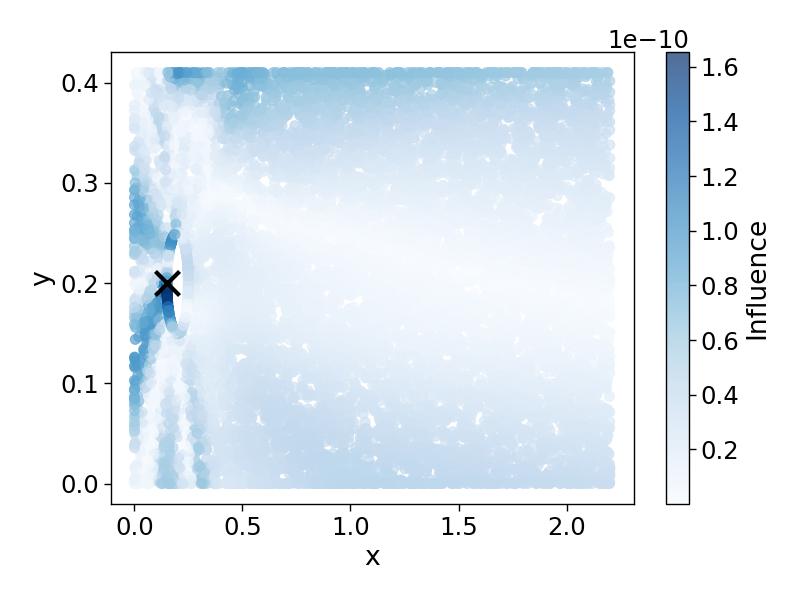}
        \end{subfigure}
        \hfill
        \begin{subfigure}[t]{0.48\textwidth}
            \centering
            \includegraphics[width=\textwidth]{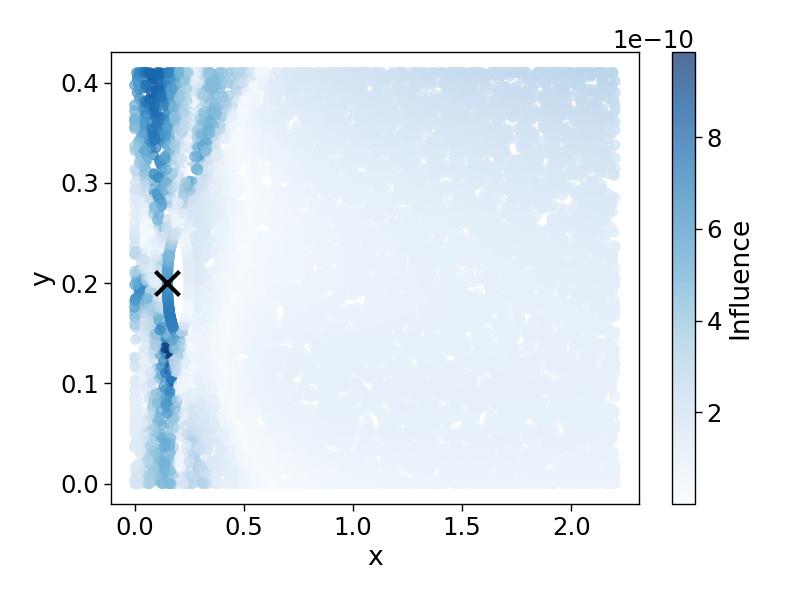}
        \end{subfigure}
    \end{minipage}
    \hfill
    \vrule 
    \hfill
    \begin{minipage}[t]{0.48\textwidth}
        \centering
        \begin{subfigure}[t]{0.48\textwidth}
            \centering
            \includegraphics[width=\textwidth]{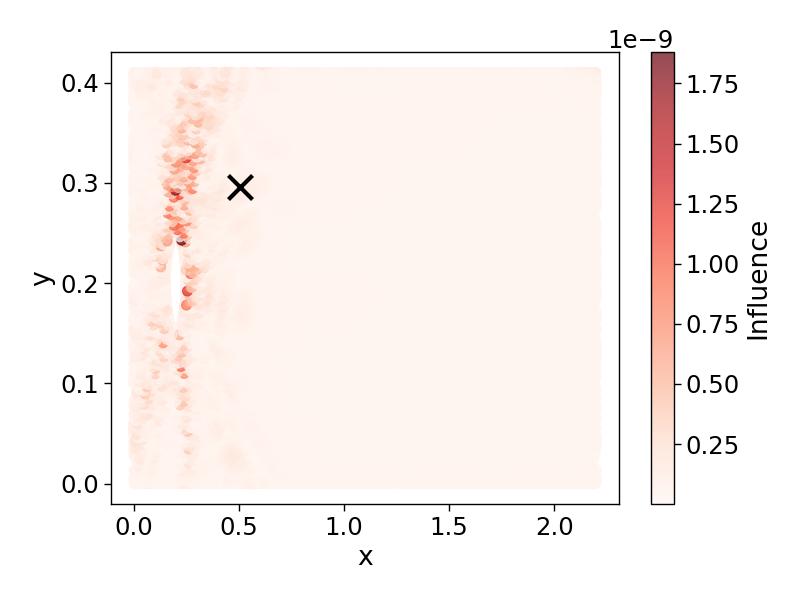}
        \end{subfigure}
        \hfill
        \begin{subfigure}[t]{0.48\textwidth}
            \centering
            \includegraphics[width=\textwidth]{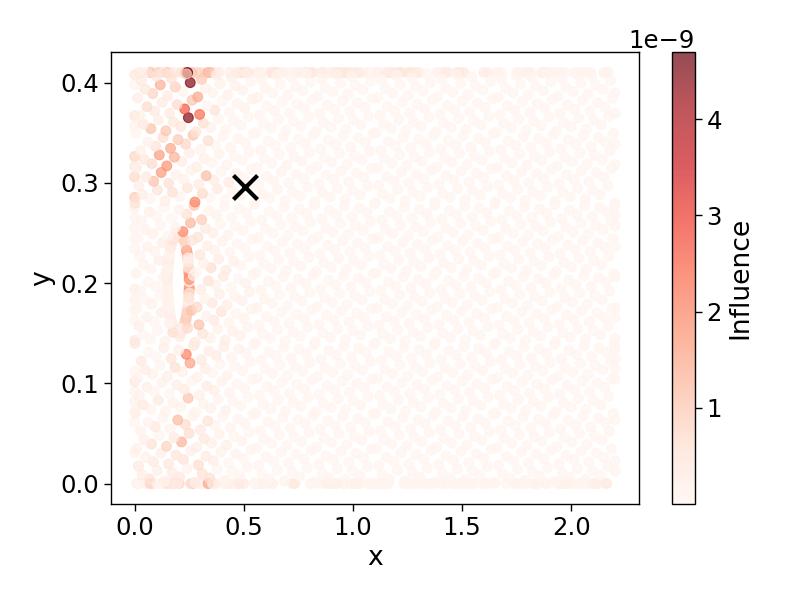}
        \end{subfigure}
        \\[0.5ex]
        \begin{subfigure}[t]{0.48\textwidth}
            \centering
            \includegraphics[width=\textwidth]{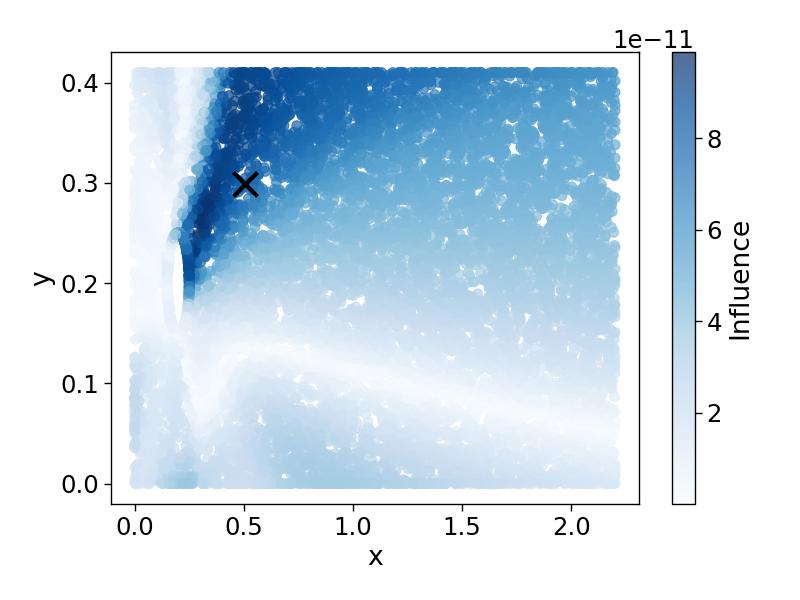}
        \end{subfigure}
        \hfill
        \begin{subfigure}[t]{0.48\textwidth}
            \centering
            \includegraphics[width=\textwidth]{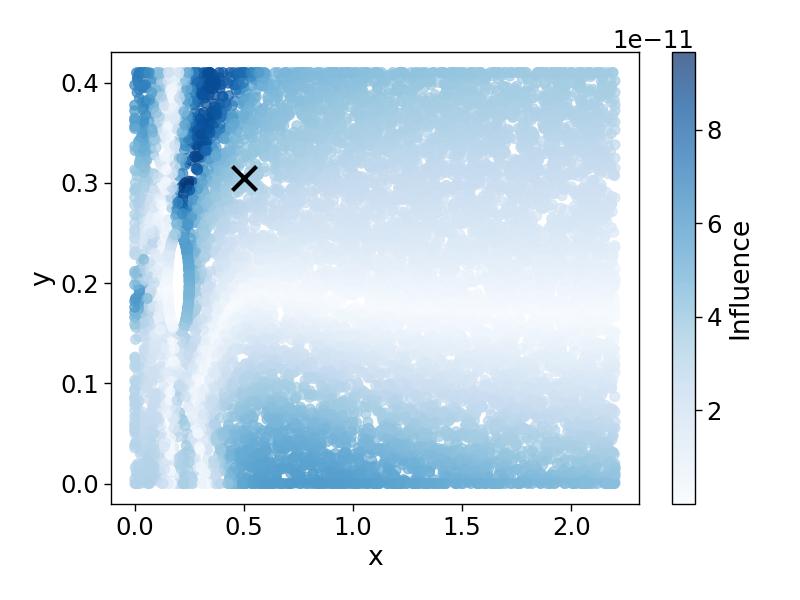}
        \end{subfigure}
    \end{minipage}
    \caption{Navier-Stokes equations ($x$-Velocity): Influence scores for well-trained (left) and poorly-trained (right) models. Top: training point influences for fixed test point. Bottom: influences for a fixed training point onto the domain}
    \label{fig:navier_stokes_influences_u}
\end{figure}

\begin{figure}
    \centering
    \begin{minipage}[t]{0.48\textwidth}
        \centering
        \begin{subfigure}[t]{0.48\textwidth}
            \centering
            \includegraphics[width=\textwidth]{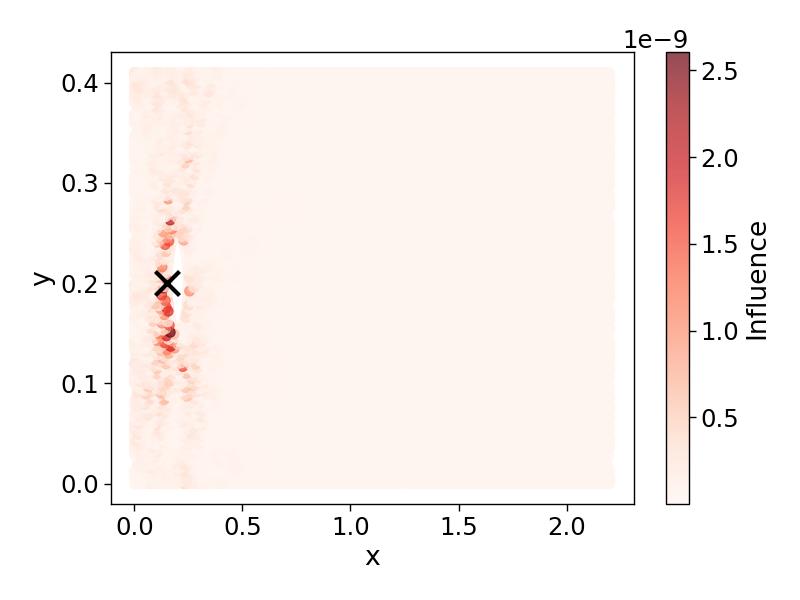}
        \end{subfigure}
        \hfill
        \begin{subfigure}[t]{0.48\textwidth}
            \centering
            \includegraphics[width=\textwidth]{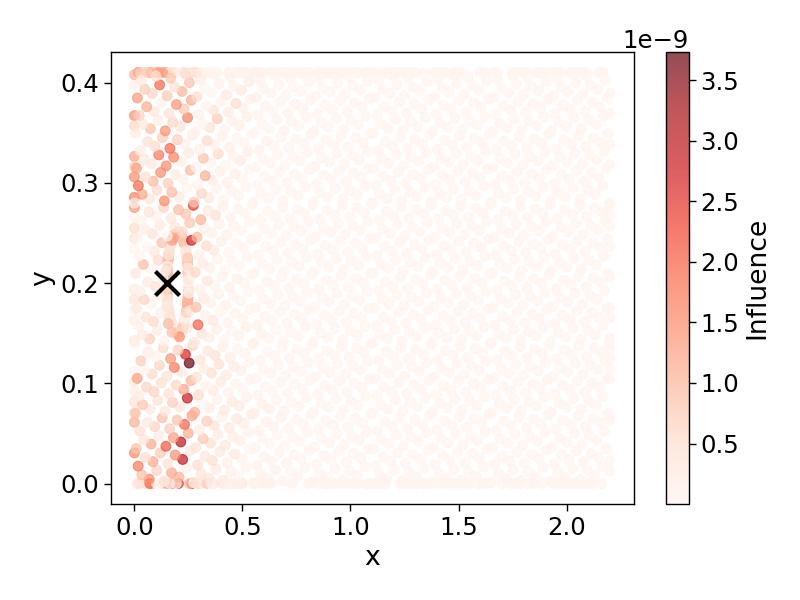}
        \end{subfigure}
        \\[0.5ex]
        \begin{subfigure}[t]{0.48\textwidth}
            \centering
            \includegraphics[width=\textwidth]{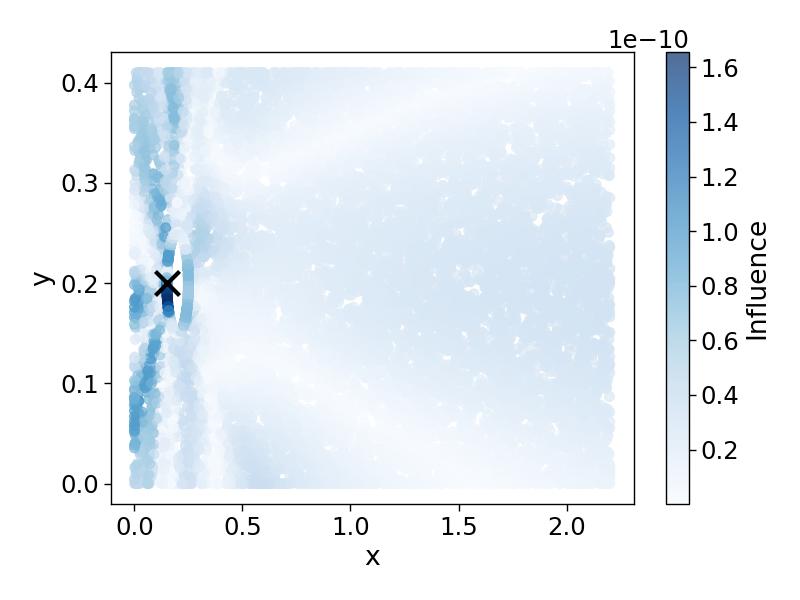}
        \end{subfigure}
        \hfill
        \begin{subfigure}[t]{0.48\textwidth}
            \centering
            \includegraphics[width=\textwidth]{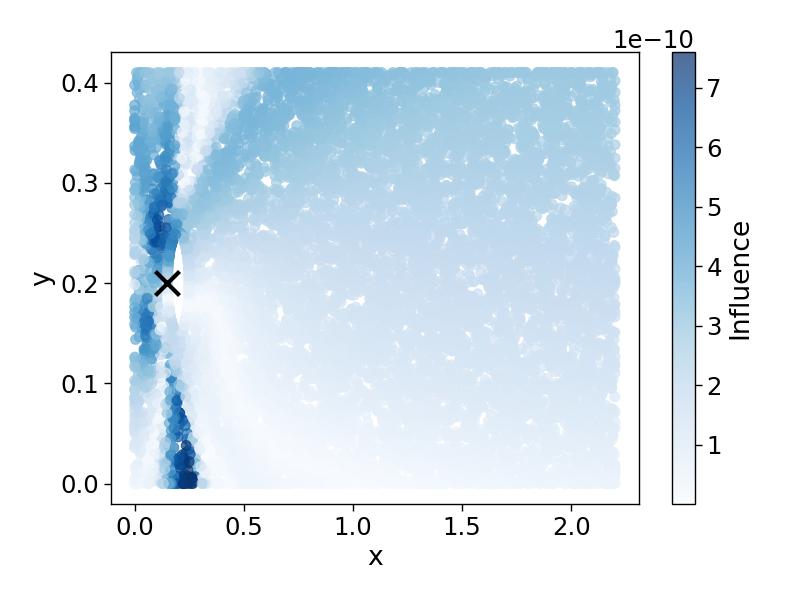}
        \end{subfigure}
    \end{minipage}
    \hfill
    \vrule 
    \hfill
    \begin{minipage}[t]{0.48\textwidth}
        \centering
        \begin{subfigure}[t]{0.48\textwidth}
            \centering
            \includegraphics[width=\textwidth]{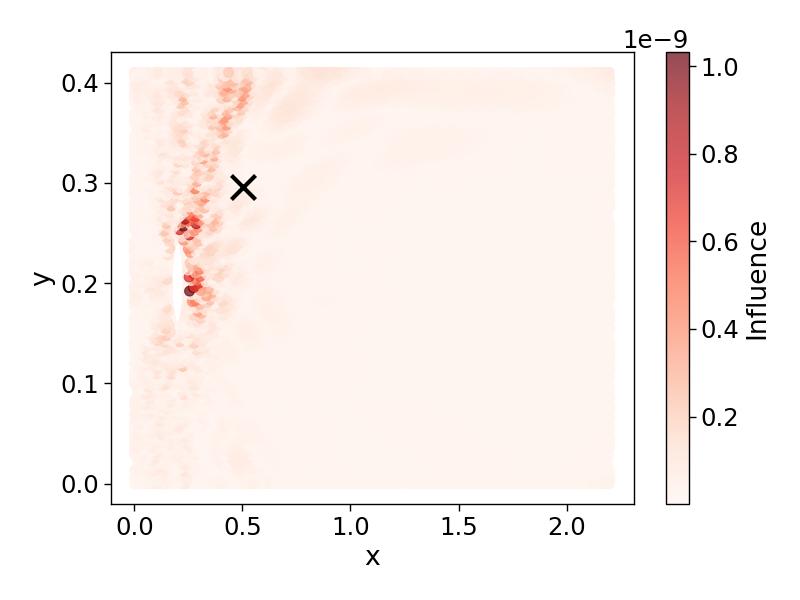}
        \end{subfigure}
        \hfill
        \begin{subfigure}[t]{0.48\textwidth}
            \centering
            \includegraphics[width=\textwidth]{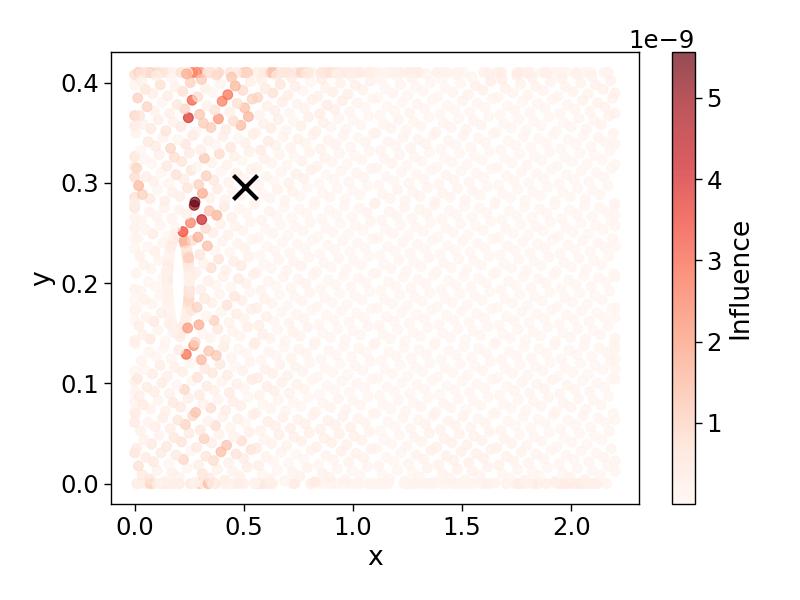}
        \end{subfigure}
        \\[0.5ex]
        \begin{subfigure}[t]{0.48\textwidth}
            \centering
            \includegraphics[width=\textwidth]{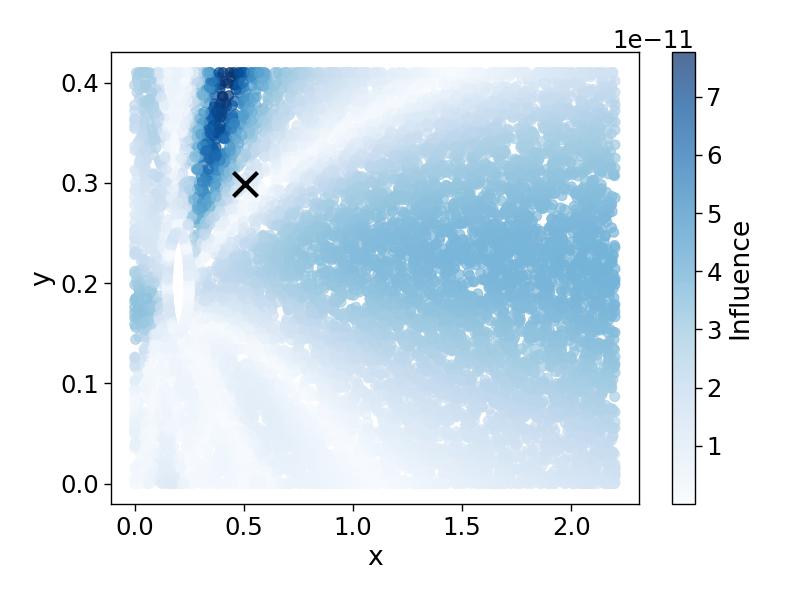}
        \end{subfigure}
        \hfill
        \begin{subfigure}[t]{0.48\textwidth}
            \centering
            \includegraphics[width=\textwidth]{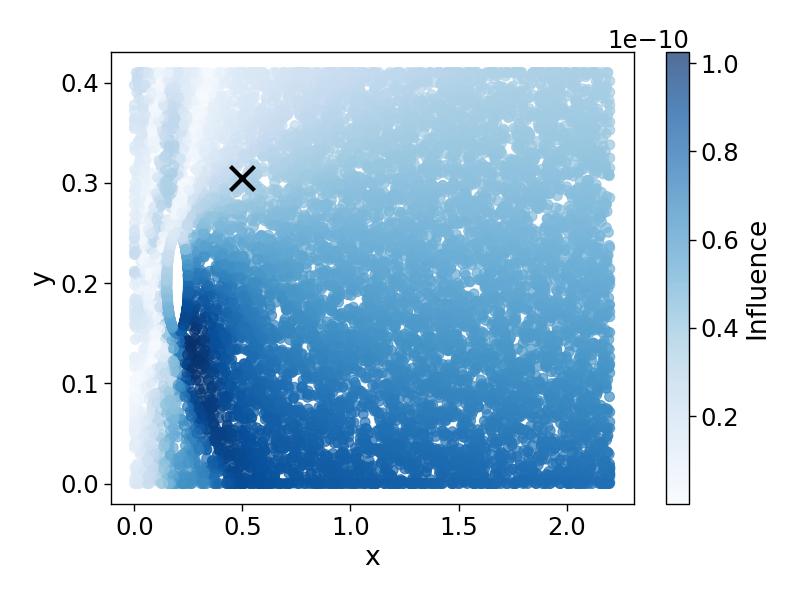}
        \end{subfigure}
    \end{minipage}
    \caption{Navier-Stokes equations ($y$-Velocity): Influence scores for well-trained (left) and poorly-trained (right) models. Top: training point influences for fixed test point. Bottom: influences for a fixed training point onto the domain.}
    \label{fig:navier_stokes_influences_v}
\end{figure}

\begin{figure}
    \centering
    \begin{minipage}[t]{0.48\textwidth}
        \centering
        \begin{subfigure}[t]{0.48\textwidth}
            \centering
            \includegraphics[width=\textwidth]{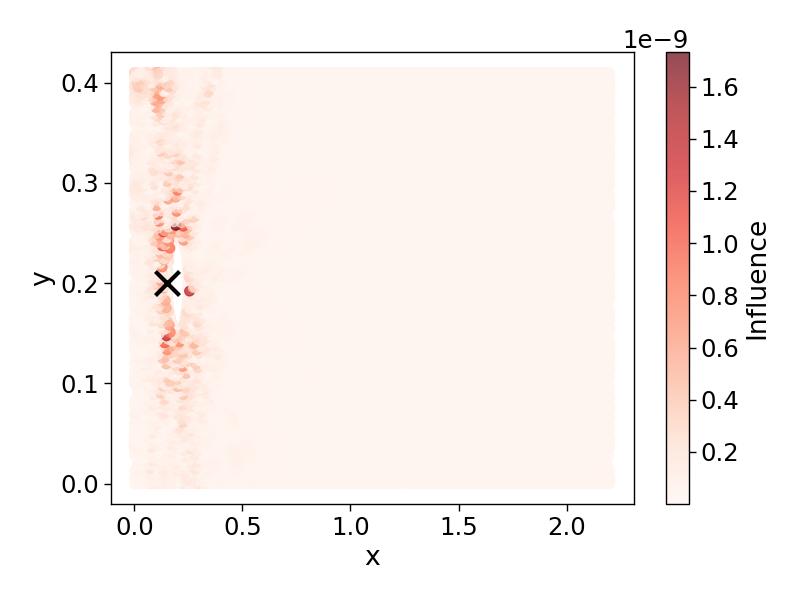}
        \end{subfigure}
        \hfill
        \begin{subfigure}[t]{0.48\textwidth}
            \centering
            \includegraphics[width=\textwidth]{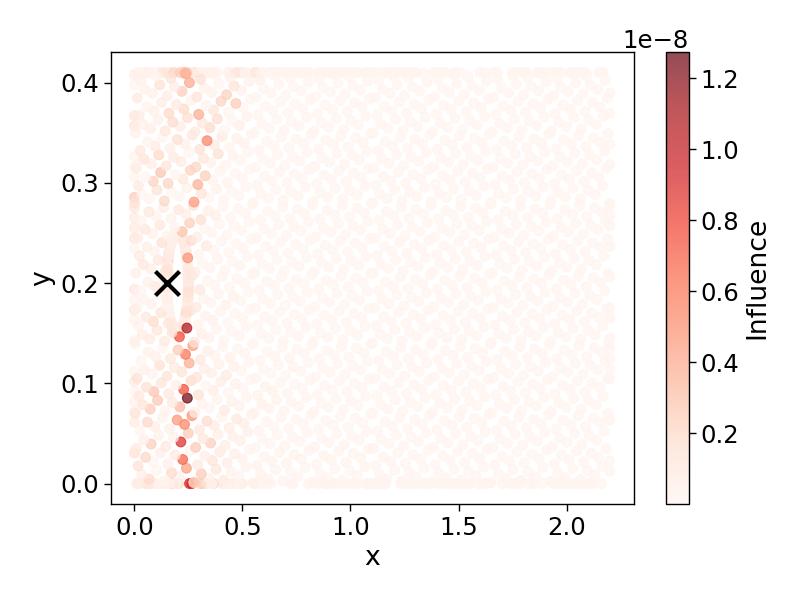}
        \end{subfigure}
        \\[0.5ex]
        \begin{subfigure}[t]{0.48\textwidth}
            \centering
            \includegraphics[width=\textwidth]{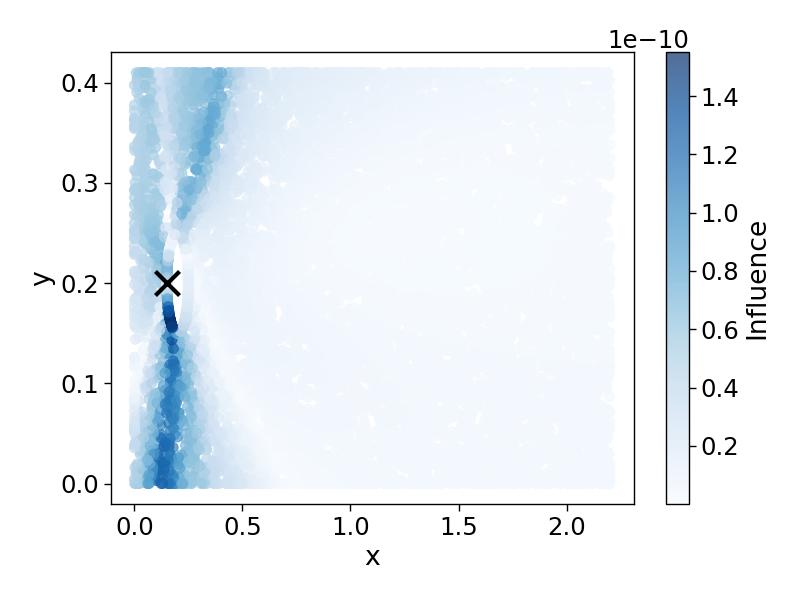}
        \end{subfigure}
        \hfill
        \begin{subfigure}[t]{0.48\textwidth}
            \centering
            \includegraphics[width=\textwidth]{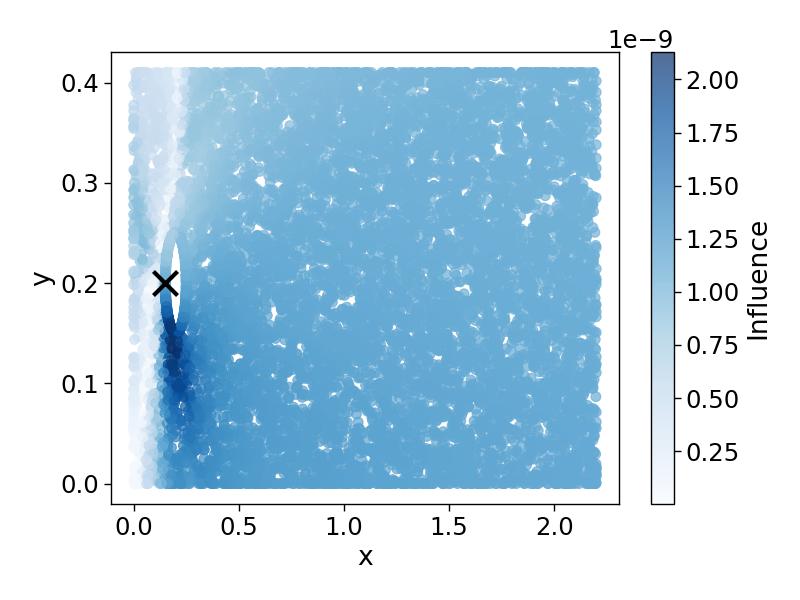}
        \end{subfigure}
    \end{minipage}
    \hfill
    \vrule 
    \hfill
    \begin{minipage}[t]{0.48\textwidth}
        \centering
        \begin{subfigure}[t]{0.48\textwidth}
            \centering
            \includegraphics[width=\textwidth]{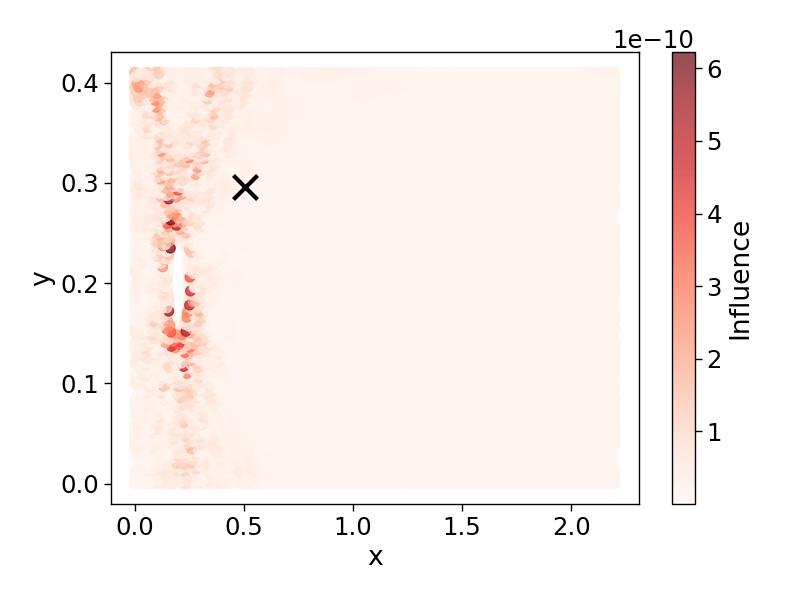}
        \end{subfigure}
        \hfill
        \begin{subfigure}[t]{0.48\textwidth}
            \centering
            \includegraphics[width=\textwidth]{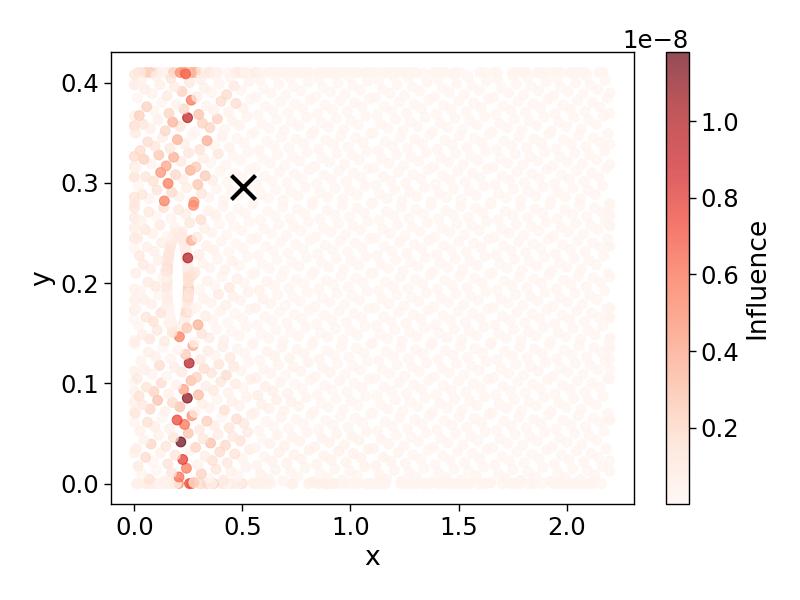}
        \end{subfigure}
        \\[0.5ex]
        \begin{subfigure}[t]{0.48\textwidth}
            \centering
            \includegraphics[width=\textwidth]{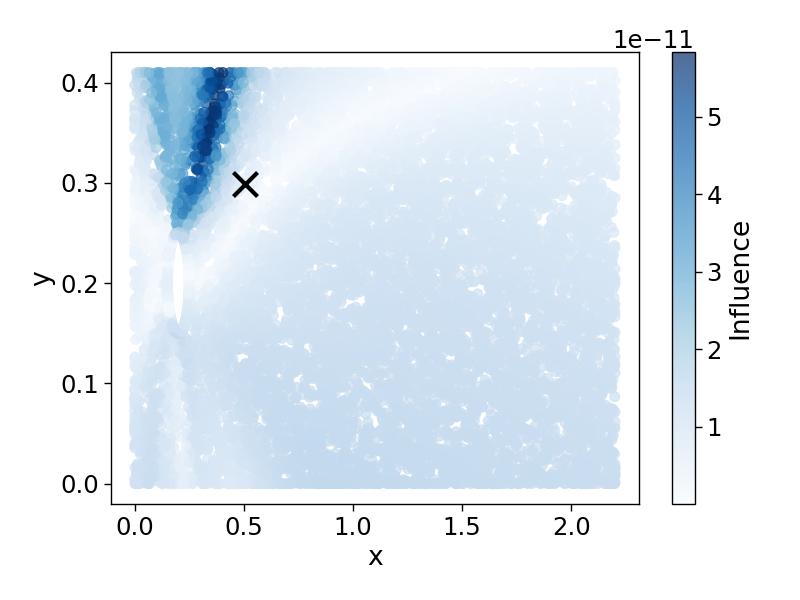}
        \end{subfigure}
        \hfill
        \begin{subfigure}[t]{0.48\textwidth}
            \centering
            \includegraphics[width=\textwidth]{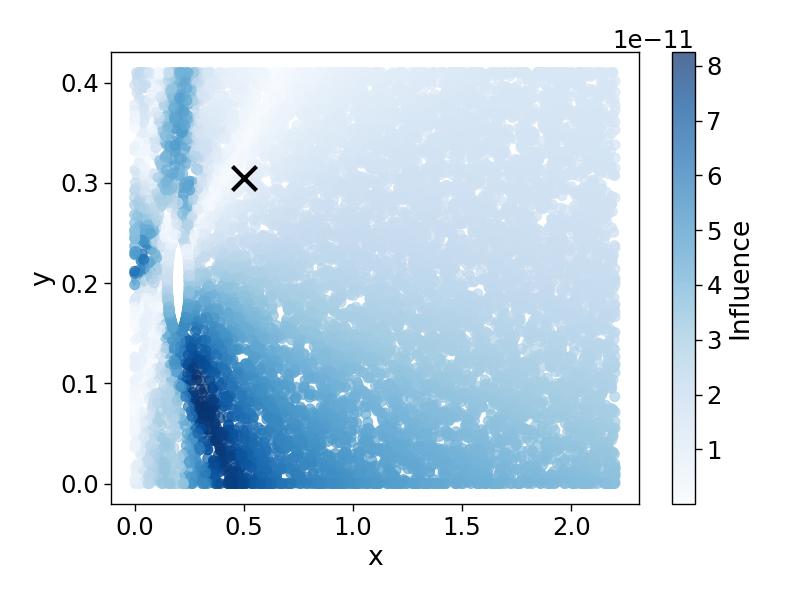}
        \end{subfigure}
    \end{minipage}
    \caption{Navier-Stokes equations (Pressure). Influence scores for well-trained (left) and poorly-trained (right) models. Top: training point influences for fixed test point. Bottom: influences for a fixed training point onto the domain.}
    \label{fig:navier_stokes_influences_p}
\end{figure}

\end{document}